\documentclass[12pt, a4paper,doubleside]{Thesis} 

\usepackage[square, comma, sort&compress, numbers]{natbib} 

\usepackage[T1]{fontenc}
\usepackage[]{palatino} 
\normalfont
\usepackage{anyfontsize}

\divide\pdfpxdimen by 50

\renewcommand\frontmatter{%
   \cleardoublepage
  \pagenumbering{Roman}}

\hypersetup{urlcolor=blue, colorlinks=true} 
\title{\ttitle} 

\begin{document}

\setstretch{1.3} 

\fancyhf{} 
\pagestyle{fancy} 
\newcommand{\HRule}{\rule{\linewidth}{0.5mm}} 

\hypersetup{pdftitle={\ttitle}}
\hypersetup{pdfsubject=\subjectname}
\hypersetup{pdfauthor=\authornames}
\hypersetup{pdfkeywords=\keywordnames}


\begin{titlepage}
\begin{center}

\large \textsc{ \DEunivname}\\[1.5cm] 

\HRule \\[0.4cm] 
{\huge \bfseries \ttitle}\\[0.3cm] 
\HRule \\[1.4cm] 

\Large  {\bf Doctoral Thesis}\\[0.4cm]
 \large{by}\\[0.4cm]
\Large{\bf \authornames}\\[0.4cm]
\large {born on the 28th of June 1984 in Barcelona,}\\[2cm]
\large{approved in fulfillment of the requirements for the degree of}\\[0.4cm] 
\large {\bf \degreename}\\[0.4cm]
\large {at the \facname\\ of the \DEunivname}\\[1.5cm]

\includegraphics[width=0.3\linewidth]{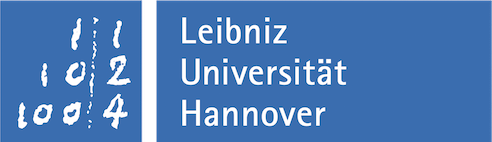}\\[1.4cm] 

\large{ 2014}\\ 
\end{center}
\end{titlepage}

\newpage
\thispagestyle{empty}
\mbox{}


\begin{titlepage}
\begin{center}

\textsc{\large \DEunivname}\\[1.5cm] 

\HRule \\[0.4cm] 
{\huge \bfseries \ttitle}\\[0.4cm] 
\HRule \\[1.5cm] 

\large {Von der \DEfacname\\ der \DEunivname  \\ zur Erlangung des akademischen Grades }\\[0.4cm] 
\large {\bf \DEdegreename}\\[0.4cm]
\large {genehmigte}\\[0.4cm] 
\Large {\bf Dissertation}\\[1.5cm]
\large {von}\\[0.4cm]
\Large{\bf \authornames}\\[0.4cm]
\large {geboren am 28. Juni 1984 in Barcelona.}\\[1.5cm]

\includegraphics[width=0.3\linewidth]{logo}\\[1.4cm] 

{\large 2014}\\ 
\end{center}
\end{titlepage}

\newpage
\thispagestyle{empty}
\mbox{}


\begin{titlepage}
{\bf Doktorvater / Supervisor: }\\[0.4cm] 
{\supname}\\
{\DEunivname, Germany}\\[1cm]

{\bf Gutachter / Reviewers: }\\[0.4cm] 
{\supname}\\
{\DEunivname, Germany}\\[0.4cm] 
{Prof. Dr. Daniel Cremers}\\
{\texorpdfstring{\href{http://www.tum.de/}
                {Technical University of Munich (TUM)}, Germany} 
                {Technical University of Munich (TUM)}}\\[0.4cm] 
{Prof. Dr.-Ing. J\"{o}rn Ostermann}\\
{\DEunivname, Germany}\\[1cm] 

\vfill

{\bf Datum des Kolloquiums / Date of Defense: }\\[0.4cm] 
{11.02.2014}\\

\vfill

{\bf Autor / Author: }\\[0.4cm] 
{\authornames}\\
{\groupname}\\
{\addressnames}\\
{\emailnames}\\

\end{titlepage}

\newpage
\thispagestyle{empty}
\mbox{}

\newpage


\frontmatter 

\addtotoc{Abstract} 

\abstract{\addtocontents{toc}{\vspace{1em}} 

Multiple people tracking is a key problem for many applications such as surveillance, animation or car navigation, and a key input for tasks such as activity recognition. In crowded environments occlusions and false detections are common, and although there have been substantial advances in recent years, tracking is still a challenging task.
Tracking is typically divided into two steps: detection, \ie, locating the pedestrians in the image, and data association, \ie, linking detections across frames to form complete trajectories.
 For the data association task, approaches typically aim at developing new, more complex formulations, which in turn put the focus on the optimization techniques required to solve them. However, they still utilize very basic information such as distance between detections. 
In this thesis, I focus on the data association task and argue that there is contextual information that has not been fully exploited yet in the tracking community, mainly  {\it social context} and {\it spatial context} coming from different views. 
As tracking framework I use a global optimization method that finds the best solution for all pedestrian trajectories and all frames using Linear Programming. This is the perfect setup to include contextual information that can be used to improve all trajectories. 
Firstly, I present an efficient way to include social and grouping behavior to improve monocular tracking. Incorporating this source of information leads to much more accurate tracking results, especially in crowded scenarios.
Secondly, I present a formulation to perform 2D-3D assignments (reconstruction) and temporal assignments (tracking) in a single global optimization. I show that linking the reconstruction and tracking processes in a tight formulation leads to a significant boost in tracking accuracy. 
Overall, I show that context is an extremely rich source of information that can be exploited to obtain more accurate tracking results.\\[1cm] 

}
\newpage
\thispagestyle{empty}


\addtotoc{Zusammenfassung} 

\zusammenfassung{\addtocontents{toc}{\vspace{1em}} 

Die kamerabasierte Verfolgung (Tracking) von Personen ist von
wesentlicher Bedeutung f{\"u}r viele Anwendungen aus den Bereichen der Sicherheitstechnik, der Fahrerassistenzsysteme oder der Animation, und eine wichtige Grundlage f{\"u}r die Aktivit{\"a}tserkennung. In komplexen Umgebungen mit gro{\ss}en
Menschenmengen treten regel- m{\"a}{\ss}ig Verdeckungen und Falscherkennungen auf, und
obwohl in den letzten Jahren erhebliche Fortschritte erzielt wurden, ist
das Tracking nach wie vor eine anspruchs- volle Aufgabe. 
{\"U}blicherweise
wird das Tracking in zwei Schritte unterteilt: Erstens die Erkennung, \dhgerman die Lokalisierung der Fu{\ss}g{\"a}nger im Bild, und zweitens die
Datenassoziation, \dhgerman die Verkn{\"u}pfung der Detektionen {\"u}ber alle Einzelbilder, um Trajektorien zu bilden.
Ans{\"a}tze zur L{\"o}sung des Problems der Datenassoziation sind oftmals bestrebt neue, komplexere Formulierungen zu entwickeln,
um vollst{\"a}ndigere Trajektorien zu erhalten. Diese lenken wiederum den
Fokus auf Optimierungsmethoden, die notwendig sind um sie zu l{\"o}sen.
Dabei werden {\"u}blicherweise nur grundlegende Informationen wie der
Abstand zwischen Detektionen genutzt. In dieser Schrift liegt der Fokus auf der Datenassoziation. Ich argumentiere, dass
kontextabh{\"a}ngige Informationen verf{\"u}gbar sind und effizient in Trackingalgorithmen eingebaut werden k{\"o}nnen, in Form von sozialem und r{\"a}umlichem Kontext.
Als Werkzeug zum Tracking benutze ich die Lineare Programmierung als globale
Optimierungsmethode, die eine eindeutige L{\"o}sung f{\"u}r alle
Fu{\ss}g{\"a}ngertrajektorien und alle Einzelbilder findet. Das ist der perfekte
Aufbau, um kontextabh{\"a}ngige Informationen zu integrieren. Zuerst
pr{\"a}sentiere ich ein effizientes Verfahren zum Erfassen von Sozial- und
Gruppenverhalten, um das monokulare Tracking zu verbessern.
Die Ber{\"u}cksichtigung dieser Informationsquelle f{\"u}hrt zu einem viel
genaueren Tracking-Ergebnis, besonders in Szenarien mit Menschenmengen.
Zweitens pr{\"a}sentiere ich eine Formulierung, um 2D-3D Zuordnungen
(Rekonstruktion) und temporale Zuordnungen (Tracking) in einer einzigen,
globalen Optimierung durchzuf{\"u}hren. Ich zeige, dass das Verkn{\"u}pfen von
 Rekonstruktion und Tracking in einer gemeinsamen
Formulierung zu einem erheblichen Anstieg der
Genauigkeit f{\"u}hrt.
Insgesamt zeige ich, dass der Kontext eine reiche
Informationsquelle ist, die genutzt werden kann um genauere Ergebnisse beim Tracking zu erzielen.


}
\clearpage


\addtotoc{Acknowledgements}

\acknowledgements{\addtocontents{toc}{\vspace{1em}} 

I would first like to thank my advisor, for always believing in my ideas, pushing me to reach my limits and for allowing me to follow my own research path. 

Special thanks to the all the members of my thesis committee, for dedicating the time to read my thesis and evaluate my work. 

A warm thanks to all the members of the Hannover group, both the ones who have already moved on and the ones who are still there. I want to thank them for the fun memories, the fruitful collaborations, the late deadline nights, the amazing Doktorhut you created, and the many hours of discussion and games during coffee time! I will never forget my time on the 13th floor. 

Thanks to the people from the Michigan laboratory, who gave me the warmest welcome during Midwest winter. I learned so much from all you guys and had tons of fun during my internship. 

Thanks to the many friends I made in the computer vision community during conferences, for the many interesting discussions, suggestions, collaborations, opportunities offered, and also for the all the fun we had. 

I would like to thanks my friends from Barcelona. Although most of us are abroad now, we still regularly meet and talk as if no time had passed. I hope we can do that for many years to come. 

A special thanks to all my family, cousins, aunts and uncles, grandparents... for their infinite support and love, and for the amazing holidays we always spend together. 

And the warmest biggest thanks goes to my parents, for their endless love, unconditional support and for teaching me everything I know.  
 
}

\newpage\null\thispagestyle{empty}\newpage


\pagestyle{fancy} 

\fancyhf{} 
\fancyhead[LO,RE]{Contents} 
\fancyhead[RO,LE]{\thepage} 
\tableofcontents 
%
%

\newpage\null\thispagestyle{empty}\newpage


\pagenumbering{gobble}
\fancyhf{} 
\pagestyle{fancy} 
\dedicatory{Dedicated to my parents} 

\newpage\null\thispagestyle{empty}\newpage


\mainmatter 

\fancyhf{} 
\fancyhead[RE]{\leftmark}
\fancyhead[LO]{\rightmark}
\fancyhead[RO,LE]{\thepage} 
\pagestyle{fancy} 



\chapter{Introduction} 

\label{introduction} 

\graphicspath{{./Figures/intro/}}

\section{Motivation}

Video cameras are increasingly present in our daily lives: webcams, surveillance cameras and other imaging devices are being used for multiple purposes. As the number of data streams increases, it becomes more and more important to develop methods to automatically analyze this type of data. People are usually the central characters of most videos, therefore, it is particularly interesting to develop techniques to analyze their behavior. 
Either for surveillance, animation or activity recognition, multiple people tracking is a key problem to be addressed. In crowded environments occlusions and false detections are common, and although there have been substantial advances in the last years, tracking is still a challenging task.
The task is typically divided into two steps: detection and data association. Detectors are nowadays very robust and provide extremely good detection rates for normal scenes, but still struggle with partial and full occlusions common in crowded scenes. Data association or tracking, on the other hand, is also extremely difficult in crowded scenarios, especially due to the high rate of missing data and common false alarms.
In this thesis, we argue that there are two main sources of pedestrian context that have not been fully exploited in the tracking community, namely {\it social context} and {\it spatial context} coming from different views. 

Typically, matching is solely based on appearance and distance information, \ie, the closest detection in the following frame is matched to the detection in the current frame. But this can be completely wrong: let us imagine a queue of people waiting at a coffee shop and a low frame rate camera, as is typical for surveillance scenarios. In one frame we might have 4 persons waiting, while in the next the first person is already out of the queue and a new person entered the queue. In this case, if we only use distance information, the 4 persons of the first frame might be matched to the 4 persons of the second frame, although they are completely different pedestrians. Though this is an extreme case, it represents an error that is common while tracking in crowded scenarios, and this is only caused by the assumption that people do not move from one frame to the next, which is clearly inaccurate.

It is therefore more natural to take into account the context of the pedestrian, which can be the activity they are performing (\eg, queueing) or the interactions that take place in a crowded scenario. It is clear that if a person is walking alone, he/she will follow a straight path towards his/her destination. But what if the environment becomes more crowded, and suddenly the straight path is no longer an option? The pedestrian will then try to find a rather short path to get to the same destination by avoiding other pedestrians and obstacles. All these pedestrian movements and reactions to the environment are ruled by what is called the Social Force Model. 

Another source of information that has not been fully exploited in the literature is the spatial context coming from different camera views. It is typical for many applications to observe the same scenario from different viewpoints. In this case, object locations in the images are temporally correlated by the system dynamics and are geometrically constrained by the spatial configuration of the cameras. 
These two sources of structure have been typically exploited separately, but splitting the problem in two steps has obviously several disadvantages, because the available evidence is not fully exploited. For example, if one object is temporarily occluded in one camera, both data association for reconstruction and tracking become ambiguous and underconstrained when considered separately. 
If, on the other hand, evidence is considered jointly, temporal correlation can potentially resolve reconstruction ambiguities and vice versa. 

In this thesis, we will show that pedestrian context is an incredibly rich source of information that should be included in the tracking procedure.

\section{Contributions and Organization}

\begin{figure*}[p]
\centering
\includegraphics[width=1\linewidth]{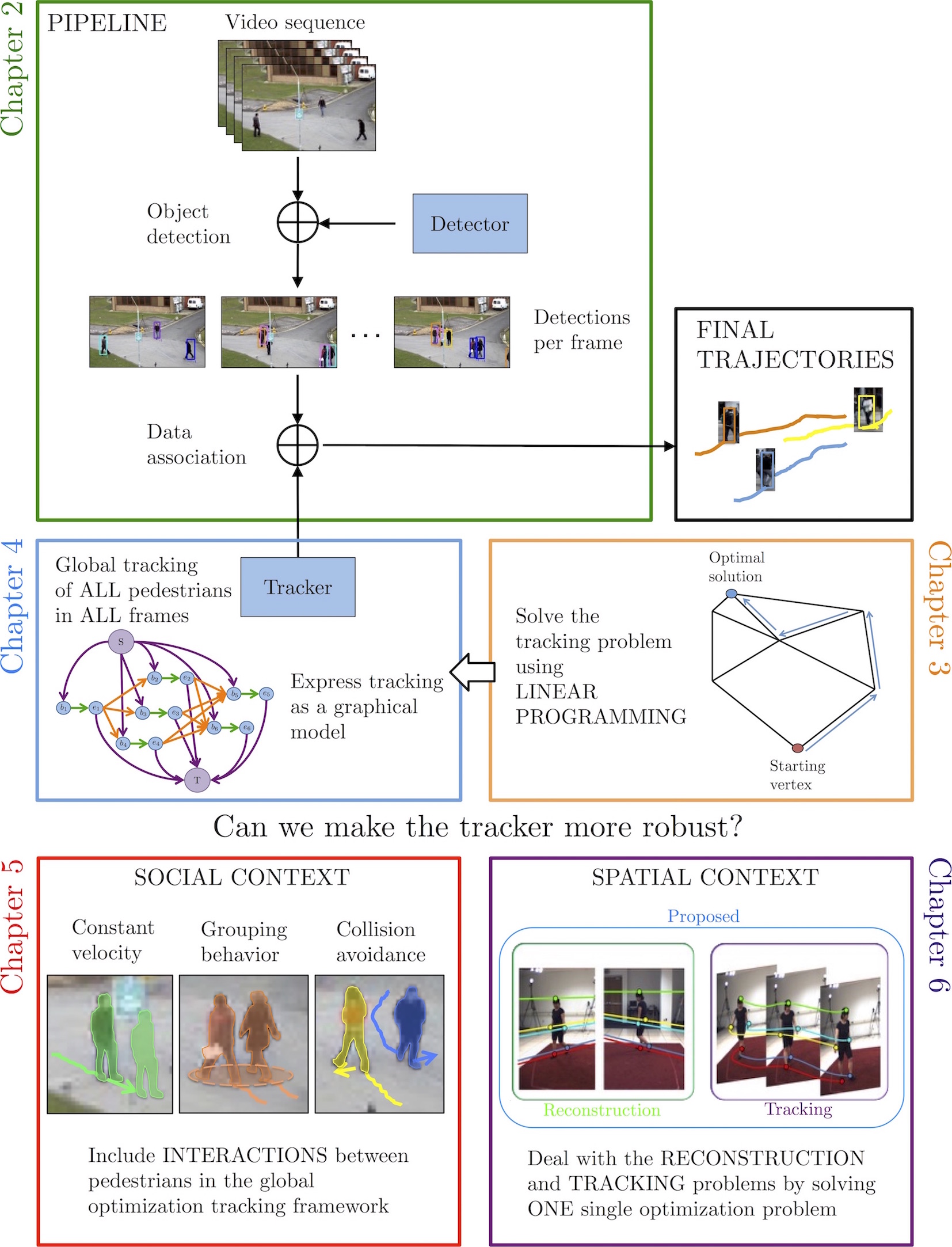} 
\caption{Organization of the thesis}
\label{fig:diagram}
\end{figure*}

As we motivated in the previous section, tracking methods still fail to capture and fully exploit much of the context of a pedestrian and his/her environment. In this thesis we mainly focus on two sources of context, {\it social} and {\it spatial}, and provide solutions which are globally optimal. 
We now detail the organization of the rest of the thesis, shown as a diagram in Figure \ref{fig:diagram}, as well as the main contributions. 

\noindent{\bf Chapter 2.} We start by introducing the problem at hand, namely pedestrian tracking or data association, and the basic paradigm we follow, \ie, tracking-by-detection. Since the rest of the thesis is focused on the tracking part, we briefly introduce here a few state-of-the-art detectors we use throughout the thesis. Finally, we discuss some of the major problems of those detectors and how they are handled during the tracking phase. 

\noindent{\bf Chapter 3.} In this chapter, we give an introduction to the basics of Linear Programming, which is the optimization technique we will use throughout the entire thesis. We describe the main solver, namely the Simplex algorithm, and its geometric intuition. We then move towards the graphical model representation of a Linear Program, so we can make use of the efficient solvers (\eg, $k$-shortest paths) present in the network flow community. 

\noindent{\bf Chapter 4.} Once we have the necessary notions of Linear Programming and graphical models, we formulate the multi-object tracking problem as a Linear Program. Unlike previous methods, this solves the tracking problem for all pedestrian trajectories and all frames at the same time, obtaining a unique and guaranteed global optimum. Furthermore, the problem can be solved in polynomial time. 
Our first contribution is to propose a small change in the graph structure which allows us to drop two parameters which have to be typically learned in an Expectation-Maximization fashion. Our graph performs tracking just depending on the actual scene information. Here we present an overview of the literature ranging from frame-by-frame matching to global methods similar to Linear Programming.

\noindent{\bf Chapter 5.} Given the Linear Programming framework presented in the previous chapter, we propose here to enhance tracking by including the {\it social context}. Interaction between pedestrians is modeled by using the well-known physical Social Force Model, used extensively in the crowd simulation community. 
The key insight is that people plan their trajectories in advance in order to avoid collisions, therefore, a graph model which takes into account future and past frames is the perfect framework to include social and grouping behavior. 
Instead of including social information by creating a complex graph structure, which then cannot be solved using classic LP solvers, we propose an iterative solution relying on Expectation-Maximization.
Results on several challenging public datasets are presented to show the improvement of the tracking results in crowded environments. An extensive parameter study as well as experiments with missing data, noise and outliers are also shown to test the robustness of the approach. In this chapter, we present an overview of state-of-the-art tracking methods that use social context.

\noindent{\bf Chapter 6.} In this chapter, we describe a method to include yet another source of context, in this case {\it spatial context}. As discussed in the previous section, spatial information between cameras and temporal information are still regarded in the literature as two separate problems, namely reconstruction and tracking. 
In this chapter, we argue that it is not necessary to separate the problem in two parts, and we present a novel formulation to perform 2D-3D assignments (reconstruction) and temporal assignments (tracking) in a single global optimization. When evidence is considered jointly, temporal correlation can potentially resolve reconstruction ambiguities and vice versa. 
The proposed graph structure contains a huge number of constraints, therefore, it can not be solved with typical Linear Programming solvers such as Simplex. In order to solve this problem, we rely on multi-commodity flow theory. We propose to exploit the specific structure of our problem and use Dantzig-Wolfe decomposition and branching to find the guaranteed global optimum. Here we present an overview of relevant literature of multiple view multiple target tracking.

\noindent{\bf Chapter 7.} Our final chapter includes the conclusions of the thesis, as well as a discussion for possible future research directions. 

\noindent{\bf Appendix A.} In the Appendix A of this thesis, we present a case study where Computer Vision is proven to be useful for the field of marine biology and chemical physics. An automatic method is presented for the tracking and motion analysis of swimming microorganisms. This includes early work done by the author at the beginning of the PhD. 
A new method for improved tracking, named multi-level Hungarian, is presented and compared with the Linear Programming formulation. Since microorganisms do not act according to the same social forces as humans, a method based on Hidden Markov Models is developed in order to analyze the motion of the microorganisms.
The final software for tracking and motion analysis has proven to be a helpful tool for biologists and physicists as it provides a vast amount on analyzed data in an easy fast way.

\section{Papers of the author}

In this section, the publications of the author are detailed by topic and chronological order. The core parts of the thesis are based on four main publications of the author:

\leftskip 0.3in
\parindent -0.3in

\citep{lealcvpr2014} {\bf L. Leal-Taix\'e}, M. Fenzi, A, Kuznetsova, Bodo Rosenhahn, Silvio Savarese. Learning an image-based motion context for multiple people tracking. {\it IEEE Conference on Computer Vision and Pattern Recognition (CVPR)}, June 2014.

\noindent{In this work, we present a method for multiple people tracking that leverages a generalized model for capturing interactions among individuals. 
At the core of our model lies a learned dictionary of interaction feature strings which capture relationships between the motions of targets.
These feature strings, created from low-level image features, lead to a much richer representation of the physical interactions between targets compared to hand-specified social force models that previous works have introduced for tracking. One disadvantage of using social forces is that all pedestrians must be detected in order for the forces to be applied, while our method is able to encode the effect of undetected targets, making the tracker more robust to partial occlusions.
The interaction feature strings are used in a Random Forest framework to track targets according to the features surrounding them.}

\citep{lealcvprw2014} {\bf L. Leal-Taix\'e}, M. Fenzi, A, Kuznetsova, Bodo Rosenhahn, Silvio Savarese. Multi-target tracking with context from interaction feature strings. {\it IEEE Conference on Computer Vision and Pattern Recognition Workshops (CVPR). SUNw: Scene Understanding Workshop}, June 2014.

\noindent{This is the 2 page abstract version of \citep{lealcvpr2014}, presented at the same conference as invited paper in the Scene Understanding workshop.}

\citep{lealbookchapter2012} {\bf L. Leal-Taix\'e}, Bodo Rosenhahn. Pedestrian interaction in tracking: the social force model and global optimization methods. {\it Modeling, Simulation and Visual Analysis of Crowds: A multidisciplinary perspective}. Springer, 2012.

\noindent{In this work, we present an approach for multiple people tracking in semi-crowded environments including interactions between pedestrians in two ways: first, considering social and grouping behavior, and second, using a global optimization scheme to solve the data association problem.
This is an extended text of the conference paper \citep{lealiccv2011} in book chapter format. It is intended to be an exhaustive introduction to Linear Programming for multiple people tracking, providing the necessary background on both graphical models and optimization to allow students to start programming such a tracking system.}

\citep{lealdagstuhl2012} {\bf L. Leal-Taix\'e}, G. Pons-Moll, B. Rosenhahn. Exploiting pedestrian interaction via global optimization and social behaviors. {\it Theoretic Foundations of Computer Vision: Outdoor and Large-Scale Real-World Scene Analysis}. Springer, 2012.

\noindent{In this work, we present an approach for multiple people tracking in semi-crowded environments including interactions between pedestrians in two ways: first, considering social and grouping behavior, and second, using a global optimization scheme to solve the data association problem.
This is an extended text of the conference paper \citep{lealiccv2011}, which includes more experiments, detailed evaluation of the effect of the method's parameters, detailed implementation details and extended theoretical background on graphical models.}

\citep{lealcvpr2012} {\bf L. Leal-Taix\'e}, G. Pons-Moll, B. Rosenhahn. Branch-and-price global optimization for multi-view multi-object tracking. {\it IEEE Conference on Computer Vision and Pattern Recognition (CVPR)}, June 2012.

\noindent{In this work, we present a new algorithm to jointly track multiple objects in multi-view images. While this has been typically addressed separately in the past, we tackle the problem as a single global optimization.
We formulate this assignment problem as a min-cost problem by defining a graph structure that captures both temporal correlations between objects as well as spatial correlations enforced by the configuration of the cameras. 
This leads to a complex combinatorial optimization problem that we solve using Dantzig-Wolfe decomposition and branching. Our formulation allows us to solve the problem of reconstruction and tracking in a single step by taking all available evidence into account. 
In several experiments on multiple people tracking and 3D human pose tracking, we show that our method outperforms state-of-the-art approaches.}

\citep{lealiccv2011} {\bf L. Leal-Taix\'e}, G. Pons-Moll, B. Rosenhahn. Everybody needs somebody: modeling social and grouping behavior on a linear programming multiple people tracker. {\it IEEE International Conference on Computer Vision Workshops (ICCV). 1st Workshop on Modeling, Simulation and Visual Analysis of Large Crowds}, November 2011.

\noindent{In this work, we present an approach for multiple people tracking in semi-crowded environments. Most tracking methods make the assumption that each pedestrian's motion is independent, thereby ignoring the complex and important interaction between subjects. On the contrary, our method includes the interactions between pedestrians in two ways: first, considering social and grouping behavior, and second, using a global optimization scheme to solve the data association problem.
Results are presented on three challenging, publicly available datasets to show that our method outperforms several state-of-the-art tracking systems.}

\par
\leftskip 0in
\noindent{The five publications related to the appendix section of the thesis are detailed below. This work was partially funded by the German Research Foundation, DFG projects RO 2497/7-1 and RO 2524/2-1 and the EU project AMBIO, and done in collaboration with the Institute of Functional Interfaces of the Karlsruhe Institute of Technology. Digital in-line holography is a microscopy technique which has gotten an increasing amount of attention over the last few years in the fields of microbiology, medicine and physics, as it provides an efficient way of measuring 3D microscopic data over time. In the following works, we explore detection, tracking and motion analysis on this challenging data, as well as ways for extending the method to a multiple camera system.}
\par
\leftskip 0.3in
\parindent -0.3in

\citep{maleschlijskibio2012} S. Maleschlijski, G. H. Sendra, A. Di Fino, {\bf L. Leal-Taix\'e}, I. Thome, A. Terfort, N. Aldred, M. Grunze, A.S. Clare, B. Rosenhahn, A. Rosenhahn. Three dimensional tracking of exploratory behavior of barnacle cyprids using stereoscopy. {\it Biointerphases. Journal for the Quantitative Biological Interface Data}. Springer, 2012.

\noindent{In this work, we present a low-cost transportable stereoscopic system consisting of two consumer camcorders. We apply this novel apparatus to behavioral analysis of barnacle larvae during surface exploration and extract and analyze the three-dimensional patterns of movement. The resolution of the system and the accuracy of position determination are characterized. In order to demonstrate the biological applicability of the system, three-dimensional swimming trajectories of the cypris larva of the barnacle Semibalanus balanoides are recorded in the vicinity of a glass surface. Parameters such as swimming direction, swimming velocity and swimming angle are analyzed.}

\citep{maleschlijskibio2011} S. Maleschlijski, {\bf L. Leal-Taix\'e}, S. Weisse, A. Di Fino, N. Aldred, A.S. Clare, G.H. Sendra, B. Rosenhahn, A. Rosenhahn. A stereoscopic approach for three dimensional tracking of marine biofouling microorganisms. {\it Microscopic Image Analysis with Applications in Biology (MIAAB)}, September 2011.

\noindent{In this work, we describe a stereoscopic system to track barnacle cyprids and an algorithm to extract 3D swimming patterns for a common marine biofouling organism - Semibalanus balanoides. The details of the hardware setup and the calibration object are presented and discussed. In addition we describe the algorithm for the camera calibration, object matching and stereo triangulation.
Several trajectories of living cyprids are presented and analyzed with respect to statistical swimming parameters.}

\citep{lealdagstuhl2011} {\bf L. Leal-Taix\'e}, M. Heydt, A. Rosenhahn, B. Rosenhahn. Understanding what we cannot see: automatic analysis of 4D digital in-line holography data. {\it Video Processing and Computational Video}. Springer, July 2011.

\noindent{In this work, we present a complete system for the automatic analysis of digital in-line holographic data; we detect the 3D positions of the microorganisms, compute their trajectories over time and finally classify these trajectories according to their motion patterns. This work includes the contributions presented in \citep{lealwmvc2009} and \citep{lealdagm2010}, extended experiments, theoretical background and implementation details.}

\citep{lealdagm2010} {\bf L. Leal-Taix\'e}, M. Heydt, S. Weisse, A. Rosenhahn, B. Rosenhahn. Classification of swimming microorganisms motion patterns in 4D digital in-line holography data. {\it 32nd Annual Symposium of the German Association for Pattern Recognition (DAGM)}, September 2010.

\noindent{In this work, we present an approach for automatically classifying complex microorganism motions observed with digital in-line holography. Our main contribution is the use of Hidden Markov Models (HMMs) to classify four different motion patterns of a microorganism and to separate multiple patterns occurring within a trajectory.
We perform leave-one-out experiments with the training data to prove the accuracy of our method and to analyze the importance of each trajectory feature for classification. We further present results obtained on four full sequences, a total of 2500 frames. The obtained classification rates range between 83.5\% and 100\%.}

\citep{lealwmvc2009} {\bf L. Leal-Taix\'e}, M. Heydt, A. Rosenhahn, B. Rosenhahn. Automatic tracking of swimming microorganisms in 4D digital in-line holography data. {\it IEEE Workshop on Motion and Video Computing (WMVC)}, December 2009.

\noindent{In this work, we approach the challenges of a high throughput analysis of holographic microscopy data and present a system for detecting particles in 3D reconstructed holograms and their 3D trajectory estimation over time. 
Our main contribution is a robust method, which evolves from the Hungarian bipartite weighted graph matching algorithm and allows us to deal with newly entering and leaving particles and compensate for missing data and outliers. In the experiments we compare our fully automatic system with manually labeled ground truth data and we can report an accuracy between 76\% and 91\%.}

\par
\leftskip 0in
\parindent 0in
\noindent{Aside from the previous publications related to multiple object tracking and motion analysis, the author was also involved in other projects, mainly related to pose estimation:}
\par
\leftskip 0.3in
\parindent -0.3in

\citep{kuznetsovaiccvw2013} A. Kuznetsova, {\bf L. Leal-Taix\'e}, B. Rosenhahn. Real-time sign language recognition using a consumer depth camera. {\it IEEE International Conference on Computer Vision Workshops (ICCV). 3rd IEEE Workshop on Consumer Depth Cameras for Computer Vision (CDC4CV)}, December 2013.

\noindent{In this work, we propose a precise method to recognize hand static gestures from a depth data provided from a depth sensor.
Hand sign recognition is performed using a multi-layered random forest (MLRF), which requires less the training time and memory when compared to a simple random forest with equivalent precision.
We evaluate our algorithm on synthetic data, on a publicly available Kinect dataset containing 24 signs from American Sign Language (ASL) and on a new dataset, collected using the Intel Creative Gesture Camera.}

\citep{fenzicvpr2013} M. Fenzi, {\bf L. Leal-Taix\'e}, B. Rosenhahn, J. Ostermann. Class generative models based on feature regression for pose estimation of object categories. {\it IEEE Conference on Computer Vision and Pattern Recognition (CVPR)}, June 2013.

\noindent{In this work, we propose a method for learning a class representation that can return a continuous value for the pose of an unknown class instance, using only 2D data and weak 3D labelling information. Our method is based on generative feature models, i.e., regression functions learnt from local descriptors of the same patch collected under different viewpoints. We evaluate our approach on two state-of-the-art datasets showing that our method outperforms other methods by 9-10\%.}

\citep{fenzidagm2012} M. Fenzi, R. Dragon, {\bf L. Leal-Taix\'e}, B. Rosenhahn, J. Ostermann. 3D Object Recognition and Pose Estimation for Multiple Objects using Multi-Prioritized RANSAC and Model Updating. {\it 34th Annual Symposium of the German Association for Pattern Recognition, DAGM}. August 2012.

\noindent{In this work, we present a feature-based framework that combines spatial feature clustering, guided sampling for pose generation and model updating for 3D object recognition and pose estimation. We propose to spatially separate the features before matching to create smaller clusters containing the object. Then, hypothesis generation is guided by exploiting cues collected off- and on-line, such as feature repeatability, 3D geometric constraints and feature occurrence frequency. The evaluation of our algorithm on challenging video sequences shows the improvement provided by our contribution.}

\citep{ponsdagstuhl2012} G. Pons-Moll, {\bf L. Leal-Taix\'e}, B. Rosenhahn. Data-driven manifold for outdoor motion capture. {\it Theoretic Foundations of Computer Vision: Outdoor and Large-Scale Real-World Scene Analysis}. Springer, 2012.

\noindent{In this work, we present a human motion capturing system that combines video input with sparse inertial sensor input under a particle filter optimization scheme. It is an extension of the work presented in \citep{ponsiccv2011} which includes a thorough theoretical introduction, extended experimental section and implementation details.}

\citep{ponsiccv2011} G. Pons-Moll, A. Baak, J. Gall, {\bf L. Leal-Taix\'e}, M. Mueller, H.-P. Seidel and B. Rosenhahn. Outdoor human motion capture using inverse kinematics and von Mises-Fisher sampling. {\it IEEE International Conference on Computer Vision (ICCV)}, November 2011.

\noindent{In this paper, we introduce a novel hybrid human motion capturing system that combines video input with sparse inertial sensor input. Employing an annealing particle-based optimization scheme, our idea is to use orientation cues derived from the inertial input to sample particles from the manifold of valid poses. Then, visual cues derived from the video input are used to weigh these particles and to iteratively derive the final pose. Our method can be used to sample poses that fulfill arbitrary orientation or positional kinematic constraints. In the experiments, we show that our system can track even highly dynamic motions in an outdoor environment with changing illumination, background clutter and shadows.}

\citep{ponsdagm2011} G. Pons-Moll, {\bf L. Leal-Taix\'e}, T. Truong, B. Rosenhahn. Efficient and robust shape matching for model based human motion capture. {\it 33rd Annual Symposium of the German Association for Pattern Recognition (DAGM)}, September 2011.

\noindent{In this work, we present an approach for Markerless Motion Capture using string matching. We find correspondences between the model predictions and image features using global bipartite graph matching on a pruned cost matrix.
Extracted features such as contour, gradient orientations and the turning function of the shape are embedded in a string comparison algorithm. The information is used to prune the association cost matrix discarding unlikely correspondences. This results in significant gains in robustness and stability and reduction of computational cost.
We show that our approach can stably track fast human motions where standard articulated Iterative Closest Point algorithms fail. This work was done by a Master's student whom the author co-supervised.}

\par
\leftskip 0in
\parindent 0in
\noindent{The following publication was presented out of the author's Master's Thesis completed at Northeastern University in Boston, USA:}
\par
\leftskip 0.3in
\parindent -0.3in

\citep{lealmsc2009} {\bf L. Leal-Taix\'e}, A.U. Coskun, B. Rosenhahn, D. Brooks. Automatic segmentation of arteries in multi-stain histology images. {\it World Congress on Medical Physics and Biomedical Engineering}, September 2009.

\noindent{Atherosclerosis is a very common disease that affects millions of people around the world. Currently most of the studies conducted on this disease use Ultrasound Imaging (IVUS) to observe plaque formation, but these images cannot provide any detailed information of the specific morphological features of the plaque. 
Microscopic imaging using a variety of stains can provide much more information although, in order to obtain proper results, millions of images must be analyzed. 
In this work, we present an automatic way to find the Region of Interest (ROI) of these images, where the atherosclerotic plaque is formed. 
Once the image is well-segmented, the amount of fat and other measurements of interest can also be determined automatically.}

\par
\leftskip 0in
\parindent 0in
\noindent{Aside from the aforementioned publications, the author edited a post-proceedings book of the Dagstuhl Seminar organized in 2012:}
\par
\leftskip 0.3in
\parindent -0.3in

\citep{lealbook2012} F. Dellaert, J.-M. Frahm, M. Pollefeys, B. Rosenhahn, {\bf L. Leal-Taix\'e}. {\it Theoretic Foundations of Computer Vision: Outdoor and Large-Scale Real-World Scene Analysis}. Springer, April 2012.

\noindent{The topic of the meeting was Large-Scale Outdoor Scene Analysis, which covers all aspects, applications and open problems regarding the performance or design of computer vision algorithms capable of working in outdoor setups and/or large-scale environments. Developing these methods is important for driver assistance, city modeling and reconstruction, virtual tourism, telepresence and outdoor motion capture. After the meeting, this post-proceedings book was edited with the collaboration of all participants who sent a paper that was peer-reviewed by three reviewers.}

\leftskip 0in
\parindent 0in


\chapter{Tracking-by-Detection} 

\label{trackingbydetection} 

\fancyhead[RE,LO]{Chapter 2. \emph{Tracking-by-Detection}} 

\graphicspath{{./Figures/Multiview/},{./Figures/trackingbydetection/},{./Figures/Holography/}} 


Tracking is commonly divided into two steps: object detection and data association. First, objects are detected in each frame of the sequence and second, the detections are matched to form complete trajectories. This is called the tracking-by-detection paradigm, and is the framework that will be used throughout the thesis. 
In this chapter we introduce the paradigm and give a brief overview of some of the most popular state-of-the-art detectors, while the main content of the thesis lies on the data association part. 
Here we also discuss the type of scenarios we are working with and give an overview of the literature that deals with high-density scenarios where people cannot be individually detected. 

\section{The scale of tracking}

Videos of walking pedestrians can vary in an infinite number of ways. Camera position, camera distance and type of environment are a few of the characteristics that define the type of video that will be created. 
Before introducing the problem that we are dealing with in this thesis, we first need to introduce the types of scenarios and the types of videos we will be working with.

\begin{figure*}[ht]
\centering
\subfigure[Sparse]{
\includegraphics[height=3.8cm]{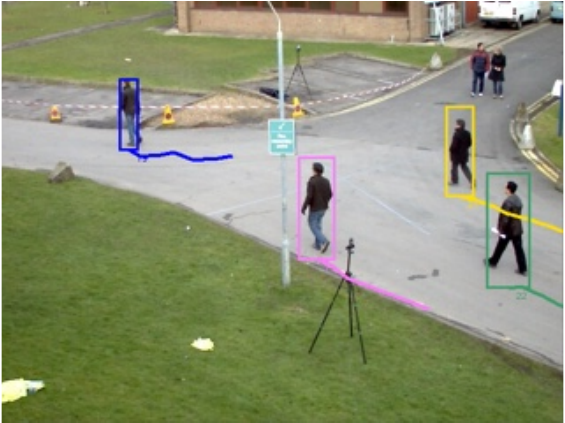} 
\label{fig:sparse}}
\subfigure[Semi-crowded]{
\includegraphics[height=3.8cm]{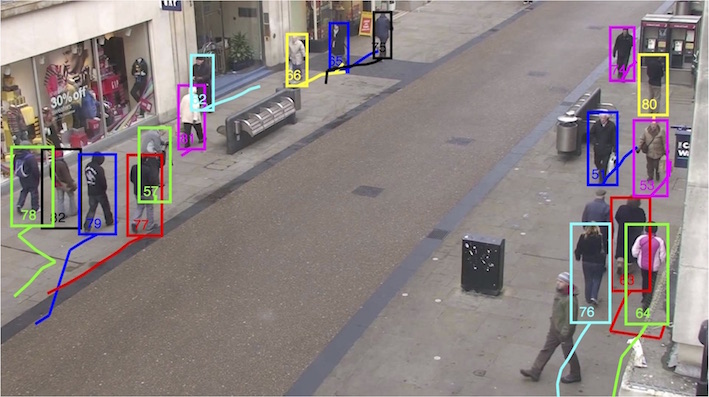} 
\label{fig:semicrowded}}
\subfigure[Crowded]{
\includegraphics[height=4.05cm]{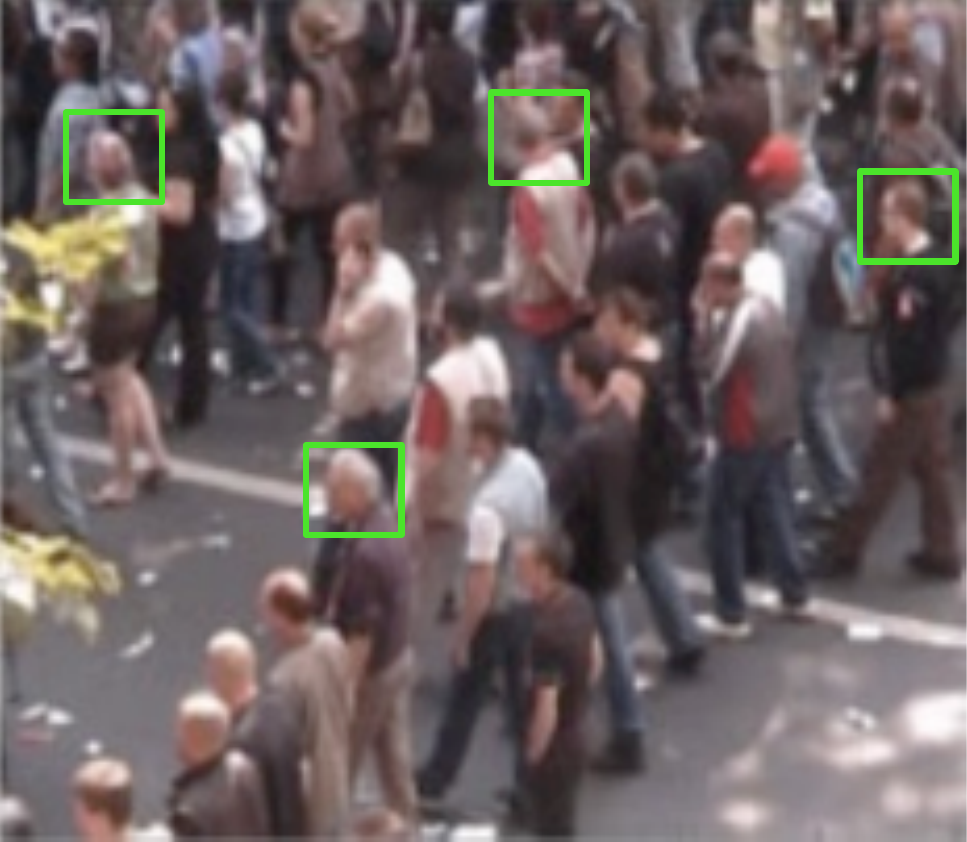} 
\label{fig:crowded}}
\subfigure[Macroscopic]{
\includegraphics[height=4.05cm]{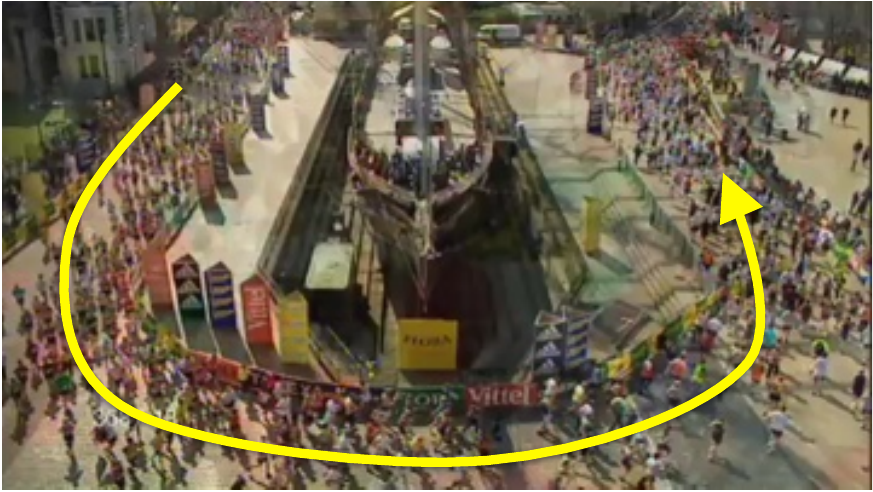} 
\label{fig:macro}}
\caption[Microscopic \vs macroscopic tracking]{Scenarios with different crowdness levels. (a) Sparse: individuals are detected and tracked throughout the scene. (b) Semi-crowded: it is still possible to detect individuals, but occlusions and missed detections are very common, making tracking challenging. (c) Crowded: tracking full-body pedestrians is no longer possible, but detection and tracking of heads is still performed. Person counting is a common task for videos of this crowdness level. Image from \citep{rodrigueziccv2011poster}. (d) Macroscopic scenario: individuals cannot be properly detected, therefore the goal in these scenarios is typically to find the overall flow of the crowd. Image from \citep{alieccv2008}.}
\label{fig:microvsmacro}
\end{figure*}

In Figure \ref{fig:microvsmacro}, we can see four examples of different scenarios with varying crowdness levels. The first example in Figure \ref{fig:sparse}, from the well-known PETS2009 dataset \citep{pets2009}, shows a scene with few pedestrians. The small size of the pedestrians, similar clothing and occlusions behind the pole or within themselves make this a challenging scenario. Nonetheless, recent methods have shown excellent results on this video, which is why more difficult datasets have been introduced. 

One example from the Town Center dataset \cite{benfoldcvpr2011} is shown in Figure \ref{fig:semicrowded}. This semi-crowded environment is challenging for the high amount of occlusions, but it is well-suited for the study of social behaviors as we will see in Chapter \ref{SFM}. Pedestrian detection is very challenging in these scenarios, since pedestrians are almost never fully visible and tracking is difficult due to the high amount of crossing trajectories. 

Even more crowded scenarios, like the one shown in Figure \ref{fig:crowded}, can still be analyzed with special methods which take into account the high density of the crowd \citep{rodrigueziccv2011poster}. For this category of videos, either the target to follow is manually initialized \citep{rodrigueziccv2009} or only head positions are tracked, since other parts of the body are rarely visible for a detector to work. Other approaches rely on feature tracking and motion factorization \citep{brostowcvpr2006}, conveying the idea that if two points move in a similar way they belong to the same object. In this last case, there is no need for a detection step. 

Finally, we have extremely crowded scenarios like marathons, demonstrations, etc. which are filmed from an elevated point of view, as in Figure \ref{fig:macro}. In these cases, individuals cannot be detected and identified, and therefore the task changes from individual person tracking towards analysis of the overall flow of the crowd \citep{alieccv2008,rodrigueziccv2011}.

Therefore, depending on the amount of people present in the scene, we can perform two types of tasks for video analysis: 

\begin{itemize}
\item {\it Microscopic} tracking focuses on the detection and tracking of individuals. Behavior analysis is centered around each individual and possibly their interactions. It uses individual motion and appearance features and is not too concerned with the overall motion in the scene. 
\item {\it Macroscopic} tracking, on the other hand, focuses on capturing the ``flow" of the crowd, the global behavior and motion tendencies. It is not focused on observing individual behavior but rather network behavior. Individual tracking can be performed if a target is manually initialized, since detection is not possible in this type of videos. 
\end{itemize}

Throughout this thesis, we work on sparse and semi-crowded scenarios as shown in Figures \ref{fig:sparse} and \ref{fig:semicrowded}. We use the tracking-by-detection paradigm, which we detail in the following section.

\section{Tracking-by-detection paradigm}

As mentioned in the previous section, there are several approaches for pedestrian video analysis. We focus on {\it microscopic} tracking, which means we are interested in detecting and tracking individuals. For this task, the {\it tracking-by-detection paradigm} has become increasingly popular in recent years, driven by the progress in object detection. 
Such methods involve two independent steps: (i) object detection on all individual frames and (ii) tracking or association of those detections across frames. 

\begin{figure*}[ht]
\centering
\includegraphics[width=1\linewidth]{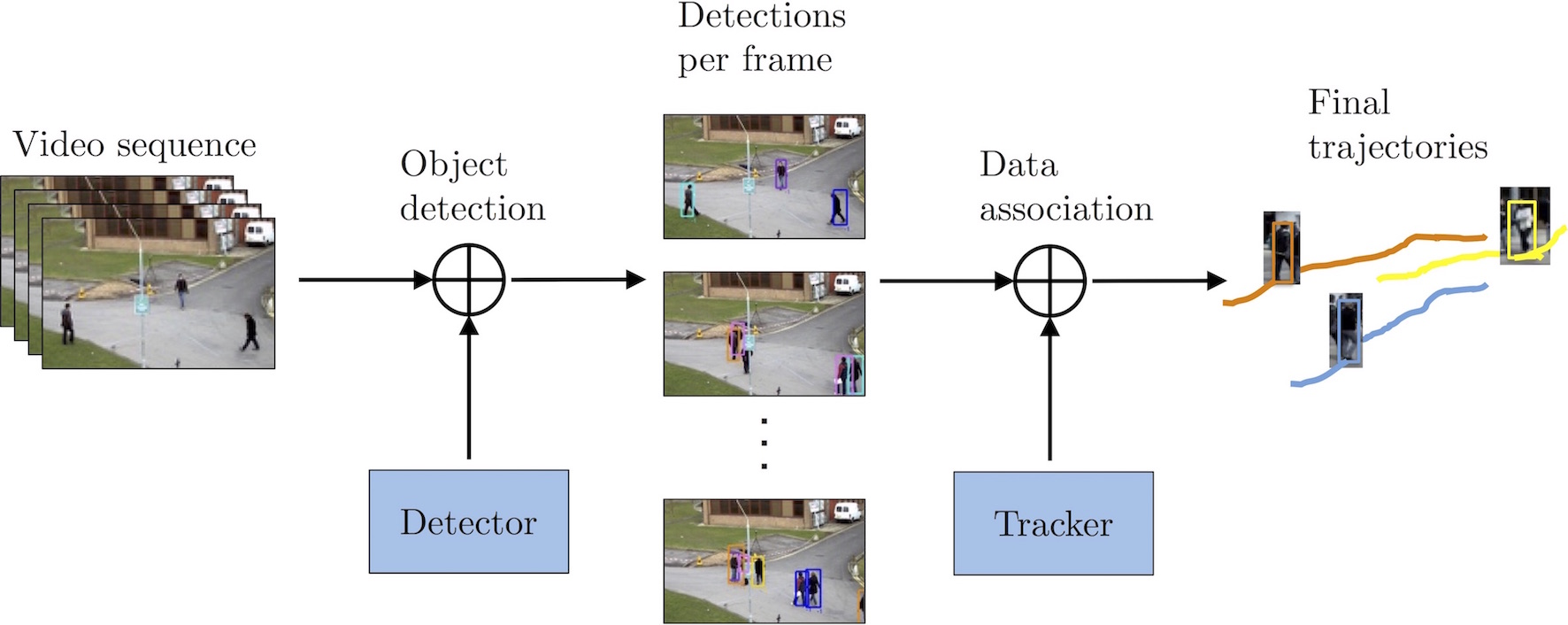} 
\caption[Tracking-by-detection paradigm.]{Tracking-by-detection paradigm. Firstly, an independent detector is applied to all image frames to obtain likely pedestrian detections. Secondly, a tracker is run on the set of detections to perform data association, \ie, link the detections to obtain full trajectories.}
\label{fig:trackingbydetection}
\end{figure*}

We can see a diagram of the tracking-by-detection paradigm in Figure \ref{fig:trackingbydetection}. There are two main components: the {\it detector} and the {\it tracker}. State-of-the-art detectors are discussed in Section \ref{detectors}, while the tracker is the main focus of the thesis. 
As we will see later on in this chapter, detectors are not perfect and often return false alarms or miss pedestrians. This makes tracking or data association a challenging task. Some of the most important challenges include:

\begin{itemize}
\item {\it Missed detections}: long-term occlusions are usually present in semi-crowded scenarios, where a detector might lose a pedestrian for 1-2 seconds. In this case, it is very hard for the tracker to re-identify the pedestrian without distinctive appearance information, and therefore, the track is usually lost. That is why in recent literature, researchers are opting for global optimization methods \citep{zhangcvpr2008,berclaztpami2011,lealiccv2011}, which are very good at dealing with long-term occlusions. 
\item {\it False alarms}: the detector can be triggered by regions in the image that actually do not contain any pedestrian, creating false positive. A tracker might follow the false alarms and create what is called a ghost trajectory. 
\item {\it Similar appearance}: one source of information commonly used for pedestrian identification is appearance. However, in some videos similar clothing can lead to virtually identical appearance models for two different pedestrians. Many methods in recent literature focus on the motion of the pedestrian rather than his/her appearance \citep{pellegriniiccv2009,lealiccv2011}.
\item {\it Groups and other special behaviors}: when dealing with semi-crowded scenarios, it is very common to observe social behaviors like grouping or waiting at a bus stop or stopping to talk to a person. All these behaviors do not fit classic tracking models like Kalman Filter \citep{kalman}, which consider pedestrians motion to be rather constant. 
\end{itemize}

These are a few of the challenges that tracking has to address. In this thesis, we make the observation that there is a lot of context that is not being used for tracking, especially social context or pedestrian interaction and spatial context coming from multiple views of the same scene. The proper use of these two sources of context in a global optimization scheme will be the center of the thesis.

\section{Detectors}
\label{detectors}

There are many pedestrian detectors in literature. Even though this thesis is focused on the data association part of multiple object tracking, we want to give a brief overview of three of the most used methods for pedestrian detection. 
Such methods can be classified in many ways; we detail one possible classification scheme: 

{\bf Model-based detectors.} A model of the background is created, and then a pixel-wise or block-wise comparison of a new image against the background model is performed to detect regions that do not fit the model \citep{detectionsurvey3}. This method is commonly used for video surveillance, since the camera is static and therefore the background model can be learned accurately. The drawback of these techniques is that they are very sensitive to changes in illumination and occlusions. 

{\bf Template-based detectors.} These detectors use a pre-learned set of templates, based for example on image features such as edges \citep{dalalcvpr2005}. 
The detector is triggered if the image features inside the local search window meet certain criteria. The drawback of this approach is that its performance can be significantly affected by background clutter and occlusions; if a person is partly occluded, the overall detection score will be very low because part of the image will be completely different from the learned examples. 

{\bf Part-based detectors.} One downside of holistic detectors like the one presented in \citep{dalalcvpr2005} is that they are easily affected by occlusions and local deformations. We would need a lot of training data to cover all the deformations that a body can undergo. In order to reduce the amount of data needed, recent works \citep{felzenszwalbcvpr2008,felzenszwalbcvpr2010} have proposed to use part-based methods, in which a template for each body part is learned separately. This way, deformations can be learned locally for each part and later combined. Another advantage of this method is that it is more robust to occlusions, since if one part is occluded, all the others can still be detected and combined for an overall high detection score. 
Other detectors based on parts have also been presented and created specifically to address occlusions \citep{shucvpr2012,wojekcvpr2011}.

Similar to these are block-based detectors, either based on HoG features \citep{galltpami2011} or SIFT features \citep{leibeijcv2005}. The objective is to learn the appearance of blocks inside the bounding box of a detection. At testing time, each block votes for the position of the center of the object to be detected. 

There is a different family of detectors, namely online detectors, that formulate the problem of tracking as that of re-detection. The combination of both types of detectors can be very beneficial as shown in \citep{shucvpr2013}, specially to account for appearance variations which might not be captured by the learned templates. 

We refer the reader to the following survey \citep{detectionsurvey1} regarding Adaboost and HOG-based pedestrian detectors for monocular videos; \citep{detectionsurvey3} for background subtraction techniques based on a mixture of gaussian background modeling and \citep{felzenszwalbtpami2010} for a detailed description of part-based models for object detection.

\subsection{Background modeling using Mixture-of-Gaussians}
\label{bkgsub}

While the most basic background subtraction methods are based on a frame-by-frame image difference, we detail here the model-based method presented in \citep{backgroundsubtraction} and used in the OpenCV implementation \citep{opencv}. The basic idea is to model each pixel's intensity by using a Gaussian Mixture Model (GMM). A simple heuristic determines which intensities most probably belong to the background, and pixels which do not match these are called foreground pixels. Foreground pixels are grouped using 2D connected component analysis.

\begin{figure*}[ht]
\centering
\subfigure[]{
\includegraphics[height=3.4cm]{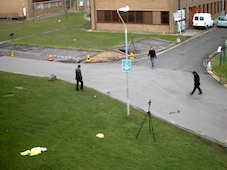} 
\label{fig:sub1}}
\subfigure[]{
\includegraphics[height=3.4cm]{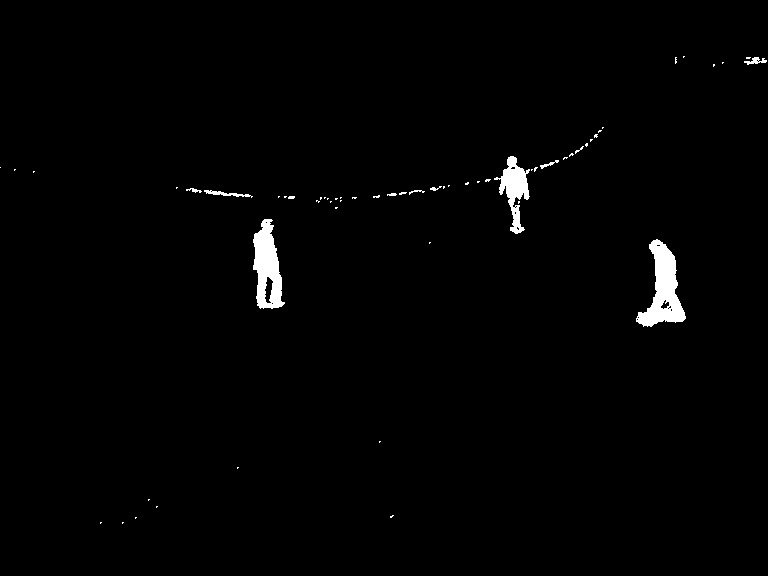} 
\label{fig:sub2}}
\subfigure[]{
\includegraphics[height=3.4cm]{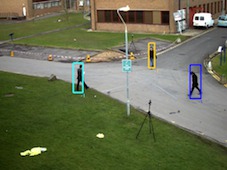} 
\label{fig:sub3}}
\caption[Background subtraction detector.]{(a) Original input image. (b) Background subtraction using the pre-learned model. White pixels are classified as foreground, black pixels as background. (c) Final detected bounding boxes.}
\label{fig:bkg_all}
\end{figure*}

An example of this process is shown in Figure \ref{fig:bkg_all}. As we can see, the background model is not perfect, which often leads to spurious foreground pixels around the scene as in Figure \ref{fig:sub2}. Of course this method detects all kinds of moving objects, and therefore it is a method prone to false detections.
In the experiments for this thesis, we use the homography provided by the camera calibration in order to determine the approximate size of a pedestrian on each pixel position. This allows us to determine a rough bounding box size and to discard groups of foreground pixels which are too small to be a pedestrian.

\subsection{Histogram of Oriented Gradients (HOG)}
\label{hog}

\begin{figure*}[ht]
\centering
\includegraphics[width=1\linewidth]{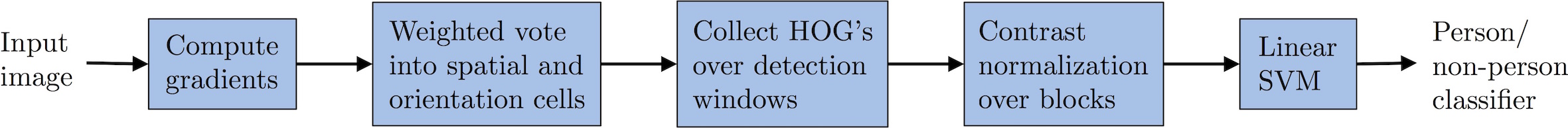} 
\caption{Overview of the feature extraction and object detection chain.}
\label{fig:HOG}
\end{figure*}

The essential thought behind the Histogram of Oriented Gradients (HOG) descriptor is that local object appearance and shape within an image can be described by the distribution of intensity gradients or edge directions. 
An overview of the pedestrian detection process as described in \citep{dalalcvpr2005} is shown in Figure \ref{fig:HOG}. 

The first step is to compute the gradients, then divide the image into small spatial windows or cells. For each cell we accumulate a local 1-D histogram of gradient directions or edge orientations over the pixels of the cell. 
The histograms can also be contrast-normalized for better invariance to changes in illumination or shadowing. Normalization is done over larger spatial regions called blocks. 
The detection window is covered with an overlapping grid of HOG descriptors, and the resulting feature vector is used in a conventional SVM classifier \citep{svm,PRML} that learns the appearance of a pedestrian vs. non-pedestrian.

\begin{figure*}[ht]
\centering
\subfigure[]{
\includegraphics[height=3.4cm]{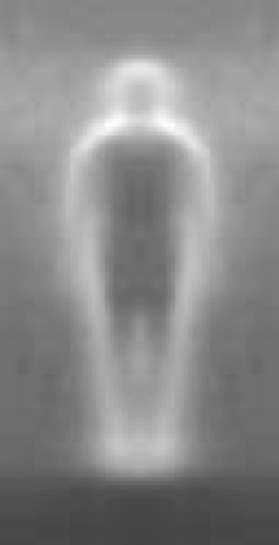} 
\label{fig:hog1}}
\subfigure[]{
\includegraphics[height=3.4cm]{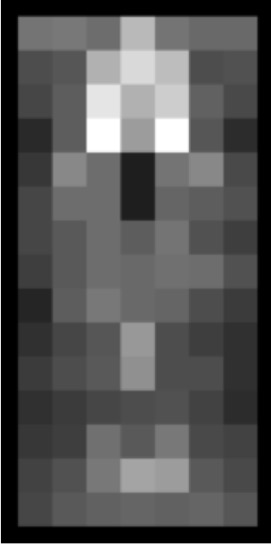} 
\label{fig:hog2}}
\subfigure[]{
\includegraphics[height=3.4cm]{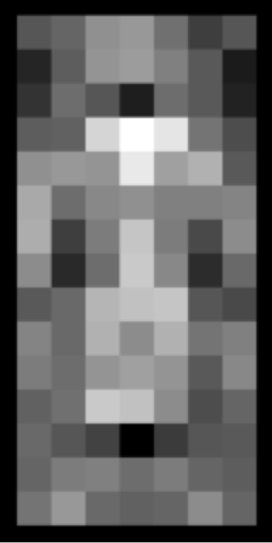} 
\label{fig:hog3}}
\subfigure[]{
\includegraphics[height=3.4cm]{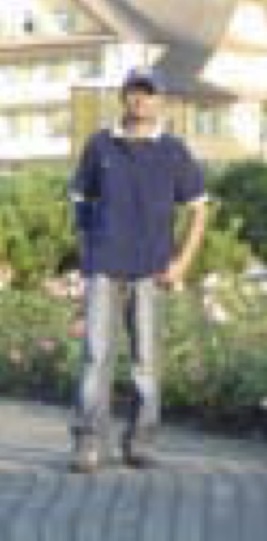} 
\label{fig:hog4}}
\subfigure[]{
\includegraphics[height=3.4cm]{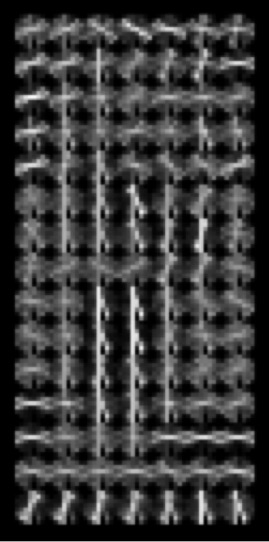} 
\label{fig:hog5}}
\subfigure[]{
\includegraphics[height=3.4cm]{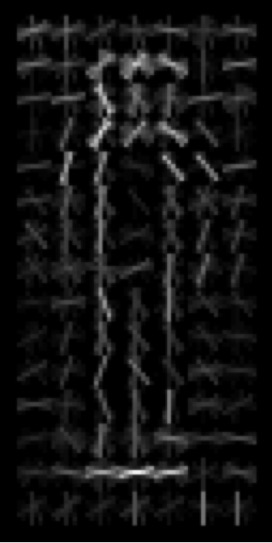} 
\label{fig:hog6}}
\subfigure[]{
\includegraphics[height=3.4cm]{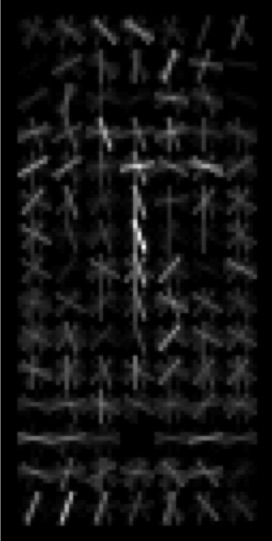} 
\label{fig:hog7}}
\caption[HOG pedestrian detector.]{The HOG detector is based mainly on silhouette contours. As we can see, the most active blocks are centered on the image background just outside the contour. (a) Average gradient image over the training samples. (b) Each ``pixel" shows the maximum positive SVM weight in the block centered in the pixel. (c) Likewise for the negative SVM weights. (d) Test image. (e) HOG descriptor. (f) HOG descriptor weighted  by positive SVM weights. (g) Likewise for negative weights. Images from \citep{dalalcvpr2005}.}
\label{fig:hog_all}
\end{figure*}

The HOG descriptor is particularly suited for human detection in images. This is because coarse spatial sampling, fine orientation sampling, and strong local photometric normalization allow the individual body movement of pedestrians to be ignored so long as they maintain a roughly upright position. 

\subsection{Part-based model}
\label{part}

Recent works have proved that modeling objects as a deformable configuration of parts \citep{felzenszwalbijcv2005,felzenszwalbcvpr2010} leads to increased detection performance compared to rigid templates \citep{dalalcvpr2005}. In the case of human detection, this is specially useful as the body can assume a large number of different poses.  
This model can also be used to estimate the 2D human pose of humans \citep{yangcvpr2011}.

\begin{figure*}[ht]
\centering
\subfigure[]{
\includegraphics[height=3.4cm]{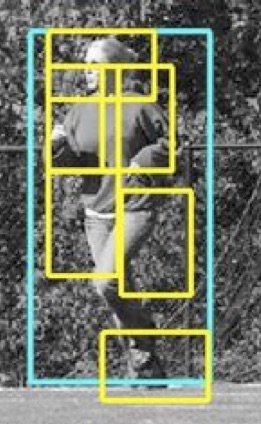} 
\label{fig:dpm1}}
\subfigure[]{
\includegraphics[height=3.4cm]{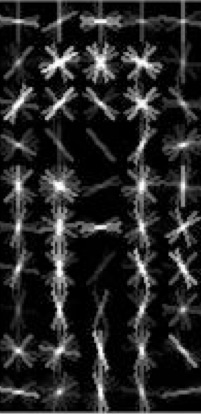} 
\label{fig:dpm2}}
\subfigure[]{
\includegraphics[height=3.4cm]{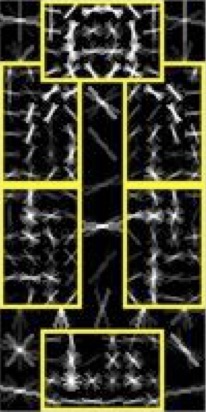} 
\label{fig:dpm3}}
\subfigure[]{
\includegraphics[height=3.4cm]{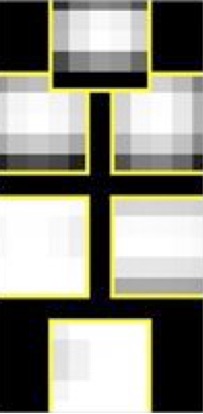} 
\label{fig:dpm4}}
\caption[Deformable part-based model detector.]{Deformable part-based model detector. (a) Example of a detection of a person. Green box represents the root filter detection while the yellow boxes represent the part detections. (b) Coarse template or root filter. (c) Templates of the parts. (d) Spatial model for the location of each part. Images from \citep{felzenszwalbcvpr2008}.}
\label{fig:dpm}
\end{figure*}

The basic idea is to have a model based on several HOG feature filters. The model for each object consists of one global root filter (see Figure \ref{fig:dpm2}),  which is equivalent to the rigid template as presented before, and several part models. The features of the part filters are computed at twice the spatial resolution of the root filter in order to capture smaller details. 
Each part model specifies a spatial model (see Figure \ref{fig:dpm4}) and a part filter (see Figure \ref{fig:dpm3}). The spatial model defines a set of allowed placements for a part relative to the detection window and a deformation cost for each placement. 

Detection is done using a sliding window approach. The score is computed by adding the score of the root filter and the sum over all parts, taking into account the placement of each part, the filter score and the deformation cost. 
Usually both part-based and rigid template-based approaches are prone to double detections, therefore a non-maxima suppression step is necessary to avoid too many false detections around one pedestrian. We will show examples of this phenomenon in Section \ref{detectionresults}.

Training is done by using a set of images with an annotated bounding box around each instance of an object (a pedestrian in our case). Learning is done in a similar way as in \citep{dalalcvpr2005}, only now, apart from learning the model parameters, the part placements also need to be learned. These are considered as latent values and therefore Latent SVM is used to learn the model.

\section{Detection results}
\label{detectionresults}

\begin{figure*}[ht]
\centering
\subfigure[Background subtraction]{
\includegraphics[height=3.4cm]{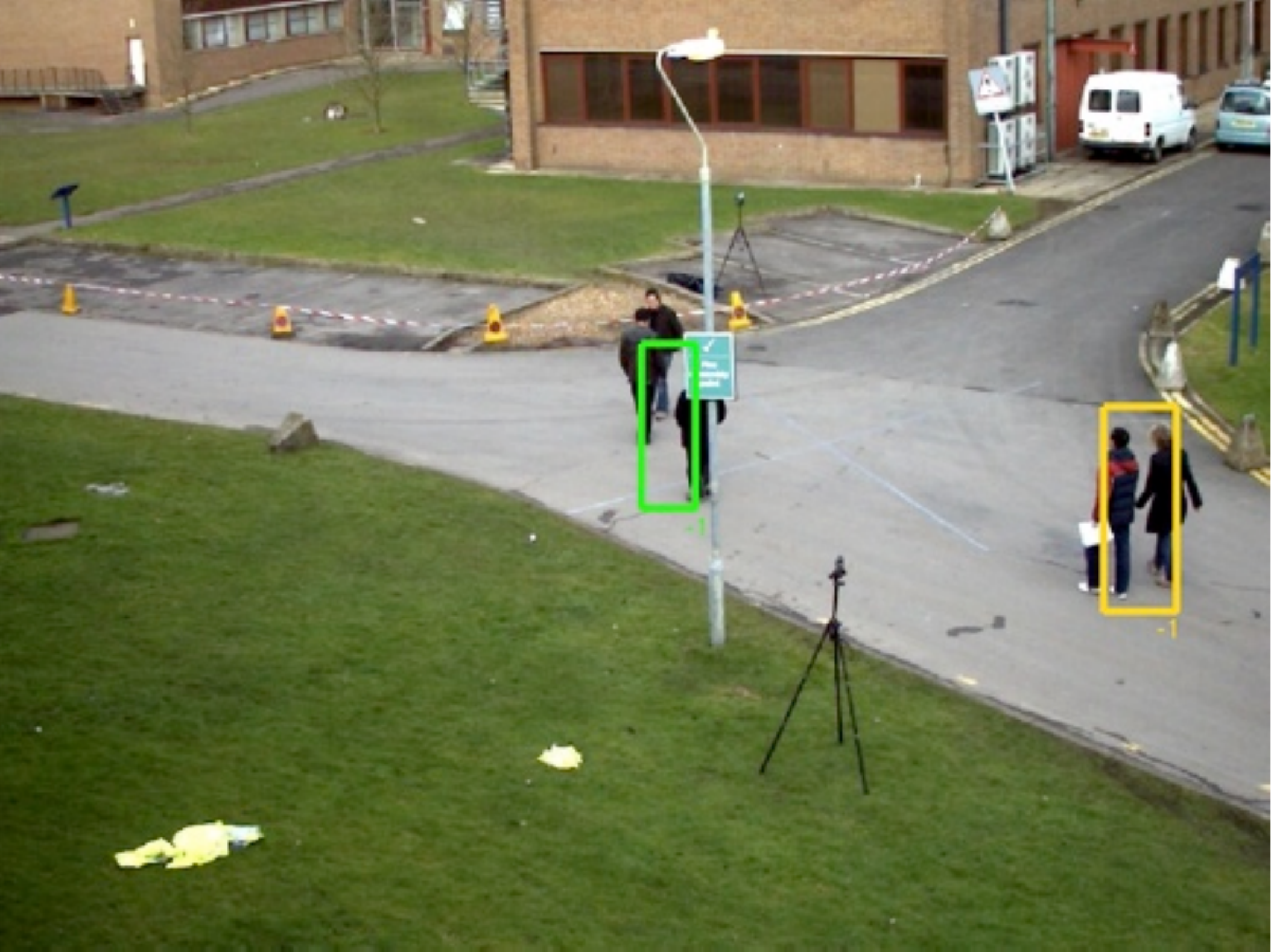} 
\label{fig:26bkgsub}}
\subfigure[HOG detector]{
\includegraphics[height=3.4cm]{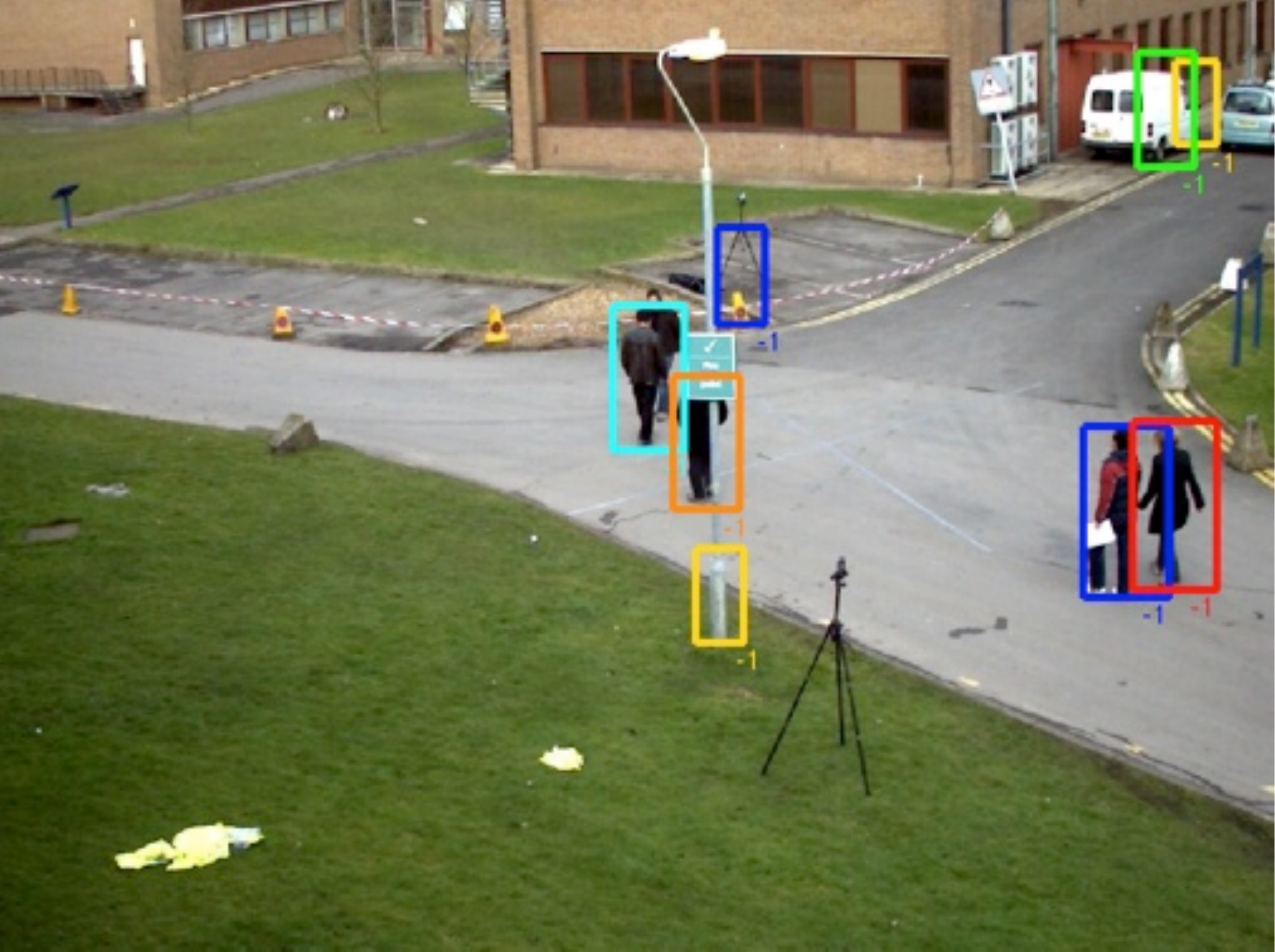} 
\label{fig:26hog}}
\subfigure[Part-based detector]{
\includegraphics[height=3.4cm]{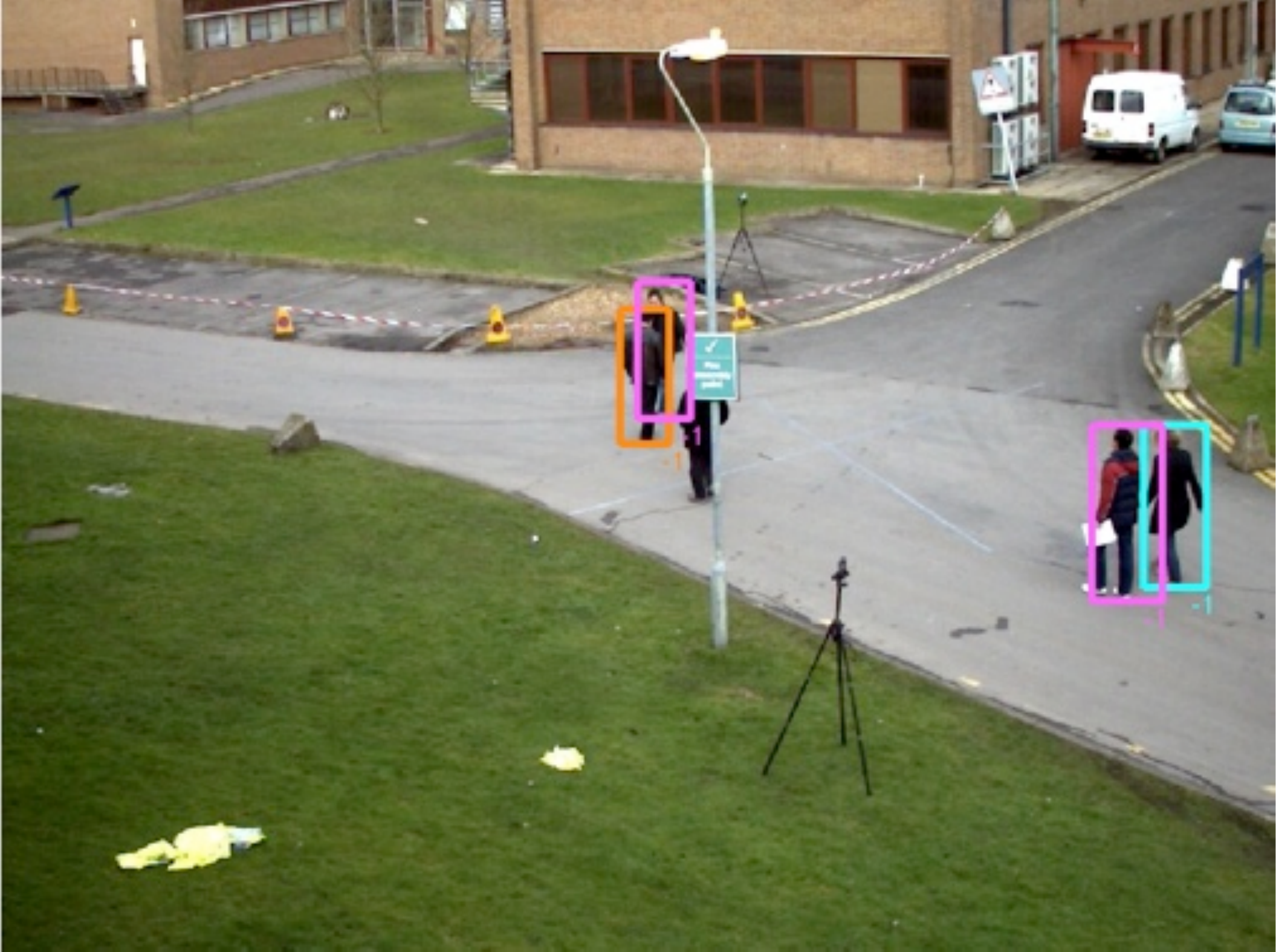} 
\label{fig:26part}}
\caption[Example of detection results on one frame of the PETS2009 sequence.]{Example of detection results on one frame of the PETS2009 sequence. (a) Using background subtraction (Section \ref{bkgsub}). (b) Using HOG features and SVM learning (Section \ref{hog}). (c) Using part-based model, HOG features and Latent SVM learning (Section \ref{part}).}
\label{fig:det_PETS}
\end{figure*}

In this section, we discuss some detection results, show common failure cases and present some further methods proposed in recent literature. In Figure \ref{fig:det_PETS}, we plot some results of the three methods referenced in the previous section on the publicly available dataset PETS2009 \citep{pets2009}. 

In Figure \ref{fig:26bkgsub}, we show a common failure case of background subtraction methods. As we can see, the three pedestrians in the center of the image are very close to each other, which means the background subtraction method obtains a single blob in that region. In some cases it is possible to determine the presence of more than one pedestrian based on blob size and a knowledge of the approximate size of the pedestrian in pixels. In this case, though, the partial occlusion of one of the pedestrians makes it hard to determine exactly how many pedestrians belong to the foreground blob. The resulting detection is therefore placed in the middle of the group of pedestrians, which means we do not only have missed detections but also an incorrect position of the detection, which in fact will be considered as a false alarm. 
A similar situation is shown for the two pedestrians on the right side of the image. The final detection is positioned in the middle between them. 

In Figure \ref{fig:26hog}, we show results of a HOG detector with SVM learning. Here, the major problems are double detections and the threshold of the score that determines what is a detection and what is not. As we can see in Figure \ref{fig:26hog}, if the threshold is too low we can get a lot of false detections. The advantage is that we can detect half-occluded pedestrians like the orange pedestrian behind the pole. 

A part-based detector returns the result shown in Figure \ref{fig:26part}. As we can see, it is successful in finding partially occluded people or people who are close together. It only fails to detect the pedestrian occluded by the pole; this is mainly because one of the most distinctive parts for detections is the one of the head and shoulders, forming an omega shape.

\begin{figure*}[ht]
\centering
\subfigure[HOG detector]{
\includegraphics[height=3.8cm]{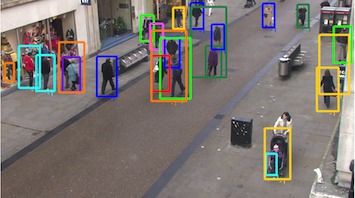} 
\label{fig:townhog1}}
\subfigure[Part-based detector]{
\includegraphics[height=3.8cm]{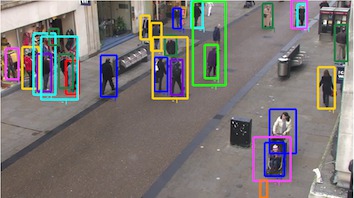} 
\label{fig:townpart1}}
\subfigure[HOG detector]{
\includegraphics[height=3.8cm]{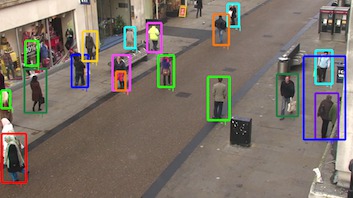} 
\label{fig:townhog2}}
\subfigure[Part-based detector]{
\includegraphics[height=3.8cm]{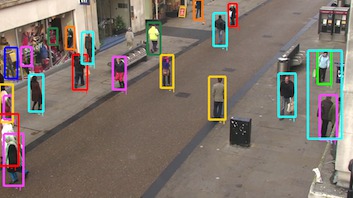} 
\label{fig:townpart2}}
\caption[Example of detection results on the Town Center sequence.]{Example of detection results on the Town Center sequence. (a,c) Using HOG features and SVM learning (Section \ref{hog}). (b,d) Using part-based model, HOG features and Latent SVM learning (Section \ref{part}).}
\label{fig:det_TOWN}
\end{figure*}

We show results on the Town Center dataset \citep{benfoldcvpr2011} in Figure \ref{fig:det_TOWN}. This is a high resolution dataset of a busy town center, where partial occlusions and false alarms are very common. As we can see, double detections are specially problematic, for both the simple HOG detector and the part-based detector. 
It is common that two pedestrians trigger a single detection with a bigger bounding box, which means the non-maxima suppression is a key step in this case. Nonetheless, these methods still present two key advantages for this dataset: (i) most false detections can be easily removed using camera calibration and an approximate size of a pedestrian; (ii) there are few missed pedestrians. As we will see in Chapter \ref{SFM}, the Linear Programming algorithms for tracking are capable of handling false alarms better than missing data.

\begin{figure*}[ht]
\centering
\subfigure[]{
\includegraphics[height=2.5cm]{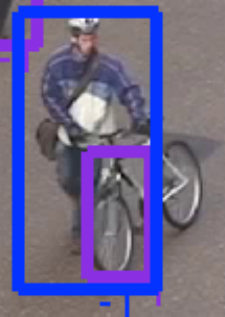} 
\label{fig:obj1}}
\subfigure[]{
\includegraphics[height=2.5cm]{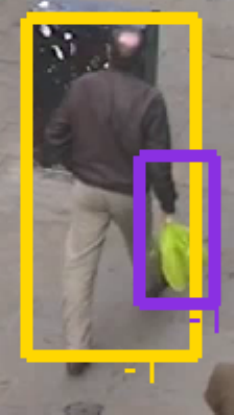} 
\label{fig:obj2}}
\subfigure[]{
\includegraphics[height=2.5cm]{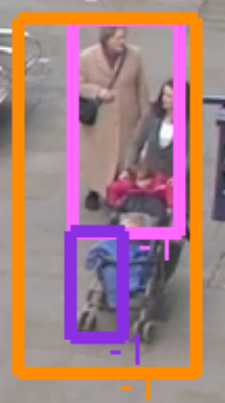} 
\label{fig:obj3}}
\subfigure[]{
\includegraphics[height=2.5cm]{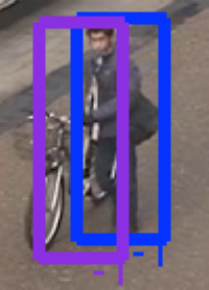} 
\label{fig:obj4}}
\caption[Example of pedestrians walking with objects.]{Example of pedestrians walking with objects. This often leads to double detections or missed detections, but pedestrian-object interactions can be a useful source of information to improve tracking.}
\label{fig:ped_objs}
\end{figure*}

It is common to see pedestrians walking with objects, either pushing a bicycle, carrying a bag or pushing a trolley or a stroller, as we can see in Figure \ref{fig:ped_objs}. The close proximity to those objects often leads to double detections or can even lead to the complete misdetection of the pedestrian. In recent works researchers proposed to include those objects in the tracking system. In \citep{mitzeleccv2012} a tracker for unknown shapes was proposed in order to deal not only with pedestrians but also with carried objects. 3D information was used to create a model of unknown shapes which was then tracked through time. 
Furthermore, in \citep{baumgartnercvpr2013} pedestrians interaction with objects was included to support tracking hypotheses. This confirms the argument presented in this thesis, that context from a pedestrian's environment (in this case pedestrian-object interaction) can be extremely useful to improve tracking. 
Finally, tracking systems for complete scene understanding are becoming more and more important in the literature \citep{wojektpami2013}.


\chapter{Introduction to Linear Programming} 

\label{linearprogramming} 

\fancyhead[RE,LO]{Chapter 3. \emph{Introduction to Linear Programming}} 

\graphicspath{{./Figures/LP/}} 

In this chapter, we give an introduction to the theory of Linear Programming (LP), defining all the basic concepts used in further chapters. We start by formally defining a Linear Program and its geometry. We then put special focus on the Simplex method, the most common LP solver and finally we introduce the concept of duality and the relation between LP and graphical models. 
We refer the interested reader to two books on Linear Programming \citep{LP1,LP2} and one on Network Flows \citep{networkflows} to delve deeper into the subject.

\section{What is Linear Programming?}

\begin{colbox}
A linear program consists of a linear objective function 
\begin{align}
\label{eq:introlp1}
\textstyle
c_1 x_1 + c_2 x_2 + \ldots + c_n x_n 
\end{align}
subject to linear constraints
\begin{align}
\label{eq:introlp2}
&a_{11} x_1 + a_{12} x_2 + \ldots + a_{1n} x_n \leq b_1 \\
&a_{21} x_1 + a_{22} x_2 + \ldots + a_{2n} x_n \leq b_2 \nonumber\\
&\quad \vdots  \qquad\qquad \vdots \qquad\qquad\quad \vdots  \nonumber\\
 &a_{m1} x_1 + a_{m2} x_2 + \ldots + a_{mn} x_n \leq b_m. \nonumber 
\end{align}
\end{colbox}

Solving the program means finding the $x_1, \ldots, x_n \in \mathbb{R}$ that maximize (or minimize) the objective function while satisfying the linear constraints. The linear program can be expressed as
\begin{align}
\label{eq:canonicallp}
\max \ \{  \mathbf{c}^\intercal \mathbf{x} : \mathbf{x} \in \mathbb{R}^n, \mathbf{A} \mathbf{x} \leq \mathbf{b} \}
\end{align}

where $\mathbf{A} \in \mathbb{R}^{m\times n}$ is the matrix of coefficients and $\mathbf{b} \in \mathbb{R}^{m}$ the vector that defines the constraints of the LP. The problem constraints can be written as equalities or inequalities ($\leq,\geq$), as these can always be converted to a standard form without changing the semantics of the problem.

\begin{figure*}[htbp]
\centering
\subfigure[]{
\includegraphics[width=0.31\linewidth]{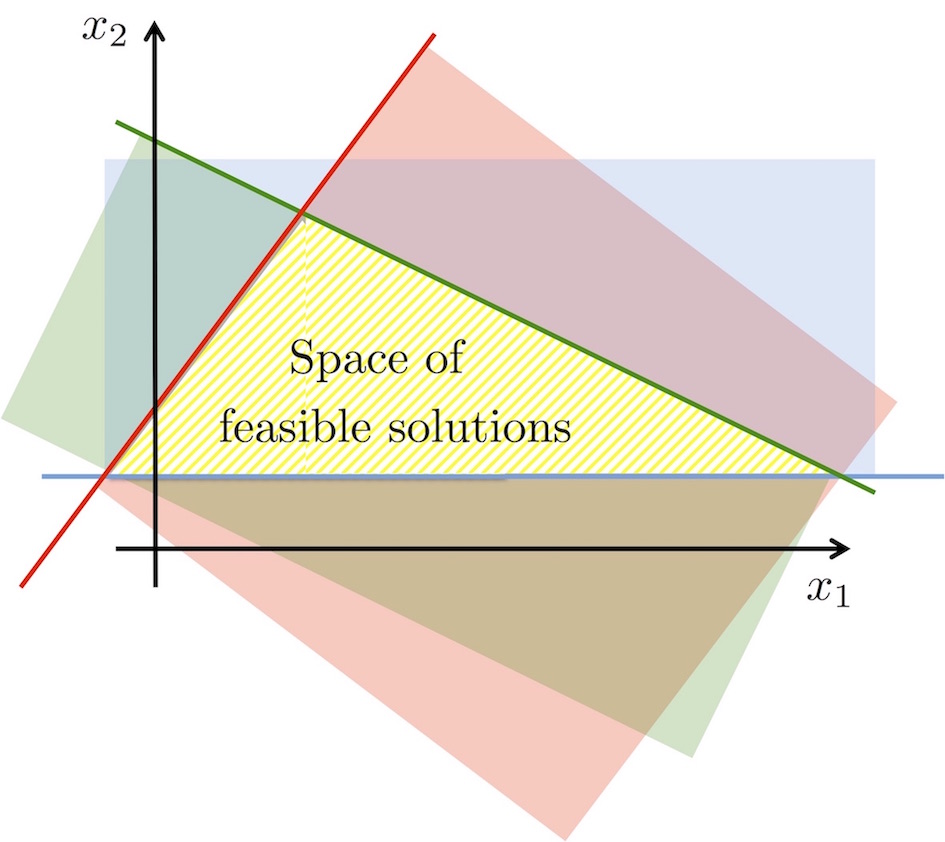} 
\label{fig:LP}
}
\subfigure[]{
\includegraphics[width= 0.31\linewidth]{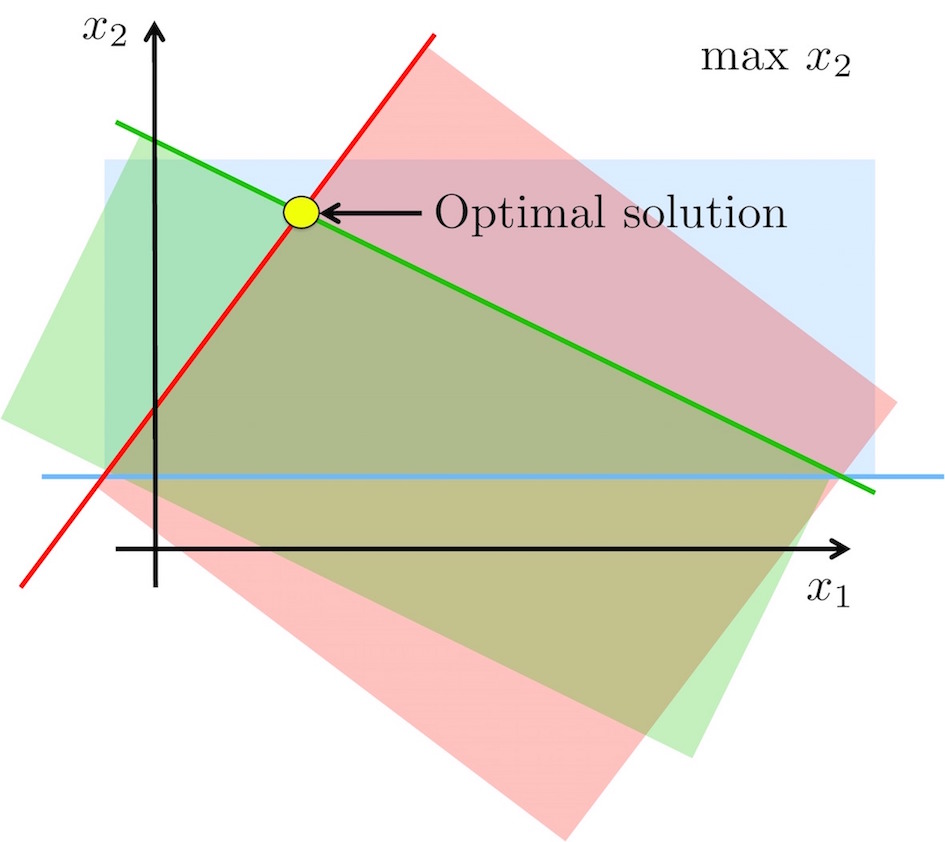} 
\label{fig:optimal}
}
\subfigure[]{
\includegraphics[width= 0.31\linewidth]{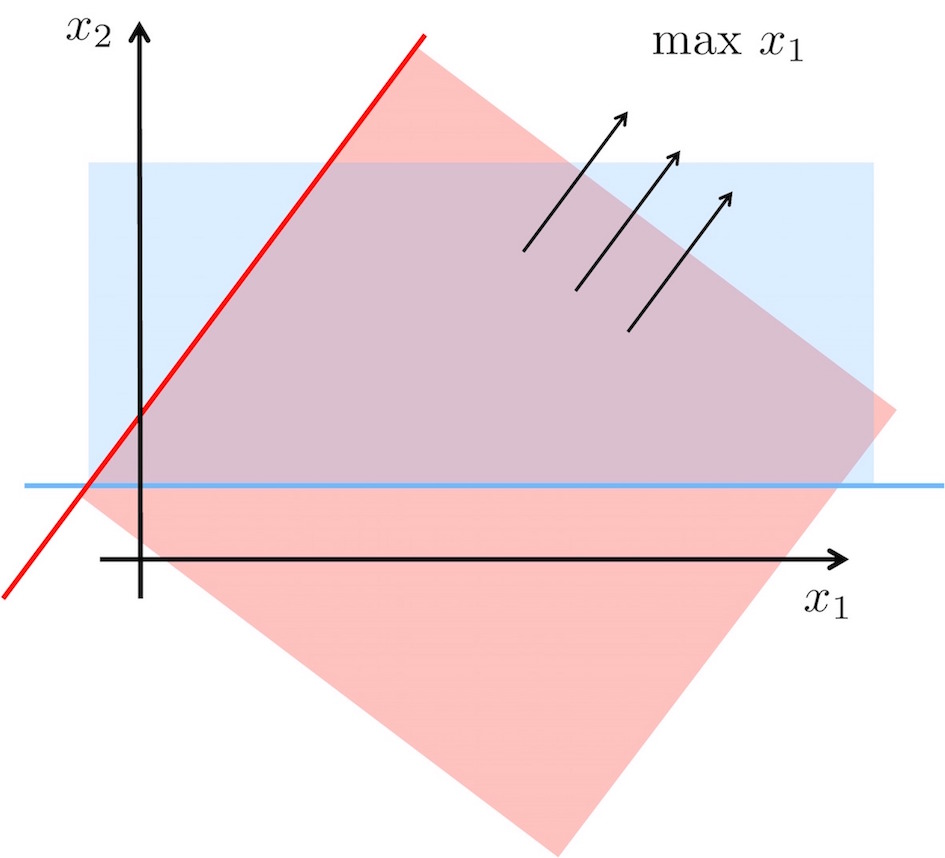} 
\label{fig:unbounded}
}
\caption[Linear Programming representation]{Representations of Linear Programs with constraints represented by colored lines and half-spaces represented by colored regions. (a) Representation of a Linear Program with three constraints represented in blue, green and red, and its space of possible solutions (in yellow). (b) Optimal solution for the Linear Program of maximizing $x_2$ (yellow dot). (c) An unbounded Linear Program, when trying to maximize $x_1$ the solution space is infinite, as indicated by the arrows pointing towards the unbounded direction.}
\label{fig:LPdef}
\end{figure*}

A point $\mathbf{x} \in \mathbb{R}^n$ is called {\it feasible} if it satisfies all linear constraints, see Figure~\ref{fig:LP}. If there are feasible solutions to a linear program, then it is called a {\it feasible} program. A problem can be infeasible if its constraints are contradictory, \eg, $x_1>1$ and $x_1<-1$.

A feasible $\mathbf{x} \in \mathbb{R}^n$ is an {\it optimal solution} to a linear program if $\mathbf{c}^\intercal \mathbf{x} \geq \mathbf{c}^\intercal \mathbf{y}$ for all feasible $\mathbf{y} \in \mathbb{R}^n$, see Figure~\ref{fig:optimal}.

A linear program is {\it bounded} if there exists a constant $M \in \mathbb{R}$ such that $\mathbf{c}^\intercal \mathbf{x} \leq M$ for all feasible $\mathbf{x} \in \mathbb{R}^n$. An example of an unbounded problem can be seen in Figure~\ref{fig:unbounded}.

\subsection{Linear Programming forms}
\label{sec:LPforms}

A Linear Program can be expressed in different forms, namely, Standard Form 1, Inequality Form, Standard Form 2 and General Form. In order to solve a problem with the Simplex method, for example, we need to have the problem in Standard Form 1, therefore, it is useful to know how to easily go from one form to another. 
All forms share the same objective function, which is a minimization, but change the way in which the constraints are expressed. Remember that we have $n$ variables, $\mathbf{x} \in \mathbb{R}^n$, and $m$ constraints, $\mathbf{A} \in \mathbb{R}^{m\times n}$, $\mathbf{b} \in \mathbb{R}^m$.

\noindent{\bf Standard Form 1}. The constraints are expressed as equalities and it is implied that the variables are nonnegative. 
\begin{align}
\label{eq:standardform1}
\min  \quad &\mathbf{c}^\intercal \mathbf{x} \\
\textrm{s.t.} \quad &\mathbf{A} \mathbf{x} = \mathbf{b}\nonumber\\
&\mathbf{x}\geq \mathbf{0}. \nonumber
\end{align}
\noindent{\bf Inequality Form}. The constraints are expressed as inequalities and we need to explicitly define the non-negativity constraints (if any). 
\begin{align}
\label{eq:inequalityform}
\min  \quad &\mathbf{c}^\intercal \mathbf{x} \\
\textrm{s.t.} \quad &\mathbf{A} \mathbf{x} \leq \mathbf{b}\nonumber
\end{align}
\noindent{\bf Standard Form 2}. The constraints are expressed as inequalities and it is implied that the variables are nonnegative. 
\begin{align}
\label{eq:standardform2}
\min  \quad &\mathbf{c}^\intercal \mathbf{x} \\
\textrm{s.t.} \quad &\mathbf{A} \mathbf{x} \leq \mathbf{b}\nonumber\\
&\mathbf{x}\geq \mathbf{0}. \nonumber
\end{align}
\noindent{\bf General Form}. The constraints are expressed both as equalities and inequalities. We need to explicitly define the non-negativity constraints (if any). 
\begin{align}
\label{eq:generalform}
\min  \quad &\mathbf{c}^\intercal \mathbf{x} \\
\textrm{s.t.} \quad &\mathbf{A} \mathbf{x} \leq \mathbf{b}\nonumber\\
&\mathbf{G} \mathbf{x} = \mathbf{f}\nonumber
\end{align}

Once we have all the forms defined, we are interested in knowing how to go from one form to the other. We are specially interested in converting a problem to the Standard Form 1, which is the one we need to use the Simplex algorithm.
Any LP can be converted into the Standard Form 1 by performing a series of operations. Let us consider the following example of an LP problem:
\begin{align}
\label{eq:exampleLP}
\max \quad & x_1 + x_2 + 2x_3 \\ \nonumber
\textrm{s.t.} \quad &2x_1 + 3x_2 \leq 12 \\ \nonumber
&x_2+x_3 \geq 5 \\ \nonumber
& x_1 \geq 4 \\ \nonumber
& x_2 \geq 0 \nonumber
\end{align}
In order to express this problem in Standard Form 1, we can follow a set of simple transformations:
\begin{itemize}
\item To convert a maximization problem into a minimization one, we simply negate the objective function:
\begin{align}
\max \qquad & x_1 + x_2 + 2x_3  \quad \rightarrow \quad \min \qquad  -x_1 - x_2 - 2x_3
\end{align}
\item To convert inequalities into equalities, we introduce a set of {\it slack variables} which represent the difference between the two sides of the inequality and are assumed to be nonnegative. The cost on the objective function for these variables is zero:
\begin{align}
&2x_1 + 3x_2 \leq 12 \quad \rightarrow \quad 2x_1 + 3x_2 + s_1 = 12  \hspace{0.1cm} , \hspace{0.1cm} s_1 \geq 0 \\ \nonumber
&x_2+x_3 \geq 5  \quad \rightarrow \quad x_2+x_3 - s_2 = 5  \hspace{0.1cm} , \hspace{0.1cm} s_2 \geq 0 \nonumber
\end{align}
\item If the lower bound of a variable is not zero, we introduce another variable and perform substitution:
\begin{align}
x_1 \geq 4 \quad \rightarrow \quad  y_1 = x_1 - 4  \hspace{0.1cm} , \hspace{0.1cm} y_1 \geq 0
\end{align}
\item We can replace unrestricted variables by the difference of two restricted variables:
\begin{align}
x_3 \quad \rightarrow \quad  x_3 = x_4 - x_5  \hspace{0.1cm} , \hspace{0.1cm} x_4 \geq 0  \hspace{0.1cm} , \hspace{0.1cm} x_5 \geq 0
\end{align}
\end{itemize}
After all the transformations, we obtain the following LP in standard form:
\begin{align}
\min \qquad & - y_1 - 4 -  x_2 - 2x_4 + 2x_5 \\ \nonumber
\st \qquad &2y_1 + 8 + 3x_2 + s_1 = 12 \\ \nonumber
&x_2+x_4 - x_5 - s _2 = 5 \\ \nonumber
& y_1,x_2,x_4,x_5,s_1,s_2 \geq 0 \\ \nonumber
\end{align}

\section{Geometry of a Linear Program}

The geometry of a Linear Program (LP) is important since most solvers exploit this geometry in order to obtain the optimal solution of an LP efficiently. A region defined by the LP, like the yellow striped region in Figure~\ref{fig:LP}, has a set of corners, also called vertices. If an LP is feasible and bounded, then the optimal solution lies on a vertex. 
More formally, a set $P$ of vectors in $\mathbb{R}^n$ is a {\it polyhedron} if $P=\{\mathbf{x} \in \mathbb{R}^n : \mathbf{A} \mathbf{x} \leq \mathbf{b}\}$ for some matrix $\mathbf{A}$ and some vector $\mathbf{b}$. $P$ defines the set of feasible solutions, as shown in Figure~\ref{fig:LP}.

An inequality $\mathbf{a}^\intercal \mathbf{x} \leq {\beta}$ is {\it valid} for a polyhedron $P$ if each $\mathbf{x}^* \in P$ satisfies $\mathbf{a}^\intercal \mathbf{x}^* \leq {\beta}$. The inequality is {\it active} at $\mathbf{x}^* \in \mathbb{R}^n$ if $\mathbf{a}^\intercal \mathbf{x}^* = {\beta}$. For example, in Figure~\ref{fig:optimal} we can see that the optimal solution, depicted as a yellow dot, is active for the green and red constraints.

\begin{figure*}[htbp]
\centering
\subfigure[]{
\includegraphics[width=0.31\linewidth]{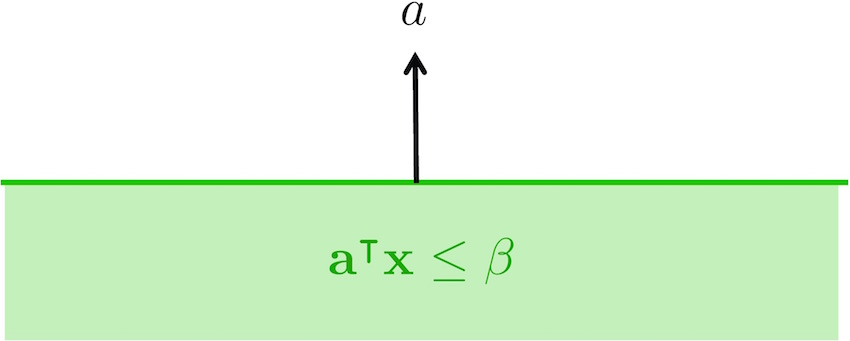} 
\label{fig:constraint1}
}
\subfigure[]{
\includegraphics[width= 0.395\linewidth]{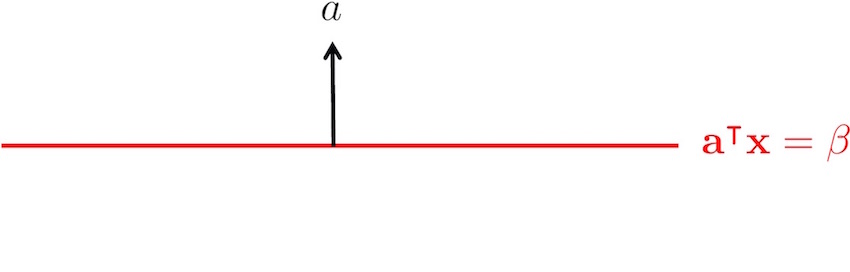} 
\label{fig:constraint2}
}
\caption[Representation of the constraints of a Linear Program]{(a) Half-space defined by the inequality constraint. (b) Hyperplane defined by the equality constraint.}
\label{fig:constraints}
\end{figure*}

Considering $\mathbf{x} \in \mathbb{R}^n$, $\mathbf{a} \in \mathbb{R}^n \setminus \{ \boldsymbol 0\}$ and $\beta \in \mathbb{R}$, we can see the representation of the inequality constraint in Figure~\ref{fig:constraint1} as a half-space and the equality constraint in Figure~\ref{fig:constraint2} as a hyperplane.

Let us now consider the notion of a {\it vertex}. Looking at Figure~\ref{fig:optimal}, we can see that the optimal solution in yellow is a point inside $P$ where the green and red constraints are active. In this 2D space, we need two constraints to define a vertex.
More formally, a point $\mathbf{x}^* \in P$ is a {\it vertex} of $P$ if there exist $n$ or more inequalities $\mathbf{a}^\intercal \mathbf{x} \leq {\beta}$ that are valid for $P$ and active at $\mathbf{x}^*$ and not all active at any other point in $P$. 

Another interpretation of the definition of vertices is that 
the point $\mathbf{x}^* \in \mathbb{R}^n$ is a {\it basic solution} if $\operatorname{rank}(\mathbf{A}_\textrm{I}) = n$, where $\mathbf{A}_\textrm{I} \mathbf{x} = \mathbf{b}_\textrm{I}$ is a sub-system of active inequations at $\mathbf{x}^*$. If $\mathbf{x}^* \in P$, then it is a {\it basic feasible solution}. In this case, $\mathbf{x}^*$ is a vertex of $P$ iff it is a basic feasible solution.

\begin{colbox}
\begin{theorem}
If a linear program $\max \{\mathbf{c}^\intercal \mathbf{x} : \mathbf{x} \in \mathbb{R}^n, \mathbf{A} \mathbf{x} \leq \mathbf{b}\}$ is feasible and bounded and if $\operatorname{rank}(\mathbf{A})=n$, the LP has an optimal solution that is a vertex.
\label{th:LPtheorem1}
\end{theorem}
\end{colbox}

Recall from linear algebra that a system of equations with $m$ constraints and $n$ variables can either be directly solvable if $m=n$ and $\mathbf{A}$ is full-rank, which means it is invertible. 
If $m<n$, we have an underdetermined system which leads to more than one optimal solution. For example, we can have several solutions that lie on an edge instead of only one solution on a vertex. 
Finally, if $m>n$, we have an overdetermined system, in which case it is possible that there exists no solution. Usually these problems are solved by using least-squares (see \citep{PRML}).

We can draw an important consequence from Theorem \ref{th:LPtheorem1}, which is that an LP can be solved by enumerating all vertices and picking the best one. As the dimensionality of our search space and number of constraints increase, enumerating all solutions quickly becomes unmanageable. In the following Section, we present the Simplex algorithm developed by George B. Danzig in 1947, which drastically reduces the number of possible optimal solutions that must be checked.

\section{The Simplex method}
\label{simplexsection}

If we know that the optimal solution lies on a vertex, we could simply evaluate the objective function on each of the vertices and just pick the optimum one. Nonetheless, the number of vertices of an LP is typically too large, therefore we need to find a clever way to move towards the optimum vertex. 

\begin{figure*}[th]
\centering
\includegraphics[height=5cm]{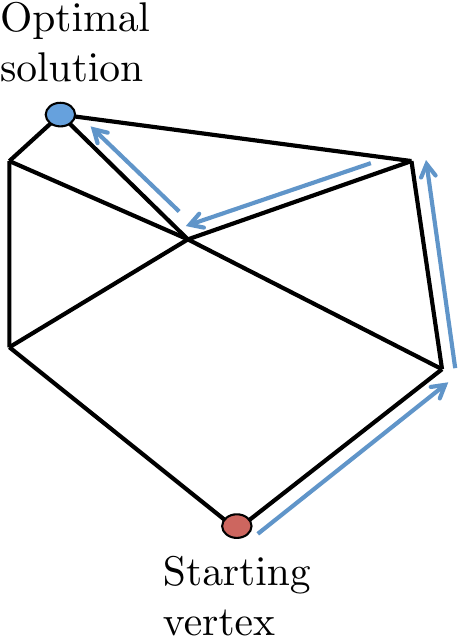} 
\caption[Idea of the Simplex algorithm.]{Idea of the Simplex algorithm. Starting from a vertex of the polyhedron, we move along the edges until we reach the optimum solution.}
\label{fig:simplexidea}
\end{figure*}

The Simplex method is an iterative method to efficiently solve an LP. The basic intuition behind the algorithm is depicted in Figure~\ref{fig:simplexidea}. Starting from a vertex in the feasible region, the idea is to move along the edges of the polyhedron until the optimum solution is reached. Each move from one vertex to another shall increase the objective function (in case we have a maximization problem), so that convergence is guaranteed.
In other words, the Simplex algorithm maintains a basic feasible solution at every step. Given a basic feasible solution, the method first applies an optimality criterion to test its optimality. If the current solution does not fulfill this condition, the algorithm obtains another solution with a higher value of the objective function (which is closer to the optimum in the case of a maximization problem). 
Let us now define some more useful concepts. 

Two distinct vertices $\mathbf{x}_1$ and $\mathbf{x}_2$ of $P=\{\mathbf{x} \in \mathbb{R}^n : \mathbf{A} \mathbf{x} \leq \mathbf{b}\}$ are {\it adjacent}, if there exist $n-1$ {\it linearly independent inequalities} of $\mathbf{A} \mathbf{x} \leq \mathbf{b}$ active at both $\mathbf{x}_1$ and $\mathbf{x}_2$.

\begin{theorem}
$\mathbf{x}_1 \neq \mathbf{x}_2 \in P$ are adjacent iff there exists $\mathbf{c} \in \mathbb{R}^n$ such that a set of solutions of $\max \{ \mathbf{c}^\intercal \mathbf{x} : \mathbf{x} \in P \}$ is the line segment spanned by $\mathbf{x}_1$ and $\mathbf{x}_2$.
\end{theorem}

Let us assume we start with a solution vertex $\mathbf{x}^*$. While $\mathbf{x}^*$ is not optimal, the algorithm finds another vertex $\mathbf{x}'$ adjacent to $\mathbf{x}^*$ with $\mathbf{c}^\intercal \mathbf{x}' > \mathbf{c}^\intercal \mathbf{x}^*$, and update $\mathbf{x}^* \vcentcolon= \mathbf{x}'$. If no vertex can be found, we can assert that the LP is unbounded.
This is summarized in Algorithm \ref{alg:simplex}.

\begin{algorithm}
\caption{Basic idea of the Simplex algorithm}          
\label{alg:simplex}                           
\begin{algorithmic}                    
\vspace{0.13cm}
\STATE Start with vertex $\mathbf{x}^*$
\vspace{0.13cm}
\WHILE{$\mathbf{x}^*$ is not optimal}
\vspace{0.13cm}
\IF{We find a vertex $\mathbf{x}'$ adjacent to $\mathbf{x}^*$ with $\mathbf{c}^\intercal \mathbf{x}' > \mathbf{c}^\intercal \mathbf{x}^*$}
\vspace{0.13cm}
\STATE $\mathbf{x}^* \vcentcolon= \mathbf{x}'$
\vspace{0.13cm}
\ELSE 
\vspace{0.13cm}
\STATE Assert that LP is unbounded.
\vspace{0.13cm}
\ENDIF
\vspace{0.13cm}
\ENDWHILE
\vspace{0.13cm}
\end{algorithmic}
\end{algorithm}

As we can see, there are two key aspects to be defined: firstly, how to assert that a vertex is optimal, and secondly, how to find an adjacent vertex with a better cost. Both will be detailed in the next subsections. 

\subsection{Optimality criteria}
\label{sec:optimality}

Again, let us start by defining some concepts, namely bases and degeneracy. 

A subset $B \subseteq \{1, \ldots, m \}$ of the rows-indices of $\mathbf{A}$, with $|B|=n$ and $\mathbf{A}_B$ invertible, is called a {\it basis} of the LP. If in addition the point $\mathbf{A}^{-1}_B\mathbf{b}_B$ is feasible, then $B$ is called a {\it feasible basis}. 

\begin{figure*}[htbp]
\centering
\includegraphics[width=0.5\linewidth]{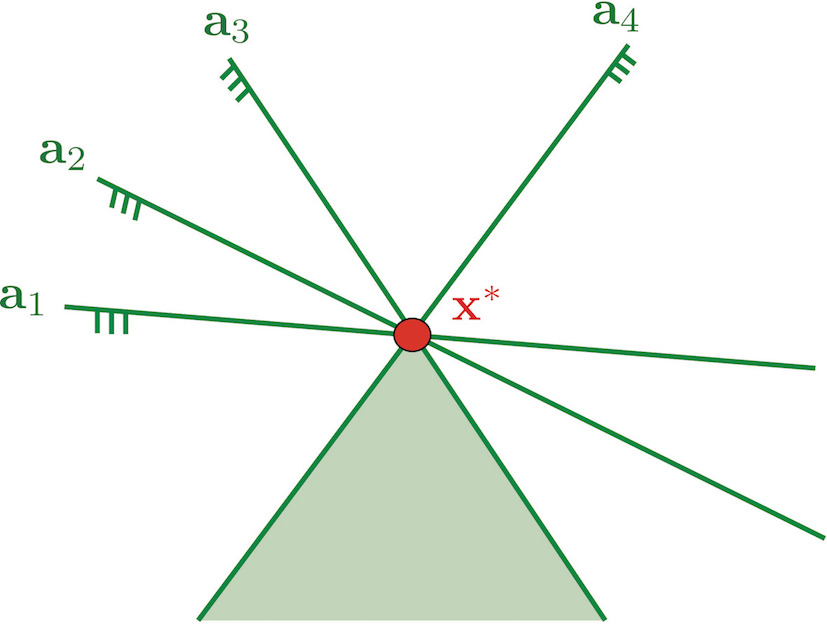} 
\caption[LP with 4 constraints]{LP with 4 constraints identified by their coefficients $\{\mathbf{a}_1,\mathbf{a}_2,\mathbf{a}_3,\mathbf{a}_4\}$ and depicted by green lines. There are 6 bases that represent $\mathbf{x}^*$, namely, $\{\{\mathbf{a}_1,\mathbf{a}_2\},\{\mathbf{a}_2,\mathbf{a}_3\},\{\mathbf{a}_1,\mathbf{a}_3\},\{\mathbf{a}_3,\mathbf{a}_4\},\{\mathbf{a}_1,\mathbf{a}_4\},\{\mathbf{a}_2,\mathbf{a}_4\}\}$. The half-space of constraints $\{\mathbf{a}_3,\mathbf{a}_4\}$ is the area depicted in light green.}
\label{fig:bases}
\end{figure*}

If a vertex $\mathbf{x}^* \in P$ is represented by a basis $B$, then $\mathbf{x}^*=\mathbf{A}^{-1}_B\mathbf{b}_B$. But a vertex can be represented by many bases. Let us consider the LP problem $\max \mathbf{c}^\intercal \mathbf{x}$ depicted in Figure~\ref{fig:bases}, where $\mathbf{x} \in \mathbb{R}^2$. There are 4 constraints in this LP, identified by their coefficients $\{\mathbf{a}_1,\mathbf{a}_2,\mathbf{a}_3,\mathbf{a}_4\}$ and depicted by green lines. Since we are in a 2D space, each pair of constraints forms a basis for $\mathbf{x}^*$. A possible set of feasible solutions created by constraints $\mathbf{a}_3$ and $\mathbf{a}_4$ is painted in light green. In total, we have 6 bases that represent $\mathbf{x}^*$, namely, $\{\{\mathbf{a}_1,\mathbf{a}_2\},\{\mathbf{a}_2,\mathbf{a}_3\},\{\mathbf{a}_1,\mathbf{a}_3\},\{\mathbf{a}_3,\mathbf{a}_4\},\{\mathbf{a}_1,\mathbf{a}_4\},\{\mathbf{a}_2,\mathbf{a}_4\}\}$.

An LP $\max \{ \mathbf{c}^\intercal \mathbf{x} : \mathbf{x} \in \mathbb{R}^n , \mathbf{A} \mathbf{x} \leq \mathbf{b}\}$ is {\it degenerate} if there exists an $\mathbf{x}^* \in \mathbb{R}^n$ such that there are more than $n$ constraints of $\mathbf{A} \mathbf{x} \leq \mathbf{b}$ active at $\mathbf{x}^*$. The LP depicted in Figure~\ref{fig:bases} is degenerate, since $n=2$ and there are 4 active constraints at $\mathbf{x}^*$.

A basis $B$ is {\it optimal} if it is feasible and the unique $\boldsymbol \lambda \in \mathbb{R}^m$ with $\boldsymbol \lambda^\intercal \mathbf{A} = \mathbf{c}^\intercal$ and $\lambda_i = 0, \forall i \notin B$ satisfies $\boldsymbol \lambda \geq \mathbf{0}$.

If all components of $\boldsymbol \lambda$ outside of $B$ are zero, then we can write the following equality $\boldsymbol \lambda_B^\intercal \mathbf{A}_B = \mathbf{c}^\intercal$, since all rows of $\mathbf{A}$ with indices outside of $B$ will not contribute to the dot product. Since $\mathbf{A}_B$ is invertible, we can then write $\boldsymbol \lambda_B^\intercal  = \mathbf{c}^\intercal \mathbf{A}_B^{-1}$. From this, two theorems emerge.

\begin{theorem}
If $B$ is an optimal basis, then $\mathbf{x}^*=\mathbf{A}_B^{-1} \mathbf{b}_B$ is an optimal solution of the LP.
\end{theorem}

\begin{theorem}
Suppose the LP is non-degenerate and $B$ is a feasible but not optimal basis, then $\mathbf{x}^*=\mathbf{A}_B^{-1} \mathbf{b}_B$ is not an optimal solution.
\label{theorem:LP}
\end{theorem}

Basically, for every vertex $\mathbf{x}^*$, we can quickly check if it is an optimal solution by checking if the basis $B$ that represents this vertex is optimal or not.
The proof of the theorem will help us see how to move closer to the optimal solution. 

\begin{proof}
Let us assume that $B$ is a feasible but not optimal basis. We can split the constraints of the LP into active and inactive ones with respect to $B$.
\begin{align}
\max \quad \mathbf{c}^\intercal \mathbf{x}, \quad \st \quad \mathbf{A} \mathbf{x} \leq \mathbf{b} \quad
\begin{cases}
\mathbf{A}_B \mathbf{x} \leq \mathbf{b}_B, \mbox{ active at } \mathbf{x}^* \\
\mathbf{A}_{\bar{B}} \mathbf{x} \leq \mathbf{b}_{\bar{B}}, \mbox{ inactive}
\end{cases}
\label{eq:ineq}
\end{align}

For a unique $\boldsymbol \lambda \in \mathbb{R}^m$ with $\boldsymbol \lambda^\intercal \mathbf{A} = \mathbf{c}^\intercal$, we have that $\lambda_j=0, \forall j \notin B$. Since $B$ is feasible but not optimal, we know that there will be some $\lambda_i < 0$ for some $i \in B$.

We now compute a $\mathbf{d} \in \mathbb{R}^n$ such that $\mathbf{A}_{B \setminus \{i\}} \mathbf{d} = \mathbf{0}$ and $\mathbf{a}_i^\intercal \mathbf{d} = -1$. That means $\mathbf{d}$ is orthogonal to all rows of $\mathbf{A}_B$ except the one that represents constraint $i$. 

Now we want to move from $\mathbf{x}^*$ in the direction given by $\mathbf{d}$, as depicted in Figure~\ref{fig:dirD}. Let us first take a look at what happens to the objective function if we move along $\mathbf{d}$:
\begin{align}
\mathbf{c}^\intercal  \mathbf{d} =  \boldsymbol \lambda_B^\intercal  \mathbf{A}_B  \mathbf{d} = \underbrace{\lambda_i}_{<0}  \underbrace{\mathbf{a}_i^\intercal  \mathbf{d}}_{-1}
\end{align}

\begin{figure*}[tbp]
\centering
\subfigure[]{
\includegraphics[height=4cm]{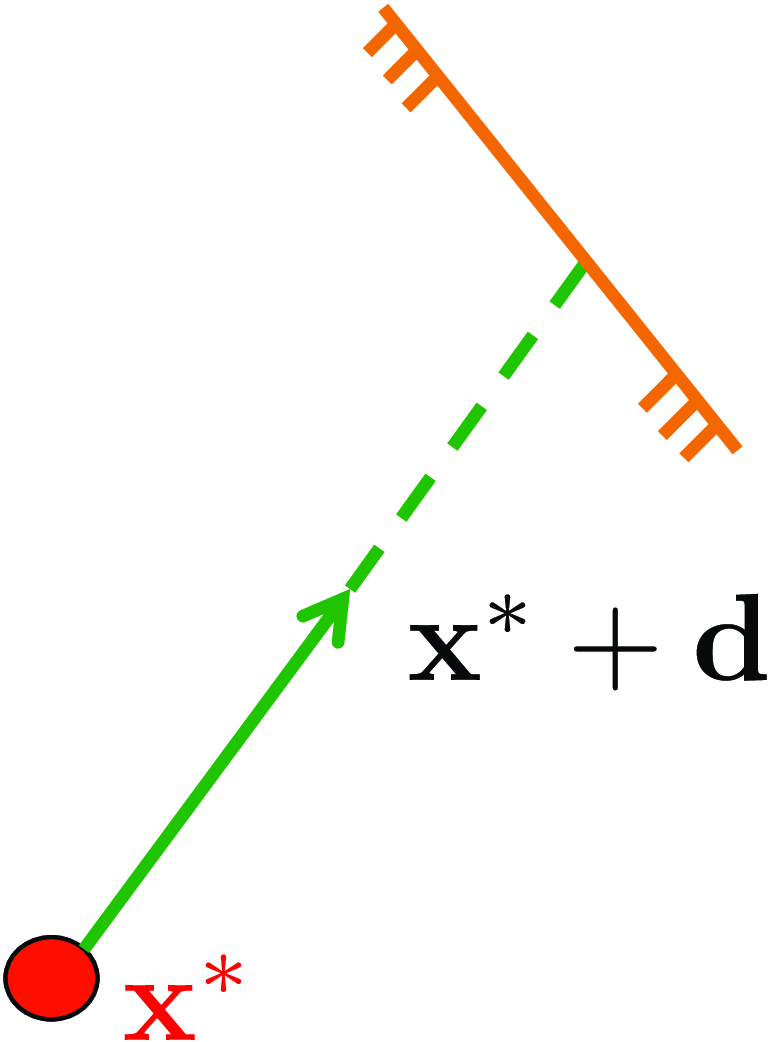} 
\label{fig:dirD}
}\qquad \qquad
\subfigure[]{
\includegraphics[height=4cm]{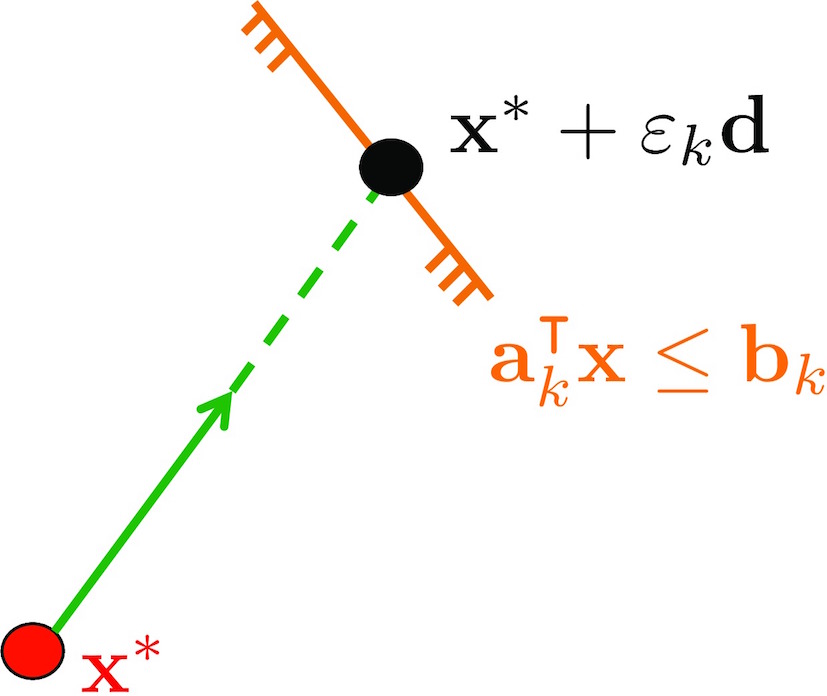} 
\label{fig:dirDconstraint}
}
\caption[Move in the direction $\mathbf{d}$ as shown in the proof of Theorem \ref{theorem:LP}.]{(a) Move in the direction $\mathbf{d}$ as shown in the proof of Theorem \ref{theorem:LP}. (b) Reaching constraint $k$ by moving $\varepsilon_k$ in the direction $\mathbf{d}$.}
\label{fig:dirsD}
\end{figure*}

Given $\lambda_i < 0$ and the conditions in which we defined $\mathbf{d}$, for which $\mathbf{a}_i^\intercal \mathbf{d}=-1$, we can see that $\mathbf{c}^\intercal \mathbf{d} > 0$. This means that if we move in the direction $\mathbf{d}$, we will improve our objective function value.

Let us now consider a given quantity $\varepsilon >0$, which represents how much we move along direction $\mathbf{d}$. Now we are interested in knowing if the new point $\mathbf{x}^* + \varepsilon  \mathbf{d}$ is feasible, \ie if it satisfies Equation \eqref{eq:ineq}. We can see that the inequations are certainly satisfied:
\begin{align}
\mathbf{A}_B  (\mathbf{x}^* + \varepsilon  \mathbf{d} ) \leq \mathbf{b}_B, 
\end{align}
since the product $\varepsilon \mathbf{A}_B  \mathbf{d} <0$ because of the previous definition $\mathbf{a}_i^\intercal \mathbf{d}=-1$. This means that there exists an $\varepsilon^*$ such that $\mathbf{x}^* + \varepsilon^*  \mathbf{d}$ is feasible because it satisfies all inequalities expressed in Equation \eqref{eq:ineq}. 
But the value of the objective function at this new point will be
\begin{align}
\mathbf{c}^\intercal  (\mathbf{x}^* + \varepsilon  \mathbf{d} ) = \mathbf{c}^\intercal  \mathbf{x}^* + \underbrace{\varepsilon   \underbrace{\mathbf{c}^\intercal   \mathbf{d} }_{>0}}_{>0},
\end{align}
which is greater than the objective value of $\mathbf{x}^*$, proving this is not an optimal solution.
\label{proof:dirD}
\end{proof}

\subsection{Moving to a better neighbor}

Now we have an $\varepsilon >0$ with which we can move from $\mathbf{x}^*$ in the direction $\mathbf{d}$ to a vertex close to the optimum, namely, to a better neighbor. The question now is how large can $\varepsilon$ be. We need to find out how far we can go before we hit a constraint for the first time, because past a constraint, the feasible region ends. This is depicted in Figure~\ref{fig:dirsD} where the constraint is represented in orange. 
 
Remember we had $m$ constraints in $\mathbf{A} \mathbf{x} \leq \mathbf{b}$. We denote $\mathcal{K}$ as the set of indices that represent the constraints that might be hit by $\mathbf{x}^* + \varepsilon  \mathbf{d}$, and it is formally defined as
\begin{align}
\mathcal{K} = \{ k : 1 \leq k \leq m , \mathbf{a}_k^\intercal \mathbf{d} > 0\}.
\end{align}
$\mathbf{a}_k^\intercal \mathbf{d}$ needs to be larger than zero, otherwise we would never hit the constraint $\mathbf{a}_k^\intercal \mathbf{x} \leq \mathbf{b}_k$. The set of constraints $\mathcal{K}$ will contain constraints not in basis $B$, since all $\mathbf{A}_B^\intercal \mathbf{d} \leq \mathbf{0}$.
There are now two cases:
\begin{enumerate}
\item{$\mathcal{K}=\emptyset$, which means we can move indefinitely in direction $\mathbf{d}$, and therefore the LP is {\it unbounded}.}
\item{$\mathcal{K}\neq\emptyset$, which means there is a constraint with index $k$ which we will hit while moving $\mathbf{x}^*$ in the direction $\mathbf{d}$, as depicted in Figure~\ref{fig:dirDconstraint}.
Let us now compute the value of $\varepsilon_k$ for which we hit constraint $k$:
\begin{align}
\mathbf{a}^\intercal_k \left( \mathbf{x}^* + \varepsilon_k \mathbf{d} \right) = \mathbf{b}_k \iff \varepsilon_k = \frac{\mathbf{b}_k - \mathbf{a}^\intercal_k \mathbf{x}^*}{\mathbf{a}^\intercal_k \mathbf{d}}
\end{align}
We know this division can be done because the denominator is greater than zero. The optimal $\varepsilon^*$ will be the smallest of all the $\varepsilon_k$:
\begin{align}
\varepsilon^*=\min_{k \in \mathcal{K}} \varepsilon_k,
\end{align}
where $k^* \in \mathcal{K}$ is the index for which we find $\varepsilon^*$. The optimal $\varepsilon^*$ must be the minimum, because all greater $\varepsilon_k$ violate at least the constraint $k^*$, and therefore go out of the feasible region.
To know that there is, in fact, a new vertex $\mathbf{x}' = \mathbf{x}^* + \varepsilon^* \mathbf{d}$, which is adjacent to $\mathbf{x}^*$ and with higher objective value, we have to prove that $B'$ defined as
\begin{align}
B'=B\setminus \{i\} \cup \{k^*\}
\end{align}
is a basis. Note that we are incorporating the new constraint $k^*$ and taking out the $i$ that did not make our basis $B$ optimal (recall that $\lambda_i<0$).
Remember that $\mathbf{d} \perp \mathbf{a}_{B \setminus \{i\}}$, but not $\mathbf{d}  \perp \mathbf{a}_{k^*}$ since $\mathbf{a}_{k^*}^\intercal \mathbf{d}>0$. This means that $\mathbf{a}_{k^*}$ is not a linear combination of $\mathbf{a}_{B \setminus \{ i \}}$, proving $B$ is a basis.
Furthermore, the inequalities $\mathbf{A}^\intercal_{B'} \mathbf{x} \leq \mathbf{b}_{B'}$ are active at $\mathbf{x}'$, which means $\mathbf{x}'$ is a vertex and in fact adjacent to $\mathbf{x}^*$.
}
\end{enumerate}

We have seen so far that the concepts of basic feasible solution and feasible basis are interchangeable, therefore we can rewrite the Simplex algorithm in basis notation, as shown in Algorithm \ref{alg:simplex_all}.

\begin{algorithm}
\caption{The Simplex algorithm}          
\label{alg:simplex_all}                           
\begin{algorithmic}                    
\vspace{0.13cm}
\STATE Start with a feasible basis $B$
\vspace{0.13cm}
\WHILE{$B$ is not optimal}
\vspace{0.13cm}
\STATE{Let $i \in B$ be the index with $\lambda_i<0$ (remember $\boldsymbol \lambda^\intercal \mathbf{A} = \mathbf{c}^\intercal$ and $\lambda_j=0, \forall j \notin B$)}
\vspace{0.13cm}
\STATE{Compute $\mathbf{d} \in \mathbb{R}^n$ with $\mathbf{A}^\intercal_{B \setminus \{i \}} \mathbf{d} = \mathbf{0}$ and $\mathbf{a}^\intercal_i \mathbf{d} =-1$}
\vspace{0.13cm}
\STATE{Determine $\mathcal{K} = \{ k : 1 \leq k \leq m , \mathbf{a}_k^\intercal \mathbf{d} > 0\}$}
\vspace{0.13cm}
\IF{$\mathcal{K}=\emptyset$}
\vspace{0.13cm}
\STATE Assert that LP is unbounded.
\vspace{0.13cm}
\ELSE 
\vspace{0.13cm}
\STATE{Let $k^* \in \mathcal{K}$ be the index where $\underset{k \in \mathcal{K}}\min \frac{\mathbf{b}_k - \mathbf{a}^\intercal_k \mathbf{x}^*}{\mathbf{a}^\intercal_k \mathbf{d}}$ is attained}\vspace{0.13cm}
\STATE Update $B \vcentcolon= B \setminus \{i \} \cup \{ k^* \}$
\vspace{0.13cm}
\ENDIF
\vspace{0.13cm}
\ENDWHILE
\vspace{0.13cm}
\end{algorithmic}
\end{algorithm}

\begin{theorem}
If the Linear Program is non-degenerate, then the Simplex algorithm terminates.
\end{theorem}

\begin{colbox}
The idea of the Simplex algorithm is to jump from one base to another (equivalently from vertex to vertex), making sure no base is revisited. We have proven before that when we move in direction $\mathbf{d}$ from point $\mathbf{x}^*$ to $\mathbf{x}'$, we obtain $\mathbf{c}^\intercal \mathbf{x}' > \mathbf{c}^\intercal \mathbf{x}^*$, which means that we are making progress at each iteration of the Simplex, proving it will eventually terminate.
\end{colbox}

\subsection{The degenerate case: Bland's pivot rule}
\label{sec:bland}

The Simplex algorithm as described in Algorithm \ref{alg:simplex_all} can be applied to degenerate Linear Programs, but we can encounter the problem of cycling, which is when we move from one basis to another without progress and end up returning to one of the bases we already visited. This means that the algorithm would never terminate. 
In order to avoid this, we need to carefully choose the indices that are leaving and entering the basis at each iteration, an operation that is called {\it pivoting}. In Algorithm \ref{alg:simplex_bland}, we highlight in orange the changes to the Simplex algorithm according to Bland's pivot rule \citep{bland1977}, which allows Simplex to solve degenerate LP. 

\begin{algorithm}
\caption{The Simplex algorithm with Bland's pivot rule}          
\label{alg:simplex_bland}                           
\begin{algorithmic}                    
\vspace{0.13cm}
\STATE Start with a feasible basis $B$
\vspace{0.13cm}
\WHILE{$B$ is not optimal}
\vspace{0.13cm}
\STATE{Let $i \in B$ be the \textcolor{orange}{\bf smallest} index with $\lambda_i<0$ ($\boldsymbol \lambda^\intercal \mathbf{A} = \mathbf{c}^\intercal$ and $\lambda_j=0,  \forall j \notin B$)}
\vspace{0.13cm}
\STATE{Compute $\mathbf{d} \in \mathbb{R}^n$ with $\mathbf{A}^\intercal_{B \setminus \{i \}} \mathbf{d} = \mathbf{0}$ and $\mathbf{a}^\intercal_i \mathbf{d} =-1$}
\vspace{0.13cm}
\STATE{Determine $\mathcal{K} = \{ k : 1 \leq k \leq m , \mathbf{a}_k^\intercal \mathbf{d} > 0\}$}
\vspace{0.13cm}
\IF{$\mathcal{K}=\emptyset$}
\vspace{0.13cm}
\STATE Assert that LP is unbounded.
\vspace{0.13cm}
\ELSE 
\vspace{0.13cm}
\STATE{Let $k^* \in \mathcal{K}$ be the \textcolor{orange}{\bf smallest} index where $\underset{k \in \mathcal{K}}\min \frac{\mathbf{b}_k - \mathbf{a}^\intercal_k \mathbf{x}^*}{\mathbf{a}^\intercal_k \mathbf{d}}$ is attained}
\vspace{0.13cm}
\STATE Update $B \vcentcolon= B \setminus \{i \} \cup \{ k^* \}$
\vspace{0.13cm}
\ENDIF
\vspace{0.13cm}
\ENDWHILE
\vspace{0.13cm}
\end{algorithmic}
\end{algorithm}

\begin{theorem}
If Bland's rule is applied, the Simplex algorithm terminates.
\end{theorem}

For the interested reader, the proof of the theorem can be found in \citep{Schrijver1998}.

\subsection{Finding an initial vertex}
\label{initialvertex}

In all descriptions of Simplex in Algorithms \ref{alg:simplex}, \ref{alg:simplex_all} and \ref{alg:simplex_bland}, it always starts by choosing a feasible initial vertex or basis. But how do we find this initial vertex? Finding a feasible solution of a Linear Program is almost as difficult as finding an optimal solution. Fortunately, by using a simple technique, we can find a feasible solution of a related auxiliary LP and use it to initialize the Simplex method on our LP.
Let us consider our initial LP to be in the standard form 2:
\begin{align}
\label{eq:canonicallp2}
\max  \quad &\mathbf{c}^\intercal \mathbf{x} \\
\textrm{s.t.} \quad &\mathbf{A} \mathbf{x} \leq \mathbf{b}\nonumber\\
&\mathbf{x}\geq \mathbf{0}. \nonumber
\end{align}

We can split the conditions according to whether $b_i$ has a positive or negative value:
\begin{align}
\label{eq:conditions}
 \mathbf{A} \mathbf{x} \leq \mathbf{b}
 \begin{cases}
  \mathbf{A}_1 \mathbf{x} \leq \mathbf{b}_1, \mathbf{b}_1\geq 0, \mathbf{b}_1 \in \mathbb{R}^{m_1}\\
 \mathbf{A}_2 \mathbf{x} \leq \mathbf{b}_2, \mathbf{b}_2< 0, \mathbf{b}_2 \in \mathbb{R}^{m_2} 
 \end{cases}
 \end{align}

and define a new artificial variable $\mathbf{y}$. We now create an auxiliary LP where we minimize the sum of the new artificial variables:
\begin{align}
\label{eq:auxiliar}
\min \quad &\sum\limits_{i=1}^{m_2} y_i \\
\textrm{s.t.} \quad &\mathbf{A}_1 \mathbf{x} \leq \mathbf{b}_1 \nonumber\\
&\mathbf{A}_2 \mathbf{x} \leq \mathbf{b}_2 + \mathbf{y} \nonumber\\
 &\mathbf{x}, \mathbf{y}\geq \mathbf{0}. \nonumber \\
&\mathbf{y}\leq |\mathbf{b}_2 |. \nonumber
\end{align}

We can show that this auxiliary problem is always feasible, since we can always find an initial feasible solution like $\mathbf{x}^*=\mathbf{0}, \mathbf{y}^*=|\mathbf{b}_2|$, \ie each $y_i$ is bounded by the absolute value of the corresponding component of $\mathbf{b}_2$. It fulfills all conditions of the auxiliary LP of Equation \eqref{eq:auxiliar}, therefore it is a feasible initial vertex. From here, we can apply the Simplex as described in Algorithm \ref{alg:simplex_bland} to find the optimal solution. 
If we find an optimal solution with variables $\mathbf{x}^*,\mathbf{y}^*$ which yields the optimal value of the objective function of Equation \eqref{eq:auxiliar} to be zero, we can assert that the vertex $\mathbf{x}^*$ is a feasible solution of the original LP problem in Equation \eqref{eq:canonicallp2}. We will then use this initial vertex to start the Simplex algorithm to solve the original LP. 
On the other hand, if we find that the minimum value of the auxiliary problem is larger than zero, we can assert that the original LP is infeasible. 

The final complete description of the Simplex algorithm is found in Algorithm \ref{alg:simplex_final}. Finding the initial vertex is commonly called Phase I of the Simplex algorithm, while the optimization towards the final solution through pivoting is commonly referred to as Phase II. A hands-on example on how to solve a problem practically with Simplex will be presented in Section \ref{sec:dictionaries}.

\begin{algorithm}
\caption{The complete Simplex algorithm with Bland's pivot rule}          
\label{alg:simplex_final}                           
\begin{algorithmic}                    
\vspace{0.13cm}
\STATE {Create the auxiliary problem of the LP and find the optimal solution with basis $B$ and objective function value $z$.}
\vspace{0.13cm}
\IF {$z=0$}
\vspace{0.13cm}
\STATE {$B$ is a feasible basis of the initial LP. Start with the feasible basis $B$.}
\vspace{0.13cm}
\WHILE{$B$ is not optimal}
\vspace{0.13cm}
\STATE{Let $i \in B$ be the smallest index with $\lambda_i<0$ ($\boldsymbol \lambda^\intercal \mathbf{A} = \mathbf{c}^\intercal$ and $\lambda_j=0,  \forall j \notin B$)}
\vspace{0.13cm}
\STATE{Compute $\mathbf{d} \in \mathbb{R}^n$ with $\mathbf{A}^\intercal_{B \setminus \{i \}} \mathbf{d} = \mathbf{0}$ and $\mathbf{a}^\intercal_i \mathbf{d} =-1$}
\vspace{0.13cm}
\STATE{Determine $\mathcal{K} = \{ k : 1 \leq k \leq m , \mathbf{a}_k^\intercal \mathbf{d} > 0\}$}
\vspace{0.13cm}
\IF{$\mathcal{K}=\emptyset$}
\vspace{0.13cm}
\STATE Assert that LP is unbounded.
\vspace{0.13cm}
\ELSE 
\vspace{0.13cm}
\STATE{Let $k^* \in \mathcal{K}$ be the smallest index where $\underset{k \in \mathcal{K}}\min \frac{\mathbf{b}_k - \mathbf{a}^\intercal_k \mathbf{x}^*}{\mathbf{a}^\intercal_k \mathbf{d}}$ is attained}
\vspace{0.13cm}
\STATE Update $B \vcentcolon= B \setminus \{i \} \cup \{ k^* \}$
\vspace{0.13cm}
\ENDIF
\vspace{0.13cm}
\ENDWHILE
\vspace{0.13cm}
\ELSE
\vspace{0.13cm}
\STATE{Assert that the LP is infeasible.}
\ENDIF
\end{algorithmic}
\end{algorithm}

%
%
%

\subsection{Complexity}

The Simplex method is remarkably efficient, specially compared to earlier methods such as Fourier-Motzkin elimination. However, in 1972 it was proven that the Simplex method has exponential worst-case complexity \citep{simplexworst}. 
Nonetheless, following the observation that the Simplex algorithm is efficient in practice, it has been found that it has polynomial-time average-case complexity under various probability distributions.
In order for the Simplex to perform in polynomial time, we have to use certain pivoting rules that allow us to go from one vertex of the polyhedron to another in a small number of steps. 
We will better understand this concept when we introduce the graphical model representation of a polyhedron in Section \ref{graphicalmodel}.

\section{The dual Linear Program}

\begin{figure*}[h]
\centering
\includegraphics[height=3.6cm]{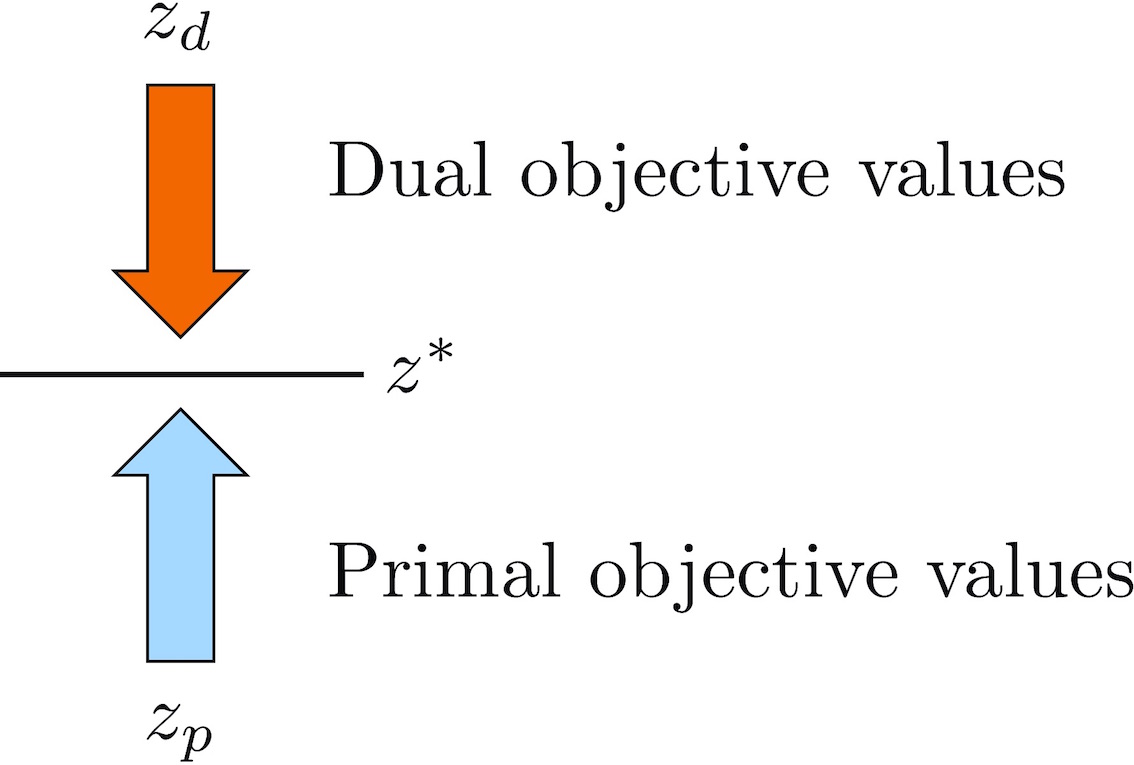} 
\caption[Motivation for dualization]{The motivation to find the dual of a problem is to find an upper bound on the objective function of the primal.}
\label{fig:dualmotivation}
\end{figure*}

In this section, we introduce a very important property of Linear Programs: {\it duality}. Given any general optimization problem, or {\it primal problem}, we can always convert it to a {\it dual problem}. For LPs the dual problem is also an LP. The motivation to use dualization, depicted in Figure~\ref{fig:dualmotivation}, is that the dual problem gives us an upper bound on the objective function of the primal problem.

As we saw in Section \ref{simplexsection}, the Simplex algorithm starts from a suboptimal solution and performs gradient ascent to iteratively find solutions with increasing objective value, until the optimum is reached. 
In the case of dual linear programs, we can find an upper bound and iteratively make it more stringent until it reaches the optimum. It is guaranteed for LPs that the smallest upper bound will correspond to the optimum solution $z^*$ of the primal problem. 

Let us consider the following LP:
\begin{align}
\max  \qquad &x_1+2x_2 \nonumber \\
\st  \qquad &-2x_1+x_2 \leq -2 \nonumber \\
&x_2 \leq 4 \nonumber \\ 
& x_1-2x_2 \leq -2\nonumber \\ 
& x_1 \leq 4 \nonumber \\
&  x_1,x_2 \leq 0 \nonumber
\end{align}
We can try to find an upper bound on the value of the objective function. One way to do this, is by linearly combining the constraints of the problem, to obtain an expression of the form $\mathbf{c}^\intercal \mathbf{x} \leq \mathbf{y}^\intercal \mathbf{b}$, where $\mathbf{y}$ are the coefficients of this linear combination. 
Let us multiply the first constraint by $2$, the fourth by $5$ and sum them up:
\begin{align}
-2x_1+x_2 \leq -2 \quad &\Longrightarrow \quad \times 2\quad  \Longrightarrow \quad -4x_1+2x_2 \leq -4 \nonumber \\
 x_1 \leq 4\quad & \Longrightarrow \quad \times 5 \quad  \Longrightarrow \quad  \hspace{0.3cm} 5x_1 \hspace{1.05cm}  \leq 20 \nonumber  \\
& \hspace{3cm} \rule{3.5cm}{0.7pt} \nonumber \\
& \hspace{3.82cm} x_1+ 2x_2 \leq 16 \nonumber
\end{align}
Note how we obtained our objective function after the sum, and therefore we can say that $16$ is an upper bound. We can also try another combination, summing the fourth constraint and the second multiplied by $2$:
\begin{align}
x_1 \leq 4 \quad &\Longrightarrow \quad \times 1\quad  \Longrightarrow \quad x_1 \hspace{1.05cm} \leq 4 \nonumber \\
 x_2 \leq 4\quad & \Longrightarrow \quad \times 2 \quad  \Longrightarrow \quad  \hspace{0.85cm} 2x_2   \leq 8 \nonumber  \\
& \hspace{3cm} \rule{3cm}{0.7pt} \nonumber \\
& \hspace{3.3cm} x_1+ 2x_2 \leq 12 \nonumber
\end{align}
In this case, we obtain an upper bound of $12$, which turns out to be the smallest upper bound and therefore corresponds to the optimum of the objective function. 

The general principle to find the dual problem is to multiply each of the constraints by a new positive variable, namely the {\it dual variable} and sum the constraints up:
\begin{alignat*}{14}
\max \qquad c_1 x_1  &+& c_2 x_2 &+& \cdots &+& c_n x_n \quad \\
\st \qquad a_{11} x_1 &+& a_{12} x_2 &+& \cdots &+& a_{1n} x_n \quad&\leq&\quad &b_1& \quad  &\longrightarrow& \quad &e_1 \leq b_1& \quad &\longrightarrow& \quad  &\textcolor{orange}{y_1}   \\
a_{21} x_1 &+& a_{22} x_2 &+& \cdots &+& a_{2n} x_n \quad&\leq& \quad &b_2&  \quad &\longrightarrow& \quad &e_2 \leq b_2& \quad &\longrightarrow& \quad  &\textcolor{orange}{y_2}  \\
 & &&& \vdots \\
a_{m1} x_1 &+& a_{m2} x_2 &+& \cdots &+& a_{mn} x_n \quad &\leq& \quad &b_m& \quad &\longrightarrow& \quad &e_m \leq b_m& \quad &\longrightarrow& \quad &  \textcolor{orange}{y_m}  \\
-x_1 &&&&& & \quad&\leq&\quad &0& &&\quad && &\longrightarrow &\quad  &\textcolor{orange}{y_{m+1}}  \\
 && -x_2 &&& & \quad&\leq& \quad &0&&& \quad& &&\longrightarrow &\quad  &\textcolor{orange}{y_{m+2}} \\
 & &&& \ddots \\
 & & &&&& -x_n \quad&\leq& \quad &0& &&\quad& &&\longrightarrow &\quad  & \textcolor{orange} {y_{m+n}} . 
\end{alignat*}
Note that we already used this trick in Section \ref{sec:optimality}, with $\boldsymbol\lambda$ as our new variables. The variables have to be positive in order not to change the inequality sign. Now we want to make this sum equal to our objective function:
\begin{align*}
\centering
z=c_1 x_1 + c_2 x_2 + \cdots + c_n x_n \equiv \sum\limits_{i=1}^m y_i e_i + \ldots + y_m e_m - y_{m+1} x_1 - \ldots - y_{m+n} x_n
\end{align*}
which, by the constraints of the primal problem, is upper bounded by 
\begin{align*}
z \leq y_1b_1+y_2b_2+\ldots+y_mb_m. 
\end{align*}
Recall that our objective is to find the {\it smallest} upper bound. 
Let us express this in a matrix notation. To make the notation clearer, we separate the new variables between the ones associated to the constraints of the primal ${\mathbf{y}=\{y_{1}, \ldots, y_{m}\}}$
and the ones associated with the implicit positivity constraints, ${\mathbf{y_s}=\{y_{m+1}, \ldots, y_{m+n}\}}$. 
\begin{align*}
\min  \quad & \mathbf{b}^\intercal \mathbf{y} \\
\st \quad &\mathbf{A}^\intercal \mathbf{y} - \mathbf{y_s}= \mathbf{c}  \\
& \mathbf{y} \geq \mathbf{0}  \\
& \mathbf{y_s} \geq \mathbf{0} 
\end{align*}
We can eliminate $\mathbf{y_s}$ by substitution, $\mathbf{y_s}= \mathbf{A}^\intercal \mathbf{y} - \mathbf{c}$, obtaining the final equations for the primal and dual problems:

\begin{figure}[H]
\begin{minipage}[b]{0.49\linewidth}
\centering
PRIMAL
\begin{align}
\max \quad & \mathbf{c}^\intercal \mathbf{x} \nonumber \\
\st \quad &\mathbf{A} \mathbf{x} \leq \mathbf{b} \nonumber \\
&\mathbf{x} \geq \mathbf{0} \nonumber
\end{align}
\end{minipage}
\begin{minipage}[b]{0.49\linewidth}
\centering
DUAL
\begin{align}
\min  \quad & \mathbf{b}^\intercal \mathbf{y} \nonumber \\
\st \quad&\mathbf{A}^\intercal \mathbf{y} \geq \mathbf{c} \nonumber \\
& \mathbf{y} \geq \mathbf{0} \nonumber
\end{align}
\end{minipage}
\end{figure}

So far, we have seen the relationship between a Linear Program and its dual. This is summarized in the following theorem:

\begin{theorem}
{\bf Weak Duality.} Consider a Linear Program $\max \{ \mathbf{c}^\intercal \mathbf{x} : \mathbf{x} \in \mathbb{R}^n , \mathbf{A} \mathbf{x} \leq \mathbf{b}, \mathbf{x} \geq \mathbf{0}\}$ and its dual $\min \{ \mathbf{b}^\intercal \mathbf{y} : \mathbf{y} \in \mathbb{R}^m , \mathbf{A}^\intercal \mathbf{y} \geq \mathbf{c}, \mathbf{y} \geq \mathbf{0}\}$. If $\mathbf{x}^* \in \mathbb{R}^n$ and $\mathbf{y}^* \in \mathbb{R}^m$ are primal and dual feasible respectively, then $\mathbf{c}^\intercal \mathbf{x}^* \leq \mathbf{b}^\intercal \mathbf{y}^*$. 
\label{th:weakdual}
\end{theorem}

This can be easily seen by the inequalities $\mathbf{c}^\intercal\mathbf{x} \leq \mathbf{y}^\intercal \mathbf{A} \mathbf{x} \leq \mathbf{y}^\intercal \mathbf{b} = \mathbf{b}^\intercal \mathbf{y}$, the first of which comes from the constraints of the dual problem, and the second from the constraints of the primal provided that $\mathbf{y} \geq \mathbf{0}$.

An even more important theorem is:

\begin{colbox}
\begin{theorem}
{\bf Strong Duality.} Consider a Linear Program $\max \{ \mathbf{c}^\intercal \mathbf{x} : \mathbf{x} \in \mathbb{R}^n , \mathbf{A} \mathbf{x} \leq \mathbf{b}, \mathbf{x} \geq \mathbf{0}\}$ and its dual $\min \{ \mathbf{b}^\intercal \mathbf{y} : \mathbf{y} \in \mathbb{R}^m , \mathbf{A}^\intercal \mathbf{y} \geq \mathbf{c}, \mathbf{y} \geq \mathbf{0}\}$.If the primal is feasible and bounded, then there exist a primal feasible $\mathbf{x}^*$ and a dual feasible $\mathbf{y}^*$ with $\mathbf{c}^\intercal \mathbf{x}^* = \mathbf{b}^\intercal \mathbf{y}^*$. 
\label{th:strongdual}
\end{theorem}
\end{colbox}

This means that with the dual we can find an upper bound that is tight at the optimal solution of the primal. This can be used to prove optimality of primal solutions and, as a consequence, optimality of dual solutions.

\begin{proof}
The proof of Theorem \ref{th:strongdual} is divided into two cases. 
\begin{enumerate}
\item{{\bf $\mathbf{A}$ has full column rank}.

If this is the case, then we can use the Simplex algorithm to obtain an optimal basis $B \in \{1,\ldots,m \}$. By the optimality of $B$, we know that $\mathbf{y} \in \mathbb{R}^m$ subject to $\mathbf{y} _B^\intercal \mathbf{A} _B = \mathbf{c} ^\intercal$, and that $y_i  \geq {0}$ for all $i \notin B$.
We then know that the condition $\mathbf{y}  \geq \mathbf{0}$ is fulfilled, and therefore $\mathbf{y} $ is dual feasible.

Now consider that $\mathbf{x}^*=\mathbf{A} _B^{-1}\mathbf{b} _B$ is the current primal solution returned by the Simplex. We can compare the value of the objective function at $\mathbf{x}^*$ with the value of the dual objective function at $\mathbf{y} $ to check if they are, in fact, equal. 

$\underbrace{\mathbf{c}^\intercal \mathbf{x}^*}_{primal} = \mathbf{y} _B^\intercal \mathbf{A} _B \mathbf{x}^* = \mathbf{y} _B^\intercal \underbrace{\mathbf{A}_B \mathbf{A} _B^{-1}}_{\mathbf{I}}\mathbf{b} _B =  \mathbf{y} _B^\intercal \mathbf{b} _B = \underbrace{\mathbf{y} ^\intercal \mathbf{b} }_{dual}$
}
\item{{\bf $rank(\mathbf{A}) < n$}.

First, we need to make sure our constraint matrix has full column rank, which is why we replace the vector of variables $\mathbf{x}$ with $\mathbf{x}_1-\mathbf{x}_2$. Now the Linear Program looks like:
\begin{align}
\max  \quad & \mathbf{c}^\intercal (\mathbf{x}_1-\mathbf{x}_2) \nonumber \\
&\mathbf{A} (\mathbf{x}_1-\mathbf{x}_2) \leq \mathbf{b} \nonumber \\
&\mathbf{x}_1, \mathbf{x}_2 \geq \mathbf{0} \nonumber 
\end{align}
Note, that the new LP will be equivalent to the old one in the sense that any solution will also be a solution of the initial LP with the same objective value. If we consider the new variable to be $\mathbf{x}'=\left( \begin{array}{cc}
\mathbf{x}_1  \\
\mathbf{x}_2 \\
\end{array} \right)$
The constraint matrix that also incorporates the positiveness constraints is: 
$\mathbf{A}'=\left( \begin{array}{cc}
\mathbf{A} & -\mathbf{A} \\
-\mathbf{I} & \mathbf{0} \\
\mathbf{0} & -\mathbf{I} \end{array} \right)$, 
where $\mathbf{I}$ is an $n\times n$ identity matrix, and the objective function vector is 
$\mathbf{c}'^\intercal=\left( \begin{array}{c c}
\mathbf{c}^\intercal &  -\mathbf{c}^\intercal \end{array} \right)$,
while the right-hand side term is 
$\mathbf{b}'=\left( \begin{array}{c}
\mathbf{b}  \\
\mathbf{0}\\
\mathbf{0}  \end{array} \right)$.
The new constraint matrix $\mathbf{A}'$ does have full column rank, since the column vectors are now all independent thanks to the placement of the new identity matrices.  
We can now use the Simplex algorithm to find a solution. Let us denote the primal solution returned as $(\mathbf{x}^*_1,\mathbf{x}^*_2)$, while $\mathbf{y} ^\intercal = \left( \begin{array}{c c c} \mathbf{y} _1^\intercal & \mathbf{y} _2^\intercal & \mathbf{y} _3^\intercal  \end{array} \right)$ is the dual returned by the Simplex to verify the optimality of the primal solution. 
Let us write the conditions that should be verified by the dual, taking into account that $\mathbf{y}  \geq \mathbf{0}$:
\begin{align}
\begin{rcases}
\mathbf{y} _1^\intercal \mathbf{A} - \mathbf{y} _2^\intercal = \mathbf{c}^\intercal \hspace{0.1cm} & \Rightarrow  \hspace{0.1cm} \mathbf{y} _1^\intercal \mathbf{A} \geq \mathbf{c}^\intercal  \nonumber \\
\mathbf{y} _1^\intercal (-\mathbf{A}) - \mathbf{y} _3^\intercal = -\mathbf{c}^\intercal  \hspace{0.1cm} & \Rightarrow \hspace{0.1cm}  -\mathbf{y} _1^\intercal \mathbf{A} \geq -\mathbf{c}^\intercal  \nonumber 
\end{rcases}
\Rightarrow \mathbf{y} _1^\intercal \mathbf{A} = \mathbf{c}^\intercal.
\end{align}
We have just proven that $\mathbf{y}_1$ is dual feasible. Now the Simplex algorithm can check the condition of optimality for the primal solution by verifying that:

$\left( \begin{array}{c c}
\mathbf{c}^\intercal &  -\mathbf{c}^\intercal \end{array} \right)  \left( \begin{array}{c}
\mathbf{x}_1^* \\
\mathbf{x}_2^*\\ \end{array} \right)  = \mathbf{y} ^\intercal \mathbf{b}'=\left( \begin{array}{c c c} \mathbf{y} _1^\intercal & \mathbf{y} _2^\intercal & \mathbf{y} _3^\intercal  \end{array} \right) \left( \begin{array}{c}
\mathbf{b}  \\
\mathbf{0}\\
\mathbf{0}  \end{array} \right) = \mathbf{y} _1^\intercal \mathbf{b} $

And this proves the theorem, because we have found one possible primal feasible solution $\mathbf{x}_1^* - \mathbf{x}_2^*$ and one dual feasible solution $\mathbf{y}_1$ whose objective function values coincide.

}
\end{enumerate}
\end{proof}

\subsection{Proving optimality and infeasibility}

So far we have seen that there is a close relationship between the dual and primal problems and between the dual and primal optimum solutions. But what happens, for example, if the dual problem is infeasible?
Let us consider the following example:
\begin{figure}[H]
\begin{minipage}[b]{0.49\linewidth}
\centering
PRIMAL
\begin{align}
\max  \quad &x_1+2x_2+x_3 \nonumber \\
\st \quad &x_1+x_2 \leq 1 \nonumber \\
&x_1+x_3 \leq 4 \nonumber \\ \nonumber \\ \nonumber 
\end{align}
\end{minipage}
\begin{minipage}[b]{0.49\linewidth}
\centering
DUAL
\begin{align}
\min \quad &y_1+4y_2 \nonumber \\
\st \quad&y_1+y_2 = 1 \nonumber \\
&y_1 = 2 \nonumber\\
&y_2 = 1 \nonumber\\
&y_1,y_2 \geq 0 \nonumber
\end{align}
\end{minipage}
\end{figure}

If we check the primal problem carefully, we can identify $\mathbf{c}^\intercal=(1,2,1)$, 
$\mathbf{A}=\left( \begin{array}{ccc}
1 & 1 & 0 \\
1 & 0 & 1 \end{array} \right)$, 
and $\mathbf{b}=\left( \begin{array}{c}
1  \\
4  \end{array} \right)$,
and therefore the dual problem is defined as shown. Nonetheless, we can quickly see that the dual problem is infeasible, since the conditions set $y_1=2$ and $y_2=1$ which means $y_1+y_2$ will never be 1. 
An infeasible dual implies that we cannot determine a bound for the primal. 
If we take a closer look at the primal problem we see that it is, in fact, unbounded. For any $\alpha\geq 0$ that we choose, if we assign $\mathbf{x}=(-\alpha,\alpha,\alpha)$, the problem is feasible and the objective value is $2\alpha$, which means the objective function can be maximized to infinity, making the problem unbounded. 

We summarize the relationship between primal and dual problems in Table \ref{tab:primaldual}. 

\begin{table}[h]
\small
\begin {center}
  \begin{tabular}{ | c| c|c|c|} \hline
    {\bf Primal/Dual} & {Optimal} & {Unbounded} & {Infeasible}  \\ \hline
    Optimal & X & & \\ \hline
    Unbounded & & & X \\  \hline
    Infeasible & & X  & X \\ \hline
    \end{tabular}
  \end{center}
    \caption{Possible combinations of properties of the primal and dual problems.}
\label{tab:primaldual}
\end{table}

In the first case, by the strong duality Theorem \ref{th:strongdual}, if a primal has an optimal solution, the dual will also have an optimal solution. The second case is when the primal is unbounded. In that case, by the weak duality Theorem \ref{th:weakdual}, the dual problem is infeasible. 

We can then ask ourselves what would happen if we dualized the dual. It can be proven that the dual of the dual is a Linear Program that is equivalent to the primal. This proves that when the primal is infeasible, then the dual is unbounded. There is also another possible case, where both primal and dual are infeasible. 

Now to recap what the Simplex algorithm does: the algorithm returns a primal solution $\mathbf{x}^*$ and a dual $\mathbf{y}^*$, and the only thing that needs to be done to prove the optimality of the solution is to check whether the equality $\mathbf{c}^\intercal \mathbf{x}^* = \mathbf{b}^\intercal \mathbf{y}^*$ is fulfilled. 
The optimality proof is clear, and additionally, infeasibility can be proven by Farkas' lemma.

\begin{lemma}
{\bf Farka's Lemma.} A system of inequalities $\mathbf{A}\mathbf{x} \leq \mathbf{b}$ is infeasible if and only if there exists a vector $\boldsymbol\lambda$ with elements $\lambda_i \geq 0$ such that $\boldsymbol\lambda^\intercal \mathbf{A} = \mathbf{0}$ and $\boldsymbol\lambda^\intercal \mathbf{b} = \mathbf{-1}$.
\end{lemma} 

Intuitively, if we consider such a vector $\boldsymbol\lambda$ to exist, then the inequality $(\boldsymbol\lambda^\intercal \mathbf{A}) \mathbf{x} \leq \boldsymbol\lambda^\intercal \mathbf{b}$ would be valid, but since $\boldsymbol\lambda^\intercal \mathbf{A} = \mathbf{0}$ and $\boldsymbol\lambda^\intercal \mathbf{b} = \mathbf{-1}$, there would be no point $\mathbf{x} \in \mathbb{R}^n$ that could satisfy the inequality $\mathbf{A}\mathbf{x} \leq \mathbf{b}$, making the problem infeasible.

\section{Simplex in practice}
\label{sec:dictionaries}

So far, we have presented the theory behind Linear Programming and the Simplex algorithm, and a step-by-step explanation of the initialization and optimization phases that lead to the complete algorithm described in Algorithm \ref{alg:simplex_final}.
But how does Simplex work in practice? How can we implement a Simplex solver?

If we want to code a Simplex solver, we need, first of all, a convenient data structure for Linear Programs and their solutions. Such data structure is called a {\it Dictionary} of an LP. 

\subsection{Dictionaries}

A dictionary is a simple way to represent an LP. We can obtain it starting from the Standard Form 2 presented in Section \ref{sec:LPforms}, and performing a series of simple steps. 
Let us recall Standard Form 2 of an LP:
\begin{align*}
\max  \quad &\mathbf{c}^\intercal \mathbf{x} \\
\textrm{s.t.} \quad &\mathbf{A} \mathbf{x} \leq \mathbf{b}\\
&\mathbf{x}\geq \mathbf{0}. \nonumber
\end{align*}
The first thing we do it add {\it slack variables} on the constraint equations to convert them to equalities:
\begin{align*}
\max  \quad &\mathbf{c}^\intercal \mathbf{x} \\
\textrm{s.t.} \quad &\mathbf{A} \mathbf{x} + \textcolor{orange}{\mathbf{x_B}}= \mathbf{b}\\
&\mathbf{x}\geq \mathbf{0}. \nonumber
\end{align*}
Now we basically rearrange the equations into the following dictionary form:
 \begin{table*}[h]
 \begin{tabular}{c}
 $   \mathbf{x_B}= \mathbf{b} - \mathbf{A}\mathbf{x_I}$ \\[1.5ex] \hline \\[-1.5ex]
$z = c_0 + \mathbf{c}^\intercal \mathbf{x_I}$
\end{tabular}
    \end{table*}
    
which brings us to the dictionary form we will use throughout this section:
\begin{figure*}[h]
\begin{minipage}[b]{0.49\linewidth}
\end{minipage}
\begin{minipage}[b]{0.49\linewidth}
  \begin{align*}
   x_{B1} &=& &b_1& \quad& + & a_{11}x_{I1} \quad& + &\cdots \quad& + &a_{1n}x_{In}  \\
   x_{B2} &=& &b_2& \quad& + & a_{21}x_{I1} \quad& + &\cdots \quad& + &a_{2n}x_{In}   \\
   \vdots  \\
    x_{Bm} &= & &b_m& \quad& + &a_{m1}x_{I1} \quad& + &\cdots \quad& + &a_{mn}x_{In}  \\ \cline{1-12}
z &=& &c_0& \quad&+ &c_1 x_{I1} \quad& + & \cdots \quad& + &c_n x_{In} . 
    \end{align*}
\end{minipage}
\end{figure*}

The variables $\mathbf{x_B} = x_{B1}, \ldots, x_{Bm}$ are called {\it basic} variables, while $\mathbf{x_I} = x_{I1}, \ldots, x_{In}$ are called {\it non-basic} variables. 
There is a solution associated with each dictionary, which is obtained by setting all non-basic variables to zero and reading out the values of the basic variables from the equations in the dictionary. If all variables of the solution have values which respect the non-negativity constraints, the dictionary is said to be {\it feasible}.
The indices of the basic variables form the {\it basis} $B$ of our solution, a concept we presented in previous sections.
As we can see, there is already an advantage of representing LPs with dictionaries: by reading out the resulting values of basic variables, we directly obtain candidate solutions to the problem.

\begin{figure}[h]
\centering
\includegraphics[width=0.8\linewidth]{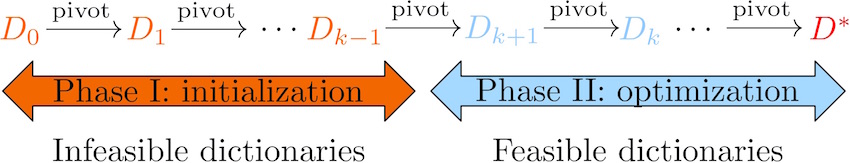} 
\caption[Overview of the Simplex method with dictionaries]{Overview of the Simplex method with dictionaries}
\label{fig:simplex_dict}
\end{figure}

Once we have a way to structure the data of our problem, we can define how the Simplex method works with dictionaries. An overview is presented in Figure~\ref{fig:simplex_dict}.

The main operation we perform on dictionaries is {\it pivoting}, which is a way to go from one dictionary to another.
As we mentioned before, Simplex is divided into two phases: 
\begin{itemize}
\item Phase I, or Initialization Phase: we start with an infeasible dictionary and pivot until we reach a feasible dictionary or determine the problem is infeasible. 
\item Phase II, or Optimization Phase: we optimize our feasible dictionary and our solution until we reach the optimum or determine the problem is unbounded.
\end{itemize}

In the following subsections, we describe both phases and how they work with dictionaries.

\subsection{Phase II: Pivoting}

The idea of the pivoting operation is, given a feasible initial dictionary, to obtain a new dictionary which has a corresponding solution with a higher objective value. 
Recall that the solution associated with a dictionary is represented by the basic variables. During pivoting, we consider whether inserting some of the non-basic variables to the basis would actually lead to an objective value increase. Of course, if a variable enters the basis, another variable has to leave it. But how do we choose the entering and leaving variables?

Let us consider the following example:
\begin{align*}
\max \quad & 5x_1 + 4x_2 + 3x_3  \\
\st \quad& 2x_1 + 3x_2 + x_3 \leq 5  \\
& 4x_1 + x_2 + 2x_3 \leq 11  \\
& 3x_1 + 4x_2 + 2x_3 \leq 8 \\
& x_1,x_2,x_3 \geq 0
\end{align*}

which has the following corresponding dictionary:
\begin{figure*}[h]
\begin{minipage}[b]{0.59\linewidth}
\end{minipage}
\begin{minipage}[b]{0.4\linewidth}
  \begin{align*}
   x_4 &=& &5& &-& 2&x_1& &-& 3&x_2& &-& &x_3&  \\
   x_5 &=& &11&&-& 4&x_1& &-& &x_2& &-& 2&x_3&   \\
   x_6 &=& &8& &-& 3&x_1& &-& 4&x_2& &-& 2&x_3&  \\ \cline{1-16}
z &=& &0& &+& 5&x_1& &+& 4&x_2& &+& 3&x_3&. 
    \end{align*}
\end{minipage}
\end{figure*}

We can immediately read the solution associated with this dictionary, which is $x_1=0,x_2=0,x_3=0,x_4=5,x_5=11,x_6=8$, and the corresponding objective value $z=0$.
Let us consider the possible {\it entering variables}. Remember that an entering variable should increase the objective value, therefore, we are looking for non-basic variables with positive coefficients $c_j>0$. The variable with the highest coefficient in our example is $x_1$. Ideally, we want to increase this variable as much as possible so as to increase the value $z$ as much as possible. In the current dictionary, $x_1$ has value $0$. Let us set $x_1=10$. What happens to the basic variables? 
If we look at the equations of the dictionary, we see that if $x_1=10,x_2=0,x_3=0$ then $x_4 = -15$, which violates the non-negativity constraints. Therefore, we can intuitively see that the basic variables will limit how much we can increase the entering variable. The basic variable that puts the tightest restriction will be chosen as {\it leaving variable}.

Let us see which variable is limiting the increase of the value of $x_1$. If we increase $x_1$, then variable $x_4$ will decrease since the coefficient associated with $x_1$ is negative. We can only increase $x_1$ up to $\frac{5}{2}$, in which case $x_4=0$. $x_5$ limits $x_1 \leq \frac{11}{4}$ and $x_6$ limits $x_1 \leq \frac{8}{3}$.
In this case, $x_4$ limits $x_1$ to the lowest value, and therefore will be chosen as leaving variable.

The next step is to modify the dictionary according to the entering and leaving variables. In order to do that, we first solve the equation of the leaving variable for the entering variable. In our example:
\begin{align*}
x_4 = 5 - 2x_1 - 3x_2 - x_3   \qquad \longrightarrow \qquad  x_1 = \frac{5}{2} - \frac{3}{2} x_2 - \frac{1}{2} x_3 - \frac{1}{2}x_4
\end{align*}
 Now we substitute $x_1$ by the obtained expression in all the other equations to obtain a new dictionary:
 \begin{figure*}[h]
\begin{minipage}[b]{0.59\linewidth}
\end{minipage}
\begin{minipage}[b]{0.4\linewidth}
  \begin{align*}
   x_1 &=& &\frac{5}{2}& &-& \frac{3}{2} &x_2& &-& \frac{1}{2} &x_3& &-& \frac{1}{2}&x_4&\\
   x_5 &=& &1& &+& 5&x_2& && && &+& 2&x_4&   \\
   x_6 &=& &\frac{1}{2}& &+& \frac{1}{2} &x_2& &-& \frac{1}{2} &x_3& &+& \frac{3}{2} &x_4&  \\ \cline{1-16}
z &=& &\frac{25}{2}& &-& \frac{7}{2}&x_2& &+& \frac{1}{2}&x_3& &-& \frac{5}{2}&x_4&. 
    \end{align*}
\end{minipage}
\end{figure*}

 We can read the new solution associated with this new dictionary, which is $x_1=\frac{5}{2},x_2=0,x_3=0,x_4=0,x_5=1,x_6=\frac{1}{2},z=\frac{25}{2}$. As we can see, pivoting has brought us to a new dictionary with higher objective value that the initial one. 
 
 We can pivot one more time, with entering variable $x_3$ and leaving variable $x_6$. Note that $x_5$ imposes no constraint on the increase of $x_3$ because they are not related by any equation. 
The new dictionary we will obtain after pivoting is:
 \begin{figure*}[h]
\begin{minipage}[b]{0.59\linewidth}
\end{minipage}
\begin{minipage}[b]{0.4\linewidth}
  \begin{align*}
  x_3 &=& &1& &+& &x_2& &+& 3&x_4& &-& 2&x_6&   \\
   x_1 &=& &2& &-& 2 &x_2& &-& 2 &x_4& &-& &x_6&\\
   x_6 &=& &1& &+& 5&x_2& &+&2&x_4&   \\ \cline{1-16}
z &=& &13& &-&3&x_2& &-&&x_4& &-&&x_6&. 
    \end{align*}
\end{minipage}
\end{figure*}

If we look at the last dictionary obtained, we see that all the coefficients $c_j \leq 0$, which means we do not have a choice for entering variable. There is no non-basic variable we can choose that will increase the value of the objective function, which means we have reached the optimum at $z=13$ with $x_1=2, x_2=0, x_3=1, x_4=0, x_5=1, x_6=0$.

In Algorithm \ref{alg:phase2} we present an overview of Simplex Phase II:

\begin{algorithm}
\caption{Phase II: optimization phase}          
\label{alg:phase2}                           
\begin{algorithmic}                    
\vspace{0.13cm}
\STATE {Input: a {\it feasible} dictionary $D$}
\vspace{0.13cm}
\WHILE{There exists an entering variable with the largest $c_j \geq 0$}
\vspace{0.13cm}
\STATE{Select a corresponding leaving variable with the lowest $\frac{b_{i}}{-a_{ij}}$ that limits the value of the entering variables $x_{Ij} \leq \frac{b_{i}}{-a_{ij}}$}
\vspace{0.13cm}
\IF{ There exists no leaving variable. }
\vspace{0.13cm}
\STATE {The problem is unbounded.  }
\vspace{0.13cm}
\ELSE
\vspace{0.13cm}
\STATE{ Perform pivoting to obtain $D'$.}
\vspace{0.13cm}
\ENDIF
\vspace{0.13cm}
\ENDWHILE
\vspace{0.13cm}
\STATE{Return dictionary $D'$ as final.}
\end{algorithmic}
\end{algorithm}

In the following subsection, we will discuss the feasibility of dictionaries after pivoting as well as degenerate dictionaries. Most importantly, we will discuss how to identify unbounded problems.

\subsubsection{Proving feasibility}

One could ask if by pivoting we will always end up with a feasible dictionary. In other words, does pivoting maintain feasibility of the dictionaries? Here is a small proof. Let us again consider the general dictionary:
  \begin{align*}
   x_{B1} &=& &b_1& & + & a_{11}x_{I1} \quad& + &\cdots \quad  &+& \textcolor{blue}{a_{1j}}x_{Ij}  \quad& + &\cdots \quad& + &a_{1n}x_{In}   &\ \rightarrow\  x_{Ij} \leq \frac{b_1}{-a_{1j}}&\\
   x_{B2} &=& &b_2& & + & a_{21}x_{I1} \quad& + &\cdots \quad &+& \textcolor{blue}{a_{2j}}x_{Ij}  \quad& + &\cdots \quad& + &a_{2n}x_{In}   &\ \rightarrow\   x_{Ij} \leq \infty&\\
   \vdots  \\
     \textcolor{orange}{x_{Bi}} &= & &\textcolor{orange}{b_i}& & + &\textcolor{orange}{a_{i1}}x_{I1} \quad& + &\cdots \quad &+& \textcolor{orange}{\mathbf{ a_{ij}}}x_{Ij}  \quad& + &\cdots \quad& + &\textcolor{orange}{a_{in}}x_{In}  & \ \rightarrow\  \textcolor{orange}{x_{Ij}} \textcolor{orange}{\leq} \textcolor{orange}{\frac{b_i}{-a_{ij}}}& \\
      \vdots  \\
    x_{Bm} &= & &b_m& & + &a_{m1}x_{I1} \quad& + &\cdots \quad &+& \textcolor{blue}{a_{mj}}x_{Ij}  \quad& + &\cdots \quad & + &a_{mn}x_{In}  &\ \rightarrow\   x_{Ij} \leq \frac{b_m}{-a_{mj}}&\\ \cline{1-16}
z &=& &c_0& &+ &c_1 x_{I1} \quad& + & \cdots \quad   &+& \textcolor{blue}{c_j} x_{Ij}  \quad& + & \cdots \quad& + &c_n x_{In} 
    \end{align*}
where $x_{Ij}$ is the entering variable and $x_{Bi}$ the leaving variable. Let us now analyze what happens with variable $x_{B1}$, assuming it is not the leaving variable. 
Given the new entering variable $x_{Ij}$, it will be assigned a value of $x_{B1} = b_1 + a_{1j} \left( \frac{b_i}{-a_{ij}} \right)$. In order for the new dictionary to be feasible, $x_{B1} \geq 0$. Can we prove that it will never be negative? 
Firstly, we can see that $b_1 \geq 0$, because the current dictionary is feasible, \ie $x_{B1} = b_1$ when all non-basic variables are set to zero. Secondly, we know $a_{ij} < 0$, otherwise $x_{Bi}$ would not be the leaving variable related to the entering variable $x_{Ij}$, because it would not constraint the increase of $x_{Ij}$.
The only thing we need to determine now, is the value of $a_{1j}$, for which we have two possibilities:
\begin{itemize}
\item $a_{1j} \geq 0$, we can directly determine  that $x_{B1} \geq 0$.
\item $a_{1j} < 0 $, we cannot directly determine if $x_{B1}$ will be nonnegative, but we do know that $\frac{b_i}{- a_{ij}} \leq \frac{b_1}{- a_{1j}}$, otherwise $x_{B1}$ would be the leaving variable. From this, we can derive: 
\begin{align*}
\frac{b_i}{- a_{ij}} \leq \frac{b_1}{- a_{1j}} \quad \longrightarrow \quad a_{1j} \frac{b_i}{- a_{ij}} \geq - b_1 \quad \longrightarrow \quad b_1 + a_{1j} \frac{b_i}{- a_{ij}} \geq 0.
\end{align*}
\end{itemize}

\subsubsection{Degeneracy}

We have established that the pivoting operation maintains feasibility. The only question we need to answer now is what happens with the value of the objective function during pivoting.
We know that the entering variable will take value $x_{Ij}=\frac{b_i}{- a_{ij}}$ after pivoting, while the leaving variables will take value $x_{Bi}=0$. Given that all other non-basic variables will remain zero, the objective value of the new dictionary will be:
\begin{align*}
z = c_0 + c_j \left(\frac{b_i}{- a_{ij}} \right).
\end{align*}
We know $c_j > 0$, otherwise $x_{Ij}$ would not be the entering variable. On the other hand, $a_{ij} < 0$, otherwise $x_{Bi}$ would not be the leaving variable. If $b_1 > 0$, it would mean the objective value $z$ can only increase, but we know that it can be that $b_1=0$, in which case the objective value would remain constant. A dictionary with these characteristics is called a {\it degenerate} dictionary. We can see an example below:
 \begin{figure*}[h]
\begin{minipage}[b]{0.59\linewidth}
\end{minipage}
\begin{minipage}[b]{0.4\linewidth}
  \begin{align*}
  x_3 &=& &\frac{1}{2}&  && && && && &+& \frac{1}{2}&x_4&   \\
   x_5 &=& &0& &-& 2 &x_1& &+&4 &x_2& &+&3 &x_4&\\
   x_6 &=& &0& &+& &x_1& &-&3&x_2& &+& 2&x_4&  \\ \cline{1-16}
z &=& &4& &+&2&x_1& &-&&x_2& &-&4&x_4&. 
    \end{align*}
\end{minipage}
\end{figure*}

In this case, pivoting would bring us from one dictionary to another without ever increasing the objective value, which means the algorithm would cycle and not terminate. In order to avoid cycling, we can apply Bland's rule as explained in Section \ref{sec:bland}.

\subsubsection{Unbounded problems}

There is only one case that needs to be analyzed by the Phase II algorithm, and that is the case of unbounded LPs. Let us consider the following dictionary:
 \begin{figure*}[h]
\begin{minipage}[b]{0.59\linewidth}
\end{minipage}
\begin{minipage}[b]{0.4\linewidth}
  \begin{align*}
  x_4 &=& &5&  &-& &x_1& &+& &x_2&    \\
   x_5 &=& &6& &+& &x_1& && && &-& &x_3&\\
   x_6 &=& &2& &+& 2&x_1& && && &-& &x_3&  \\ 
   x_7 &=& &4& &+& &x_1& &-& &x_2& \\ \cline{1-16}
z &=& &0& &+&2&x_1& &+&3&x_2& &-&5&x_3&. 
    \end{align*}
\end{minipage}
\end{figure*}

At first glance, we cannot say if the problem is unbounded or not, so we just start pivoting. We choose $x_2$ as entering variable, which means $x_7$ is the leaving variable. The new dictionary we obtain is therefore:\\ \\

 \begin{figure*}[h]
\begin{minipage}[b]{0.59\linewidth}
\end{minipage}
\begin{minipage}[b]{0.4\linewidth}
\begin{align*}
  x_2 &=& &4& &+& &x_1& &-& &x_7& \\ 
  x_4 &=& &9&    \\
   x_5 &=& &6& &+& &x_1& && && &-& &x_3&\\
   x_6 &=& &2& &+& 2&x_1& && && &-& &x_3&  \\ \cline{1-16}
z &=& &12& &+&5&x_1& &-&3&x_7& &-&5&x_3&. 
    \end{align*}
    \end{minipage}
\end{figure*}

The new entering variable of this dictionary should be $x_1$, but let us look at what happens with the leaving variable. Remember that the leaving variable should limit the increase of $x_1$, but in this case, when we increase $x_1$, $x_2, x_5$ and $x_6$ all increase without limits, and $x_4$ does not depend on $x_1$. This means that we could arbitrarily increase $x_1$ and the non-negativity constraints would still be respected. 
When we cannot find any leaving variable, we can conclude that the problem is unbounded. Alternatively, we can say that the problem is unbounded when all entries of the column corresponding to the entering variable are nonnegative.

\subsection{Phase I: Initialization}

Up to now we described how to solve an LP given an initial feasible dictionary. Let us now consider the following LP
\begin{align*}
\max \quad & x_1 + 2x_2   \\
\st \quad& -2x_1 + x_2 \leq -2  \\
& x_2 \leq 4  \\
& x_1-2x_2 \leq -2 \\
& x_1 \leq 4 \\
& x_1,x_2 \geq 0
\end{align*}
and corresponding dictionary
 \begin{figure*}[h]
\begin{minipage}[b]{0.59\linewidth}
\end{minipage}
\begin{minipage}[b]{0.4\linewidth}
  \begin{align*}
  x_3 &=& -&2& &+& 2&x_1& &-& &x_2& \\ 
  x_4 &=& &4&   && && &-& &x_2& \\
   x_5 &=& -&2& &-& &x_1& &+& 2&x_2& \\
   x_6 &=& &4& &-& &x_1&  \\ \cline{1-12}
z &=& &0& &+& &x_1& &+&2&x_2&. 
    \end{align*}
\end{minipage}
\end{figure*}

\begin{figure*}[bth]
\centering
\includegraphics[height=5cm]{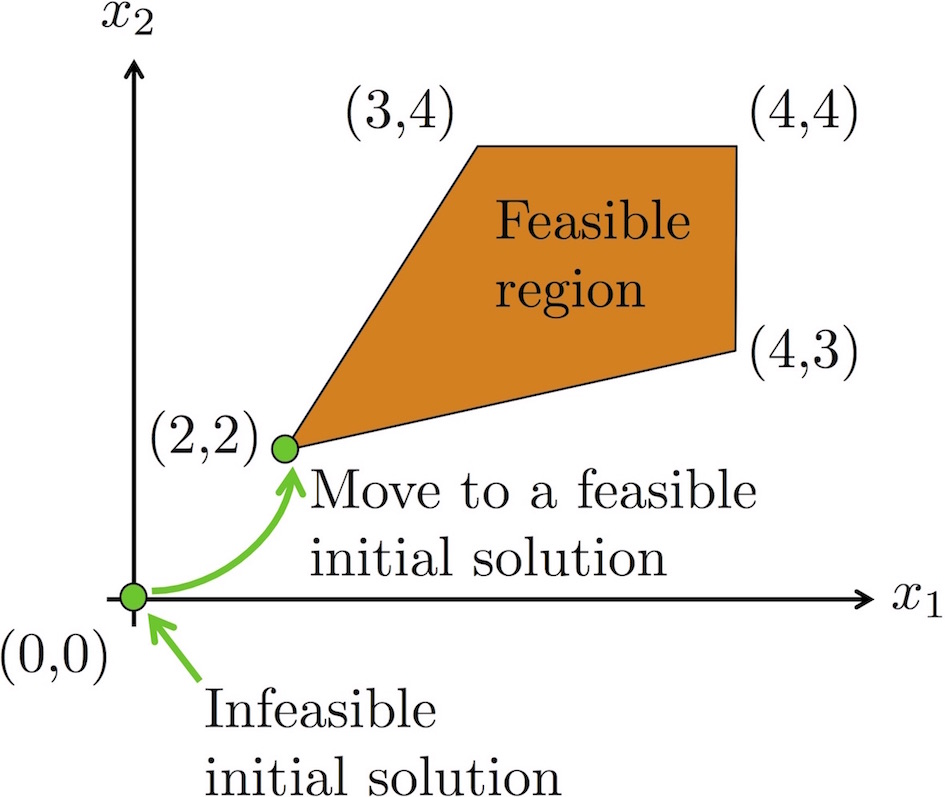} 
\caption[Infeasible initial solution]{Feasible region of the LP depicted in orange. The solution associated with the initial dictionary (0,0) is not feasible. During initialization, we move towards a feasible initial solution, (2,2) in this example.}
\label{fig:infeasible}
\end{figure*}

If we analyze the corresponding solution $x_1=0,x_2=0,x_3=-2;x_4=4,x_5=-2,x_6=4$, we see that variables $x_3$ and $x_5$ do not respect the non-negativity constraints, and therefore the initial solution is infeasible. In Figure~\ref{fig:infeasible}, we plot the feasible region of the LP in orange. As we can see, the solution associated with the dictionary, $x_1=0,x_2=0$ is outside of the feasible region.

The question is, what do we do when the initial dictionary is infeasible? The strategy we follow is to slightly modify the initial LP and create the auxiliary problem. We then perform a pivoting on the auxiliary problem, and given the solution we find, we can draw a conclusion about the feasibility of the original LP.

Let us first describe how to construct the auxiliary problem. If we look at the previous LP, we can see that the reason why the solution was not feasible, was that $x_3$ and $x_5$ had negative values. We could make those values positive by adding a certain quantity $x_0$, which would have to be at least $2$.  
Let us forget about the objective function for a moment and focus on the new constraints created by adding the variable $x_0$. They would look like:
\begin{align*}
& -2x_1 + x_2 \leq -2 + \textcolor{orange}{x_0} \\
& x_2 \leq 4  + \textcolor{orange}{x_0}\\
& x_1-2x_2 \leq -2 + \textcolor{orange}{x_0} \\
& x_1 \leq 4 + \textcolor{orange}{x_0}\\
& x_1,x_2, \textcolor{orange}{x_0} \geq 0
\end{align*}
If we set $x_0=2$, the new problem would have solution $x_1=0,x_2=0,x_3=0;x_4=6,x_5=0,x_6=6$, which is feasible. In fact, one can prove that the auxiliary problem will always be feasible.
Intuitively, we can also see that if the initial problem is feasible, then $x_0=0$, and the solution of the auxiliary problem will correspond to the solution of the initial LP. Nonetheless, if the initial problem is infeasible, then $x_0>0$. This is the intuition behind Phase I of the Simplex. We are going to work with the new constraints and the auxiliary problem's objective will be to find a minimum for $x_0$. If the final solution of the auxiliary problem is $x_0=0$, we will conclude that the original problem is feasible. An initial solution to the original problem will be obtained so we can then start with Phase II of Simplex.
Formally, the auxiliary problem has the following form:
\begin{align*}
\max  \quad &- \textcolor{orange}{x_0} \\
\textrm{s.t.} \quad &\mathbf{A} \mathbf{x} + \textcolor{blue}{\mathbf{x_s}} -  \textcolor{orange}{x_0 \mathbf{1}} = \mathbf{b}\\
&\mathbf{x},  \textcolor{blue}{\mathbf{x_s}} \geq \mathbf{0} \\
& \textcolor{orange}{x_0} \geq {0},
\end{align*}
where $\textcolor{blue}{\mathbf{x_s}}$ is the slack variable vector we added to form the dictionary, and $\textcolor{orange}{x_0}$ is the new variable we use to create the auxiliary problem. 
Before, we said that the auxiliary problem is always feasible, and we can see that by looking at the solution associated with the initial dictionary of the auxiliary problem:
\begin{align*}
\mathbf{x}= \mathbf{0}, \textcolor{orange}{x_0} = - \min (\mathbf{b},0), \textcolor{blue}{\mathbf{x_s}} = \mathbf{b} + x_0 \mathbf{1}.
\end{align*}
The value of $x_0$ is chosen so as to make the problem feasible, and therefore it must bring all the variables at least up to zero. In our previous example $ b_{\textrm{min}}=-2$ and therefore $x_0=2$ in order to make $x_3$ and $x_5$ nonnegative. If we take a look at the slack variables, we can see that $x_{s}^i=b_i - b_{\textrm{min}}$. Since by definition $b_i \geq b_{\textrm{min}}$, we know that the slack variables will also be nonnegative, and therefore the auxiliary problem will always be feasible.

Coming back to our example, this is the complete auxiliary problem corresponding to the previous LP:
\begin{align*}
\max \quad & - x_0   \\
\st \quad& -2x_1 + x_2 \leq -2 + {x_0} \\
& x_2 \leq 4  + {x_0}\\
& x_1-2x_2 \leq -2 + {x_0} \\
& x_1 \leq 4 + {x_0}\\
& x_1,x_2,{x_0} \geq 0
\end{align*}

Let us now construct the initial dictionary for the auxiliary problem:\\ \\ \\ \\

 \begin{figure*}[h]
\begin{minipage}[b]{0.59\linewidth}
\end{minipage}
\begin{minipage}[b]{0.4\linewidth}
  \begin{align*}
  x_3 &=& -&2& &+& &x_0& &+&  2&x_1& &-& &x_2& \\ 
  x_4 &=& &4&  &+& &x_0& && && &-& &x_2& \\
   x_5 &=& -&2& &+& &x_0&&-& &x_1& &+& 2&x_2& \\
   x_6 &=& &4& &+& &x_0& &-& &x_1&  \\ \cline{1-16}
w &=& &0& &-& &x_0&. 
    \end{align*}
\end{minipage}
\end{figure*}

The pivoting of the auxiliary dictionary has a couple of special rules that we need to follow: (i) the initial move will always be to make $x_0$ the entering variable, and the leaving variable will be the one with the least value $b_i$, (ii) whenever $x_0$ is one of the possible leaving variables, preferentially choose it. 
In our example, if $x_0$ enters, the leaving variable can either be $x_3$ or $x_5$. We choose $x_5$ and obtain the following dictionary:
 \begin{figure*}[h]
\begin{minipage}[b]{0.59\linewidth}
\end{minipage}
\begin{minipage}[b]{0.4\linewidth}
  \begin{align*}
  x_0 &=& &2& &+& &x_1& &-&  2&x_2& &+& &x_5& \\ 
  x_3 &=& &0&  &+& 3 &x_1& &-& 3&x_2& &+& &x_5& \\
   x_4 &=& &6& &+& &x_1&&-& 3&x_2& &+& &x_5& \\
   x_6 &=& &6& && && &-& 2&x_2& &+& &x_5&  \\ \cline{1-16}
w &=& -&2& &-& &x_1& &+& 2&x_2& &-& &x_5&. 
    \end{align*}
\end{minipage}
\end{figure*}

The next pivot is done with $x_2$ as entering and $x_3$ as leaving variable, leading to the following dictionary:
 \begin{figure*}[h]
\begin{minipage}[b]{0.59\linewidth}
\end{minipage}
\begin{minipage}[b]{0.4\linewidth}
  \begin{align*}
  x_0 &=& &2& &-& &x_1& &+&  \frac{2}{3}&x_3& &+& \frac{1}{3}&x_5& \\ 
  x_2 &=& &0&  &+& &x_1& &-& \frac{1}{3}&x_3& &+& \frac{1}{3}&x_5& \\
   x_4 &=& &6& &-& 2&x_1&&+& &x_3&  \\
   x_6 &=& &6& &-& 2&x_1& &+& \frac{2}{3}&x_3& &+& \frac{1}{3}&x_5&  \\ \cline{1-16}
w &=& -&2& &+& &x_1& &-& \frac{2}{3}&x_3& &-& \frac{1}{3}&x_5&. 
    \end{align*}
\end{minipage}
\end{figure*}

Finally, after choosing $x_1$ as entering variable, we see that $x_0$ is the leaving variable, which leads to the final dictionary of the auxiliary problem:\\ \\
 \begin{figure*}[h]
\begin{minipage}[b]{0.59\linewidth}
\end{minipage}
\begin{minipage}[b]{0.4\linewidth}
  \begin{align*}
  x_1 &=& &2& &-& &x_0& &+&  \frac{2}{3}&x_3& &+& \frac{1}{3}&x_5& \\ 
  x_2 &=& &2&  &-& &x_0& &+& \frac{1}{3}&x_3& &+& \frac{2}{3}&x_5& \\
   x_4 &=& &2& &+& 2&x_0&&-& \frac{1}{3}&x_3& &-& \frac{2}{3}&x_5& \\
   x_6 &=& &2& &+& 2&x_0& &-& \frac{2}{3}&x_3& &-& \frac{1}{3}&x_5&  \\ \cline{1-16}
w &=& &0& &-& &x_0&. 
    \end{align*}
\end{minipage}
\end{figure*}

As we can see, the solution associated with the final auxiliary dictionary is $x_0=0,x_1=2,x_2=2,x_3=0,x_4=2,x_5=0,x_6=2$ and the final objective value is $w=0$, which means that the original LP is feasible. As we can see, the point $x_1=2,x_2=2$ is inside the feasible region depicted in Figure~\ref{fig:infeasible}.

The question now is, how do we construct a feasible dictionary for the original LP, so we can start Phase II of Simplex? The answer is simple, we just eliminate $x_0$ from the constraints and rewrite the objective function $z$ with respect to the new non-basic variables. 
Here is the resulting dictionary:

 \begin{figure*}[h]
\begin{minipage}[b]{0.59\linewidth}
\end{minipage}
\begin{minipage}[b]{0.4\linewidth}
  \begin{align*}
  x_1 &=& &2& &+& \frac{2}{3}&x_3& &+&  \frac{1}{3}&x_5&  \\ 
  x_2 &=& &2& &+& \frac{1}{3}&x_3& &+& \frac{2}{3}&x_5& \\
   x_4 &=& &2& &-& \frac{1}{3}&x_3& &-& \frac{2}{3}&x_5&  \\
   x_6 &=& &2&  &-& \frac{2}{3}&x_3& &-& \frac{1}{3}&x_5&  \\ \cline{1-12}
z &=& &6&  &+& \frac{4}{3}&x_3& &+& \frac{5}{3}&x_5&. 
    \end{align*}
\end{minipage}
\end{figure*}

Recall that the original objective function was $z=x_1+2x_2$, which we just rewrite by substituting $x_1$ and $x_2$.
The solution associated with this dictionary is $x_1=2,x_2=2,x_3=0,x_4=2,x_5=0,x_6=2$, which is a feasible solution. Therefore, now we can use this dictionary to start Phase II of the Simplex algorithm to find the optimal solution.

\subsubsection{Infeasible problems}

In this section we just want to present an infeasible problem, and how the auxiliary problem helps us determine its infeasibility. Let us consider the following LP:
\begin{align*}
\max \quad & 2x_1 - 3x_2   \\
\st \quad& -x1 + x_2 \leq -3 \\
& 2x_1 + x_2 \leq 10\\
& x_1 - 2x_2 \leq -2 \\
& x_1,x_2 \geq 0
\end{align*}
with associated initial dictionary:

 \begin{figure*}[h]
\begin{minipage}[b]{0.59\linewidth}
\end{minipage}
\begin{minipage}[b]{0.4\linewidth}
  \begin{align*}
  x_3 &=& -&3& &+& &x_1& &-& &x_2& \\
   x_4 &=& &10& &-& 2&x_1& &-& &x_2&  \\
   x_5 &=& -&2&  &-& &x_1& &+& 2&x_2&  \\ \cline{1-12}
z &=& &0&  &+& 2&x_1& &-& 3&x_2&. 
    \end{align*}
\end{minipage}
\end{figure*}

The solution associated with this dictionary is $x_1=0,x_2=0,x_3=-3,x_4=10,x_5=-2$, which is infeasible. We therefore start Phase I of the Simplex algorithm by constructing the auxiliary problem:

 \begin{figure*}[h]
\begin{minipage}[b]{0.59\linewidth}
\end{minipage}
\begin{minipage}[b]{0.4\linewidth}
  \begin{align*}
  x_3 &=& -&3& &+& &x_0& &+& &x_1& &-& &x_2& \\
   x_4 &=& &10& &+& &x_0& &-& 2&x_1& &-& &x_2&  \\
   x_5 &=& -&2& &+& &x_0& &-& &x_1& &+& 2&x_2&  \\ \cline{1-16}
w &=& &0& &-& &x_0&. 
    \end{align*}
\end{minipage}
\end{figure*}

We start the pivoting for auxiliary problems by choosing $x_0$ as entering variable and $x_3$ as leaving, obtaining the following dictionary:

 \begin{figure*}[h]
\begin{minipage}[b]{0.59\linewidth}
\end{minipage}
\begin{minipage}[b]{0.4\linewidth}
  \begin{align*}
  x_0 &=& &3& &-& &x_1& &+& &x_2& &+& &x_3& \\
   x_4 &=& &13& &-& 3&x_1& && && &+& &x_3&  \\
   x_5 &=& &1& &-& 2&x_1& &+& 3&x_2& &+& &x_3&  \\ \cline{1-16}
w &=& -&3& &+& &x_1&&-& &x_2& &-& &x_3& . 
    \end{align*}
\end{minipage}
\end{figure*}

We continue pivoting by making $x_1$ enter and $x_5$ leave the basis:\\

 \begin{figure*}[h]
\begin{minipage}[b]{0.59\linewidth}
\end{minipage}
\begin{minipage}[b]{0.4\linewidth}
  \begin{align*}
  x_0 &=& &\frac{5}{2}& &-& \frac{1}{2}&x_2& &+& \frac{1}{2}&x_3& &+& \frac{1}{2}&x_5& \\
   x_1 &=& &\frac{1}{2}& &+& \frac{3}{2}&x_2& &+& \frac{1}{2}&x_3& &-& \frac{1}{2}&x_5&  \\
   x_4 &=& &\frac{23}{2}& &-& \frac{9}{2}&x_2& &-& \frac{1}{2}&x_3& &+& \frac{3}{2}&x_5&  \\ \cline{1-16}
w &=& -&\frac{5}{2}& &+& \frac{1}{2}&x_2&&-& \frac{1}{2}&x_3& &-& \frac{1}{2}&x_5& . 
    \end{align*}
\end{minipage}
\end{figure*}

Next, we make $x_2$ enter and $x_4$ leave:

 \begin{figure*}[h]
\begin{minipage}[b]{0.59\linewidth}
\end{minipage}
\begin{minipage}[b]{0.4\linewidth}
  \begin{align*}
  x_0 &=& &\frac{11}{9}& &+& \frac{5}{9}&x_3& &+& \frac{1}{9}&x_4& &+& \frac{1}{3}&x_5& \\
   x_1 &=& &\frac{13}{3}& &+& \frac{1}{3}&x_3& &-& \frac{1}{3}&x_4&  \\
   x_2 &=& &\frac{23}{9}& &-& \frac{1}{9}&x_3& &-& \frac{2}{9}&x_4& &+& \frac{1}{3}&x_5&  \\ \cline{1-16}
w &=& -&\frac{11}{9}& &-& \frac{5}{9}&x_3&&-& \frac{1}{9}&x_4& &-& \frac{1}{3}&x_5& . 
    \end{align*}
\end{minipage}
\end{figure*}

Once we reach this dictionary, we see that there are no possible entering variables, and we reach a solution where $x_0 = \frac{11}{9} > 0$. We can only conclude that the original problem is infeasible. 

If we take another look at the constraints of the original problem, and we sum the first and the third constraints, we get:
  \begin{align*}
  -x_1 + x_2 + x_1 - 2x_2 \leq -3 -2 \quad \longrightarrow \quad -x_2 \leq -5 \quad \longrightarrow \quad x_2 \geq 5.
    \end{align*}
And if we take the second equation and subtract the first one, we get:
  \begin{align*}
 2x_1 + x_2 + x_1 - x_2 \leq 10 + 3 \quad \longrightarrow \quad 3 x_1 \leq 13 \quad \longrightarrow \quad x_1 \leq \frac{13}{3}.
    \end{align*}
    If we now substitute the two variables into the first equation:
     \begin{align*}
- \frac{13}{3} + 5 = \frac{2}{3} \leq -3 
    \end{align*} 
$x_2$ has a value of $5$ or larger, while $x_1$ cannot be larger than $\frac{13}{3}$, which means that the minimum value of the first constraint $-x_1 + x_2$ will be $\frac{2}{3}$, which is clearly larger than $-3$. Therefore, we can see that the problem is, indeed, infeasible.

\section{Graph model representation}
\label{graphicalmodel}

Now that we have a clear definition of Linear Programs and its important properties, and we know how to solve a Linear Program with Simplex, we move towards the graphical model of a polyhedron. Going from LP representations to graphical model representations and vice versa is certainly useful since, for example, there will be certain LPs which will be solved faster by using network flow solvers like $k$-shortest paths. 

\begin{figure*}[bth]
\centering
\subfigure[Undirected graph]{
\includegraphics[height=3cm]{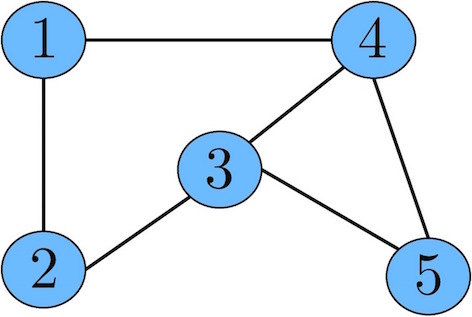} 
\label{fig:undirgraph}
}
\subfigure[Walk (green) and path (orange)]{
\includegraphics[height=3cm]{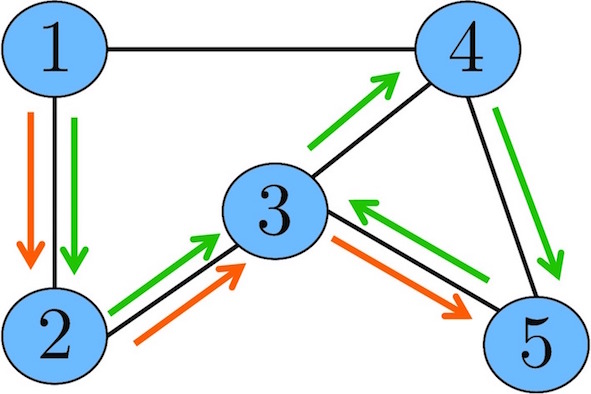} 
\label{fig:walkpath}
}
\subfigure[Diameter]{
\includegraphics[height=3cm]{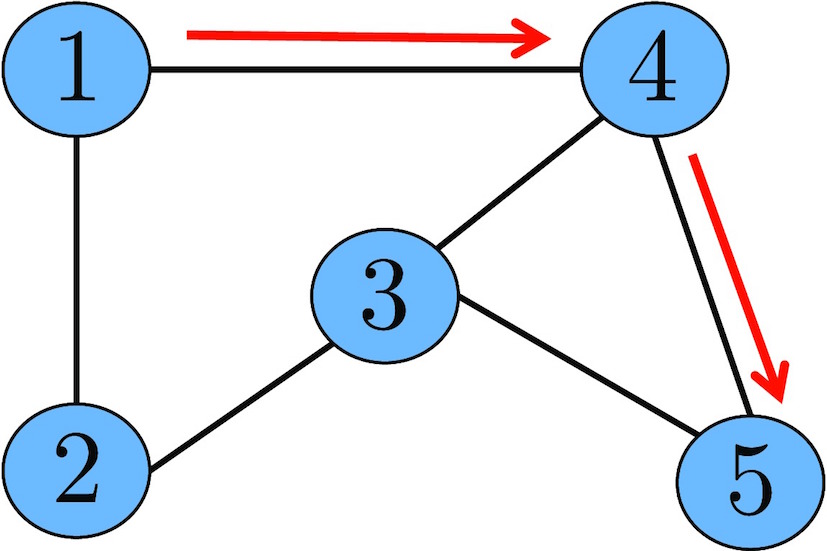} 
\label{fig:diameter}
}
\caption{Basic concepts of graph theory}
\label{fig:graphtheory}
\end{figure*}

An {\it undirected graph} $G=(V,E)$ consists of a finite set $V$ of {\it nodes} or {\it vertices} and a set $E$ of {\it edges}, where each edge $e \in E$ is a two-element subset of vertices $e=(u,v)$, where $u \neq v \in V$.
An example of such a graph is shown in Figure~\ref{fig:undirgraph}, where $V=\{ 1,2,3,4,5\}$ and $E=\left\{ (1,2), (1,4) , (2,3) , (3,4) , (3,5) , (4,5)  \right\}$.

A {\it walk} from node $i_1 \in V$ to $i_t \in V$ is a sequence $i_1,i_2, \ldots, i_t$ of nodes such that $(i_k,i_{k+1}) \in E$ for $k=1, \ldots, t-1$. A walk is called a {\it path} if it has no repeated nodes. In Figure~\ref{fig:walkpath} we show an example of a walk $1,2,3,4,5,3$ in green; note that it is not a path since node 3 is repeated. We depict a path $1,2,3,5$ in orange.

The {\it distance} between $u,v \in V$ is the smallest $t$ such that there exists a path $i_1, \ldots, i_t$ in $G$ with $i_1=u$ and $i_t=v$. The {\it diameter} of $G$ is the largest distance between two nodes of $G$. In Figure~\ref{fig:diameter} we show a case, where the longest distance between any two nodes in the graph is 2, therefore the diameter of the graph is 2.

Now that we have defined some basic concepts of graphs, we can proceed by representing polyhedra used to describe Linear Programs as graphs.

\begin{figure*}[h]
\centering
\subfigure[Polyhedron $P$]{
{\makebox[5cm][c]{\includegraphics[height=3.2cm]{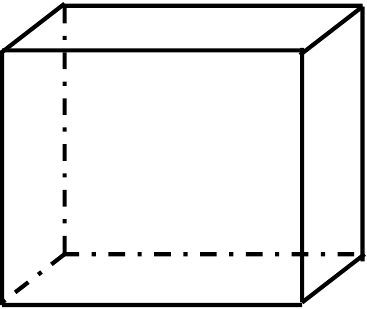}}} 
\label{fig:polyhedra}
}\qquad 
\subfigure[Graph representation of $P$, vertices in orange, edges in blue.]{
{\makebox[5cm][c]{\includegraphics[height=3.2cm]{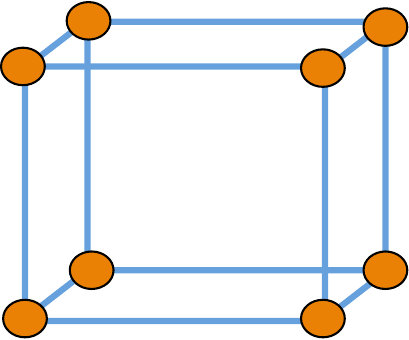}}} 
\label{fig:polygraph}
}
\caption{Conversion from polyhedron to graph}
\label{fig:polyhedratograph}
\end{figure*}

A polyhedron $P=\{\mathbf{x} \in \mathbb{R}^n : \mathbf{A} \mathbf{x} \leq \mathbf{b}\}$ with vertices defines a graph $G_P=(V,E)$ as follows. The set of nodes $V$ is the set of vertices of $P$ and $(v_1,v_2) \in E \iff v_1$ and $v_2$ are adjacent in $P$ (see Figure~\ref{fig:polyhedratograph}).

The diameter of $G_P$ is the diameter of $P$. If a version of the Simplex algorithm requires only a polynomial number of iterations (in both $n$ and $m$), then the diameter of each polyhedral graph is polynomial.

Considering the previously defined Linear Program $\max \{ \mathbf{c}^\intercal\mathbf{x} : \mathbf{x} \in \mathbb{R}^n , \mathbf{A} \mathbf{x} \leq \mathbf{b}\}$, we know that the Simplex algorithm walks along the edges of a graph $G_P$ of $P=\{\mathbf{x} \in \mathbb{R}^n : \mathbf{A} \mathbf{x} \leq \mathbf{b}\}$.
The question we asked ourselves in previous sections was whether there is a version of Simplex requiring a polynomial number of iterations. A necessary condition for this is that the diameter of $G_P$ must be polynomial. We define a new variable $\Delta(n,m)$ as the diameter of a graph $G_P$ of a polyhedron $P \subseteq \mathbb{R}^n$ described by $m$ inequalities.
The best bound for $\Delta(n,m)$ found so far was presented in 1992 by Kalai and Kleitman \citep{kalai1992} and it is defined as $\Delta(n,m) \leq m^{1+\log n}$. This bound belongs to a family of functions called quasi polynomial, which grow much slower than exponential functions, but not as slow as polynomial functions. 

Now we introduce an important property of a graph $G_P$ related to polyedron $P=\{\mathbf{x} \in \mathbb{R}^n : \mathbf{A} \mathbf{x} \leq \mathbf{b}\}$, which is that $G_P$ is {\it connected}. Furthermore, for each pair of vertices $u,v$ there exists a path connecting $u$ and $v$ such that each inequality of $\mathbf{A} \mathbf{x} \leq \mathbf{b}$ active at both $u$ and $v$ is also active at each vertex of that path.   

\begin{figure*}[h]
\centering
\includegraphics[height=3.2cm]{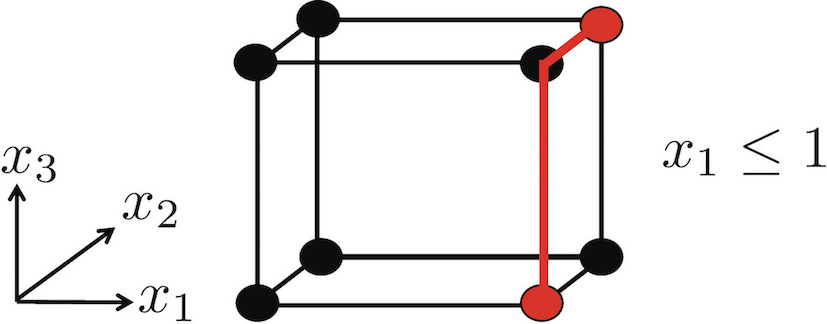} 
\caption[Connected graph]{Path between $u$ and $v$ that satisfies the inequality active at both vertices.}
\label{fig:connectedgraph}
\end{figure*}

Let us look at the example shown in Figure~\ref{fig:connectedgraph}, where we have our graph $G_P$ drawn in black and two vertices $u$ and $v$ marked in red. Inequality $x_1 \leq 1$ is active at both vertices and there is a path (marked in red) along which this constraint is also active. 

\begin{figure*}[ht]
\centering
\includegraphics[width=0.7\linewidth]{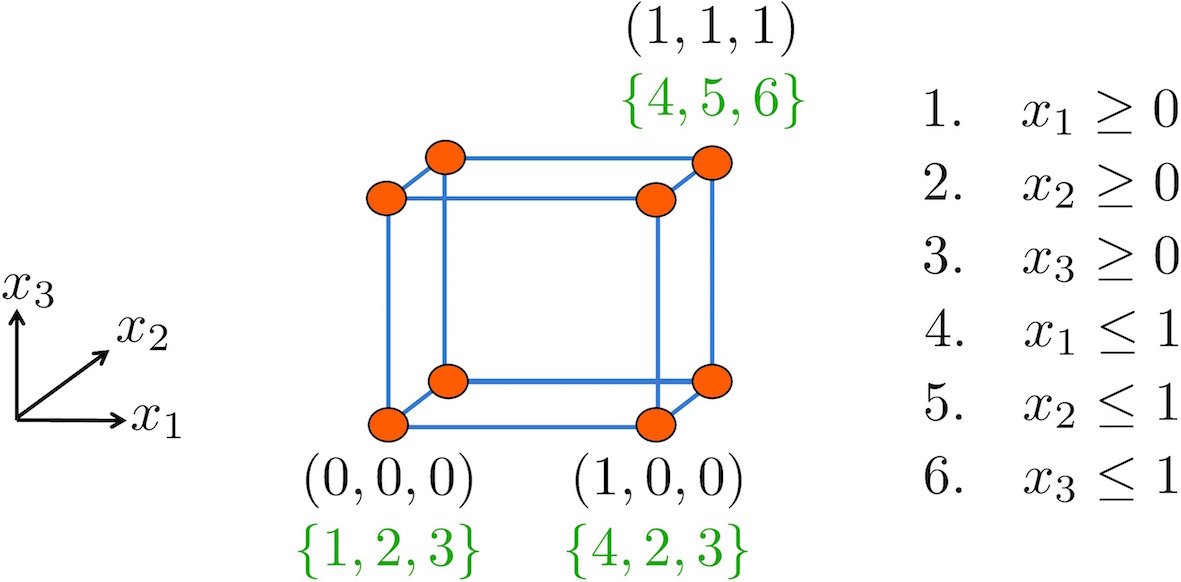} 
\caption{Identifying vertices with their feasible bases.}
\label{fig:feasiblebasis}
\end{figure*}

The most interesting thing is that vertices and feasible bases are equivalent concepts. A graph $G_P=(V,E)$ has a set of vertices or nodes which is indeed a set of feasible bases. 
Let us look at the following example shown in Figure~\ref{fig:feasiblebasis}. There are six conditions active for this polyhedron (blue edges), they are written and numbered on the right side on the Figure. For each vertex, we can determine the active constraints, and therefore form the basis of that vertex. We show some of the vertex - basis correspondence in Figure~\ref{fig:feasiblebasis}.  

\section{Matchings and vertex covers}

\begin{figure*}[h]
\centering
\includegraphics[height=5cm]{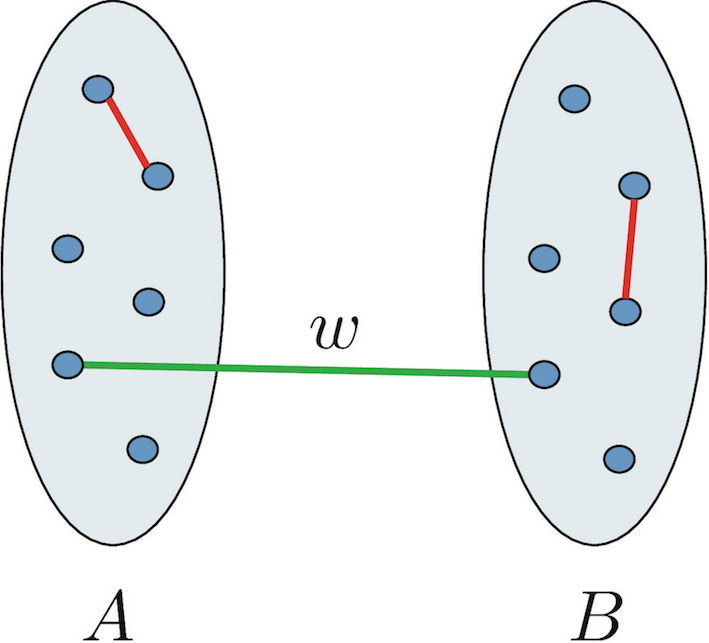} 
\caption[Bipartite graph]{Bipartite graph: two sets $A$ and $B$, edges between sets are allowed (green) while edges within the same set are not allowed (red). Edges are assigned a weight $w$.}
\label{fig:bipartite}
\end{figure*}

A graph $G=(V,E)$ is {\it bipartite} if one can partition $V$ into $V=A\cup B$ such that each edge $(u,v) \in E$ satisfies $u \in A, v\in B$. As we can see in Figure~\ref{fig:bipartite}, edges within a set are not allowed (marked in red), while edges that connect a node of one set with a node of the other set are allowed (marked in green). Each edge has a weight $w \in \mathbb{R}_0$, which can be used to represent costs, distances, etc. depending on what the graph is modeling. 

A matching is a subset $M \subseteq E$ of the edges such that each pair $e_1, e_2 \in M, e_1 \neq e_2$ satisfies $e_1 \cap e_2 = \emptyset$. 

\begin{colbox}
The maximum weight (bipartite) matching problem can be defined as follows: given a (bipartite) graph $G=(V,E)$ and edge weights $w \in \mathbb{R}_0$, determine a matching $M \subseteq E$ such that $w(M)=\sum\limits_{e \in M} w_e$ is maximal.
\end{colbox}

\begin{figure*}[ht]
\centering
\subfigure[Graph representation of the job assignment problem]{
\includegraphics[height=2.7cm]{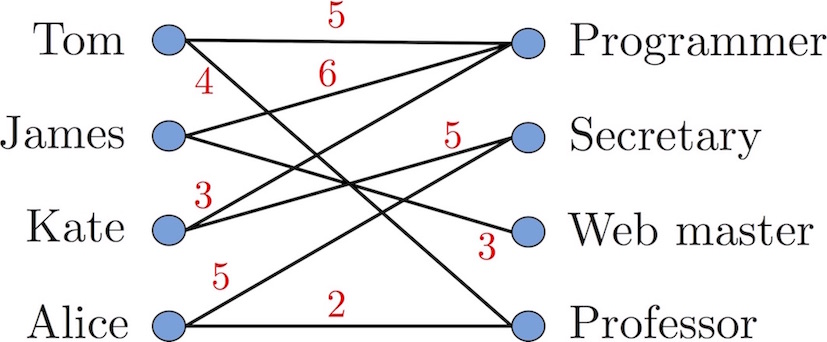} 
\label{fig:job1}
}\qquad 
\subfigure[Possible match $M$ marked by green edges]{
\includegraphics[height=2.7cm]{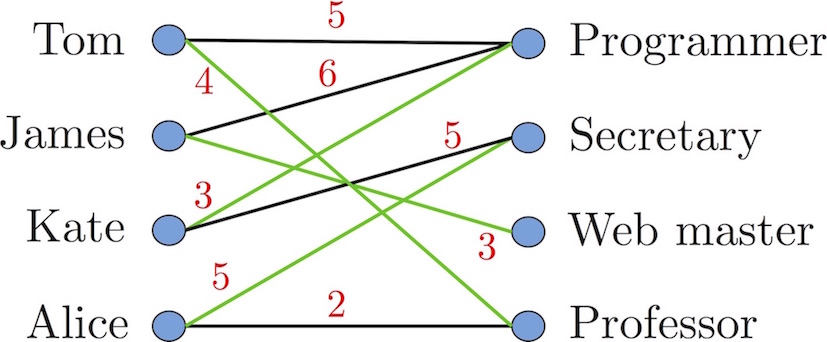} 
\label{fig:job2}
}
\subfigure[Possible w-vertex cover]{
\includegraphics[height=3cm]{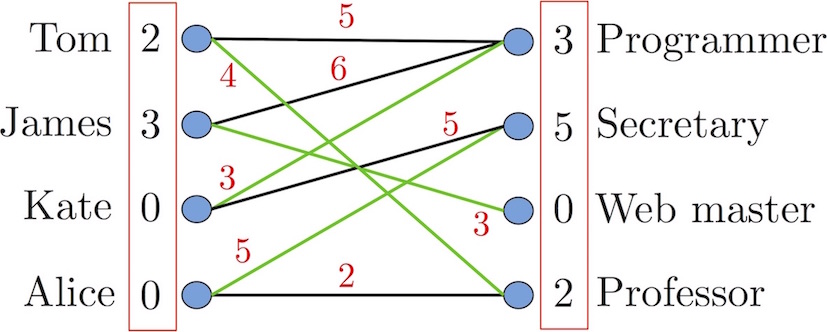} 
\label{fig:job3}
}
\caption{Bipartite graph representation of the job assignment problem.}
\label{fig:jobassignment}
\end{figure*}

Let us consider a typical example of a matching problem in bipartite graphs: the job assignment problem. The problem is very simple: we have four job openings and four applicants. Each applicant has a performance score for each job, and we want to maximize the total performance score for the company. This can be translated to a bipartite graph as shown in Figure~\ref{fig:job1}. One possible match is shown with green edges in Figure~\ref{fig:job2}, but now the question is whether this match is maximal or not. In this case, the sum of the weight of the edges used in the matching is equal to $15$.

In order to find out if this matching is optimal, we use the concept of {\it w-vertex covers}. The relation between w-vertex covers and the maximum weight matching problem is similar to the relation between primal and dual problems. 

A {\it w-vertex cover} is a vector $\mathbf{y} \in \mathbb{N}_0^{\| V \|}$ such that $\forall (u,v) \in E: y_u+y_v \geq w_{uv}$. The value of a w-vertex cover is $\sum\limits_{v\in V} y_v$.
An example of w-vertex cover for the job assignment bipartite graph can be found in Figure~\ref{fig:job3}. In this case, we need to assign a value to each node (number inside the red rectangle) so that the value of the edge that connects two nodes is smaller than or equal to the sum of the node values.
Using the concept of w-vertex cover we can prove the optimality of a matching $M$.

\begin{lemma}
Let $G=(V,E)$ be a graph and let $w \in \mathbb{R}_0$ be edge weights. If $M$ is a matching of $G$ and if $\mathbf{y}$ is a w-vertex cover of $G$, then $w(M) \leq \sum\limits_{v\in V} y_v$.
\end{lemma}

This lemma is equivalent to the weak duality of Linear Programs presented in Theorem \ref{th:weakdual}. We will know that the matching $M$ and the w-vertex cover $\mathbf{y}$ are both optimal if their values are equal, as is the case of Figure~\ref{fig:jobassignment}, where $w(M)=15=\sum\limits_{v\in V} y_v$.

\subsection{Back to Linear Programs and towards Integer Programs}

Now we can return to Linear Programs to find another way to prove the weak duality on bipartite graphs, move towards strong duality and later on discuss  integer programs and linear relaxation. These last concepts are extremely important for the multiple people tracking formulation we present in Chapter \ref{LPtracking}.

\begin{figure*}[h]
\centering
\subfigure[Example of a graph with two possible matchings (in red and green)]{
{\makebox[5cm][c]{\includegraphics[height=4cm]{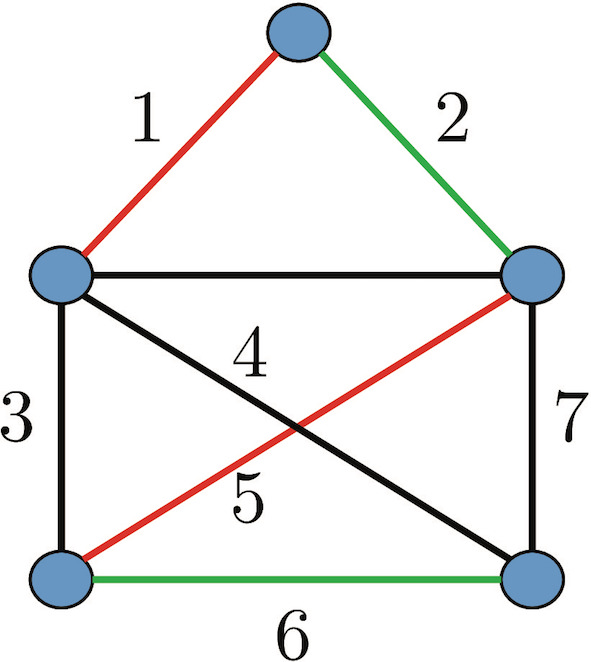}}} 
\label{fig:01vector}
}\qquad 
\centering
\subfigure[Triangle graph]{
{\makebox[5cm][c]{\includegraphics[height=3.2cm]{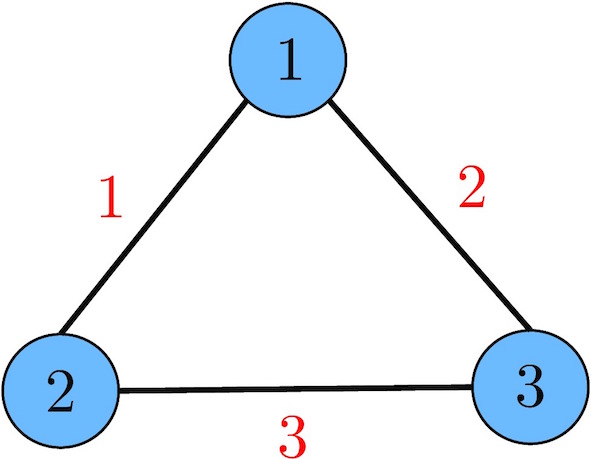}}} 
\label{fig:triangle}
}
\caption{Example graphs}
\label{fig:examples}
\end{figure*}

Let us first convert our graph to a Linear Programming formulation. We want to describe the matchings by linear constraints.
We start by enumerating all edges and describing matchings as vectors $\in \{0,1\}$. In Figure~\ref{fig:01vector} we see an example graph, in this case, the numbers identify the edges and not the nodes. We show two possible matchings, one marked by red edges and the other by green edges. The vectors corresponding to these matchings are: $\textcolor{red}{\mathbf{x}^M}=(1,0,0,0,1,0,0)$ and $\textcolor{green}{\mathbf{x}^M}=(0,1,0,0,0,1,0)$. 
The vector $\mathbf{x}^M=(1,0,0,1,0,0,1)$, for example, would not be a possible matching vector, since the edges 1 and 4 share a node, and so do edges 4 and 7. One of the characteristics of a vector that represents a matching is that one node can only be connected once by an edge. We describe this property formally now. 

\begin{figure*}[htbp]
\centering
\includegraphics[height=6cm]{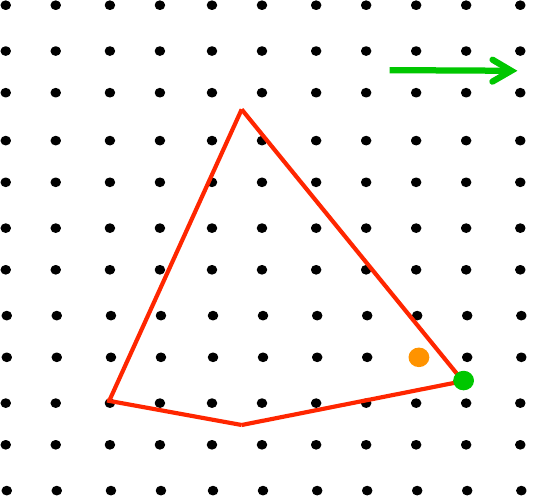} 
\caption[Representation of an integer program]{Representation of an integer program. The conditions define the red polyhedron, solutions inside it will be feasible. The green arrow points to the direction of maximization of our optimization problem, while the orange dot marks the optimal integer-valued solution.}
\label{fig:integerprogram}
\end{figure*}

For $v\in V$ we denote the set of edges {\it incident} to $v$ by $\delta(v)=\{ e \in E : v \in e\}$. The set $\{ \mathbf{x}^M : M$ is matching of $G\}$ is a set of {\it feasible solutions} that satisfies
\begin{align}
&v \in V : \sum\limits_{e \in \delta(v)} x_e \leq 1 \nonumber \\
&e \in E : x_e \in \{ 0,1\},  
\label{eq:incidence}
\end{align}
where $x_e$ is an indicator that tells us if an edge is used in the matching (1) or not (0).
Let us look at the simple graph of Figure~\ref{fig:triangle}, where nodes are denoted by black numbers and edges by red ones. The conditions defined before in Equation \eqref{eq:incidence} can be summarized by the following expression
\begin{align}
\left( \begin{array}{ccc}
1 & 0 & 1 \\
1 & 1 & 0  \\
0 & 1 & 1
\end{array} \right) 
\mathbf{x} \leq 
\left( \begin{array}{c}
1  \\
1 \\ 
1 \\
\end{array} \right) 
\quad \mbox{ where } \quad
\mathbf{x} \in \{ 0,1\}^3
\nonumber
\end{align}
These constraints have a similar expression as the ones we have seen before for Linear Programs
\begin{align}
\label{eq:integerprogram}
& \mathbf{A} \mathbf{x} \leq \mathbf{b} \nonumber \\
& \mathbf{x} \in \mathbb{Z}^n,
\end{align}
except that now $\mathbf{x}$ can now only take integer values. This defines an {\it Integer Program}, a problem like the one depicted in Figure~\ref{fig:integerprogram}, where the conditions form the red polyhedron. The black points represent integer solutions, and those within the polyhedron are feasible. The green arrow is the direction of maximization of our optimization problem. 
As we can see, if the program was a Linear Program as the ones defined in previous sections, the optimal solution would be the vertex marked by the green dot. But since our problem has decision variables which can only take integer values, we have to find the closest integer-valued solution, which is marked by the orange dot. 

The biggest drawback of Integer Programs is that they are $\mathcal{NP}$-hard, which means that they cannot be solved in polynomial time. Though $\mathcal{NP}$ complexity is not a central topic of this thesis, we make a short explanation in the following lines. We refer the interested reader to \citep{networkflows} for more details.

\begin{colbox}
{\bf A note on $\mathcal{NP}$ complexity} 
\begin{itemize}
\item{Class $\mathcal{P}$: a decision problem $P$ belongs to $\mathcal{P}$ if it can be solved by a deterministic Turing machine in polynomial time.}
\item{Class $\mathcal{NP}$: a decision problem $P$ is in $\mathcal{NP}$ if for every instance of $P$ that has a positive results, there is a certificate proving the positive result, which can be verified in polynomial time.}
\item{Class $\mathcal{NP}$-{\it complete}: a decision problem $P$ is said to be $\mathcal{NP}$-{\it complete} if: (i) $P \in \mathcal{NP}$ and (ii) all other problems in the class $\mathcal{NP}$ are reducible to $P$ in polynomial time. This implies that, if there is an efficient algorithm for some $\mathcal{NP}$-{\it complete} problem, there is an efficient algorithm for every problem in the class $\mathcal{NP}$. As a result, an $\mathcal{NP}$-{\it complete} problem is at least as hard as every other problem in the class $\mathcal{NP}$. }
\item{Class $\mathcal{NP}$-{\it hard}: a problem $P$ is said to be $\mathcal{NP}$-{\it hard} if all other problems in the class $\mathcal{NP}$ are reducible to $P$ in polynomial time. Informally, an $\mathcal{NP}$-{\it hard} problem is at least as hard as the hardest problems in $\mathcal{NP}$.}
\end{itemize}
\end{colbox}

\begin{figure*}[htb]
\centering
\includegraphics[height=7cm]{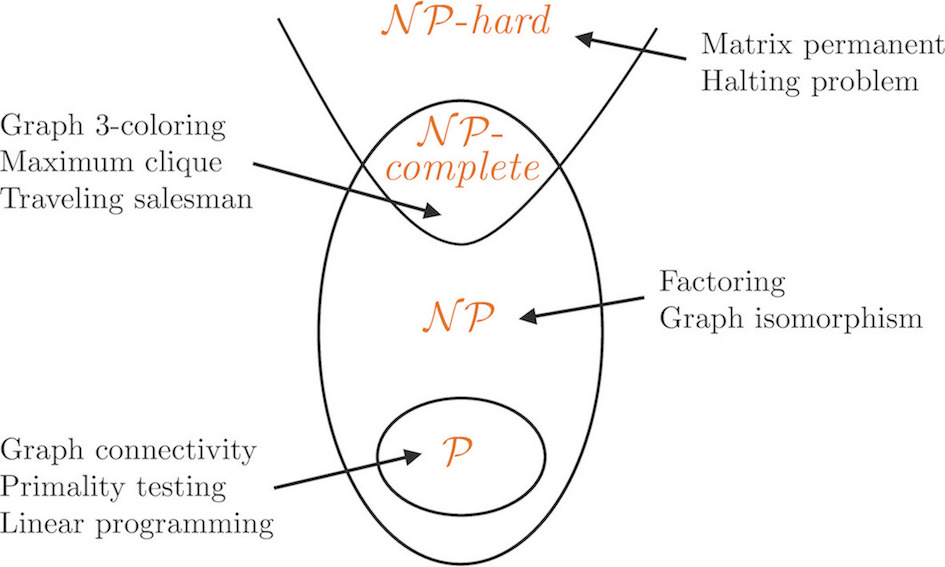} 
\caption[Representation of computational complexity classes]{Representation of computational complexity classes. Some examples of common problems belonging to each of the classes are given.}
\label{fig:npcomplexity}
\end{figure*}

Returning to the max-weight matching problem, we formulate it as an Integer Program with the constraints shown in Equation \ref{eq:integerprogram}. The variables $x_e$ are the indicators of whether an edge belongs to a matching or not, as we said before. Recall that the goal of the max-weight matching problem was to maximize the sum of $w_e$, which are the weights of the edges of the matching solution. Note that the condition $x_e \in \{ 0,1 \}$ is now expressed as $x_e \geq 0$, because together with the other condition, we only allow the variables to be bounded between 0 and 1, and if they can only take integer values, then they can only take the value 0 or 1. 

\begin{figure}[H]
\begin{minipage}[b]{0.38\linewidth}
\centering
INTEGER PROGRAM
\begin{align}
\max \quad & \sum\limits_{e \in E} w_e x_e  \nonumber \\
\st \quad &\forall v \in V :  \sum\limits_{e \in \delta(v)} x_e \leq 1 \nonumber \\
&\forall e \in E : x_e \geq 0 \nonumber \\
&\forall e \in E : \textcolor{red}{\mathbf{x}_e \in \mathbb{Z}^{| E |}} 
\label{eq:intprogram}
\end{align}
\end{minipage}
\begin{minipage}[b]{0.2\linewidth}
\hspace{1.2cm}
\includegraphics[height=1cm]{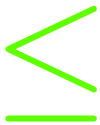} 
\vspace{2cm}
\end{minipage}
\begin{minipage}[b]{0.38\linewidth}
\centering
LP RELAXATION
\begin{align}
 \max \quad &\sum\limits_{e \in E} w_e x_e \nonumber \\
\st \quad &\forall v \in V :  \sum\limits_{e \in \delta(v)} x_e \leq 1 \nonumber \\
&\forall e \in E : x_e \geq 0 \nonumber \\
&\forall e \in E :  \textcolor{red}{\mathbf{x}_e \in \mathbb{R}^{| E |}} 
\label{eq:lprelaxation}
\end{align}
\end{minipage}
\end{figure}

We can convert the problem into a Linear Program by changing the condition marked in red in Equation \eqref{eq:intprogram} and simply considering a larger set of feasible solutions. By doing so, we would obtain the formulation of Equation \eqref{eq:lprelaxation}. Since we are considering a larger set of solutions ($\mathbb{Z} \subset \mathbb{R}$), it follows that the solution of the relaxed problem will always be an upper bound of the integer program. 

Let us look into a quick example illustrating the difference between the integer program solution and the linear relaxation solution. We take the graph of Figure~\ref{fig:triangle}, where all weights are $1$ and the maximum number of active edges connected to a node is also $1$. 
In this setting, the maximum cardinality of a matching is 1, since the use of any edge invalidates the use of all other edges. If we consider the Linear Program relaxation though, we can find a solution like $x_1=1/2,x_2=1/2,x_3=1/2$. This would make the objective value equal to $3/2$ which is indeed larger than $1$. 
To prove that this is indeed the optimal solution, we can add the three constraints and obtain $2x_1 + 2x_2 + 2x_3 \leq 3$ which brings us to $x_1+x_2+x_3 \leq 3/2$.

\begin{figure}[H]
\begin{minipage}[b]{0.36\linewidth}
\centering
INTEGER PROGRAM
\begin{align}
\min \quad &\sum\limits_{v \in V} y_v  \nonumber \\
\st \quad & \forall  (u,v) \in E :  y_u+y_v \geq w_{uv} \nonumber \\
&\forall v \in V : y_v \geq 0 \nonumber \\
&\textcolor{red}{\mathbf{y} \in \mathbb{Z}^{| V |}} 
\label{eq:wcoverinteger}
\end{align}
\end{minipage}
\begin{minipage}[b]{0.15\linewidth}
\hspace{0.5cm}
\includegraphics[height=1cm]{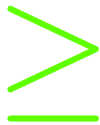} 
\vspace{2cm}
\end{minipage}
\begin{minipage}[b]{0.36\linewidth}
\centering
LP RELAXATION
\begin{align}
\min \quad & \sum\limits_{v \in V} y_v \nonumber \\
\st \quad &\forall (u,v) \in E :  y_u+y_v \geq w_{uv} \nonumber \\
&\forall v \in V : y_v \geq 0 \nonumber \\
&\textcolor{red}{\mathbf{y} \in \mathbb{R}^{| V |}} 
\label{eq:wcoverlprelax}
\end{align}
\end{minipage}
\end{figure}

We can express the minimum w-vertex cover problem as an Integer Program as well, as shown in Equation \eqref{eq:wcoverinteger}, and its LP-relaxed version in Equation \eqref{eq:wcoverlprelax}. In this case, since we are working with minimization, the LP relaxation version is a lower bound of the integer solution. 

\begin{figure*}[h]
\centering
\includegraphics[width=1\linewidth]{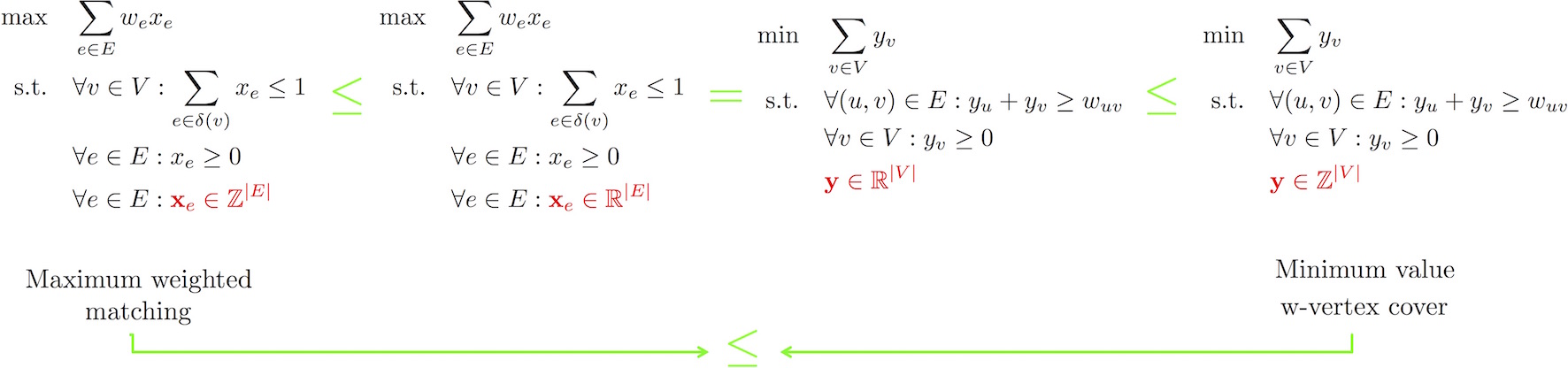} 
\caption[Weak duality proof]{Relationship between the maximum weighted matching problem, its integer and LP relaxation versions, and the minimum w-vertex cover, also with its integer and LP relaxation versions.}
\label{fig:weakdualityproof}
\end{figure*}

With these Linear Programming representations, we can again prove the weak duality between the maximum weight matching and the w-vertex cover, relationship shown in Figure~\ref{fig:weakdualityproof}.

\begin{theorem}
The maximum weight of a matching is at most the minimum value of w-vertex cover.
\end{theorem}

The only thing left to prove is that the LP relaxation representation of the maximum weighted matching is equal to the minimum w-vertex cover, which would be the proof of strong duality.

\subsection{Total unimodularity and strong duality}
\label{unimodularity}

If we again take a look at the LP relaxed versions of the problems, we can see that they have the general forms described in Equations \eqref{eq:generalmax} and \eqref{eq:gencover}. In this section we will take a closer look at the matrix $\mathbf{A}_G$ and see that these two problems are, in fact, duals of each other.

\begin{figure}[H]
\begin{minipage}[b]{0.50\linewidth}
\centering
\begin{align}
 \max \quad &\sum\limits_{e \in E} w_e x_e  \nonumber \\
\st \quad &\forall v \in V :  \sum\limits_{e \in \delta(v)} x_e \leq 1 \nonumber \\
& \forall e \in E : x_e \geq 0 \nonumber \\
& \textcolor{red}{\mathbf{x} \in \mathbb{R}^{| E |}} \nonumber \\ \nonumber \\
 \max \quad & \mathbf{w}^\intercal \mathbf{x}\nonumber \\
\st \quad & \mathbf{A}_G \mathbf{x} \leq \mathbf{1} \nonumber \\
& \mathbf{x} \geq \mathbf{0}   
\label{eq:generalmax}
\end{align}
\end{minipage}
\begin{minipage}[b]{0.48\linewidth}
\centering
\begin{align}
\min \quad &\sum\limits_{v \in V} y_v  \nonumber \\
\st \quad & \forall (u,v) \in E :  y_u+y_v \geq w_{uv} \nonumber \\
& \forall v \in V : y_v \geq 0 \nonumber \\
&\textcolor{red}{\mathbf{y} \in \mathbb{R}^{| V |}} \nonumber \\ \nonumber \\
\max \quad & \mathbf{1}^\intercal \mathbf{y}\nonumber \\
\st \quad & \mathbf{A}_G^\intercal \mathbf{y} \geq \mathbf{w} \nonumber \\
& \mathbf{y} \geq \mathbf{0}  
\label{eq:gencover}
\end{align}
\end{minipage}
\end{figure}

Let $G=(V,E)$ be a graph and suppose the nodes and edges are ordered as $v_1, \ldots, v_n$ and $e_1, \ldots, e_m$, respectively. The matrix $\mathbf{A}_G \in \{ 0,1\}^{n\times m}$ with 
\begin{align}
\mathbf{A}_G^{i,j} = \quad
\begin{cases}
1 \quad \mbox{ if } v_i \in e_j \\
0 \quad \mbox{ otherwise } \nonumber
\end{cases}
\end{align}
is the {\it node-edge incidence} matrix of $G$.

\begin{figure*}[h]
\centering
\includegraphics[width=0.8\linewidth]{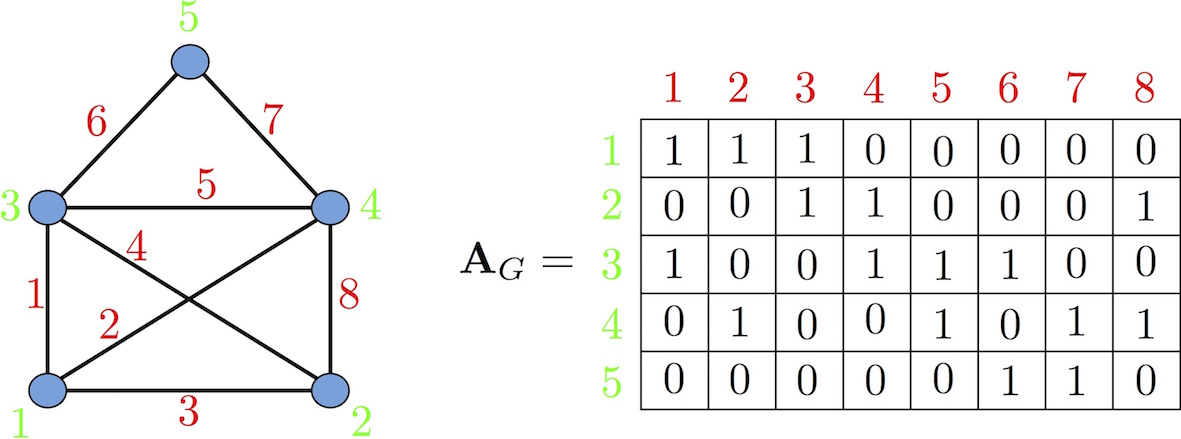} 
\caption[Node-edge incidence matrix]{Node-edge incidence matrix $\mathbf{A}_G$ of the graph represented on the left. The green numbers identify the nodes, while the red ones identify the edges.}
\label{fig:incidence}
\end{figure*}

By using Linear Programming strong duality we will be able to prove that, in fact, the optimal solutionw for the LP relaxation versions of the problems are equal to the solutions of their integer counterparts. 
This is true for bipartite graphs and in general Linear Programs defined by a node-edge incidence matrix $\mathbf{A}$ which is totally unimodular.

A matrix $\mathbf{A} \in \{ 0, \pm 1 \}$ is {\it totally unimodular} if the determinant of each square sub-matrix of $\mathbf{A}$ is equal to 0, 1 or -1. 

\begin{theorem}
Let $G=(V,E)$ be a bipartite graph. The node-edge incidence matrix $\mathbf{A}_G$ of $G$ is totally unimodular.
\label{th:unimodularbipartite}
\end{theorem}

The proof of Theorem \ref{th:unimodularbipartite} is interesting since we can use it to prove the total unimodularity of the matrix that will define our tracking problem, as we will see in Chapter \ref{LPtracking}.

\begin{proof}
We will give a proof depending on the value of $k$, where $\mathbf{B}$ is a $k \times k$ sub matrix of $\mathbf{A}_G$. Remembering that the columns of $\mathbf{A}_G$ represent edges and rows represent nodes, we can see that each column will contain two $1$'s for the nodes that edge connects, and the rest will be $0$'s. 

\begin{itemize}
\item{$k=1$. \\
If we only take a matrix of one element, this element can be $B = 0, \pm 1$ which means $\det(B)=0, \pm 1$.
}
\item{$k>1$, $\mathbf{B}$ has a column with exactly one entry equal to $1$. \\
If we develop the determinant along that column, then all coefficients are $0$ except for one, and we can derive the following expression: $\det(\mathbf{B})= \pm 1 \det (\mathbf{B'})$, where $\mathbf{B'}$ is a $(k-1) \times (k-1)$ sub matrix obtained by deleting the column and row marked by an orange line. 
\begin{figure*}[h]
\centering
\includegraphics[height=3.8cm]{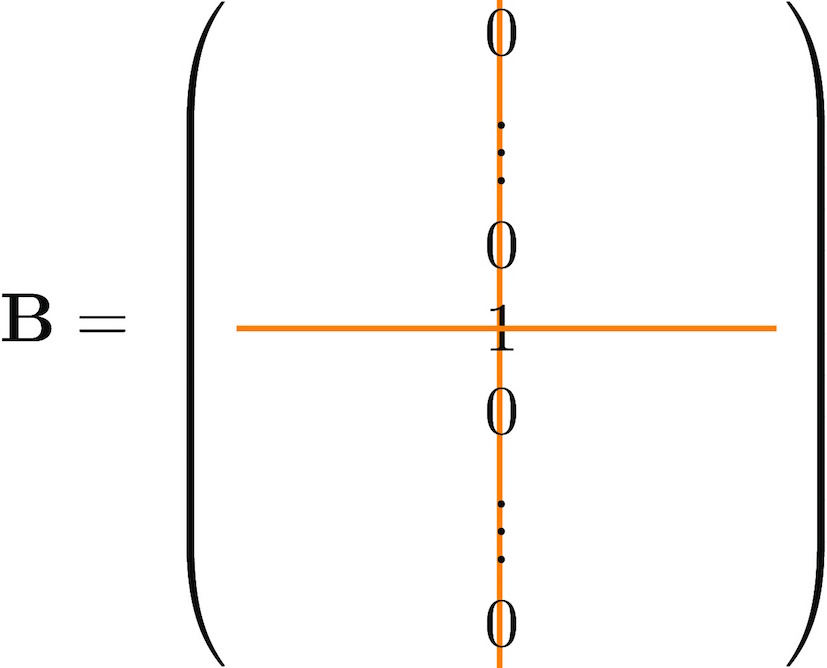} 
\end{figure*}

Of course, we can see that $\mathbf{B'}$ is a sub matrix of $\mathbf{A}_G$, so we can use induction on $\mathbf{B'}$ to prove its total unimodularity. 
}
\item{$k>1$, each column of $\mathbf{B}$ contains exactly two entries equal to $1$. \\
We can order the rows according to the set they belong to. Remember that a bipartite graph consists of two sets of vertices (for example, $F$ and $M$), and that an edge can only connect one vertex from set $F$ to a vertex of the other set $M$. 
\begin{figure*}[ht]
\centering
\includegraphics[height=4.3cm]{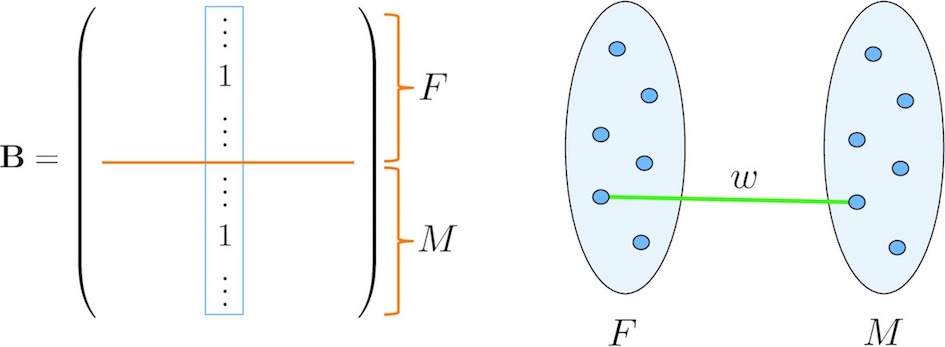} 
\end{figure*}

If we put all the rows of the first set on top and all the rows of the second set at the bottom, we can see that for each column we will have exactly one $1$ above the orange line and another below it. If we then add up all the rows above the line, we will obtain an all $1$'s vector. We will obtain the same if we add up all rows below the line. This means that these rows are not linearly independent, making $\det(\mathbf{B})=0$.
}
\end{itemize}
\end{proof}

Going back to integer and linear programs, we can state the following theorem.

\begin{colbox}
\begin{theorem}
If $\mathbf{A}\in \mathbb{Z}^{m \times n}$ is totally unimodular and $\mathbf{b} \in \mathbb{Z}^m$, then every vertex of the polyhedron $P=\{\mathbf{x} \in \mathbb{R}^n : \mathbf{A} \mathbf{x} \leq \mathbf{b}\}$ is integral. 
\end{theorem}

This basically tells us that as long as our matrix $\mathbf{A}$ is totally unimodular, even if the indicator variable $\mathbf{x}$ is not specified to be integer-valued, the optimal solution will always be integral. 
\end{colbox}

Therefore, we can state the following corollary.

\begin{corollary}
If $\mathbf{A}\in \mathbb{Z}^{m \times n}$ is totally unimodular, $\mathbf{b} \in \mathbb{Z}^m$, and if $\max \{ \mathbf{c}^\intercal \mathbf{x} :\mathbf{x} \in \mathbb{R}^n , \mathbf{A} \mathbf{x} \leq \mathbf{b}\}$ is bounded, then 

$\max \{ \mathbf{c}^\intercal \mathbf{x} :\mathbf{x} \in \textcolor{red}{\mathbb{R}^n} , \mathbf{A} \mathbf{x} \leq \mathbf{b}, \mathbf{x} \geq \mathbf{0}\}=\max \{ \mathbf{c}^\intercal \mathbf{x} :\mathbf{x} \in \textcolor{red}{\mathbb{Z}^n} , \mathbf{A} \mathbf{x} \leq \mathbf{b}, \mathbf{x} \geq \mathbf{0}\}$
\end{corollary}

From all this, we can conclude that the maximum weight of a matching is equal to the minimum value of a w-vertex cover, which is the strong duality in a bipartite graph. This is all summarized in the following theorem by Egerv{\' a}ry \citep{egervary1931}.

\begin{theorem}
Let $G=(V,E)$ be a bipartite graph and let $w \in \mathbb{N}_0$ be edge weights. The maximum weight of a matching is equal to the minimum value of a w-vertex cover.
\label{th:egervary}
\end{theorem}

If we take a look at Figure~\ref{fig:weakdualityproof}, we see that we have proven all inequalities to be equal when $\mathbf{A}_G$ is totally unimodular, which is in fact true for the bipartite graph, so we have proven Theorem \ref{th:egervary}.

A similar but less general theory was developed independently by K{\"o}nig in 1931 \citep{konig1931}. It defines a {\it vertex cover} of a graph $G=(V,E)$ to be a subset $U \subseteq V$ such that $e \cap U \neq \emptyset$ for each $e \in E$. This is the same as a w-vertex cover in the special case when $\mathbf{w}= \mathbf{1}$, \ie an all ones vector.

\begin{theorem}
Let $G=(V,E)$ be a bipartite graph. The maximum cardinality of a matching of $G$ is equal to the minimum cardinality of a vertex cover of $G$.
\label{th:konig}
\end{theorem}

\section{The shortest path problem}

After explaining the main concepts of Linear Programming and their relationship to graph theory, in this section we focus on solvers for graphs, namely shortest paths. Though we do not use this particular algorithm to solve our multiple people tracking problem, it has been widely used in the literature \citep{berclaztpami2011}, \citep{pirsiavashcvpr2011} and therefore we consider them to be a valuable concept to be included in this Chapter.
Towards the end of the section, we will see again the connection of this method to Linear Programming.

\begin{figure*}[ht]
\centering
\subfigure[Undirected graph and its adjacency matrix]{
\includegraphics[height=4cm]{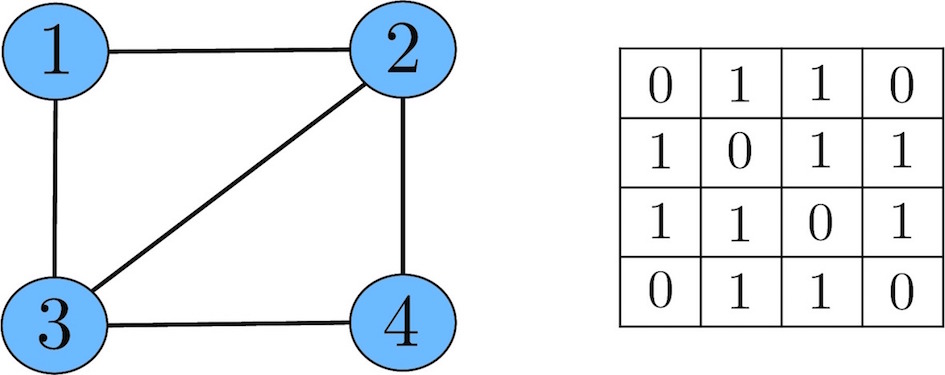} 
\label{fig:undirected}
}\qquad 
\centering
\subfigure[Directed graph and its adjacency matrix]{
\includegraphics[height=4cm]{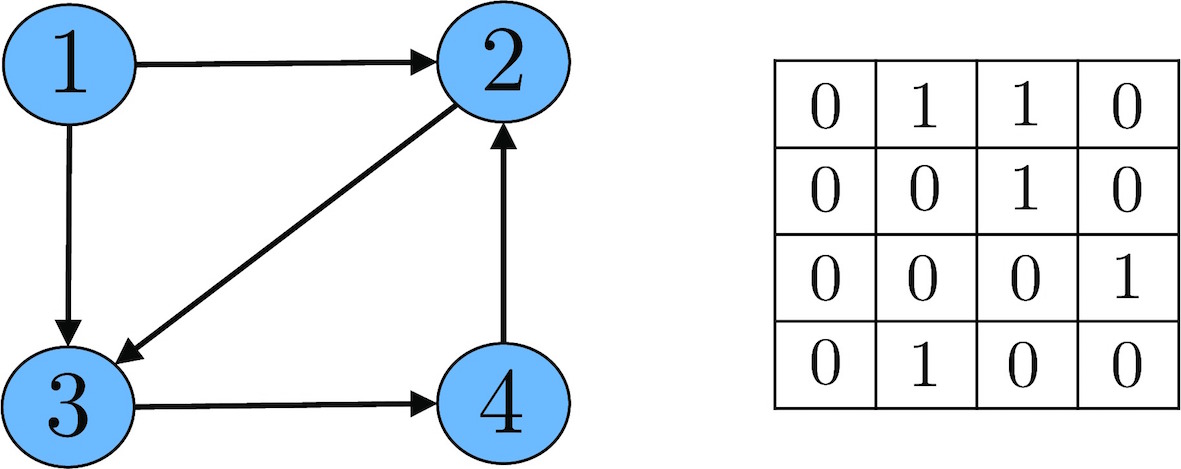} 
\label{fig:directed}
}
\caption{Undirected \vs directed graphs}
\label{fig:undirvsdirgraphs}
\end{figure*}

So far we have talked about undirected graphs $G=(V,E)$, where $V=\{1, \ldots, n \}$ is the set of vertices or nodes and $M \in \{0,1\} ^{|V| \times |V|}$ represents their adjacency matrix. Every pair of nodes connected by an edge has a $1$ entry in the matrix, \ie:
\begin{align}
M_{ij} = 
\begin{cases}
1 \quad \mbox{ if } (i,j) \in E\\
0 \quad  \mbox{ otherwise.} \nonumber
\end{cases}
\end{align}
We can see an example of this graph in Figure~\ref{fig:undirected}, note that the adjacency matrix is symmetric since the edges are undirected. 

A {\it directed graph}, on the other hand, is a tuple $D=(V,A)$, where $V$ is a finite set of vertices or nodes and $A$ is the set of arcs or {\it directed} edges of $D$. We denote a directed edge by its defining tuple $(u,v) \in A$. The nodes $u$ and $v$ are called tail and head of $(u,v)$, respectively. In the example of Figure~\ref{fig:directed}, the edge $(3,4)$ would have $3$ as head and $4$ as tail.
The adjacency matrix of directed graphs is composed by:
\begin{align}
M_{uv} = 
\begin{cases}
1 \quad \mbox{ if } (u,v) \in A\\
0 \quad  \mbox{ otherwise.} \nonumber
\end{cases}
\end{align}
This means that a $1$ represents not an undirected edge but a directed {\it arc}, so we have to pay attention to the beginning and end point, since the direction of the arc changes where the $1$ is placed within the adjacency matrix. 
We can see the same example as before in Figure~\ref{fig:directed}, but this time as a directed graph. As we can see, the adjacency matrix is no longer symmetric.

For our multiple people tracking problem, we use weighted directed graphs, where each edge has a weight related to it. Let $D=(V,A)$ be a directed graph without cycles, where $c : A \rightarrow \mathbb{R}$ are the {\it lengths} or {\it costs} of the arcs. The length of a walk $W=v_0, \ldots, v_k$ is the sum of the lengths or costs of its arcs:
\begin{align}
c(W)=\sum\limits_{i=1}^k c (v_{i-1},v_i). \nonumber
\end{align}
The {\it distance} between two nodes $s$ and $t$ is the cost of a {\it shortest path} from $s$ to $t$.

\begin{figure*}[ht]
\centering
\includegraphics[height=4cm]{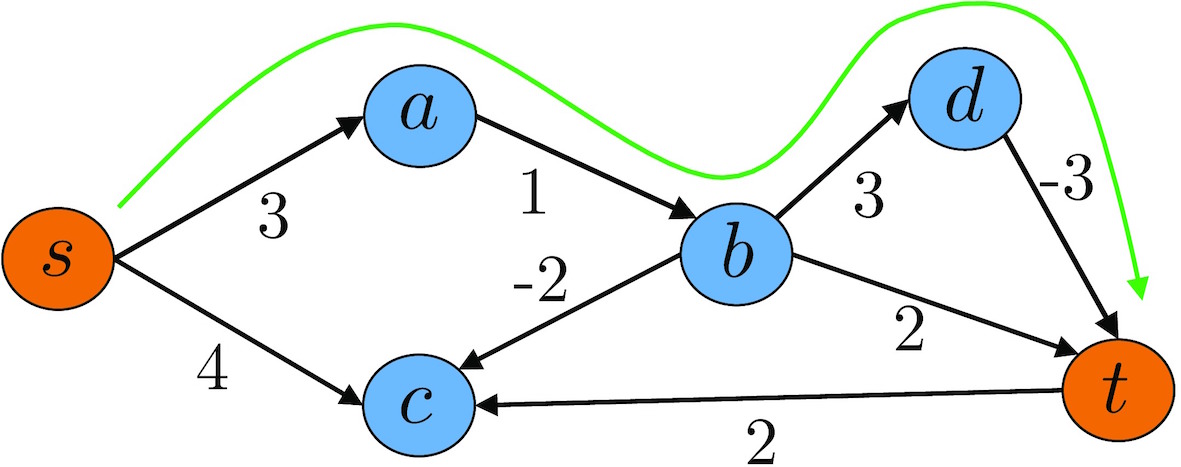} 
\caption[Weighted directed graph]{Weighted directed graph. Shortest path from $s$ to $t$ with length 4 marked in green.}
\label{fig:weighteddirected}
\end{figure*}

In Figure~\ref{fig:weighteddirected} we can see an example of a weighted directed graph. We can see, for example, that the cost of the walk $s,a,b,c$ is $3+1-2=2$. The shortest path from $s$ to $t$ is marked in green and has cost $s,a,b,d,t=3+1+3-3=4$.

The shortest path problem can be formally defined as: given a directed graph with edge costs and a designated node $s$, compute $d(s,v)$ for each $v \in V$. This is an $\mathcal{NP}$-hard problem in general but solvable in polynomial time if there are no negative cycles. 
A {\it cycle} is defined as a walk $v_0, v_1, \ldots ,v_k$  with $v_0=v_k$.

There are many solvers to find the shortest path in a graph, e.g. Dijkstra's algorithm or the Bellman-Ford algorithm. 
Like Dijkstra's Algorithm, Bellman-Ford is based on the principle of relaxation, in which an approximation to the correct distance is gradually replaced by more accurate values until eventually reaching the optimum solution. In both algorithms, the approximate distance to each vertex is always an overestimate of the true distance, and is replaced by the minimum of its old value with the cost of a newly found path. 

For the multiple people tracking problem, we can use a solver based on Dijkstra's algorithm or Bellman-Ford algorithm that will find a series of $k$-shortest paths, each one representing a valid pedestrian trajectory \citep{berclaztpami2011,pirsiavashcvpr2011}.


\subsection{The Bellman-Ford method}
\label{bellmanfordsection}

In this section, we describe a method to compute minimum length walks given a weighted directed graph $D=(V,A)$ with no cycles of negative length and a designated node $s \in V$. The goal of the method is to compute shortest path distances from $s$ to all other nodes, assuming that each node is reachable from $s$.

For $k \geq 0$ and $t \in V$, we can define $d_k(t)$ to be the minimum length of any $s-t$ walk, traversing at most $k$ arcs. For example, $d_0(s)=0$, since the length of a walk from $s$ to $s$ traversing at most $0$ arcs is, in fact, also $0$. $d_0(t)= \infty$ unless $t=s$, since we cannot reach any other node from $s$ by traversing at most $0$ arcs.

Let us assume $d_i(t)$ is known for each $i \leq k$ and each $t \in V$, and now we want to compute $d_{k+1}(t)$ for each $t \in V$. Now we can encounter two cases: the first one is when a shortest walk traversing at most $k+1$ arcs traverses exactly $k+1$ arcs; the second one is when the shortest walk traversing at most $k+1$ arcs actually traverses at most $k$ arcs, \ie $d_{k+1}(t)=d_k(t)$. Both of these are upper bounds of $d_{k+1}(t)$.

To sum up, for $k \geq 0$ and $t \in V$ : $d_{k+1}(t) = \min \{ d_k(t), \min\limits_{(u,t) \in A} d_k(u) + c (u,t)\}$. In Algorithm \ref{alg:bellmanford1} we depict the procedure to compute the values $d_{k+1}(t)$ assuming that $d_k(t)$ are pre-computed. The idea of the algorithm is to iteratively set $d_{k+1}(t)$ to the smallest value possible. Since both $d_k(t)$ and $d_k(t) + c (u,t)$ are upper bounds of $d_{k+1}(t)$, we make sure that this is set to its smallest possible value at each iteration.

\begin{algorithm}
\caption{Bellman-Ford algorithm}          
\label{alg:bellmanford1}                           
\begin{algorithmic}                    
\vspace{0.13cm}
\STATE{\textbf{initialize}}
\vspace{0.13cm}
\STATE{\hspace{0.25cm}  $\forall t \in V \setminus \{s\}, d_0(t)=\infty$}
\vspace{0.13cm}
\STATE{\hspace{0.25cm} $d_0(s)=0$}
\vspace{0.13cm}
\FOR{$k=0$ to $n-2$}
\vspace{0.13cm}
\FOR{\textbf{each} $t \in V$}
\vspace{0.13cm}
\STATE{$d_{k+1}(t) \vcentcolon= d_k(t)$}
\vspace{0.13cm}
\ENDFOR
\FOR{\textbf{each} $(u,t) \in A$}
\vspace{0.13cm}
\IF{$d_k(t) + c (u,t) < d_{k+1}(t) $}
\vspace{0.13cm}
\STATE{$d_{k+1}(t) \vcentcolon= d_k(t) + c (u,t)$}
\vspace{0.13cm}
\ENDIF
\vspace{0.13cm}
\ENDFOR
\vspace{0.13cm}
\ENDFOR
\vspace{0.13cm}
\IF{$\exists t \in V$ with $d_n(t) < d_{n-1}(t)$}
\vspace{0.13cm}
\STATE{$D$ has a negative cycle}
\vspace{0.13cm}
\ENDIF
\vspace{0.13cm}
\end{algorithmic}
\end{algorithm}

\begin{figure}[ht]
\begin{floatrow}
\ffigbox{%
  \includegraphics[height=4cm]{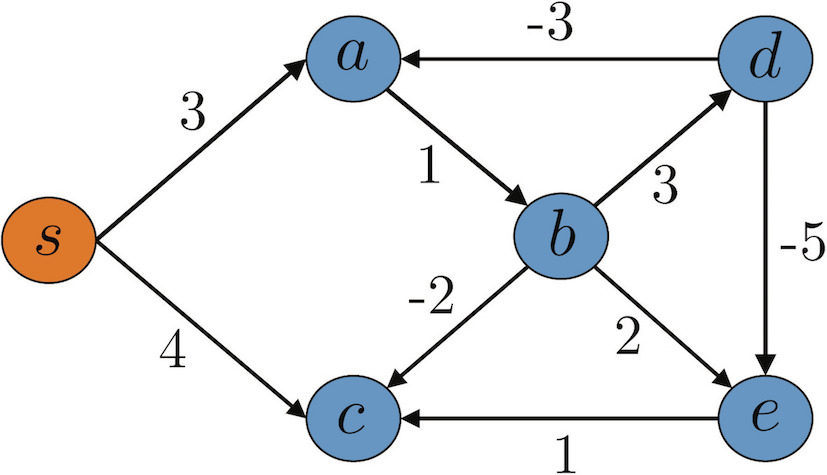} %
}{%
 \caption[Bellman-Ford algorithm example]{Bellman-Ford algorithm example.}%
 \label{fig:bellmanfordexample1}
}\qquad
\capbtabbox{%
  \begin{tabular}{ c | c|c|c|c|c||c} 
  0 & 3 & 4 & 2 & 7 & 2 & $d_5$ \\ \hline
  0 & 3 & 4 & 2 & 7 & 2 & $d_4$ \\ \hline
    0 & 3 & 4 & 2 & 7 & 6 & $d_3$ \\ \hline
    0 & 3 & 4 & 4 & $\infty$ & $\infty$ & $d_2$\\ \hline
    0 & 3 & $\infty$ & 4 & $\infty$ & $\infty$  & $d_1$\\  \hline
    0 & $\infty$ & $\infty$ & $\infty$ & $\infty$ & $\infty$ & $d_0$\\ \hline\hline
    $s$ & $a$ & $b$ & $c$ & $d$ & $e$ 
    \end{tabular}
}{%
  \caption{Distances computed by the Bellman-Ford algorithm.}
\label{tab:bellmanford1}%
}
\end{floatrow}
\end{figure}

Let us consider the following example as depicted in Figure~\ref{fig:bellmanfordexample1}. We start the computation of distances as explained in Algorithm \ref{alg:bellmanford1}, by first initializing $d_0$ to $0$ for the node $s$ and $\infty$ for all other nodes.
We start the first iteration with $k=0$, where we compute $d_1$ accordingly.  By traversing at most $1$ arc from $s$, we can reach nodes $a$ and $c$ with lengths $3 $ and $4$, respectively. We therefore obtain the distances as shown in Table \ref{tab:bellmanford1}, third row from the bottom. We keep computing distances for $k=2,3$ until we reach $k=4$, where the algorithm converges and all values of $d_5$ are equal to $d_4$.

Now we have to formally define the conditions under which the Bellman-Ford algorithm terminates (converges). 

\begin{theorem}
Given $D=(V,A), s \in V$, $d_n=d_{n-1}$ for $n=|V|$ iff $D$ does not have a cycle of negative length that is reachable from $s$.
\end{theorem}

\begin{theorem}
Given $D=(V,A), s \in V$, and suppose no negative cycle is reachable from s. Then for each $t \in V$, the computed $d_{n-1}(t)$ is the distance between $s$ and $t$.
\end{theorem}

If we do not encounter any negative length cycles, we just need to perform $n-1$ iterations, where $n$ is the number of vertices of the graph, and we are guaranteed to find the shortest path solution from $s$ to all vertices in the graph.
%
%

The Bellman-Ford algorithm runs in time $O(|V|  |A|)$. While for most of the graphs the algorithm needs much less than $|V|-1$ iterations, the algorithm still does not scale well. 

\subsection{Shortest path expressed as a Linear Program}

\begin{colbox}
There is a natural linear programming formulation for the shortest path problem. Given a directed graph $D=(V, A)$ with source node $s$, target node $t$, and cost $c (u,v)$ for each arc $(u,v) \in A$, consider the program with variables $f(u,v)$:
\begin{align}
&\min \qquad \sum\limits_{(u,v) \in A} c(u,v) f(u,v) \label{eq:shortestpath1}  \\
&\mbox{ subject to } f \ge 0  \label{eq:shortestpath2}\\
&\mbox{ and } \forall u \in V, \sum_{v \in V} f(u,v) - \sum_{v \in V} f(v,u) = \begin{cases}1, &\text{if }u=s;\\ -1, &\text{if }u=t;\\ 0, &\text{ otherwise.}\end{cases}
\label{eq:shortestpath3}
\end{align}
\end{colbox}
This LP has the special property that it is integral; more specifically, the decision variables of every basic optimal solution (when one exists) assume values of $0$ or $1$. This is because the condition matrix is totally unimodular, as explained in Section \ref{unimodularity}.

The shortest path problem can be seen from a network flow point of view \citep{networkflows}, where we are interested in sending a commodity through a network at the smallest cost possible (see Equation \eqref{eq:shortestpath1}). In this case, each commodity sent through the network is one unit of ``flow", represented by the variable $f$. The {\it capacity} of an arc is defined as the amount of flow that can be sent through that arc; this is the first condition of the LP, shown in Equation \eqref{eq:shortestpath2}. The {\it mass balance constraints} are defined for each node and make sure that all flow that enters a node also exits that node as expressed in Equation \eqref{eq:shortestpath3}.

\subsection{{\it k}-shortest paths}

To conclude this chapter, we will introduce the $k$-shortest paths algorithm. There are several types of $k$-shortest paths problems that can be solved, \eg, finding $k$ paths with decreasing costs \citep{ksp1971}, but for the multiple object tracking problem we are interested in the problem of finding {\it $k$-shortest disjoint paths} \citep{ksp,networkflows}. This problem is based on the assumption that we are interested in finding edge-disjoint paths, \ie paths that do not share common edges. This exclusion property is key to the multiple people tracking problem, since a node represents a detected person and therefore we cannot assign it to two trajectories. 

\noindent{\bf Node potentials.}

Let us first start by introducing some useful concepts. In many network flow algorithms it is useful to measure the cost of an arc relative to ``imputed" costs associated with its incident nodes. These costs are typically intermediate data that is computed within the context of an algorithm. Let $D=(V,A)$ be a directed graph; we associate to each node $i \in V$ a number $\pi(i)$, which we refer to as the {\it potential} of that node. We can define a reduced cost (or length) of an arc as:
\begin{align}
c^{\pi} (i,j) = c(i,j) - \pi (i) + \pi (j).  \nonumber 
\end{align}
Often algorithms work with these reduced costs, since they have an interesting property: minimum cost flow problems with arc costs $c(i,j)$ or $c^{\pi} (i,j)$ have the same optimal solutions since their objective functions only differ by a constant.

\noindent{\bf Residual network.}

Sometimes it is convenient to measure flow $f$ not in absolute terms, but rather in terms of incremental flow about some given feasible solution, what would be the equivalent of an intermediate solution. 
For this, we can define a new additional network called the {\it residual network} \citep{fordfulkerson}. The advantage is that the formulations of a problem in the original network and in the residual one are actually equivalent.

\begin{figure}[ht]
  \includegraphics[height=3.5cm]{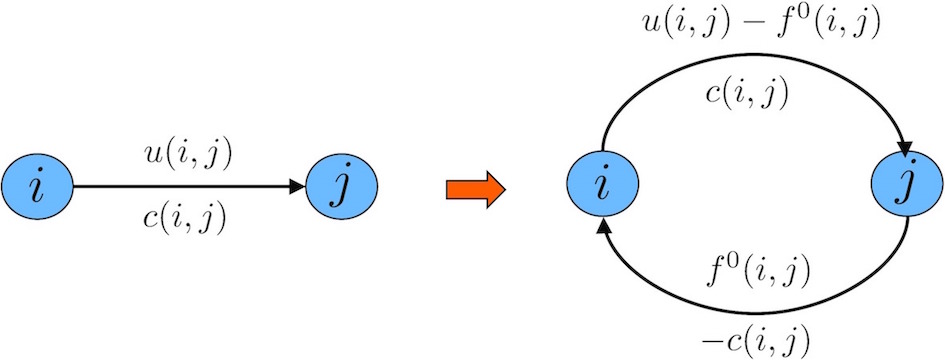} 
 \caption{How to construct a residual network.}%
 \label{fig:residualnetwork}
\end{figure}

Given an original graph $G$, we define a residual network $G(f^0)$ with respect to flow $f^0$ as follows. We replace arc $(i,j)$ in the original network with two arcs: $(i,j)$ with cost $c(i,j)$ and residual capacity $r(i,j) = u(i,j) - f^0 (i,j)$, and another arc $(j,i)$ that has cost $-c(i,j)$ and residual capacity $r(j,i) = f^0(i,j)$, as shown in Figure~\ref{fig:residualnetwork}. The residual network consists of only the arcs with a positive residual capacity. 

An interesting property of residual networks is that a flow $f$ is feasible in the network $G$ if and only if its corresponding flow $f'$, defined by $f'(i,j) - f'(j,i) = f(i,j) - f^0(i,j)$ and $f'(i,j)f'(j,i)=0$, is feasible in the residual network $G(f^0)$. Furthermore, $cf=c'f'+c f^0$. 
This provides us with the flexibility of working with a residual network, and once we determine an optimal solution for it, we can convert it to an optimal solution of the original network.

\noindent{\bf Shortest paths for multiple people tracking.}

The idea of the successive shortest path algorithm \citep{networkflows} is to maintain optimality of the solution at every step while trying to achieve feasibility. It maintains a solution $f$ that satisfies the nonnegativity and capacity constraints but violates the mass balance constraints of the node.


The details of the general $k$-shortest paths algorithm in Algorithm \ref{alg:shortestpaths}.

\begin{algorithm}
\caption{$k$-shortest paths}          
\label{alg:shortestpaths}                           
\begin{algorithmic}                    
\vspace{0.13cm}
\STATE{\textbf{initialize}}
\vspace{0.13cm}
\STATE{\hspace{0.25cm}  $f \vcentcolon= 0$ and $\pi \vcentcolon=0$}
\vspace{0.13cm}
\FOR{$1$ to $k$}
\vspace{0.13cm}

\STATE{1. Compute the shortest path from node $s$ to node $t$, computed using the Bellman-Ford algorithm of Section \ref{bellmanfordsection}.}
\vspace{0.13cm}

\STATE{2. Update $\pi \vcentcolon= -d$.}
\vspace{0.13cm}

\STATE{3. $\delta \vcentcolon= \min [ e(s) , - e(t) , \min \{ r(i,j) : (i,j) \in P\} ]$.}
\vspace{0.13cm}

\STATE{4. Create a residual graph from sending flow $\delta$ along $P$.}
\vspace{0.13cm}

\STATE{5. Compute the reduced costs $c^{\pi} (i,j) = c(i,j) - \pi (i) + \pi (j)$.}
\vspace{0.13cm}

\vspace{0.13cm}
\ENDFOR
\vspace{0.13cm}
\end{algorithmic}
\end{algorithm}

\begin{figure*}[ht]
\centering
\subfigure[Initial graph with zero potentials. First shortest path found.]{
\includegraphics[height=4cm]{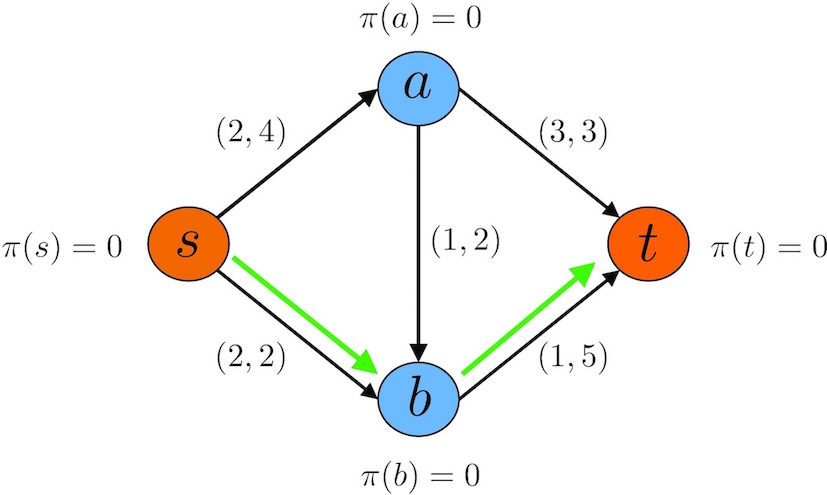} 
\label{fig:ksp1}
}\quad 
\centering
\subfigure[Compute new node potential and create the residual graph.]{
\includegraphics[height=4cm]{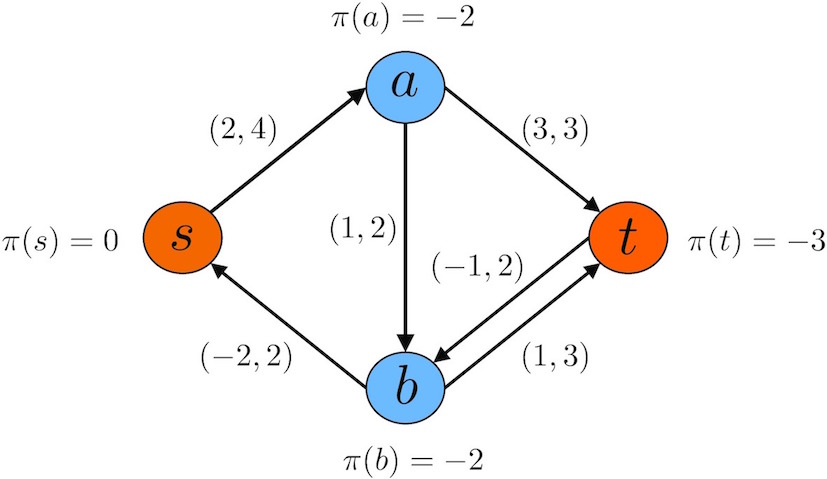} 
\label{fig:ksp2}
}
\subfigure[Reduce costs according to new potentials and find new shortest path.]{
\includegraphics[height=3.93cm]{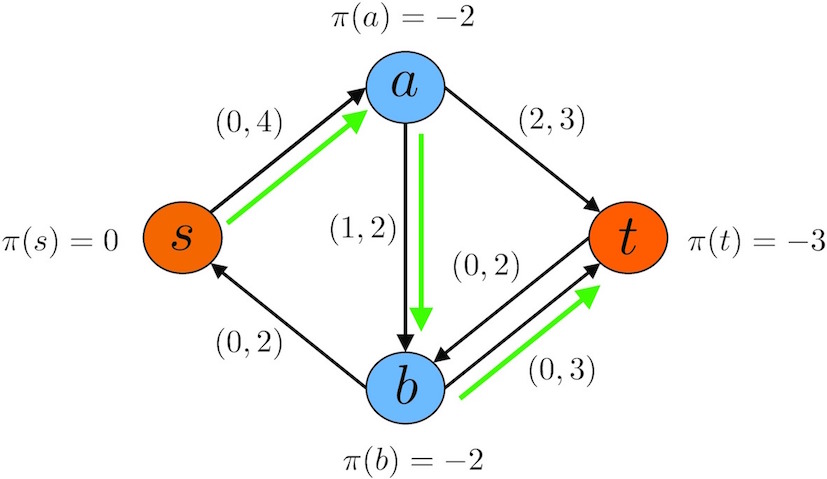} 
\label{fig:ksp3}
}\quad 
\centering
\subfigure[Compute new node potential and create the residual graph.]{
\includegraphics[height=3.93cm]{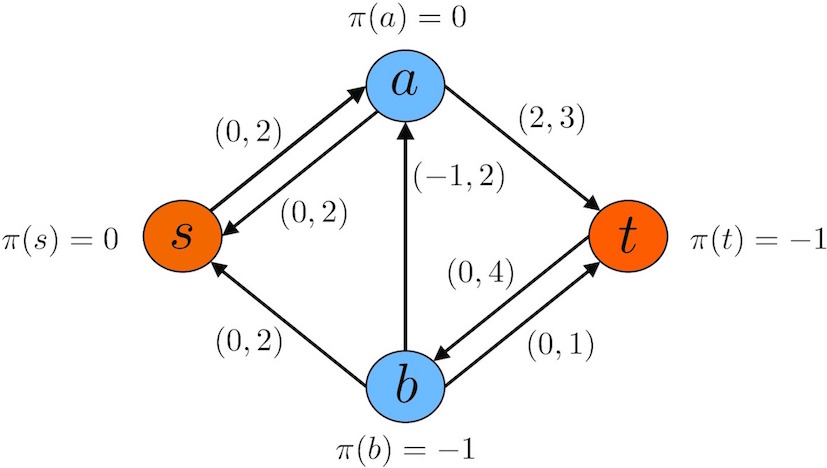} 
\label{fig:ksp4}
}
\subfigure[Reduce costs according to new potentials and find new shortest path.]{
\includegraphics[height=4cm]{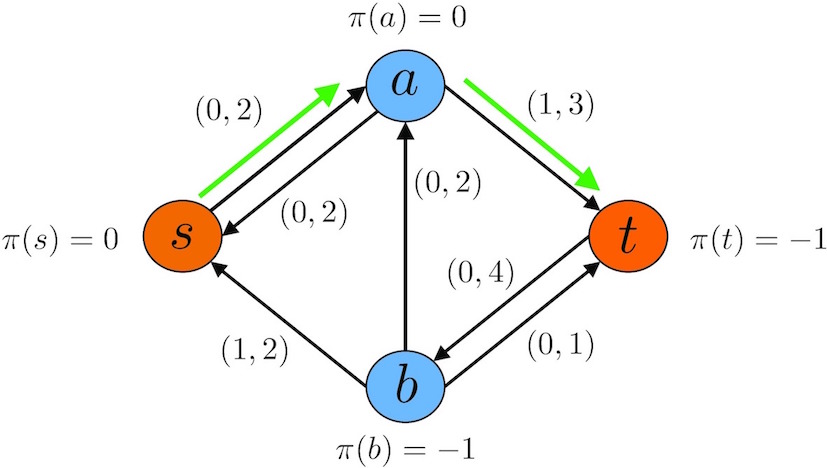} 
\label{fig:ksp5}
}\quad 
\centering
\subfigure[Three shortest paths found on this graph.]{
{\makebox[6.5cm][c]{\raisebox{0.5cm}{\includegraphics[height=3.1cm]{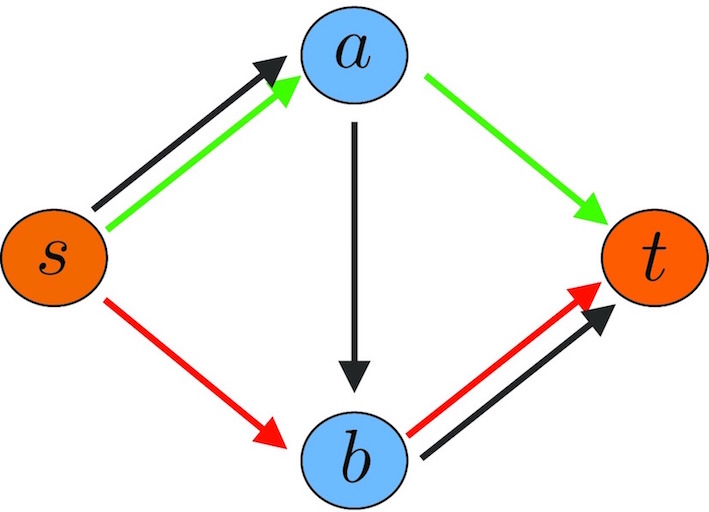}}}} 
\label{fig:ksp6}
}
\caption{Example of how $k$-shortest path algorithm works}
\label{fig:kspexample}
\end{figure*}

Let us look at the example of Figure~\ref{fig:kspexample}; we start with the initial graph depicted in Figure~\ref{fig:ksp1} where we find the first shortest path $s-b-t$. With this we compute the node potentials shown in Figure~\ref{fig:ksp2} and create the residual graph from sending flow through the path $s-b-t$. We then go to step 5 of Algorithm \ref{alg:shortestpaths} and compute the new reduced costs as shown in Figure~\ref{fig:ksp3}. Now we can start the cycle again by computing a new shortest path $s-a-b-t$, new potentials and residual graph (Figure~\ref{fig:ksp4}), new reduced costs and the final shortest path $s-a-t$ as shown in Figure~\ref{fig:ksp5}. The three shortest paths found are shown in red, green and black in Figure~\ref{fig:ksp6}.

This algorithm is described as an edge-disjoint successive shortest path, which means that a node might be used by two or more different paths, as $a$ and $b$ in the example of Figure~\ref{fig:kspexample}. In order to convert it to a node-disjoint successive shortest path algorithm, one can divide each node into two nodes and insert an extra edge in the middle with capacity equal to $1$. This procedure is done anyway for multiple people tracking as explained in Chapter \ref{LPtracking}, so we can directly compute the $k$-shortest paths as explained in Algorithm \ref{alg:shortestpaths}. 

\begin{colbox}
For multiple people tracking, we build a graph with the detections and add a source $s$ node and a sink $t$ node, from where all flows start and end. The algorithm iterates $k$ times the following two steps: (i) find the shortest path in the network; (ii) create a residual network and augment the flow along the path. Each time a flow of $1$ is pushed to the network ($\delta=1$), which can be interpreted as one trajectory. Trajectories found previously can be changed if the algorithm pushes the flow back.
\end{colbox}

\noindent{\bf Computational complexity.}

Bellman-Ford algorithm has a complexity of $O(|V| |A|)$ and while it can be applied to graphs with a wider range of inputs, it is slower than Dijkstra's algorithm which runs at $O(|A| \log |V|)$ under certain conditions. 
For the multiple people tracking problem, there exist negative costs in the network and therefore we cannot directly apply Dijkstra  to find the shortest paths. Fortunately, we can convert this initial graph into an equivalent graph by using Bellman-Ford once at the beginning and creating a graph with reduced costs using node potentials. 
For the rest of $k$ iterations, we can use Dijkstra's algorithm to find the shortest paths. This procedure is described in \citep{pirsiavashcvpr2011} for multiple people tracking with the network structure that we will present in Chapter \ref{LPtracking}.

\section{Programming Linear Programs}

There are several Linear Programming solvers available online. In this section, we give a quick overview of a C library which includes several Linear and Integer Programming solvers as well as the MatLab functions that allow us to solve LPs using Simplex.

\subsection{GLPK Library}

The GNU Linear Programming Kit (GLPK) package is intended for solving large-scale Linear Programs, Mixed Integer Programs (MIP), and other related problems. It is a set of routines written in C and organized in the form of a callable library. It can be downloaded from \url{http://www.gnu.org/software/glpk/}, where installation instructions can be found.

As an example, we will present how to write up the following LP:
\begin{align*}
\max \quad & 4x_1 - 6x_2 + 3x_2   \\
\st \quad& x_1 + 10 x_2 + x_3 \leq 5 \\
& 2x_1 - x_2 = 0\\
& x_1,x_2, x_3 \geq 0
\end{align*}

All GLP API data types and routines are defined in a header that should be included in all source files:\\

\lstset{escapechar=@,style=customc}
\begin{lstlisting}[frame=single]
#include <glpk.h>
\end{lstlisting}

The problem object contains all the information of the LP, \ie the objective function, the constraint matrix, the parameters of the solver, etc. It can be initialized in the following way: \\

\lstset{escapechar=@,style=customc}
\begin{lstlisting}[frame=single]
glp_prob *lp;
lp = glp_create_prob(); // creates an empty problem
glp_set_prob_name(lp, "tracking");  // problem object name
\end{lstlisting}

The most important characteristic of the problem is whether it is a maximization or a minimization, which we can change with the function:\\

\lstset{escapechar=@,style=customc}
\begin{lstlisting}[frame=single]
glp_set_obj_dir(lp, GLP_MAX);  // minimization or maximization
\end{lstlisting}

Now we are ready to define the actual LP. An important property of the GLPK library is that the index $0$ is not used, therefore we will always start inputing information from index $1$.

The information of the objective function is included in the {\it columns} or {\it structural variables} of the problem object. 
The structural variables contain both the coefficients $\mathbf{c}$ of the LP as well as the limits of the variables $\mathbf{x}$, for example, the non-negativity constraints. In our example: \\

\lstset{escapechar=@,style=customc}
\begin{lstlisting}[frame=single]
glp_add_cols(lp, 3); // Create Columns: Structural variables 

glp_set_col_name(lp,1,"x1");
glp_set_col_bnds(lp, 1, GLP_UP, 0.0, 0.0); // we set  x_1 >= 0
glp_set_obj_coef(lp, 1, 4 ); // objective function coefficient c_1
	
glp_set_col_name(lp,1,"x2");
glp_set_col_bnds(lp, 1, GLP_UP, 0.0, 0.0); // we set  x_2 >= 0
glp_set_obj_coef(lp, 1, -6 ); // objective function coefficients c_2

glp_set_col_name(lp,1,"x3");
glp_set_col_bnds(lp, 1, GLP_UP, 0.0, 0.0); // we set  x_3 >= 0
glp_set_obj_coef(lp, 1, 3 ); // objective function coefficients c_3 
\end{lstlisting}

The values $\mathbf{b}$ of the constraints are set using the {\it rows} or {\it auxiliary variables} of the problem object. In our example, we have to represent one equality and one inequality: \\

\lstset{escapechar=@,style=customc}
\begin{lstlisting}[frame=single]
glp_add_rows(lp, 2); // Create Rows: Auxiliary variables

glp_set_row_name(lp, 1,"constraint1");
glp_set_row_bnds(lp, 1, GLP_FX, 0.0, 0.0);  // x_1+x_2+x_3 <= 5
    
glp_set_row_name(lp, 2,"constraint2");
glp_set_row_bnds(lp, 2, GLP_FX, 0.0, 0.0);  // 2x_1 - x_2 =0    		
\end{lstlisting}

Finally, we only need to define the matrix $\mathbf{A}$, which represents the coefficients of the variables in the constraints. This information is introduced using three variables, \url{ia} and \url{ja} which contain the indices $i$ and $j$ of each matrix element, and the corresponding \url{ar} which contains the actual value of the coefficient $a_{ij}$. \\

\lstset{escapechar=@,style=customc}
\begin{lstlisting}[frame=single]
double *ar;
int *ia,*ja;

ar=new double[constraintsize+1];
ia=new int[constraintsize+1];
ja=new int[constraintsize+1];

ia[1]=1; ja[1]=1; ar[1]=1;  /* a[1,1]=1 */
ia[2]=1; ja[2]=2; ar[2]=10;  /* a[1,2]=10 */
ia[3]=1; ja[3]=3; ar[3]=1;  /* a[1,3]=1 */
ia[4]=2; ja[4]=1; ar[4]=2;  /* a[2,1]=2 */
ia[5]=2; ja[5]=2; ar[5]=-1;  /* a[2,2]=-1 */
ia[6]=2; ja[6]=3; ar[6]=0;  /* a[2,3]=0 */

\end{lstlisting}

Once we have defined the whole LP, we are ready to proceed to the solver. \\

\lstset{escapechar=@,style=customc}
\begin{lstlisting}[frame=single]
glp_load_matrix(lp,constraintsize , ia2, ja2, ar2); // load the problem
	
glp_simplex(lp, &param); // Solve with Simplex!
\end{lstlisting}

We can obtain the objective value of the optimal solution as well as the values of $\mathbf{x}$:\\

\lstset{escapechar=@,style=customc}
\begin{lstlisting}[frame=single]
z = glp_get_obj_val(lp); 

x1= glp_get_col_prim(lp,1);
x2= glp_get_col_prim(lp,2);
x3= glp_get_col_prim(lp,3);
	
glp_delete_prob(lp); // clear problem from memory
\end{lstlisting}

There exists also a MEX package in order to use the GLPK library in MatLab. It can be downloaded here \url{http://glpkmex.sourceforge.net}. It can be used in a similar way as the native MatLab LP solver, which we explain next.

\subsection{MatLab}

MatLab provides a simple way to define and solve Linear Programs using the function \url{linprog}. The inputs are directly the matrices $\mathbf{A}$, $\mathbf{c}$ and $\mathbf{b}$. If the problem has some equalities, they can be specifically defined using $\mathbf{A_{eq}}$ and $\mathbf{b_{eq}}$, as shown in the example below which represents the same LP as the one in the previous section. \\

\lstset{escapechar=@,style=custommatlab}
\begin{lstlisting}[frame=single]
c=[4,-6,3];
b=5;
beq=0;
A=[1,10,1];
Aeq=[2,-1,0];
lb=[0,0,0];
ub=[inf,inf,inf];
[x,fval] = linprog(c,A,b,Aeq,beq,lb,ub); % Solve!
\end{lstlisting}

The values of the variables at the optimum can be found in \url{x} and the final objective value is \url{fval}, which are both outputs of the function \url{linprog}.


\chapter{Linear Programming for Tracking} 

\label{LPtracking} 

\fancyhead[RE,LO]{Chapter 4. \emph{Linear Programming for Tracking}} 

\graphicspath{{./Figures/SFM/}} 

We have seen in previous chapters that tracking is commonly divided into object detection and data association. First, objects are detected in each frame of the sequence and second, the detections are matched to form complete trajectories. 
In Chapter \ref{trackingbydetection}, we presented an introduction to several state-of-the-art detection methods. In this chapter, we focus on data association, which is the core of this thesis. 
We define the data association problem formally and describe how to convert it to a minimum-cost network flow problem, which can be efficiently solved using Linear Programming. 
The idea is to build a graph in which nodes represent pedestrian detections. These nodes are fully connected to past and future observations by edges, which determine the relation between two observations with a cost.
Thereby, the matching problem is equivalent to a minimum-cost network flow problem: finding the optimal set of trajectories is equivalent to sending flow through the graph so as to minimize the cost. This can be efficiently computed using the Simplex algorithm \citep{LP} or k-shortest paths \citep{networkflows} as presented in the previous chapter. 
In this chapter, we define the multiple object tracking problem using the Linear Programming formulation, which will be the basis for the contributions introduced in Chapters \ref{SFM} and \ref{multiview}.

\section{Related work: from local to global matching}

The data association problem deals with keeping the identity of tracked objects given available detections. False alarms and missed detections mainly due to occlusions are two sources of inaccuracies in the data association problem, and these become more apparent as the density of objects to be tracked is increased.
Typically, data association is performed on a frame-by-frame basis, predicting pedestrians' motion from one frame to the next with, e.g. Kalman Filter \citep{kalman} or particle filter \citep{particlefilter3,particlefilter2,particlefilter} and then matching them with the detections using, e.g. the Hungarian algorithm \citep{hungarian} or the Auction algorithm \citep{munkres}.
While this type of approach is very useful for real-time applications \citep{breitensteiniccv2009}, the matching decisions are made individually for each pedestrian and with only the information of the previous frame, which makes it difficult to distinguish targets in crowded environments and it is completely defenseless against occlusions. 
Joint particle approaches, such as the joint probabilistic data association filter (JPDAF) \citep{rasmussentpami2001}, can be used to make a joint motion prediction for all pedestrians at the same time. Sampling can be done using, e.g. Markov Chain Monte Carlo (MCMC) \citep{schikorabook2012,khantpami2005}, but matching is still limited to be frame-by-frame.

In order to include more information from previous frames, researchers have proposed several solutions: multi-hypothesis (MHT) approaches \citep{betkecvpr2007,esscvpr2008}, which extend the prediction of a pedestrian's motion to several frames, thereby creating several hypotheses of what path the pedestrian might have followed; solving the matching problem for a small fixed number of frames \citep{lealwmvc2009}; using Bayesian networks to reason about how trajectories split and merge \citep{nilliuscvpr2006}; or dealing with difficult matching situations, such as matching people in groups, using the Nash Equilibrium of game theory \citep{yangiccv2007}.
Nonetheless, for most of these techniques computational time increases exponentially as more and more frames and objects are taken into account, since the search space of hypotheses quickly grows.

In contrast, in \citep{berclazcvpr2006} an efficient approximative Dynamic Programming (DP) scheme was presented in which trajectories are estimated in succession. The advantage is that tracking each individual is done using the information of all frames. On the other hand, if a trajectory is formed using a certain detection, the other trajectories which are computed later will not be able to use that detection anymore. This obviously does not guarantee a global optimum for all trajectories. 

Recent works show that global optimization can be more reliable in crowded scenes, as it solves the matching problem jointly for all tracks. The multiple object tracking problem is defined as a linear constrained optimization flow problem and Linear Programming (LP) \citep{LP} is commonly used to find the global optimum. 
Linear Programming is widely used for Computer Vision applications such as 3D shape matching \citep{windheusereurographics2011, windheusericcv2011}, image segmentation \citep{schoenemanniccv2009} or pose estimation \citep{benezraeccv2000}.
The idea to use it for people tracking was first published in \citep{jiangcvpr2007}, although this method requires a priori the number of targets to track, which limits its application in real tracking situations. 
In \citep{berclaztpami2011}, the scene is divided into equally-sized cells, each represented by a node in the constructed graph.  Using the information from the Probability Occupancy Map, the problem is formulated either as a max-flow and solved with Simplex, or as a min-cost and solved using k-shortest paths, which is a more efficient solution.
In \citep{andriyenkoeccv2010}, the problem is also defined as a maximum flow on a hexagonal grid, but instead of matching individual detections, they make use of tracklets. 
There also exist continuous solutions which do not work with a discrete state space, \ie a finite set of possible detection locations, but with a continuous state space which provides a more accurate pedestrian location. In \citep{andriyenkocvpr2011} the authors propose a well-designed local optimization scheme in a continuous state space. Mixed solutions have also been presented \citep{andriyenkocvpr2012} where tracking is performed in the discrete domain but trajectory estimation is performed continuously.
In \citep{wucvpr2011}, global and local methods are combined to match trajectories across cameras and across time.

Finally, in \citep{lealiccv2011,zhangcvpr2008,pirsiavashcvpr2011} the tracking problem is formulated as a Maximum A-Posteriori (MAP) problem which is mapped to a minimum-cost network flow and then efficiently solved using LP. In this case, each node represents a detection, which means the graph is much smaller compared to \citep{berclaztpami2011,andriyenkoeccv2010}.
In this chapter we detail the graph construction and creation of the system of linear equations as proposed in \citep{lealiccv2011,zhangcvpr2008,pirsiavashcvpr2011}

\section{Multiple object tracking: Problem statement}

Let $\mathcal{O}=\{{\bf o}_j^t\}$ be a set of object detections with ${\bf o}_{j}^t=({\bf p}_j^t,t)$, where ${\bf p}_j^t=(x,y,z)$ is the 3D position and $t$ is the time stamp. A trajectory is defined as a list of ordered object detections $T_k=\{{{\bf o}_{k_1}^{t_1}},{{\bf o}_{k_2}^{t_2}}, \cdots , {{\bf o}_{k_N}^{t_N}}\}$ with $t_{1} \leq t_{2} \leq \ldots \leq t_{N}$ and the goal of multiple object tracking is to find the set of trajectories $\mathcal{T}*=\{T_k\}$ that best explains the detections. 

This is equivalent to finding the $\mathcal{T}$ that maximizes the a-posteriori probability given the set of detections $\mathcal{O}$, which is known as \emph{maximum posterior} or \emph{MAP} problem.  
\begin{align}
\mathcal{T}*=\underset{\mathcal{T}}\argmax \ P(\mathcal{T}|\mathcal{O})\label{eq:mapping}
\end{align}

Further assuming that detections are conditionally independent, the equation can be rewritten as:
\begin{align}
\mathcal{T}*=  \underset{\mathcal{T}} \argmax \ P(\mathcal{O} | \mathcal{T}) P(\mathcal{T})
 = \underset{\mathcal{T}}\argmax \ \prod\limits_j P({\bf o}_j | \mathcal{T}) P(\mathcal{T})
\label{eq:mapping2}
\end{align}

Optimizing Eq. \eqref{eq:mapping2} directly is intractable since the space of $\mathcal{T}$ is huge. Nonetheless, we make the assumption trajectories cannot overlap (\ie, a detection cannot belong to two trajectories) which allows us to treat each trajectory independently, and therefore decompose the equation as: 
\begin{align}
\label{eq:mapfinal}
\mathcal{T}*=  \underset{\mathcal{T}} \argmax \prod\limits_j P({\bf o}_j)  \prod_{T_k \in \mathcal{T}} P(T_k) 
\end{align}

where $P({\bf o}_j)$ is the likelihood of detection ${\bf o}_j$ and the trajectories are represented by a Markov chain:
\begin{align}
\label{eq:markov}
P(\mathcal{T})= \prod\limits_{T_k \in \mathcal{T}} P_\textrm{in}({\bf o}_{k_1}^{t_1})P({\bf o}_{k_2}^{t_2} |{\bf o}_{k_1}^{t_1}) \ldots P({\bf o}_{k_m}^{t_m}|{\bf o}_{k_{m-1}}^{t_{m-1}})\ldots P({\bf o}_{k_n}^{t_n}|{\bf o}_{k_{n-1}}^{t_{n-1}})P_\textrm{out}({\bf o}_{k_n}^{t_n}) 
\end{align}

where $P_\textrm{in}({\bf o}_{k_1}^{t_1})$ is the probability that a trajectory $k$ is initiated with detection ${\bf o}_{k_1}^{t_1}$, $P_\textrm{out}({\bf o}_{k_n}^{t_n})$ the probability that the trajectory is terminated at ${\bf o}_{k_n}^{t_n}$ and $P({\bf o}_{k_m}^{t_m}|{\bf o}_{k_{m-1}}^{t_{m-1}})$ is the probability that ${\bf o}_{k_{m-1}}^{t_{m-1}}$ is followed by ${\bf o}_{k_m}^{t_m}$ in the trajectory.

\section{Tracking with Linear Programming}
\label{TrackingLP}

In this section, we explain how to convert the MAP problem into a Linear Program, which is particularly interesting, since it can be efficiently solved in polynomial time, as explained in Chapter \ref{linearprogramming}.

Let us recall the definition of a linear programming problem. It consists in minimizing or maximizing a linear function in the presence of linear constraints which can be both equalities and inequalities. 
\begin{align}
\label{eq:lp1}
\textstyle
\textrm{Minimize }  \qquad & c_1 f_1 + c_2 f_2 + \ldots + c_n f_n 
\end{align}
\begin{align}
\label{eq:lp2}
\textrm{Subject to} \qquad &a_{11} f_1 + a_{12} f_2 + \ldots + a_{1n} f_n \geq b_1 \\
 \qquad & a_{21} f_1 + a_{22} f_2 + \ldots + a_{2n} f_n \geq b_2 \nonumber\\
& \quad \vdots  \qquad\quad \vdots \qquad\qquad\quad \vdots  \nonumber\\
 \qquad & a_{m1} f_1 + a_{m2} f_2 + \ldots + a_{mn} f_n \geq b_m \nonumber 
\end{align}
\\
where Eq. \eqref{eq:lp1} is the \emph{objective function} and Eq. \eqref{eq:lp2} represents the \emph{constraints}. $c_1, c_2, \ldots ,c_n$ denote the known \emph{cost coefficients} and $f_1,f_2, \ldots ,f_n$ are the \emph{decision variables} to be determined.

To convert our problem into a linear program, we linearize the objective function by defining a set of flow flags $f=\{f_\textrm{in}(i),f_\textrm{out}(i),f_\textrm{t}(i,j),f_\textrm{det}(i)\}$ which are limited to the values of $\{0,1\}$. 
Flow flag $f_\textrm{t}(i,j)$ is defined as:

\begin{align}
f_\textrm{t}(i,j) =
\begin{cases}
	1, & {\bf o}_i^{t_i} \mbox{ is directly followed by }  {\bf o}_j^{t_j} \mbox{ in a trajectory.}\\
	0, & \mbox{otherwise}
	\end{cases}
\end{align}

These edges are only allowed if $\Delta f \leq F_{\textrm{max}}$, where $\Delta f$ is the frame number difference between observations ${\bf o}_j^{t_j}$ and ${\bf o}_i^{t_i}$ and $F_{\textrm{max}}$ is the maximum allowed frame gap.
Flow flag $f_\textrm{det} (i)$ is defined as:

\begin{align}
f_\textrm{det} (i)=
\begin{cases}
	1, & {\bf o}_i^t  \mbox{ belongs to a trajectory, }  \\
	0, & \mbox{otherwise}
	\end{cases}
\end{align}

while $f_{\textrm{out}}(i)$ is:

\begin{align}
f_{\textrm{in}}(i) \mbox{ (or } f_{\textrm{out}}(i)\mbox{)}=
\begin{cases}
	1, & \mbox{a trajectory starts (or ends) at } {\bf o}_i^t. \\
	0, & \mbox{otherwise}
	\end{cases}
\end{align}

In a minimum cost network flow problem, the objective is to find the values of the variables that minimize the total cost of the flows through the network. Defining the costs as negative log-likelihoods, and combining Equations \eqref{eq:mapfinal} and \eqref{eq:markov}, the following objective function is obtained:
\begin{align}
\mathcal{T}*&=\underset{\mathcal{T}} \argmin \sum_{T_k \in \mathcal{T}} -\log P(T_k) - \sum_{j} \log P({\bf o}_j ) \\ 
&=\underset{{f}} \argmin \sum_{i} C_\textrm{in}(i)f_\textrm{in}(i)  +\sum_{i,j} C_\textrm{t}(i,j) f_\textrm{t}(i,j) \\\nonumber
& + \sum_{i} C_\textrm{det}(i)f_\textrm{det}(i)  + \sum_{i} C_\textrm{out}(i) f_\textrm{out}(i) \nonumber
\end{align}

subject to the following constraints: 

\begin{itemize}
\item{Edge capacities: assuming each detection can only correspond to one trajectory, the edge capacities have an upper bound of $1$. Furthermore, two conditions have to be fulfilled in order to make sure that, if an observation is active, this is either the start or end of a trajectory, or it is in the middle of a trajectory:}
\begin{align}
\textstyle
 f_{\textrm{in}}(i) + f_\textrm{det}(i) \leq 1  \qquad f_{\textrm{out}}(i)+ f_\textrm{det}(i) \leq 1 
 \label{eq:flow1}
\end{align}

\item{Flow conservation at the nodes:}
\begin{align}
\textstyle
 f_{\textrm{in}}(i) + f_\textrm{det}(i) =  \sum\limits_{j} f_\textrm{t}(i,j)  \qquad \quad \sum\limits_{j} f_\textrm{t}(j,i) = f_{\textrm{out}}(i)+ f_\textrm{det}(i)
 \label{eq:flow2}
 \end{align}
 
 \item{Exclusion property:}
\begin{align}
\textstyle
f \in \{0,1\}
\label{eq:flow3}
\end{align}

\end{itemize}

The condition in Eq. \ref{eq:flow3} requires to solve an integer program, which is known to be NP-complete. Nonetheless, we can relax the condition to have the following linear equation:
\begin{align}
\textstyle
0 \leq f \leq 1.
\label{eq:flow4}
\end{align}

Now the problem is defined and can be solved as a linear program. If certain conditions are fulfilled, namely that the constraint matrix $A$ is totally unimodular, as explained in Chapter \ref{linearprogramming}, the solution $\mathcal{T}*$ will still be integer, and therefore it will also be the optimal solution to the initial integer program. 
If the unimodularity condition is not fulfilled, as we will see for the graph structure of Chapter \ref{multiview}, we can always use branching \citep{networkflows} to transform fractional solutions into integers.

\begin{figure}[htbp] 
   \centering
   \includegraphics[width=0.8\linewidth]{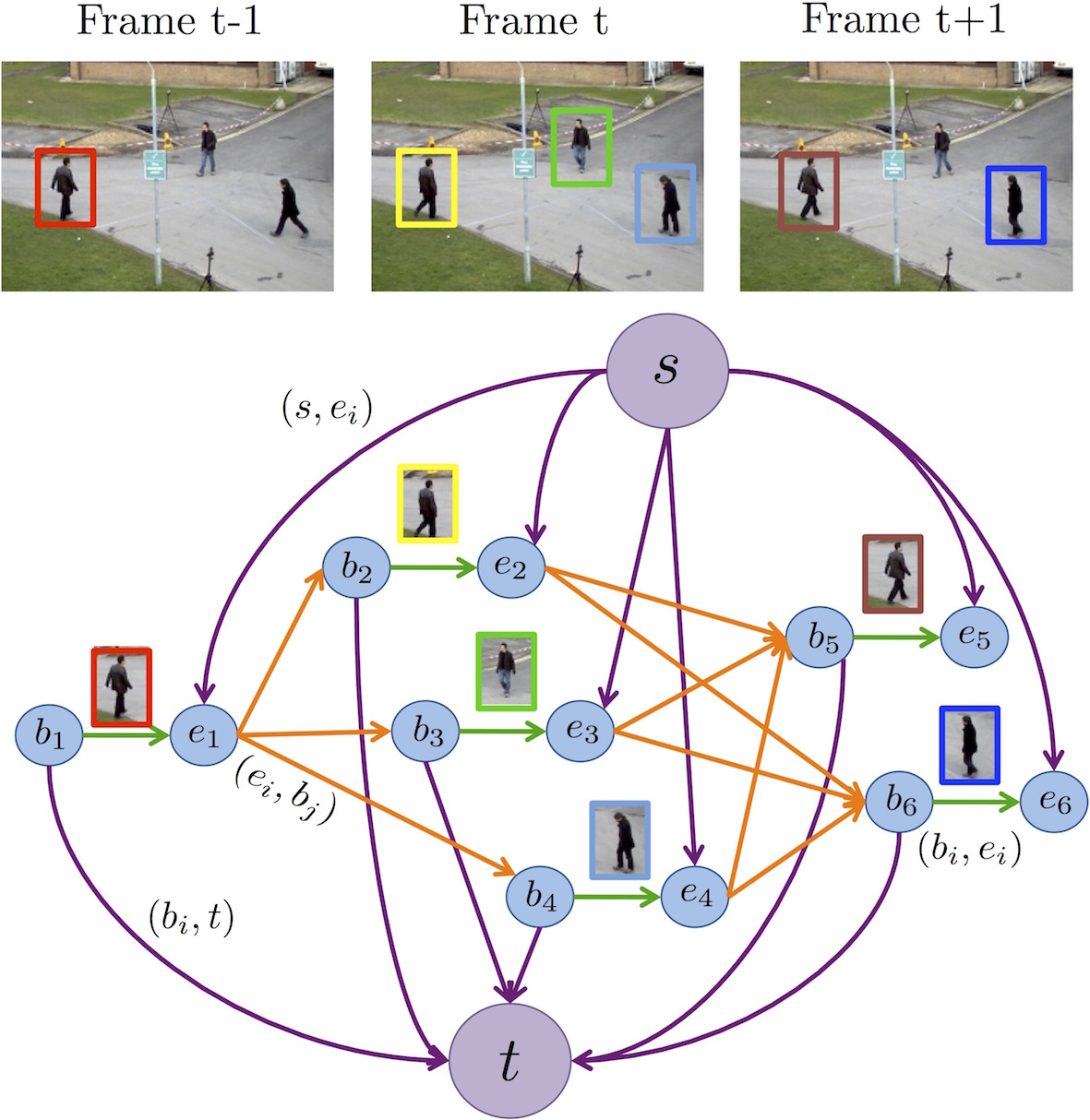} 
   \caption[Example of a graph model for multiple object tracking]{Example of a graph with the special source $s$ and sink $t$ nodes, 6 detections which are represented by two nodes each: the beginning $b_{i}$ and the end $e_{i}$.}
   \label{fig:network}
\end{figure}

\section{Graphical model representation}

To map this formulation to a cost-flow network, we define $G=(V,E)$ to be a directed network with a cost $C(i,j)$ and a capacity $u(i,j)$ associated with every edge $(i,j)\in E$, as explained in Chapter \ref{linearprogramming}.
An example of such a network is shown in Figure \ref{fig:network}; it contains two special nodes, the source $s$ and the sink $t$; all flow that goes through the graph starts at the $s$ node and ends at the $t$ node.
Thereby, each unit of flow represents a trajectory $T_k$, 
and the path that it follows indicates which observations belong to each $T_k$. Each observation ${\bf o}_i$ is represented with two nodes, the beginning node $b_i \in V$ and the end node $e_i \in V$ (see Figure \ref{fig:network}). A detection edge connects $b_i$ and $e_i$.

Below we detail the three types of edges present in the graphical model and the cost for each type:

\noindent{\bf Link edges.} The edges $(e_i,b_j)$ connect the end nodes $e_i$ with the beginning nodes $b_j$ in following frames, with cost $C_\textrm{t}(i,j)$. This cost represents the spatial relation between different subjects. Assuming that a subject cannot move a lot from one frame to the next, we define the costs to be an increasing function of the distance between detections in successive frames.
The time gap between observations is also taken into account in order to be able to work at any frame rate, therefore velocity measures are used instead of distances.
The velocities are mapped to probabilities with a Gauss error function as shown in Equation \eqref{eq:erf}, assuming the pedestrians cannot exceed a maximum velocity $V_{\textrm{max}}$. 

\begin{align}
\label{eq:erf}
E(V_{t},V_{\textrm{max}}) = \frac{1}{2} +  \frac{1}{2}\textrm{erf} \left( \frac{-V_{t} + \frac{V_{\textrm{max}}}{2}}{\frac{V_{\textrm{max}}}{4}}\right)
\end{align}

\begin{figure}[htpb] 
   \centering
   \includegraphics[width=0.6\linewidth]{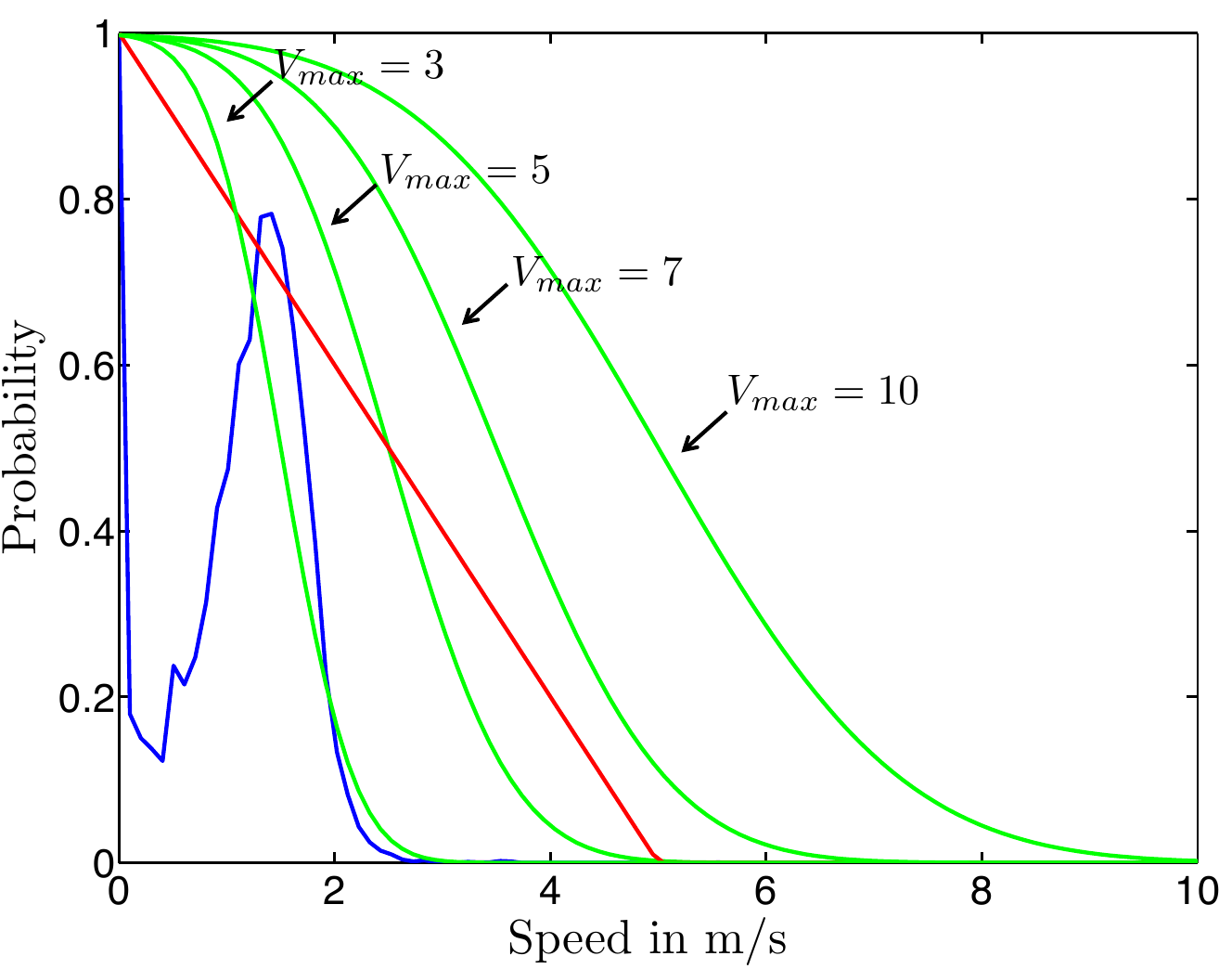} 
   \caption[Relationship between costs and pedestrian velocity]{\emph{Blue} = normalized histogram of speeds learned from training data. \emph{Red} = probability distribution if cost depends linearly on the velocity. \emph{Green} = probability distribution if the relation of cost and velocities is expressed by Equation \eqref{eq:erf}. A $V_{\textrm{max}}=$\unit[7]{m/s} is used in the experiments.}
   \label{fig:whyerf}
\end{figure}

As we can see in Figure \ref{fig:whyerf}, the advantage of using Equation \eqref{eq:erf} over a linear function is that the probability of lower velocities decreases more slowly, while the probability of higher velocities decreases more rapidly. This is consistent with the probability distribution of speed learned from training data (in our case, we use the two sequences in \citep{pellegriniiccv2009} to obtain the velocity distribution).

Therefore, the cost of a link edge is defined as:
\begin{align}
C_\textrm{t}(i,j)&=-\textstyle \log\left(\textstyle P({{\bf o}_j^{t_j} | {\bf o}_i^{t_i} }) \right) + C(\Delta f) \\ \nonumber
& = -\log E\left(\textstyle \frac{\|{\bf p}_j^{t+\Delta t} - {\bf p}_i^{t})\|}{\Delta t},V_{\textrm{max}}\right)  + \textstyle C(\Delta f)
\label{linkedge}
\end{align}

where $C(\Delta f) =- \log\left( B_j ^ {\Delta f - 1}\right)$ is the cost depending on the frame difference between detections. How high this cost is depends on parameter $B_j$ and its effects will be analyzed in section \ref{cost_bj_param}.

\noindent{\bf Detection edges.} The edges $(b_i,e_i)$ connect the beginning node $b_i$ and end node $e_i$, with cost $C_\textrm{det} (i)$. If all costs of the edges are positive, the solution to the minimum-cost problem is the trivial null flow. Consequently, we represent each observation with two nodes and a detection edge with negative cost:
\begin{align}
C_\textrm{det} (i)=\log \left(1-P_{det}({\bf o}_i^t) \right) + \log \left(\frac{\textrm{BB}_{\textrm{min}}}{\|{\bf p}_{\textrm{BB}} - {\bf p}_i^t\|}\right).
\end{align}

The higher the likelihood of a detection $P_{det}({\bf o}_i^t)$ the lower the negative the cost of the detection edge, hence, flow is likely to be routed through edges of confident detections in order to minimize the total cost.
If a map of the scene is available, we can also include this information in the detection cost. If a detection is far away from a possible entry/exit point, we add an extra negative cost to the detection edge, in order to favor the inclusion of that observation into a trajectory. The added cost depends on the distance to the closest entry/exit point ${\bf p}_{\textrm{BB}}$, and is only computed for distances higher than $\textrm{BB}_{\textrm{min}}=$\unit[1.5]{m}.
This is a simple probabilistic way of including other information present in the scene, such as obstacles or attraction points (shops, doors, etc).

\begin{figure*}[htbp]
\centering
\subfigure[]{
\includegraphics[width=0.495\linewidth]{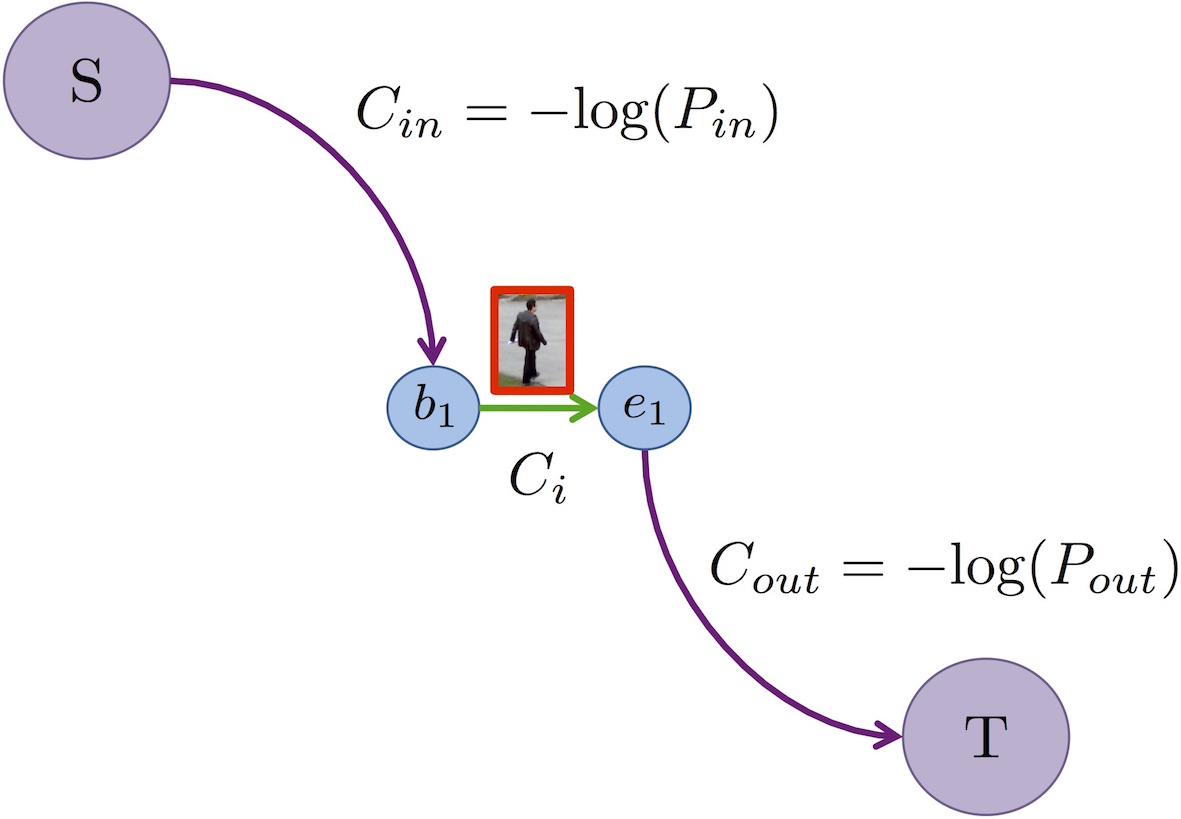} 
\label{change1}
}
\subfigure[]{
\includegraphics[width= 0.45\linewidth]{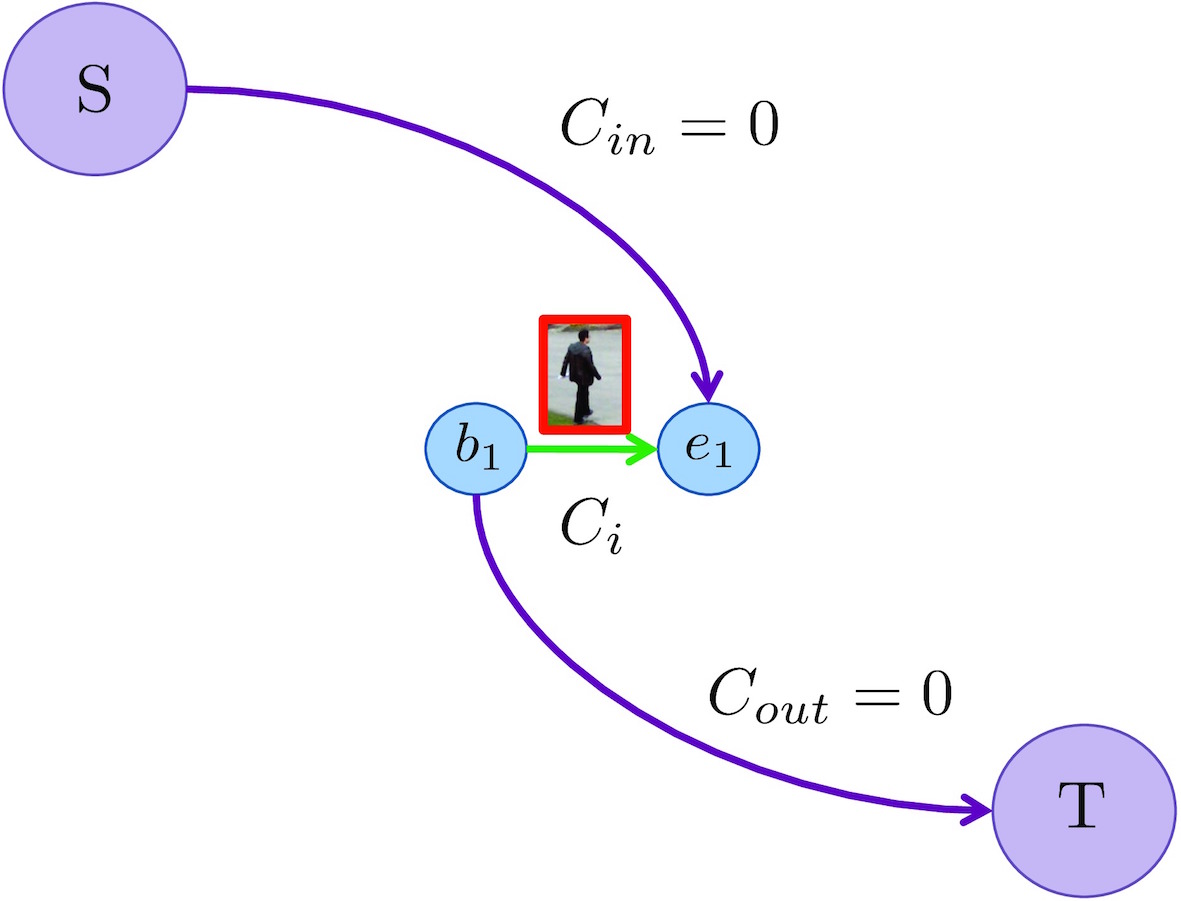} 
\label{change2}
}\caption[Different proposed approaches for source and sink connection]{(a) Graph structure as used in \citep{zhangcvpr2008}, which requires the computation of $P_\textrm{in}$ and $P_\textrm{out}$ in an Expectation-Maximization step during optimization.  (b) Graph structure as used in \citep{lealiccv2011} which does not require the computation of these two parameters; the trajectories are found only with the information of the link and detection edges.}
\label{fig:graph_change}
\end{figure*}

\noindent{\bf Entrance and exit edges.} The edges $(s,e_i)$ connect the source $s$ with all the end nodes $e_i$, with cost $C_{\textrm{in}}(i)$ and flow $f_{\textrm{in}}(i)$. Similarly, $(b_i,t)$ connects the end node $b_i$ with sink $t$, with cost $C_{\textrm{out}}(i)$.
This connection, as shown in Figure \ref{change2}, was proposed in \citep{lealiccv2011} so that when a track starts (or ends) it does not benefit from the negative cost of the detection edge. Setting $C_\textrm{in}=C_\textrm{out}=0$ and taking into account the flow constraints of Eqs. \eqref{eq:flow1} and \eqref{eq:flow2}, trajectories are only created with the information of link edges.

In contrast, the authors in \citep{zhangcvpr2008} propose to create the opposite edges $(s,b_i)$ and $(e_i,t)$, which means tracks entering and leaving the scene go through the detection node and therefore benefit from its negative cost (see Figure \ref{change1}).  
If the costs $C_\textrm{in}$ and $C_\textrm{out}$ are then set to zero, a track will be started at each detection of each frame, because it will be cheaper to use the entrance and exit edges than the link edges. 
On the other hand, if $C_\textrm{in}$ and $C_\textrm{out}$ are very high, it will be hard for the graph to create any trajectories. Therefore, the choice of these two costs is extremely important. In \citep{zhangcvpr2008}, the costs are set according to the entrance and exit probabilities $P_\textrm{in}$ and $P_\textrm{out}$, which are data dependent terms that need to be calculated during optimization.

\chapter{Tracking with social context} 

\label{SFM} 

\fancyhead[RE,LO]{Chapter 5. \emph{Tracking with social context}} 

\graphicspath{{./Figures/SFM/}}

If a pedestrian does not encounter any obstacles, the natural path to follow is a straight line. But what happens when the space gets increasingly crowded and the pedestrian can no longer follow the straight path?
Social interaction between pedestrians is especially important when the environment is crowded. 

Though each object can be tracked separately, recent works have proven that tracking objects jointly and taking their interaction into consideration can give much better results in complex scenes. Current research is mainly focused on two aspects to exploit interaction between pedestrians: the use of a global optimization strategy as presented in Chapter \ref{linearprogramming} and a social motion model \citep{SFM}. 
The focus of this chapter is to marry the concepts of global optimization and social and grouping behavior to obtain a robust tracker able to work in crowded scenarios.

\section{Related work: social forces}

Most tracking systems work with the assumption that the motion model for each target is independent. This simplifying assumption is especially problematic in crowded scenes: imagine the chaos if every pedestrian followed his or her chosen path and completely ignored other pedestrians in the scene. 
In order to avoid collisions and reach the chosen destination at the same time, a pedestrian follows a series of social rules or social forces. These have been defined in what is called the Social Force Model (SFM) \citep{SFM} which has been used for abnormal crowd behavior detection \citep{mehrancvpr2009}, crowd simulation \citep{pelechanoeurographics2007,pellegrinisim2012} and has only recently been applied to multiple people tracking. 

Most methods include these social forces or motion contexts in a predictive tracking framework. 
In \citep{scovannericcv2009}, an energy minimization approach was used to estimate the future position of each pedestrian considering all terms of the social force model. In \citep{pellegriniiccv2009} and \citep{lubericra2010}, the social forces were included in the motion model of the Kalman or Extended Kalman filter. 
In \citep{gewacv2009} a method was presented to detect small groups of people in a crowd, but it is only recently that grouping behavior has been included in a tracking framework \citep{choieccv2010,pellegrinieccv2010,yamaguchicvpr2011}. 

Predictive approaches though, are too local and unable to deal with trajectory changes (e.g. when people meet and stop to talk).
Recently, \citep{pellegrinieccv2010} included group information in a graphical model. Nonetheless, the structure created to express these group relations is a graph which contains cycles and, therefore, Dual Decomposition \citep{DD} was needed to find the solution, which obviously is computationally much more expensive than using Linear Programming. Moreover, the results presented in \citep{pellegrinieccv2010} were only for short time windows.
In \citep{buttcvpr2013} a solution is presented to include certain constant velocity conditions into a Linear Programming tracking framework. However, in that case the constraint matrix is no longer totally unimodular, so the authors propose to use Lagrangian relaxation in order to solve the problem. 
This kind of context information can also be extremely useful to track players in sports videos \citep{liucvpr2013}, given the great amount of interaction present in those sequences.

The authors of \citep{andriyenkoeccv2010} also define the problem as a maximum flow on a hexagonal grid, but instead of matching individual detections, they make use of tracklets. This has the advantage that they can precompute the social forces for each of these tracklets, nonetheless, the fact that the tracklets are chosen locally means the overall matching is not truly global, and if errors occur during the creation of the tracklets, these cannot be overcome by global optimization.
In \citep{wucvpr2011}, global and local methods are combined to match trajectories across cameras and across time.

In this chapter, we focus on the method presented in \citep{lealiccv2011} where tracking is done by taking the interaction between pedestrians into account in two ways: first, using global optimization for data association and second, including social as well as grouping behavior.
The key insight is that people plan their trajectories in advance in order to avoid collisions, therefore, a graph model which takes into account future and past frames is the perfect framework to include social and grouping behavior. 
The problem of multiple object tracking is formulated as a minimum-cost network flow problem as presented in Chapter \ref{linearprogramming}.
Instead of including social information by creating a complex graph structure which then cannot be solved using classic LP solvers, the method proposes an iterative solution relying on Expectation-Maximization.
Results on several challenging, public datasets are presented to show the improvement of tracking in crowded environments. Experiments with missing data, noise and outliers are also shown to test the robustness of the approach.

\section{The social force model}

The social force model states that the motion of pedestrians can be described as if they were subject to "social forces". These forces are not directly exerted by the pedestrians' personal environment, but they are a measure for the internal motivations of the individuals to perform certain actions, in this case, movements.
The idea is that there are certain sensory stimuli that cause a behavioral reaction that depends on personal aims. This reaction is chosen among all behavioral alternatives with the objective of utility maximization. 
In summary, one can say that a pedestrian acts as if he/she would be subject to a set of external forces.

\begin{figure*}[ht]
\centering
\subfigure[Constant velocity]{
\includegraphics[height=4.5cm]{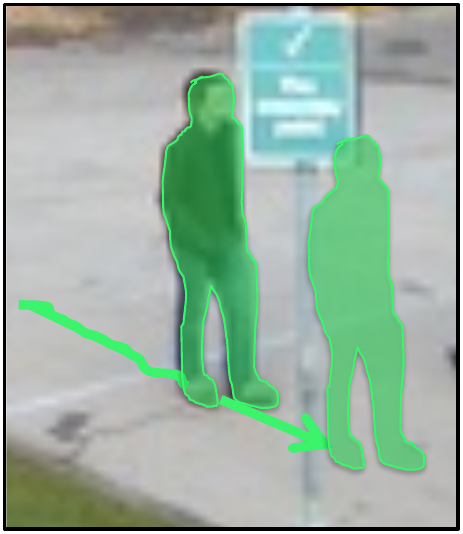} 
\label{fig:sfm1}
}
\subfigure[Collision avoidance]{
\includegraphics[height=4.5cm]{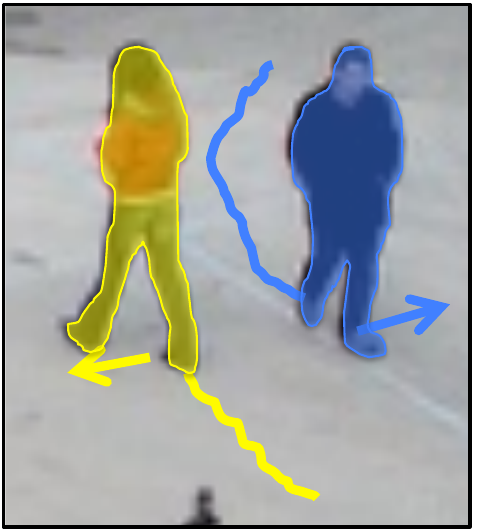} 
\label{fig:sfm2}
}
\subfigure[Group behavior]{
\includegraphics[height=4.5cm]{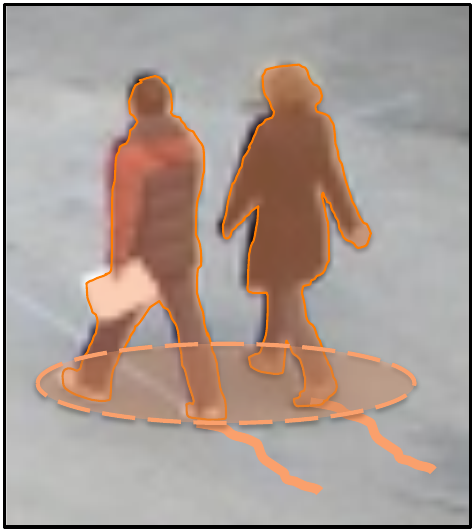} 
\label{fig:sfm3}
}
\caption{The three terms of the social force model that are included in the tracking framework}
\label{fig:sfm}
\end{figure*}

There are three main terms that need to be considered, visualized in Figure \ref{fig:sfm}:
\begin{itemize}
\item{{\it Constant velocity}: The acceleration of a pedestrian to keep a desired speed and direction.}
\item{{\it Collision avoidance}: The term reflecting that a pedestrian keeps a comfortable distance from other pedestrians and borders.} 
\item{{\it Group behavior}: The attraction forces which occur when a pedestrian is attracted to a friend, shop, etc.}
\end{itemize}

In this chapter we only consider the attractive effects of people within a group, since we do not consider any information about the static environment such as shops, entries/exists, etc. 
In contrast with \citep{pellegriniiccv2009}, we do not use the destination of the pedestrian as input, since we want to keep the tracking system as independent as possible from the environment.

In following sections we detail how to include this specific information into the Linear Programming multi-people tracking framework introduced in Chapter \ref{LPtracking}.

\section{Updated MAP and Linear Programming formulation}

The original social force model \citep{SFM} describes a physical system that estimates the position of a pedestrian in a continuous way, which has been successfully used for crowd simulation \citep{pelechanoeurographics2007,pellegrinisim2012}. Nonetheless, we use the social information within a different paradigm: in our Linear Programming system, we have a set of hypothetical pedestrian positions (in the form of nodes) and we apply the social forces to find out the probability of a certain match (\ie, a certain trajectory being followed by a pedestrian). 

When including social and grouping information in the Linear Programming formulation, we can no longer assume that the motion of each subject is independent, which means we have to deal with a much larger search space of $\mathcal{T}$. 

We extend this space by including the following dependencies for each trajectory $T_k$: 

\begin{itemize}
\item{Constant velocity assumption: the observation ${\bf o}_{k_m}^{t_m} \in T_k$ depends on the previous two observations $[{\bf o}_{k_{m-1}}^{t_{m-1}},{\bf o}_{k_{m-2}}^{t_{m-2}}]$ }
\item{Grouping behavior: If $T_k$ belongs to a group, the set of members of the group $\mathcal{T}_{k,\textrm{GR}}$ has an influence on $T_k$} 
\item{Avoidance term: $T_k$ is affected by the set of trajectories $\mathcal{T}_{k,\textrm{SFM}}$ which are close to $T_k$ at some point in time and do not belong to the same group as $T_k$}
\end{itemize}

The first and third dependencies are grouped into the SFM term. The sets $\mathcal{T}_{k,\textrm{SFM}}$ and $\mathcal{T}_{k,\textrm{GR}}$ are disjoint, \ie, for a certain pedestrian $k$, the set of pedestrians that have an attractive effect (the group to which pedestrian $k$ belongs) is different from the set of pedestrians that have a repulsive effect on pedestrian $k$. Therefore, we can assume that these two terms are independent and decompose $P(\mathcal{T})$ as:
\begin{align}
\label{eq:mapfinal_sfm}
P(\mathcal{T})&= \prod_{T_k \in \mathcal{T}} P(T_k   \cap \mathcal{T}_{k,\textrm{SFM}}\cap \mathcal{T}_{k,\textrm{GR}} ) \\
&= \prod_{T_k \in \mathcal{T}}  P(\mathcal{T}_{k,\textrm{SFM}} | T_k) P(\mathcal{T}_{k,\textrm{GR}} | T_k ) P(T_k) \nonumber
\end{align}

\begin{figure}[htbp] 
   \centering
   \includegraphics[width=0.7\linewidth]{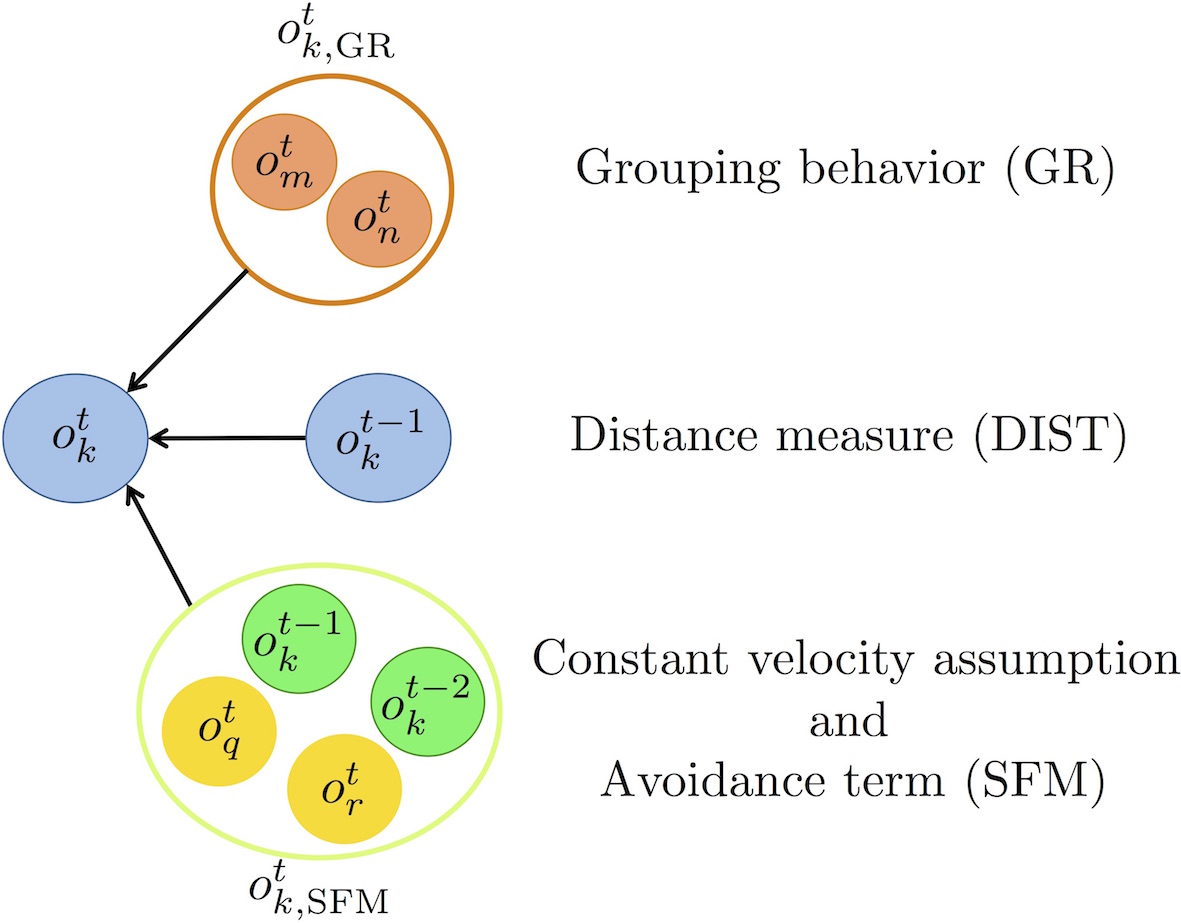} 
   \caption{Diagram of the dependencies for each observation ${\bf o}_{k}^t$. }
   \label{fig:dependencies}
\end{figure}

Let us assume that we are analyzing observation ${\bf o}_{k}^t$. In Figure \ref{fig:dependencies} we summarize which observations influence the matching of ${\bf o}_{k}^t$. Typical approaches \citep{zhangcvpr2008} only take into account distance (DIST) information, that is, the observation in the previous frame ${\bf o}_{k}^{t-1}$. We introduce the social dependencies (SFM) given by the constant velocity assumption (green nodes) and the avoidance term (yellow nodes). In this case, two observations, ${\bf o}_{q}^t$ and ${\bf o}_{r}^t$ that do not belong to the same group as ${\bf o}_{k}^t$, will be considered to create a repulsion effect on ${\bf o}_{k}^t$. On the other hand, the orange nodes which depict the grouping term (GR), are two other observations ${\bf o}_{m}^t$ and ${\bf o}_{n}^t$ which do belong to the same group as ${\bf o}_{k}^t$ and therefore have an attraction effect on ${\bf o}_{k}^t$. 
Note that all these dependencies can only be modeled by high order terms, which means that either we use complex solvers \citep{pellegrinieccv2010} to find a solution in graphs with cycles, or we keep the linearity of the problem by using an iterative approach as we explain later on.

The objective function is accordingly updated:
\begin{align}
\mathcal{T}*&=  \underset{\mathcal{T}} \argmax \ P(\mathcal{O} | \mathcal{T}) P(\mathcal{T}) \\ \nonumber
&=\underset{\mathcal{T}} \argmin \sum_{T_k \in \mathcal{T}} -\log P(T_k)  -\log P(\mathcal{T}_{\textrm{SFM}} | T_k )  \\ \nonumber
&  -\log P(\mathcal{T}_{\textrm{GR}} | T_k) + \sum_{j} -\log P({\bf o}_j ) \\ \nonumber
&=\underset{{f}} \argmin \sum_{i} C_{\textrm{in}}(i)f_{\textrm{in}}(i)  + \sum_{i} C_{\textrm{out}}(i)f_{\textrm{out}}(i) \\ \nonumber
&+\sum_{i,j} [ C_\textrm{t}(i,j)+ C_{\textrm{SFM}}(i,j) + C_{\textrm{GR}}(i,j) ] f_\textrm{t}(i,j) + \sum_{i} C_\textrm{det}(i)f_\textrm{det} (i)  
\end{align}

In the following section, we define the new cost terms according to the Social Force Model.

\section{New costs for the social terms}
\label{SFMsection}

\noindent{\bf Constant velocity assumption.} A pedestrian tries to keep a certain speed and direction, therefore we assume that at time $t+\Delta t$ we have the same speed as at time $t$ and we estimate the pedestrian's position accordingly.

\begin{equation}
{\bf \tilde{p}}_{\textrm{SFM},i}^{t+\Delta t}={\bf p}_i^{t} + {\bf v}_i^{t}\Delta t
\label{eq:cvpos}
\end{equation}


\noindent{\bf Avoidance term.} Pedestrians also try to avoid collisions and keep a comfortable distance from each other. This term is modeled as a repulsion field with an exponential distance-decay function with value $\alpha$ learned from training data.

\begin{align}
{ a}_i^{t+\Delta t}=\sum\limits_{g_m \neq g_i} \exp\left( - \frac{  \|{\bf \tilde{p}}_i^{t+\Delta t} - {\bf \tilde{p}}_m^{t+\Delta t} \| }{\alpha \Delta t} \right)
\label{avoidance_acceleration}
\end{align}


The constant velocity assumption is used to estimate the positions of all pedestrians at time $t+ \Delta t$. From these estimated positions, the repulsion acceleration they exert on each other can be computed as shown in Eq. \eqref{avoidance_acceleration}. For a pedestrian $i$, only non-members of his group ($g_m \neq g_i$) who are less than \unit[1]{m} away, that is $\| {\bf \tilde{p}}_i^{t+\Delta t} - {\bf \tilde{p}}_m^{t+\Delta t}\| \leq $ \unit[1]{m}, are used to compute the avoidance acceleration.

The estimation of the pedestrian's future position is computed using also the aforementioned avoidance acceleration term:
\begin{align}
{\bf \tilde{p}}_{\textrm{SFM},i}^{t+\Delta t}={\bf p}_i^{t} + ({\bf v}_i^{t} + {\bf a}_i^{t+\Delta t}\Delta t)\Delta t.
\label{eq:cv_pos}
\end{align}

To compute the cost of the edge connecting $(i,j)$, the distance between estimated position and real measurement is used:
\begin{align}
C_{\textrm{SFM}}(i,j) = -\log E\left(\frac{\|{\bf \tilde{p}}_{\textrm{SFM},i}^{t+\Delta t} - {\bf p}_j^{t+\Delta t}\|}{\Delta t},V_{\textrm{max}}\right)
\end{align}

where the function $E$ is detailed in Eq. \eqref{eq:erf}. 

\begin{figure*}[htbp]
\centering
\subfigure[Only distances (DIST)]{
\includegraphics[height=4.15cm]{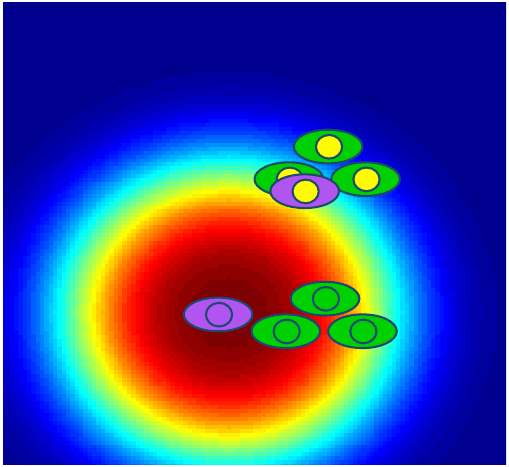} 
\label{prob1}
}
\subfigure[Only social force model term (SFM)]{
\includegraphics[height=4.15cm]{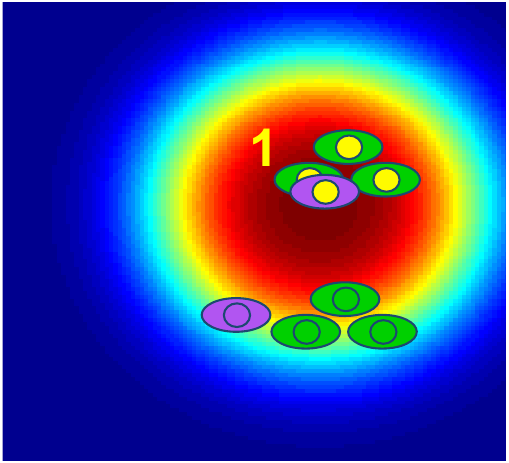} 
\label{prob2}
}
\subfigure[Only grouping term (GR)]{
\includegraphics[height=4.15cm]{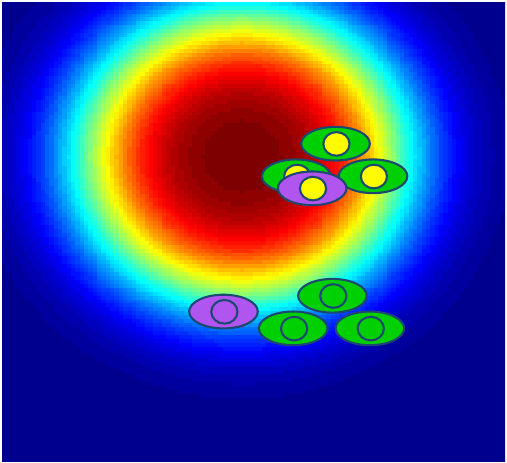} 
\label{prob3}
}
\subfigure[DIST+SFM]{
\includegraphics[height=4.15cm]{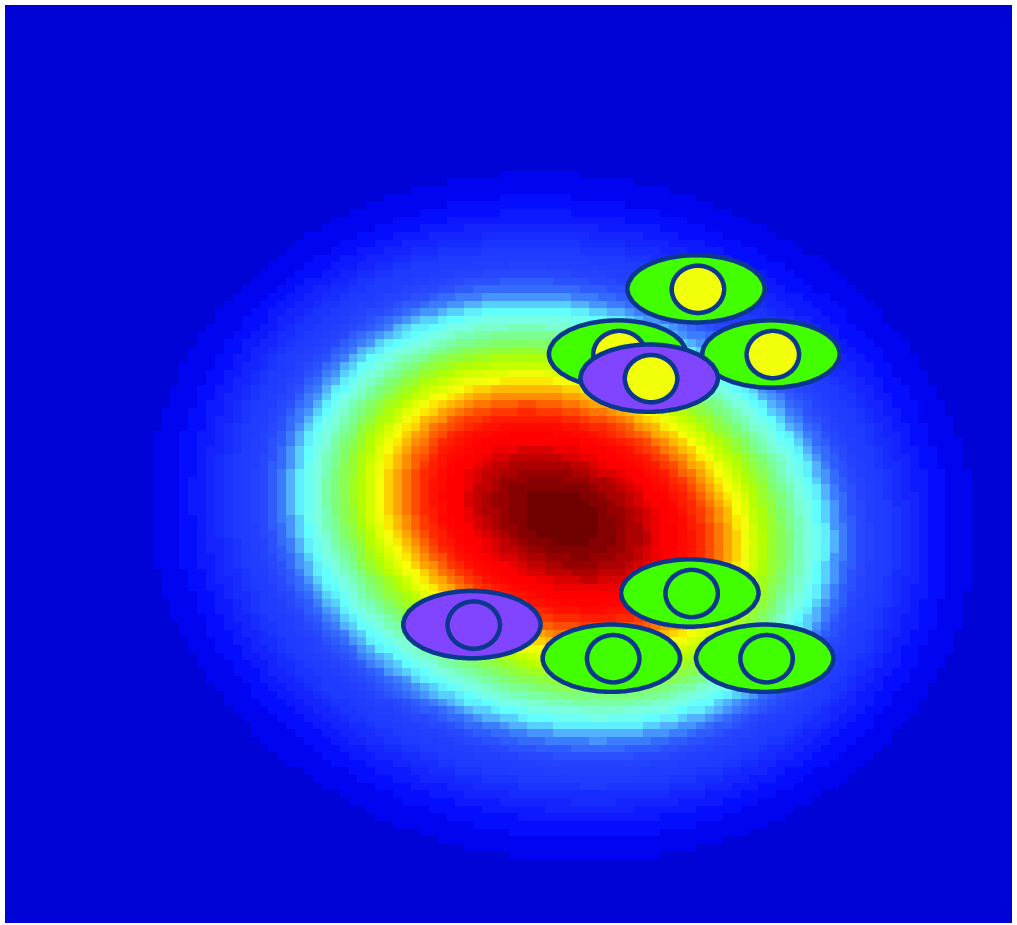} 
\label{prob4}
}
\subfigure[DIST+SFM+GR]{
\includegraphics[height=4.15cm]{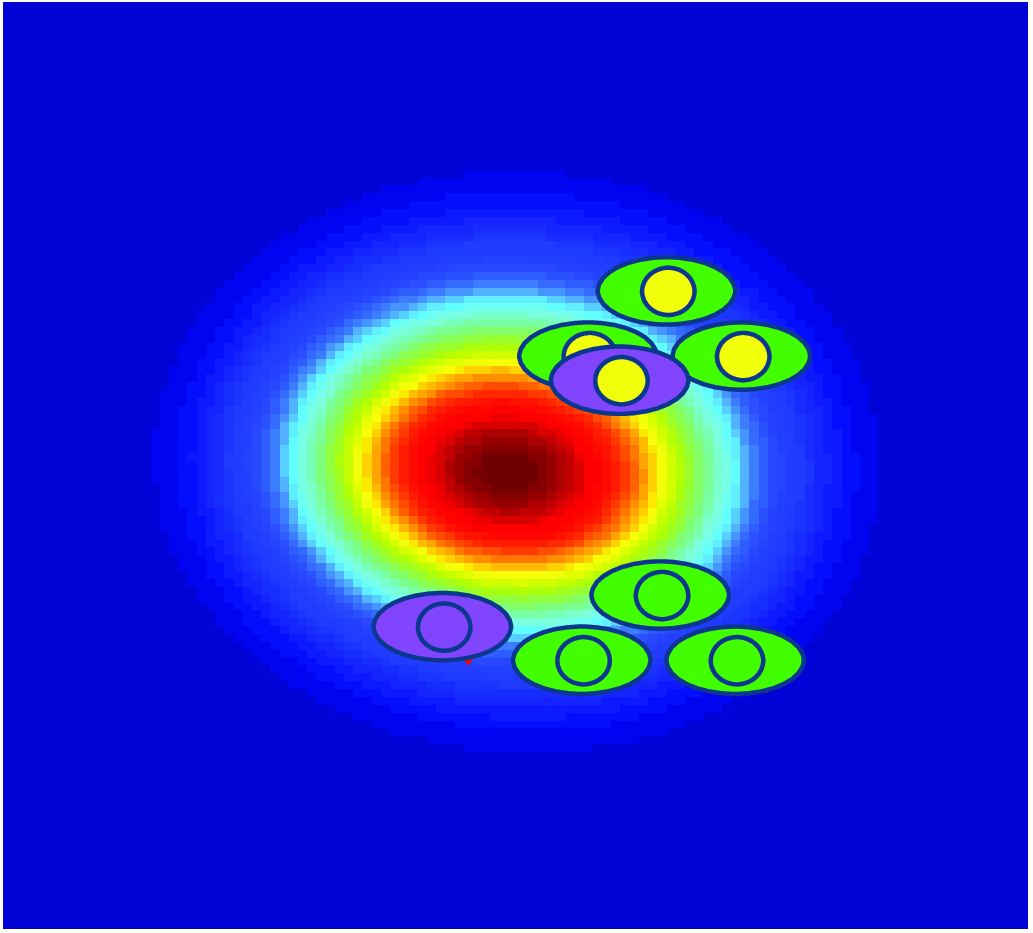} 
\label{prob5}
}
\caption[Reduction of the uncertainty of the pedestrian position using the social force model]{Three green pedestrians walk in a group, the estimated positions in the next frame are marked by yellow heads. The purple pedestrian's linearly estimated position (yellow head) clearly interferes with the trajectory of the group. Representation of the probability map (blue is 0 red is 1) for the purple pedestrian's next position using: (a) only distances, (b) only SFM (constant velocity assumption and avoidance term), (c) only GR (considering the purple pedestrian belongs to the group), (d) distances+SFM and (e) distances+SFM+GR.}
\label{fig:SFMandGROUP}
\end{figure*}

In Figure \ref{fig:SFMandGROUP} we plot the probability distributions computed using different terms. Note, this is just for visualization purposes, since we do not compute the probability for each point on the scene, but only for the positions where the detector has fired. 
There are 4 pedestrians in the scene, the purple one and 3 green ones walking in a group. As shown in \ref{prob2}, if we only use the estimated positions (yellow heads) given the previous speeds, there is a collision between the purple pedestrian and the green marked with a 1. The avoidance term shifts the probability mode to a more plausible position.

\noindent{\bf Grouping behavior.} 
Before modeling group behavior, we need to determine which tracks form each group and at which frame the group begins and ends (to deal with splitting and formation of groups). 
The idea is that if two pedestrians are close to each other over a determined period of time, they are likely to belong to the same group.
%
From the training sequence in \citep{pellegriniiccv2009}, the distance and speed probability distributions of the members of a group $P_g$ \vs individual pedestrians $P_i$ are learned. 
If $m$ and $n$ are two trajectories which appear on the scene at $t \in [0,N]$, we compute the flags $g_m$ and $g_n$ which indicate to which groups do $m$ and $n$ belong. 
If $\sum\limits_{t=0}^{N} P_g(m,n) > \sum\limits_{t=0}^{N} P_i(m,n)$, then $g_m=g_n$.


Therefore, for every observation ${\bf o}_{i}^t$, we will have a group label $g_i$ which indicates to which group the observation belongs, if any.
If several pedestrians form a group, they tend to keep a similar speed, therefore, if ${\bf o}_i^t$ belongs to a group, we can use the mean speed of all the other members of the group to estimate the next position for ${\bf o}_i^t$:

\begin{align}
{\bf \tilde{p}}_{\textrm{GR},i}^{t+\Delta t}={\bf p}_i^{t} + \frac{1}{|\{ m | g_m= g_i \} |} \sum\limits_{\{ m | g_m= g_i \}} {\bf v}_m^{t} \Delta t
\label{eq:group_pos}
\end{align}

The distance between this estimated position and real measurements is used in \eqref{eq:erf} to obtain the edge costs for the grouping term:
\begin{align}
C_{\textrm{GR}}(i,j) = -\log E\left(\frac{\|{\bf \tilde{p}}_{\textrm{GR},i}^{t+\Delta t} - {\bf p}_j^{t+\Delta t}\|}{\Delta t},V_{\textrm{max}}\right)
\end{align}

An example is shown in Figure \ref{prob3}, where we can see that the maximum probability provided by the group term keeps the group configuration.
In Figure \ref{prob4} we show the combined probability of the distance and SFM information which narrows the space of probable positions. Finally, Figure \ref{prob5} represents the combined probability of DIST, SFM and GR.
As we can see, the space of possible locations for the purple pedestrian is considerably reduced, as we add the social and grouping behaviors, which means we have less ambiguities for data association. This is specially useful to decrease the number of identity switches, as we present in Section \ref{experimental}.

\section{Optimization}
\label{implementation}

To compute the SFM and grouping costs, we need to have information about pedestrians' velocities, which can only be obtained if we already have the trajectories. 
We solve this in an expectation-maximization (EM) fashion where the parameters to estimate are the flow flags $f_i$ and the latent variables are the velocities and group flags. 
The proposed solver is presented in Algorithm \ref{alg1}; on the first iteration, trajectories are estimated only with the information defined in Section \ref{TrackingLP}, while for the rest of iterations, the SFM and GR is also used. The algorithm stops when the trajectories do not change or when a maximum number of iterations $M_i$ is reached.

\begin{algorithm}     
\small           
\caption{Iterative optimization}          
\label{alg1}                           
\begin{algorithmic}                    
\vspace{0.13cm}
\WHILE{\quad $\mathcal{T}_i\not=\mathcal{T}_{i-1} \quad \mbox{and} \quad i \leq M_i$ \quad}
\vspace{0.13cm}
\IF{$i==1$}
\vspace{0.13cm}
\STATE{1.1. Create the graph using only DIST information}
\vspace{0.13cm}
\ELSE
\vspace{0.13cm}
\STATE{1.2. Create the graph using DIST, SFM and GR information}
\vspace{0.13cm}
\ENDIF
\vspace{0.13cm}
\STATE{2. Solve the graph to find $\mathcal{T}_i$}
\vspace{0.13cm}
\STATE{3. Compute velocities and groups given $\mathcal{T}_i$}
\vspace{0.13cm}
\ENDWHILE
\end{algorithmic}
\end{algorithm}

Typically, only $4-6$ iterations are needed for the algorithm to converge to a solution.

\subsection{Computational reduction}

To reduce the computational cost, the graph can be pruned by using the physical constraints represented by the edge costs. If any of the costs $C(i,j)$, $C_{\textrm{SFM}}(i,j)$ or $C_{\textrm{GR}}(i,j)$ is infinite, the two detections $i$ and $j$ are either too far away to belong to the same trajectory or they do not match according to social and grouping rules, therefore  the edge $(i,j)$ is erased from the graphical model. 
For long sequences, the video can be divided into several batches and optimized for each batch. For temporal consistency, the batches have an overlap of $F_{\textrm{max}}=10$ frames. 
The runtime of \citep{lealiccv2011} for a sequence of 800 frames (114 seconds), 4837 detections, batches of 100 frames and 6 iterations is 30 seconds on a 3GHz machine.

\section{Experimental results}
\label{experimental}

In this section we show the tracking results of several state-of-the-art methods on three publicly available datasets and compare them using the CLEAR metrics \citep{clear}, explained below.

\subsection{Metrics used for performance evaluation}

The CLEAR metrics were presented in \citep{clear} for detection and tracking of both single objects as well as multiple objects. The framework includes guidelines for ground truth annotation, performance metrics, evaluation protocols, and tools including scoring software and baseline algorithms. 
The scores for multiple people tracking are computed in 2D using pedestrian bounding boxes. They are split into \emph{accuracy} and \emph{precision}:

{\bf Detection Accuracy (DA)}. Measures how many detections were correctly found and therefore is based on the count of missed detections $m_t$ and false alarms $f_t$ for each frame $t$.
\begin{align*}
DA = 1-\frac{\sum_{t=1}^{N_{f}} m_t + f_t}{\sum_{t=1}^{N_{f}} N_G^t}
\end{align*}
where $N_f$ is the number of frames of the sequence and $N_G^t$ is the number of ground truth detections in frame $t$. A detection is considered to be correct when the 2D bounding boxes of both ground truth and detection have some overlap. In this thesis, the overlap measure that we use is 25\% which is the standard measure taken in most of the literature.

{\bf Tracking Accuracy (TA)}. Similar to DA but also including identity switches $i_t$. In this case, the measure does not penalize identity switches as much as  missing detections or false alarms, as we use a $\mbox{log}_{10}$ weight. That is why in most papers the number of identity switches is explicitly shown in order to better compare performance with other methods.
\begin{align*}
TA = 1-\frac{\sum_{t=1}^{N_{f}} m_t + f_t + \mbox{log}_{10} (1+ i_t)}{\sum_{t=1}^{N_{f}} N_G^t}
\end{align*}

{\bf Detection Precision (DP)}. Precision measurements represent how well bounding box detections match the ground truth. For this, an overlap measure between bounding boxes is used:
 \begin{align*}
\textrm{Ov}^t=\sum\limits_{i=1}^{N^t_{\textrm{mapped}}}\frac{\left| G_i^t \cap D_i^t \right|}{\left| G_i^t \cup D_i^t \right|}
\end{align*}
where $N^t_{\textrm{mapped}}$ is the number of mapped objects in frame $t$, \ie, the number of detections that are matched to some ground truth object. $G_i^t$ is the $i$th ground truth object of frame $t$ and $D_i^t$ the detected object matched to $G_i^t$.
The DP measure is then expressed as:
\begin{align*}
DP = \frac{\sum\limits_{t=1}^{N_{f}} \frac{\textrm{Ov}^t}{N^t_{\textrm{mapped}}}}{N_f}
\end{align*}

{\bf Tracking Precision (TP)}. Measures the spatiotemporal overlap between ground truth trajectories and detected ones, considering also split and merged trajectories.
\begin{align*}
TP = \frac{ \sum\limits_{i=1}^{N^t_{\textrm{mapped}}} \sum\limits_{t=1}^{N_{f}} \frac{\left| G_i^t \cap D_i^t \right|}{\left| G_i^t \cup D_i^t \right|} }{\sum\limits_{t=1}^{N_{f}} N^t_{\textrm{mapped}}}
\end{align*}

\subsection{Analysis of the effect of the parameters}
\label{params}

All parameters defined in previous sections are learned from training data using one sequence of the publicly available dataset \citep{pellegriniiccv2009}.
In this section we study the effect of the few parameters needed in \citep{lealiccv2011} and show that the method works well for a wide range of values and, therefore, no parameter tuning is needed to obtain good performance. The analysis is done on two publicly available datasets: a crowded town center \citep{benfoldcvpr2011} and the well-known PETS2009 dataset \citep{pets2009}, to observe the different effects of each parameter on each dataset.
In order to apply the Social Force Model to pedestrians, we need their 3D position in world coordinates. Since all pedestrians walk on a 2D ground plane, we can transform the 2D image coordinates to 3D real world coordinates (with z=0) using a simple homography \citep{multiviewbook}. We use the calibration provided with each dataset.

\begin{figure*}[htbp]
\centering
\subfigure[TownCenter: iterations $M_i$]{
\includegraphics[width=0.46\linewidth]{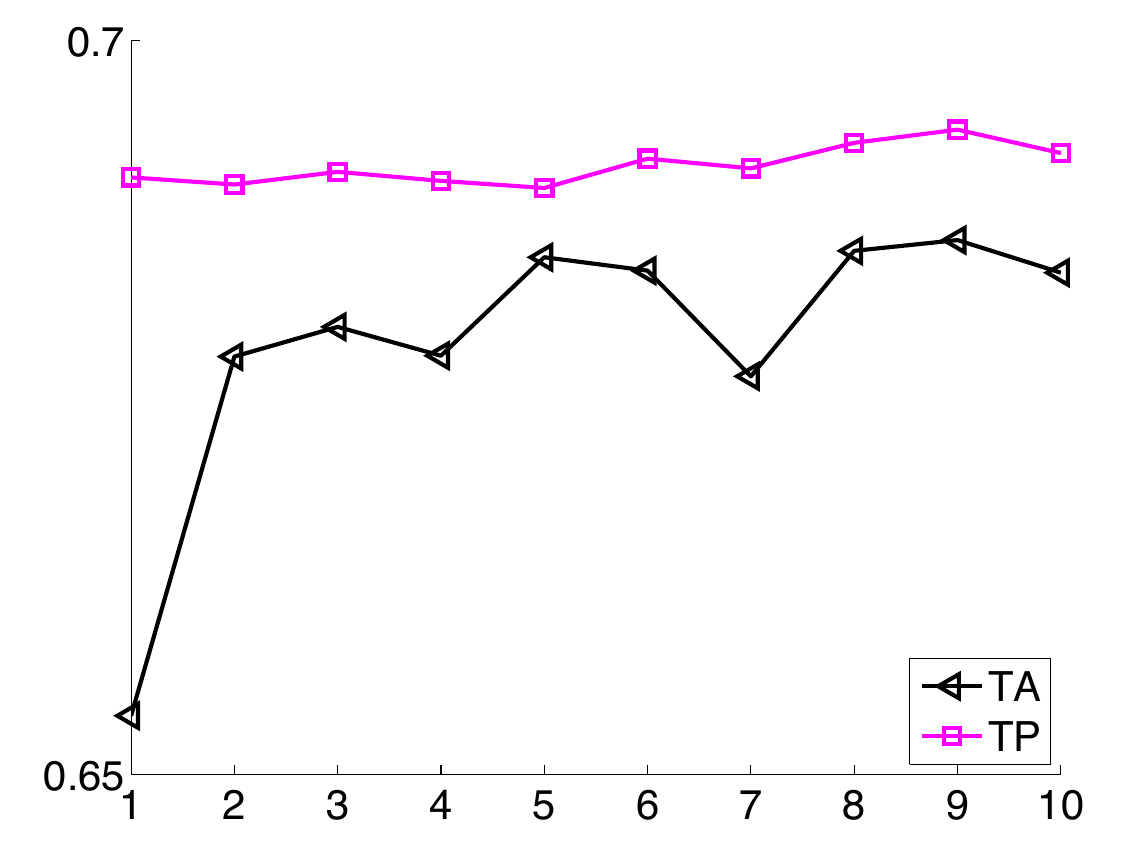} 
\label{fig:iter}
}\quad
\subfigure[PETS2009: iterations $M_i$]{
\includegraphics[width=0.46\linewidth]{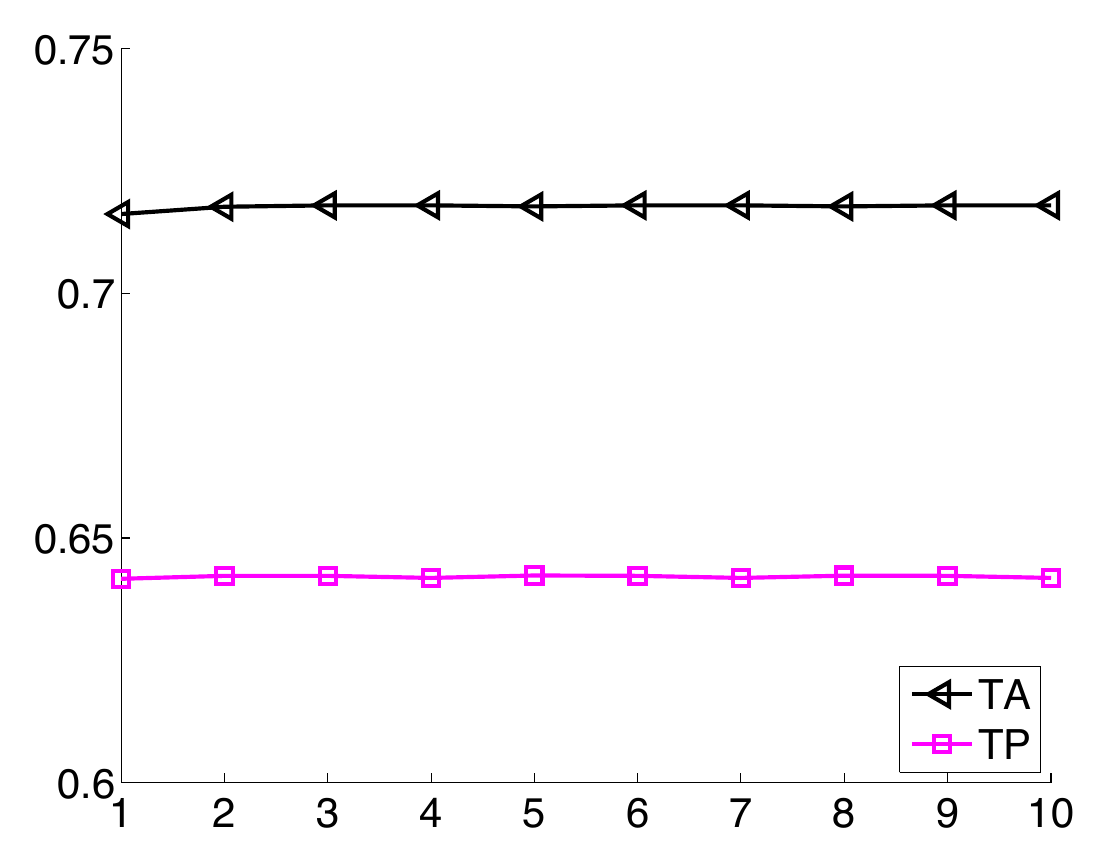} 
\label{fig:iterP}
}
\subfigure[TownCenter: maximum speed $V_\textrm{max}$]{
\includegraphics[width=0.46\linewidth]{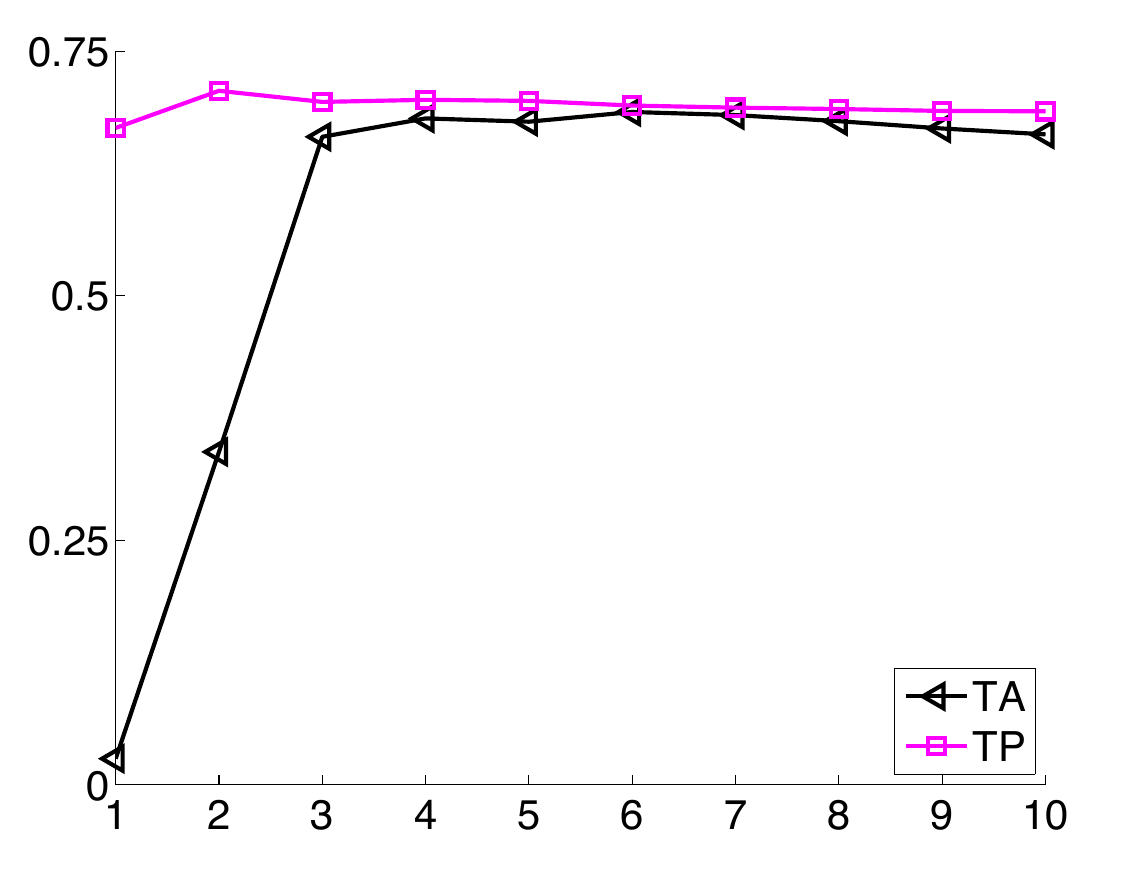} 
\label{fig:maxspeed}
}\quad
\subfigure[PETS2009: maximum speed $V_\textrm{max}$]{
\includegraphics[width=0.46\linewidth]{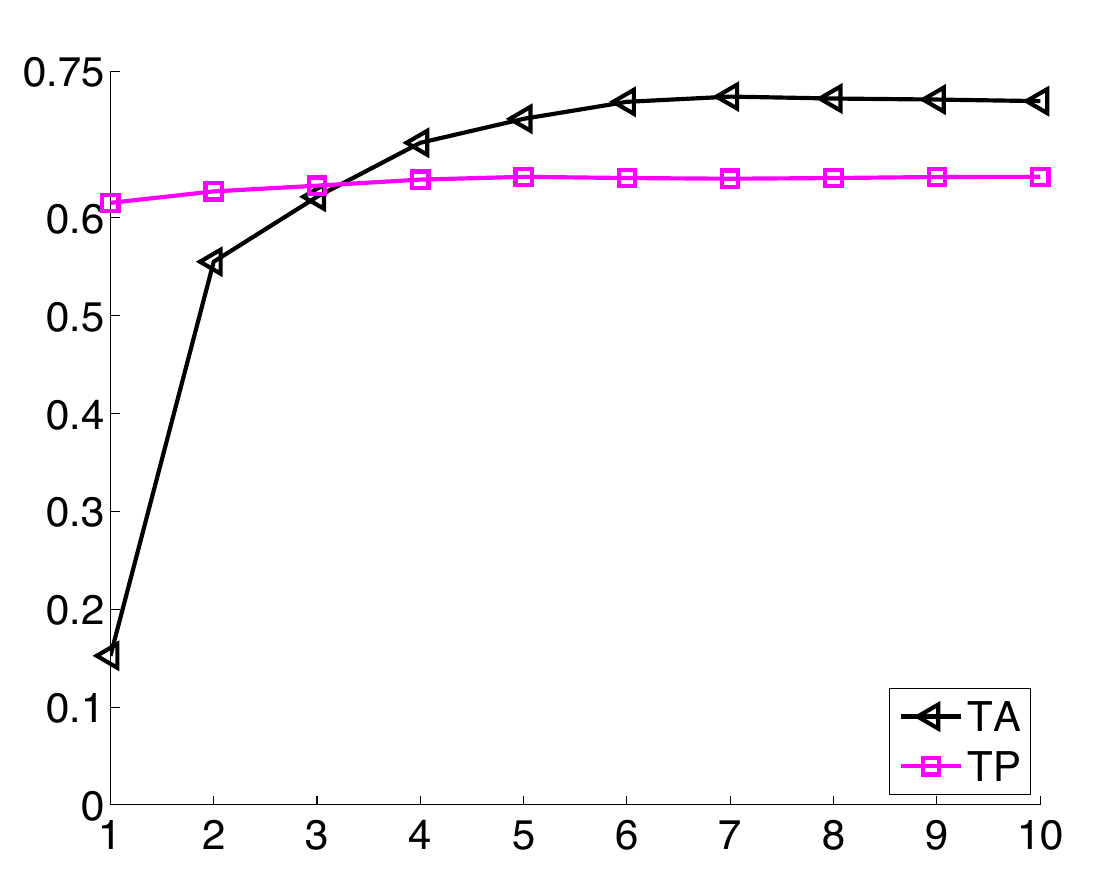} 
\label{fig:maxspeedP}
}
\subfigure[TownCenter: $B_j$]{
\includegraphics[width=0.46\linewidth]{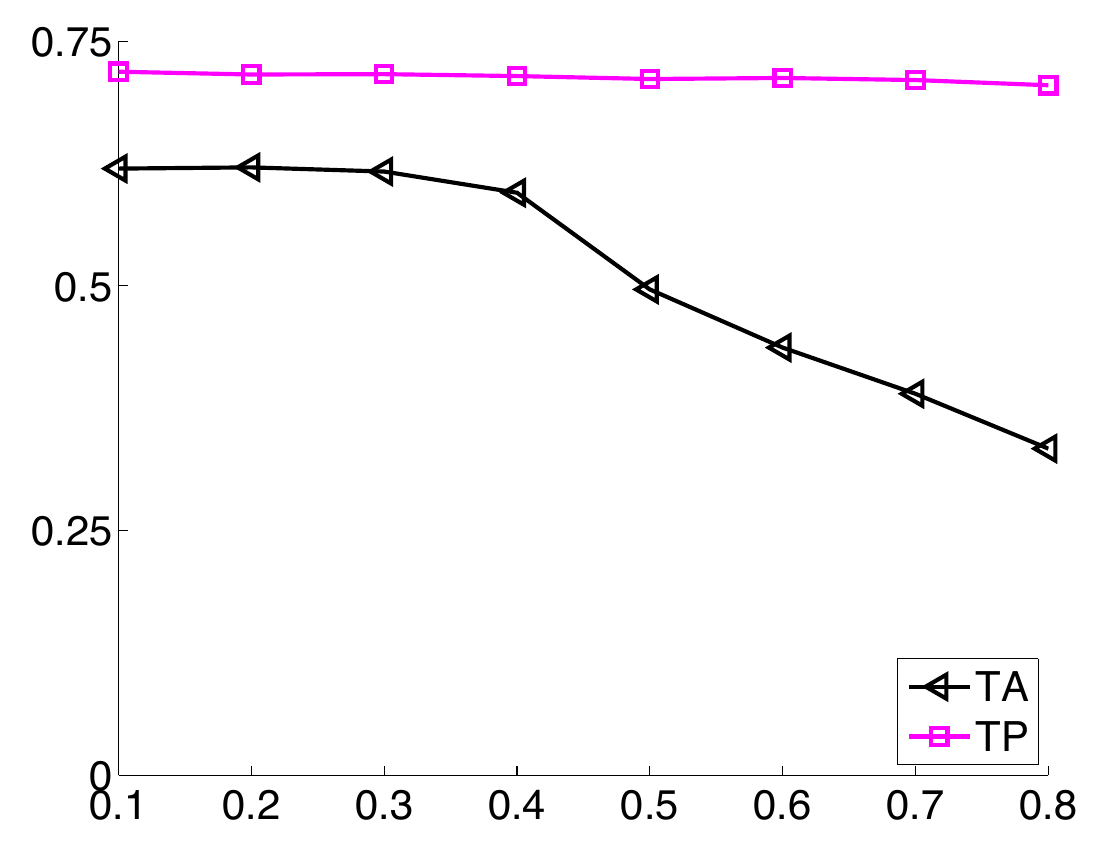} 
\label{fig:base}
}\quad
\subfigure[PETS2009: $B_j$]{
\includegraphics[width=0.46\linewidth]{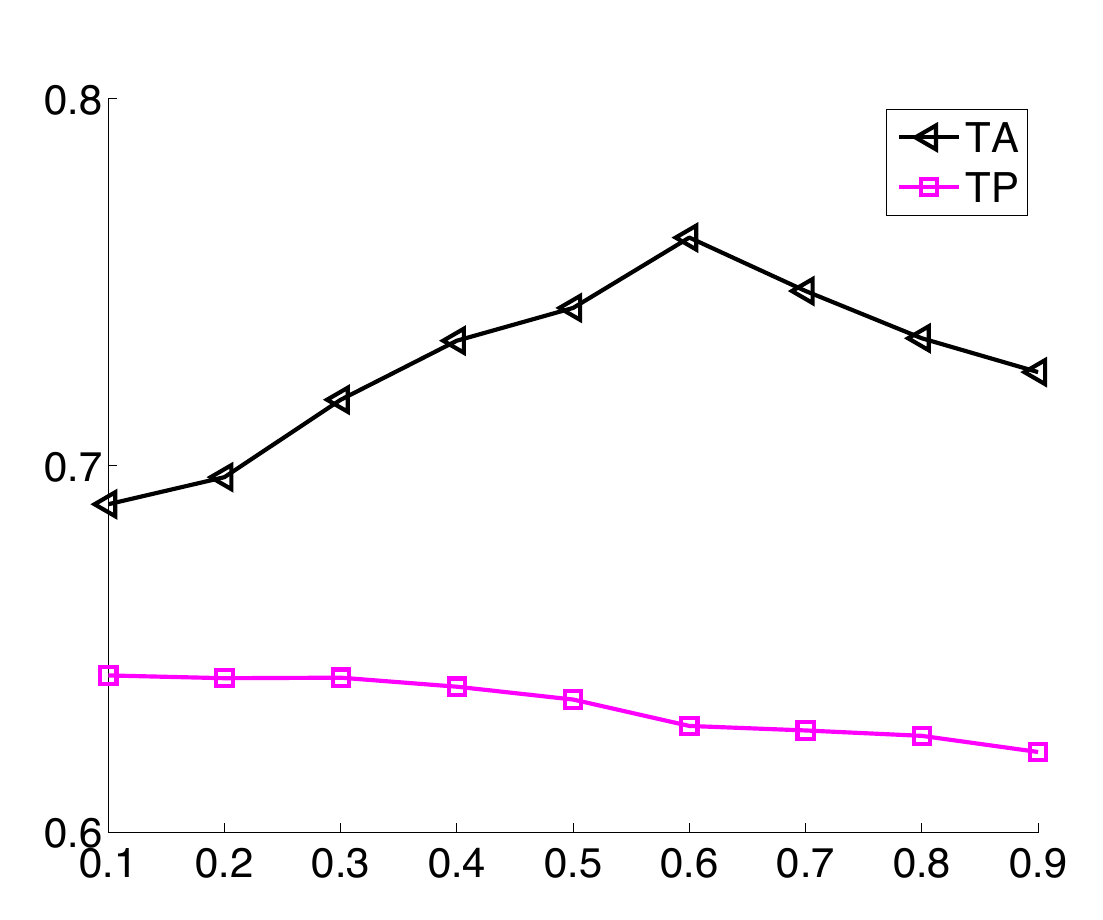} 
\label{fig:baseP}
}
\caption[Effect of the parameters on the tracking accuracy and precision]{Tracking accuracy (black) and precision (magenta) obtained for the Town Center dataset (left column) and the PETS 2009 dataset (right column) given varying parameter values.}
\label{fig:params}
\end{figure*}

\subsubsection{Number of iterations} The first parameter we analyze is $M_i$, the number of iterations allowed. This determines how many times the loop of computing social forces and trajectories is executed as explained in Algorithm \ref{alg1}. 
Looking at the results on the PETS 2009 dataset in Figure \ref{fig:iterP}, we can see that after just 2 iterations the results remain very stable. Actually, the algorithm reports no changes in the trajectories after 3 iterations, and therefore stops even though the maximum number of iterations allowed is higher. The result with 1 and 2 iterations is not very different either, which means the social and grouping behavior does not effect significantly the results for this particular dataset.
This is due to the fact that this dataset is very challenging from a social behavior point of view, with subjects often changing direction and groups forming and splitting frequently. More details and comments on these results can be found in Section \ref{pets2009}.
We observe a different effect on the TownCenter dataset, shown in Figure \ref{fig:iter}. In this case, there is a clear improvement when using social and grouping behavior (\ie, the result improves when we use more than one iteration). We also observe a pattern on how the Tracking Accuracy of the dataset evolves: there is a cycle of 3 iterations for which the accuracy increases and decreases in a similar way. 
This means that the algorithm is jumping between two solutions and will not converge to either one of them. This happens when pedestrians are close together for a long period of time but are not forming a group, which means that even with social forces it is hard to say which paths they will follow.

\subsubsection{Maximum speed} This is the parameter that determines the maximum speed we assume for the pedestrians we are observing. In this case, we can see in Figures \ref{fig:maxspeed} and \ref{fig:maxspeedP} a clear trend in which the results are very bad when we underestimate the pedestrians maximum speed, since we are artificially splitting trajectories. The results converge when the maximum speed allowed is between \unit[3]{m/s} - \unit[7]{m/s}, which makes sense since the reported mean speed of pedestrians in a normal situation is around \unit[2]{m/s}.
More interestingly, we observe that the results remain constant when using higher maximum speed values. This is a positive effect of the global optimization framework, since we can use a speed limit much above average and this will still give us good results and will allow us to track, for example, a person running through the scene.

\subsubsection{Cost for the frame difference} 
\label{cost_bj_param}
\begin{figure*}[htbp]
\centering
\includegraphics[width=1\linewidth]{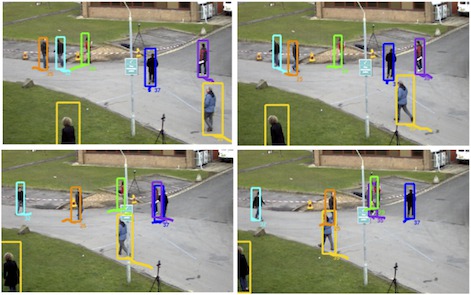} 
\caption[Tracking results under several occlusions]{Four frames of the PETS2009 sequence (separation of 9 frames), showing several occlusions, both created by the obstacle on the scene and among pedestrians. All occlusions can be overcome with the proposed method.}
\label{fig:pets_occlude}
\end{figure*}

The last parameter, $B_j$, appears in Eq. \eqref{linkedge} and represents the penalty term we apply when the frame difference between two detections that we want to match is larger than 1. 
This term is used in order to give preference to matches that are close in time. Here again we can see different effects on the two datasets. In Figure \ref{fig:base}, we see that the results are stable up to a value of 0.4. The lower the value, the higher the penalty cost for the frame difference, which means it is more difficult to match those detections which are more than 1 frame apart. 
When the value of $B_j$ is higher than 0.4, there are more ambiguities in the data association process, because it is easier to match detections across distant frames. 
In the TownCenter dataset, there is no occluding object in the scene, which means missing detections are sporadic within a given trajectory. In this scenario, a lower value for $B_j$ is better, since small gaps can be filled and there are fewer ambiguities.
Nonetheless, we see different results in the PETS 2009 dataset in Figure \ref{fig:baseP} since there is an occluding object in the middle of the scene (see Figure \ref{fig:pets_occlude}) which occludes pedestrians for longer periods of time. In this case, a higher value of $B_j$ allows to overcome these large gaps of missing data, and that is why the best value for this dataset is around 0.6.

\begin{figure*}[htpb]
\centering
\subfigure[Wrong match with DIST, corrected with SFM.]{
\includegraphics[width=0.1875\linewidth]{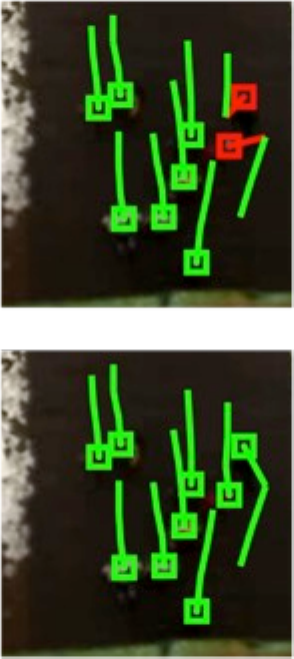} 
\label{fig:res1}
}
\qquad
\subfigure[Missing detections cause the matches to shift due the global optimization; correct result with SFM.]{
\includegraphics[width=0.43\linewidth]{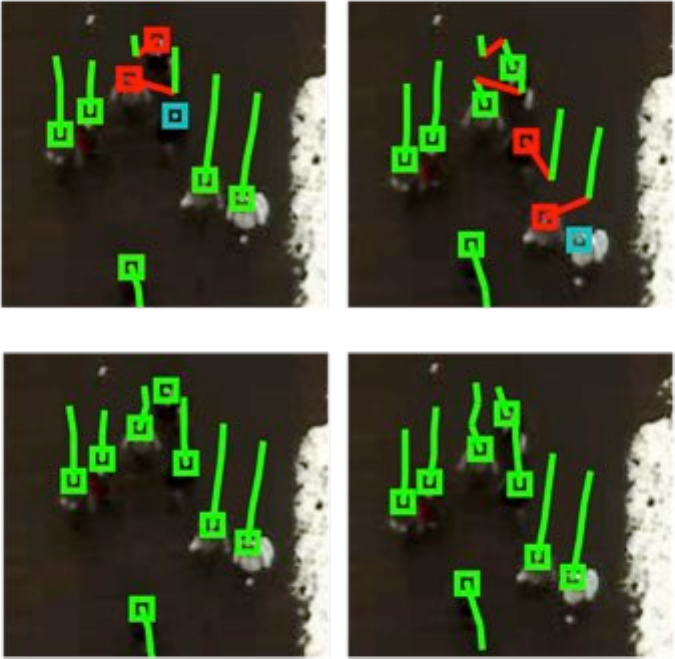} 
\label{fig:res2}
}
\subfigure[Missed detection for subject 3 on two consecutive frames. With SFM, subject 2 in the first frame (yellow arrow) is matched to subject 3 in the last frame (yellow arrow), creating an identity switch; correct result with grouping information.]{
\includegraphics[width=0.82\linewidth]{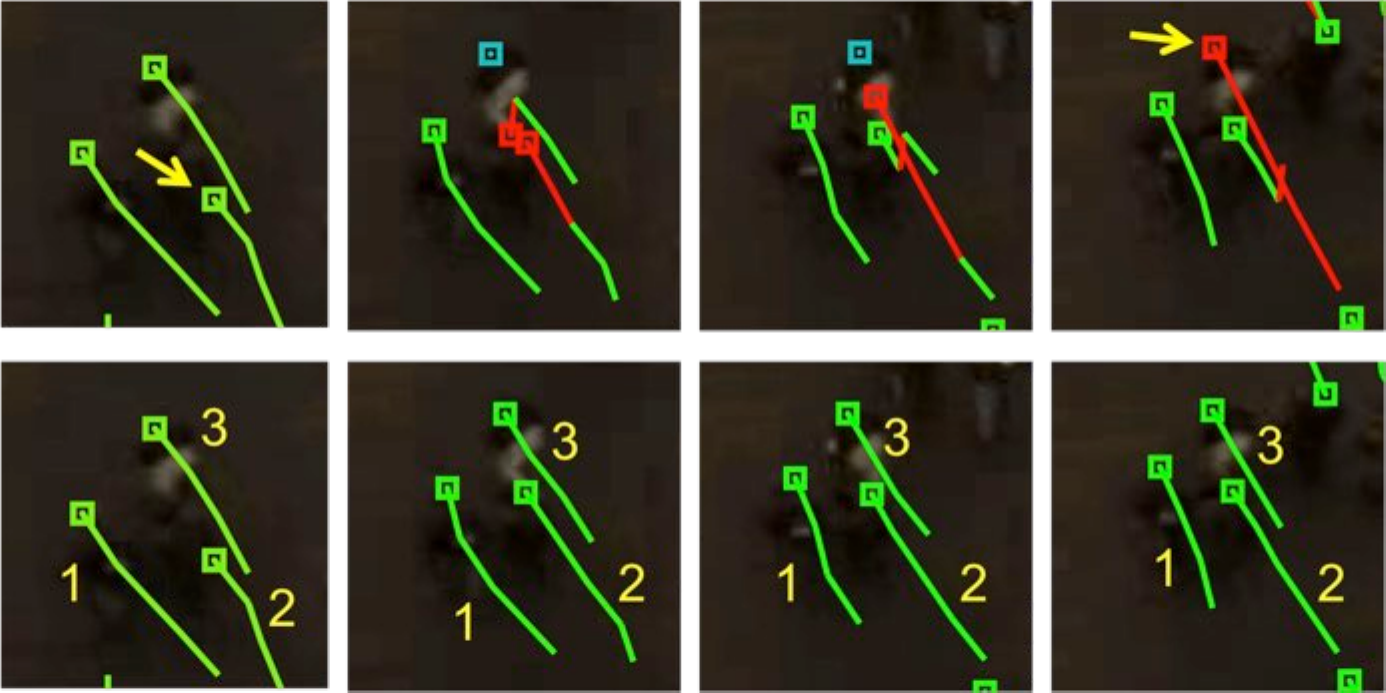} 
\label{fig:res3}
}
\caption[Example of correct tracking using the social force model \vs only distance information]{\emph{Top row}: Tracking results with only DIST. \emph{Bottom row}: Tracking results with SFM+GR.  \emph{Green} = correct trajectories, \emph{Blue} = observation missing from the set, \emph{Red} = wrong match.}
\label{fig:seqeth}
\end{figure*}

\subsection{Evaluation with missing data, noise and outliers}
\label{biwi}

We evaluate the impact of every component of the approach in \citep{lealiccv2011} with one of the sequences of the dataset \citep{pellegriniiccv2009} which contains images from a crowded public place with several groups as well as walking and standing pedestrians.
The sequence is 11601 frames long and contains more than 300 trajectories. First of all, the group detection method is evaluated on the whole sequence with ground truth detections: 61\% are correctly detected, 26\% are only partially detected and 13\% are not found. Furthermore, an extra 7\% false positive groups are detected. 
All experiments are performed with 6 iterations, a batch of 100 frames, $V_{\textrm{max}}=$\unit[7]{m/s}, $F_{\textrm{max}}=10$, $\alpha=0.5$ and $B_j=0.3$.

Using the ground truth (GT) pedestrian positions as the baseline for our experiments, we perform three types of tests: missing data, outliers and noise, and compare the results obtained with:

\begin{itemize}
\renewcommand{\labelitemi}{$\bullet$}
\item{DIST: proposed network model with distances}
\item{SFM: adding the Social Force Model (Section \ref{SFMsection})}
\item{SFM+GR: adding SFM and grouping behavior (Section \ref{SFMsection})}
\end{itemize}

\begin{figure*}[p]
\centering
\includegraphics[height=7cm]{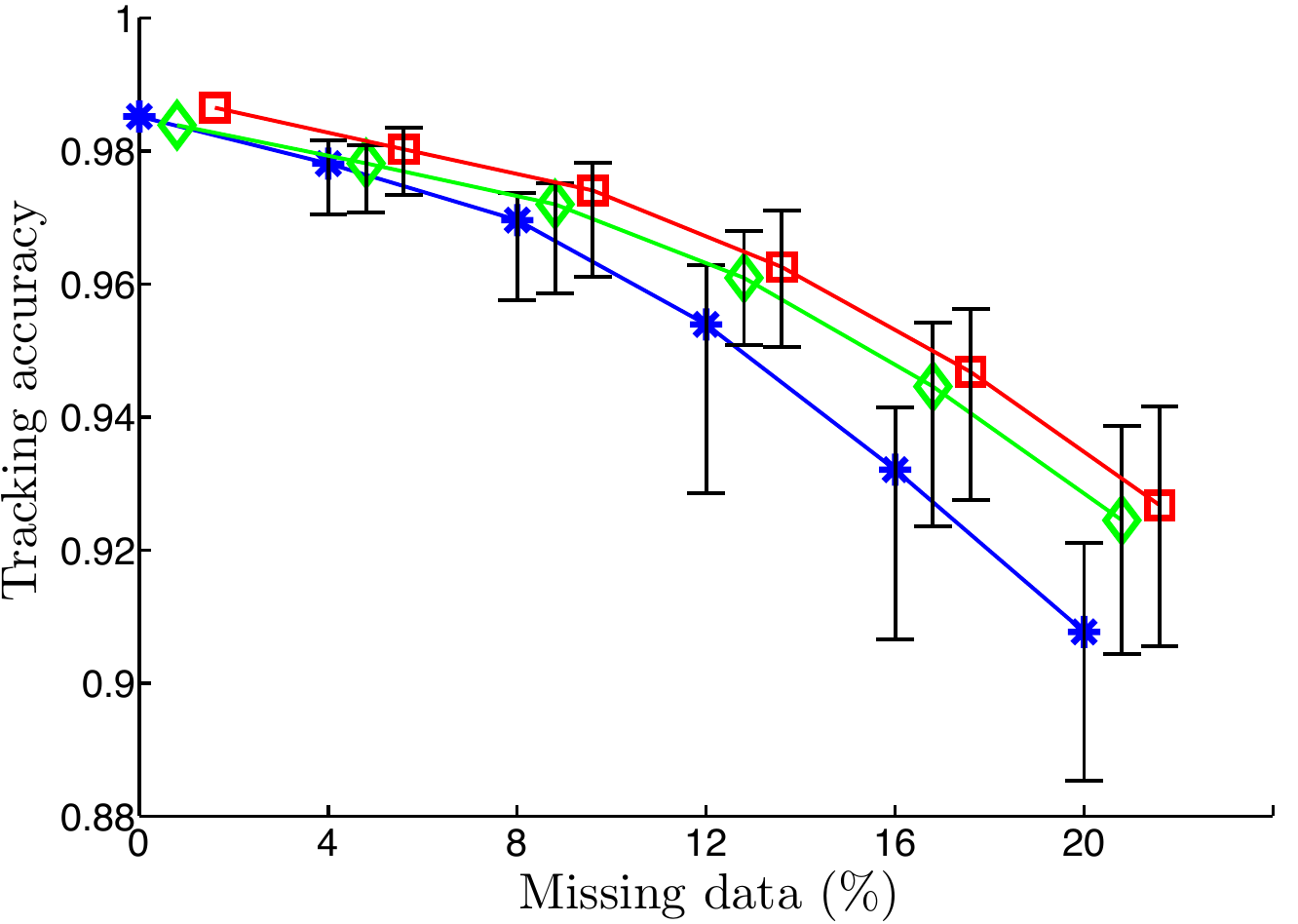}\vspace{0.4cm} 
\includegraphics[height=7cm]{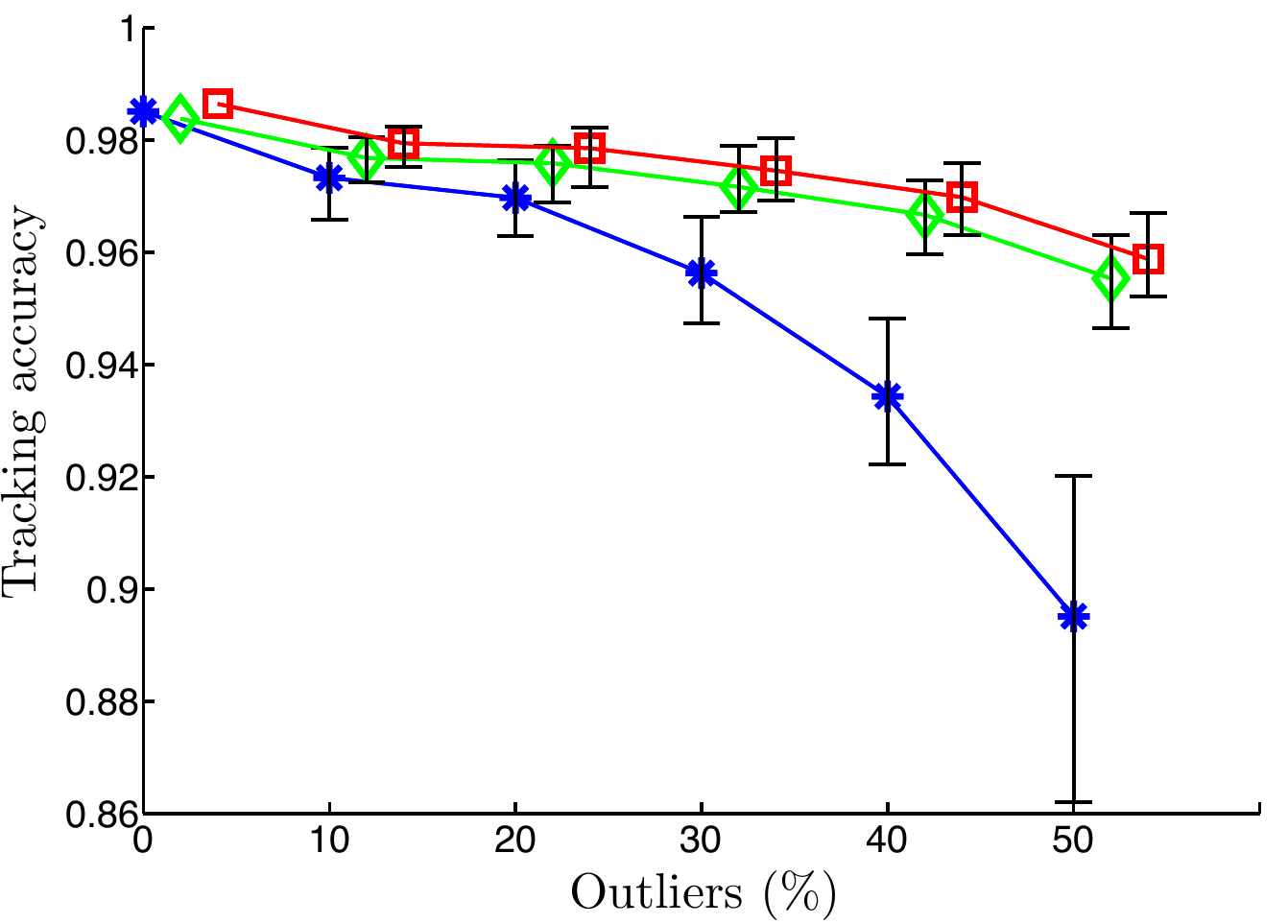}\vspace{0.4cm} 
\includegraphics[height=7cm]{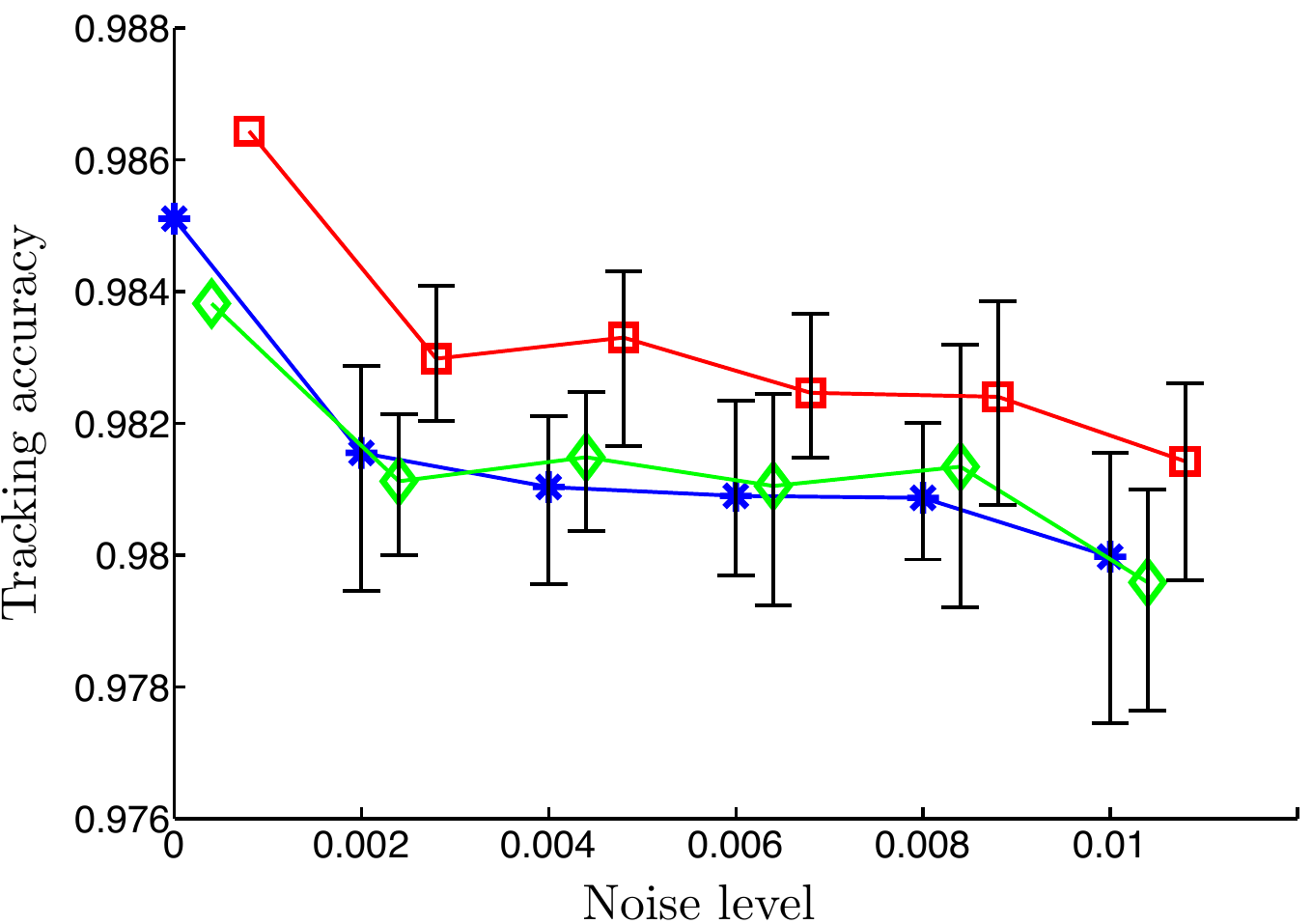}\vspace{0.2cm} 
\caption[Experiments with simulated missing data, outliers and random noise]{Experiments are repeated 50 times with random generation of outliers, missing data and noise and the average result, maximum and minimum are plotted. \emph{Blue star} = results with DIST, \emph{Green diamond} = results with SFM, \emph{Red square} = results with SFM+GR. \emph{From top to bottom}: Experiment with simulated missing data, with outliers, and with random noise.}
\label{fig:missnoiseout}
\end{figure*}

\noindent{\bf Missing data.} 
This experiment shows the robustness of our approach given missed detections. This is evaluated by randomly erasing a certain percentage of detections from the GT set. The percentages evaluated are $[0,4,8,12,16,20]$ of the total number of detections over the whole sequence.
As we can see in Figure \ref{fig:missnoiseout}, both SFM and SFM+GR increase the tracking accuracy when compared to DIST.

\noindent{\bf Outliers.}
With an initial set of detections of GT with 2\% missing data, tests are performed with $[0,10,20,30,40,50]$ percent outliers added in random positions over the ground plane.
In Figure \ref{fig:missnoiseout}, the results show that the SFM is especially important when the tracker is dealing with outliers. With 50\% of outliers, the SFM+GR terms reduce the number of identity switches by 70\% w.r.t the DIST results.

\noindent{\bf Noise.} 
This test is used to determine the performance of our approach given noisy detections, which are very common mainly due to small errors in the 2D-3D mapping. From the GT set with 2\% missing data, random noise is added to every detection. The variances of the noise tested are $[0,0.002,0.004,0.006,0.008,0.01]$ of the size of the observed scene. As expected, group information is the most robust against noise; if the position of pedestrian A is not correctly estimated, other pedestrians in the group will contribute to the estimation of the true trajectory of A.

\begin{figure*}[htbp]
\centering
\includegraphics[width=1\linewidth]{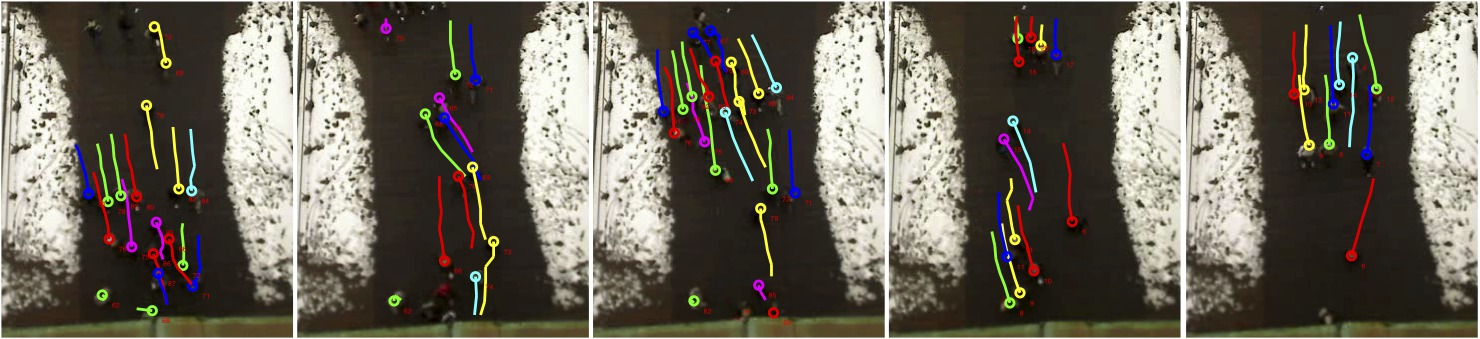} 
\caption{Visual results on the BIWI dataset (Section \ref{biwi}). The scene is heavily crowded, social and grouping behavior are key to obtaining good tracking results.}
\label{fig:biwi}
\end{figure*}

These results corroborate the assumption that having good behavioral models becomes more important as observations deteriorate.
In Figure \ref{fig:seqeth} we plot the tracking results of a sequence with 12\% simulated missing data. If we only use distance information, we can see the resulting identity switches as shown in Figure \ref{fig:res1}. In Figure \ref{fig:res2} we can see how missing data affects matching results. The matches are shifted; this chain reaction is caused by the global optimization. In both cases, the use of SFM allows the tracker to extrapolate the necessary detections and find the correct trajectories.
Finally, in Figure \ref{fig:res3} we plot the wrong result caused by track 3 having two consecutive missing detections. Even with SFM, track 2 is switched for 3 since the switch does not create extreme changes in velocity. In this case, the grouping information is key to obtaining good tracking results.
More results are shown in Figure \ref{fig:biwi}.

\begin{figure}[htbp]
\centering
\includegraphics[width=1\linewidth]{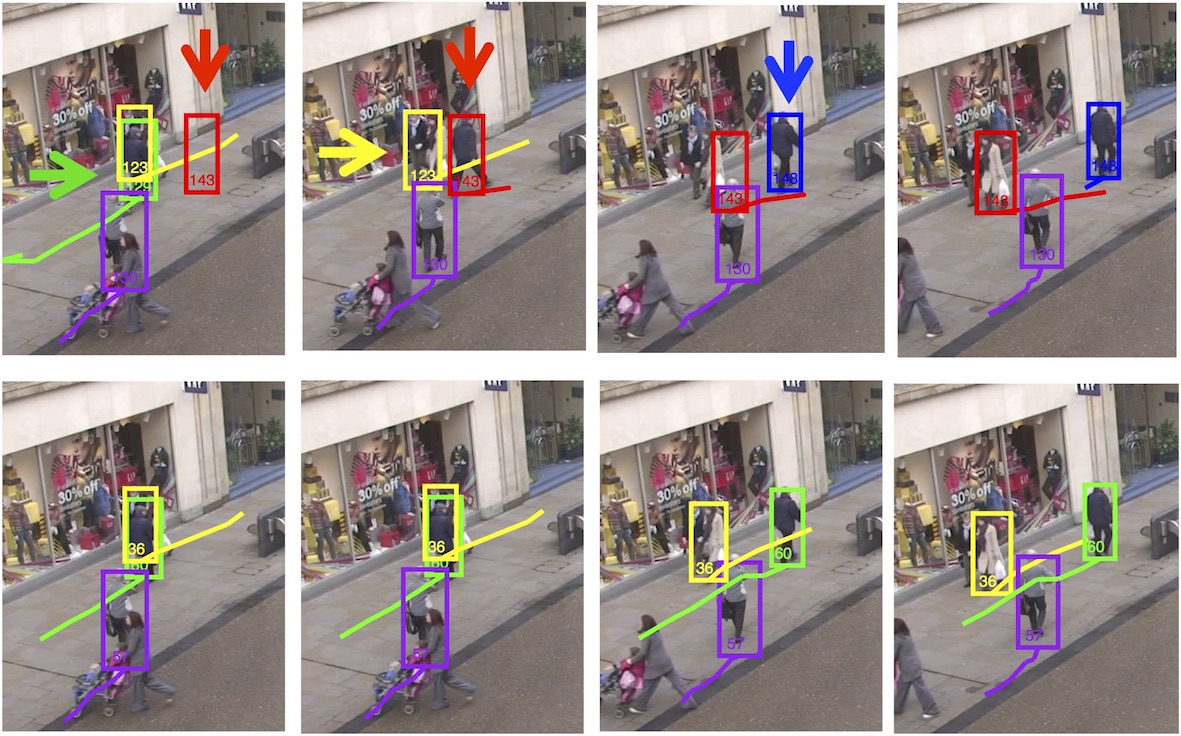} 
\caption[Using the social force model for tracking: predictive approaches \vs LP]{Predictive approaches \citep{pellegriniiccv2009,yamaguchicvpr2011} (first row) \vs Proposed method (second row)}
\label{fig:LPvspred}
\end{figure}

\subsection{Tracking results}

In this section, we compare results of several state-of-the-art methods on two publicly available datasets: a crowded town center \citep{benfoldcvpr2011} and the well-known PETS2009 dataset \citep{pets2009}.
We compare results obtained with:

\begin{itemize}
\renewcommand{\labelitemi}{$\bullet$}
\item{\citep{benfoldcvpr2011}: using the results provided by the authors for full pedestrian detections. The HOG detections are also given by the authors and used as input for all experiments.}
\item{\citep{zhangcvpr2008}: globally optimum tracking based on network flow linear programming.}
\item{\citep{pellegriniiccv2009}: tracker based on Kalman Filter which includes social behavior.}
\item{\citep{yamaguchicvpr2011}: tracker based on Kalman Filter which includes social and grouping behavior.}
\item{\citep{lealiccv2011}: globally optimum tracking based on network flow linear programming and including social and grouping behavior.}
\end{itemize}

For a fair comparison, we do not use appearance information for any method. The methods \citep{benfoldcvpr2011,pellegriniiccv2009,yamaguchicvpr2011} are online, while \citep{lealiccv2011,zhangcvpr2008} processes the video in batches. For these last two methods, all experiments are performed with 6 iterations, a batch of 100 frames, $V_{\textrm{max}}=$\unit[7]{m/s}, $F_{\textrm{max}}=10$, $\alpha=0.5$ and $B_j=0.3$.

\subsubsection{Town Center dataset}
\label{towncenter}

We perform tracking experiments on a video of a crowded town center \citep{benfoldcvpr2011} using one out of every ten frames (simulating 2.5 fps). We show detection accuracy (DA), tracking accuracy (TA), detection precision (DP) and tracking precision (TP) measures as well as the number of identity switches (IDsw). 

\begin{table}[htbp]
\small
\begin {center}
  \begin{tabular}{ cccccc }
     & {DA} & {TA} & {DP} & {TP} & {IDsw}  \\ \hline
    HOG Detections & 63.1& $-$ & 71.9& $-$ & $-$ \\ \hline
    \citep{benfoldcvpr2011} & 64.9 & 64.8 & {\bf 80.5} & {\bf 80.4} & 259 \\  
    \citep{zhangcvpr2008} & 66.1 & 65.7 & 71.5 & 71.5 & 114 \\ 
    \citep{pellegriniiccv2009}  & 64.1 & 63.4 & 70.8 & 70.7 & 183 \\ 
    \citep{yamaguchicvpr2011} & 64.0 & 63.3 & 71.1 & 70.9 & 196 \\ 
     \citep{lealiccv2011} & {\bf 67.6} & {\bf 67.3} & 71.6 & 71.5 &  {\bf 86} \\  \hline 
    \end{tabular}
  \end{center}
    \caption{CLEAR results on the Town Center sequence.}
\label{tab:towncenter}

\end{table}

\begin{figure*}[htbp]
\centering
\includegraphics[width=1\linewidth]{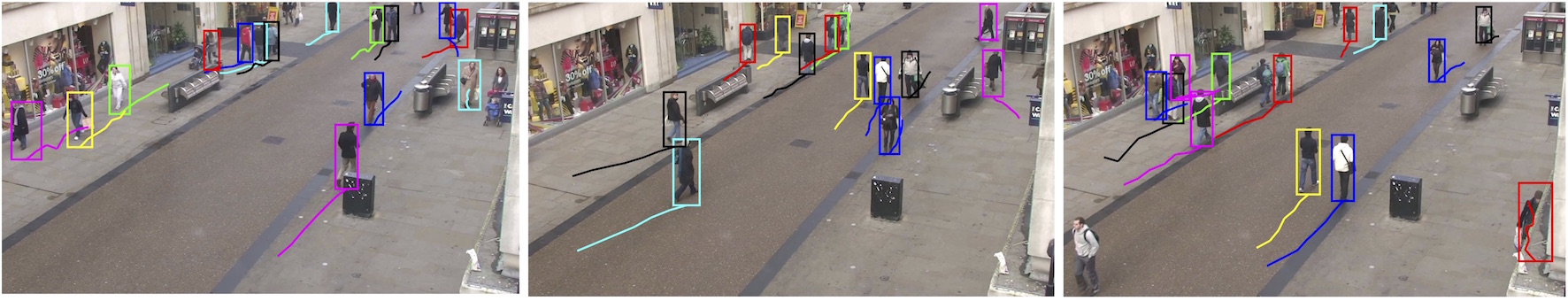} 
\includegraphics[width=1\linewidth]{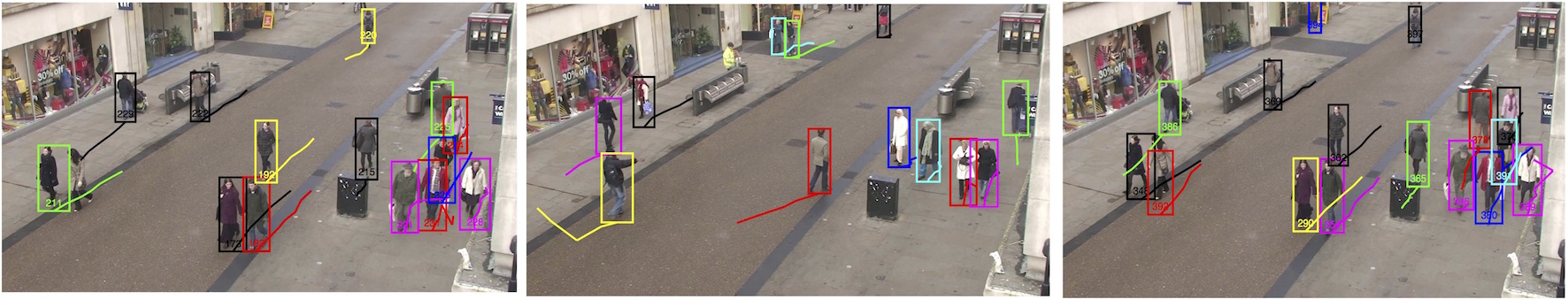} 
\caption{Visual results on the Town Center dataset (Section \ref{towncenter}).}
\label{fig:towncenter}
\end{figure*}

Note, the DP reported in \citep{benfoldcvpr2011} is about 9 percentage points higher than the input detection precision; this is because the authors use the motion estimation obtained with a KLT feature tracker to improve the exact position of the detections, while we use the raw detections. Still, our algorithm reports almost 67\% fewer ID switches.
As shown in Table \ref{tab:towncenter}, \citep{lealiccv2011} algorithm outperforms \citep{pellegriniiccv2009,yamaguchicvpr2011}, both of which include social behavior information, by almost 4 percentage points in accuracy and reduces the number of identity switches by more than 53\%.
In Figure \ref{fig:LPvspred} we can see an example where \citep{pellegriniiccv2009,yamaguchicvpr2011} fail. The errors are created in the greedy phase of predictive approaches, where trajectories compete to get assigned to detections. A red trajectory is started by a false detection in the first frame. This trajectory then takes the detection in the second frame that should belong to the green trajectory (which ends in the first frame). In the third frame, the red trajectory takes over the yellow trajectory and a new blue trajectory starts where the green should have been. None of the resulting trajectories violate the SFM and GR conditions. 
On the other hand, a global optimization framework takes full advantage of the SFM and GR information and correctly recovers all trajectories. More results of the proposed algorithm can be seen in Figure \ref{fig:towncenter}.

\subsubsection{Results on the PETS2009 dataset}
\label{pets2009}

In addition, we present results of monocular tracking on the PETS2009 sequence L1, View 1 with the detections obtained using the Mixture of Gaussians (MOG) background subtraction method. 
We compare the results with the previously described methods plus the monocular result of View 1 presented in \citep{berclaztpami2011}, where the detections are obtained using the Probabilistic Occupancy Map (POM) and the tracking is done using k-shortest paths.

\begin{figure*}[htbp]
\centering
\includegraphics[width=0.65\linewidth]{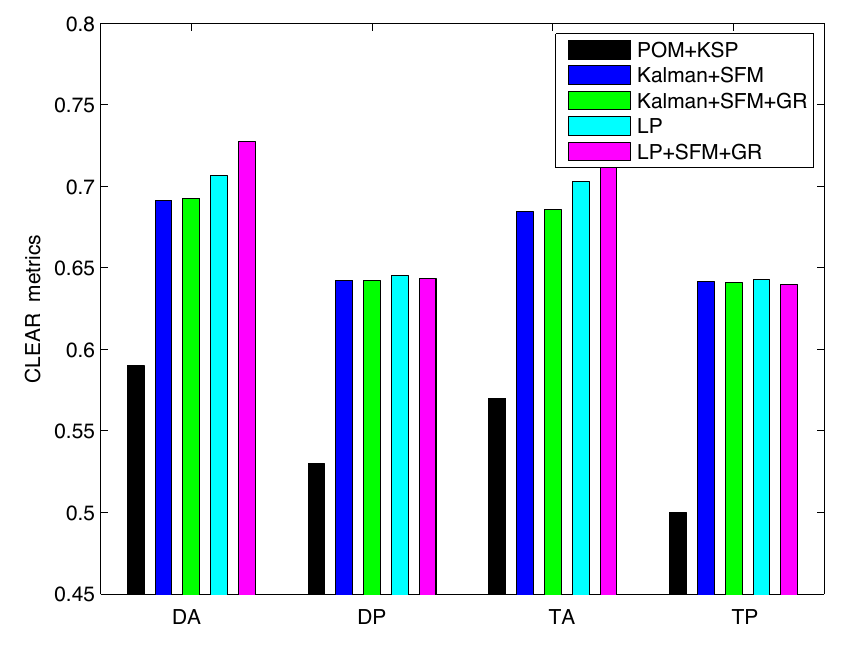} 
\caption[CLEAR results on the PETS2009 dataset]{Results of the proposed method on the PETS2009 dataset, view 1. DA=Detection accuracy. DP=Detection precision. TA=Tracking accuracy. TP=Tracking precision.}
\label{fig:finalPETS}
\end{figure*}

%

\begin{figure*}[htbp]
\centering
\includegraphics[width=1\linewidth]{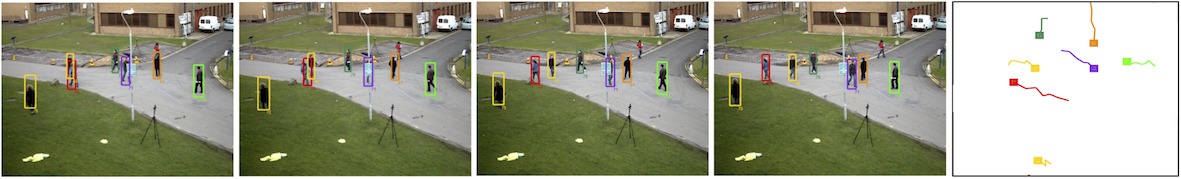} 
\includegraphics[width=1\linewidth]{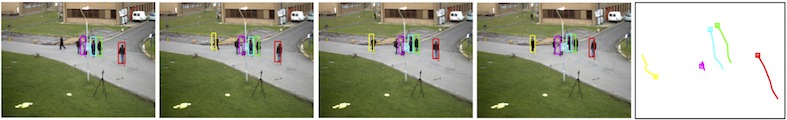} 
\caption{Visual results on the PETS2009 dataset (Section \ref{pets2009}).}
\label{fig:pets}
\end{figure*}

The first observation we make is that the linear programming methods (LP and LP+SFM+GR) clearly outperform predictive approaches in accuracy. This is because this dataset is very challenging from a social behavior point of view, because subjects often change direction and groups form and split frequently. 
Approaches based on a probabilistic framework \citep{lealiccv2011,zhangcvpr2008} are better suited for unexpected behavior changes (like destination changes), where other predictive approaches fail \citep{pellegriniiccv2009,yamaguchicvpr2011}. 
We can also see that the LP+SFM+GR method has a higher accuracy than the LP method which does not take into account social and grouping behavior. The grouping term is specially useful to avoid identity switches between members of a group (see an example in Figure \ref{fig:pets}, the cyan and green pedestrians who walk together). 
Precision is similar for all methods since the same detections have been used for all experiments and we do not apply smoothing or correction of the bounding boxes.

\section{Conclusions}

In this chapter, we presented an overview of methods that integrate pedestrian interaction into a tracking framework in two ways: using a globally optimum solver or improving the dynamic model with social forces.
Furthermore, we explained how to combine the strength of both approaches by finding the MAP estimate of the trajectories' total posterior, including social and grouping models using a minimum-cost network
flow with an improved novel graph structure that outperforms existing approaches.
Pedestrian interaction is persistent rather than transient, hence the probabilistic formulation fully exploits the power of
behavioral models, as opposed to standard predictive and recursive approaches, such as Kalman filtering.
Experiments on three public datasets reveal the importance of using social interaction
models for tracking in difficult conditions, such as crowded scenes with the presence
of missed detections, false alarms and noise.

\chapter{Tracking with multiple view context} 

\label{multiview} 

\fancyhead[RE,LO]{Chapter 6. \emph{Tracking with multiple view context}} 

\graphicspath{{./Figures/MultiView/}}

Combinatorial optimization arises in many computer vision problems such as feature correspondence, multi-view multiple object tracking, human pose estimation, segmentation, etc. 
In the case of multiple object tracking, object locations in images are temporally correlated by system dynamics and are geometrically constrained by the spatial configuration of the cameras (\ie the same object seen in two different cameras satisfies the epipolar constraints). 

These two sources of structure have been typically exploited separately by either Tracking-Reconstruction or Reconstruction-Tracking. Splitting the problem in two phases has, obviously, several disadvantages because the available evidence is not fully exploited. For example, if one object is temporarily occluded in one camera, both data association for reconstruction and tracking become ambiguous and underconstrained when considered separately. 
If, on the other hand, evidence is considered jointly, temporal correlation can potentially resolve reconstruction ambiguities and vice versa. 
However, finding the joint optimal assignment is a hard combinatorial problem that is both difficult to formulate and difficult to optimize. 
In this chapter, we argue that it is not necessary to separate the problem in two parts, and we present a novel formulation to perform 2D-3D assignments (reconstruction) and temporal assignments (tracking) in a single global optimization. 
The proposed graph structure contains a huge number of constraints, therefore, it cannot be solved with typical Linear Programming (LP) solvers such as the simplex algorithm. We rely on multi-commodity flow theory and use Dantzig-Wolfe decomposition and branching to solve the linear program. 

\begin{figure}[htbp]
\centering
\includegraphics[width =0.8 \linewidth]{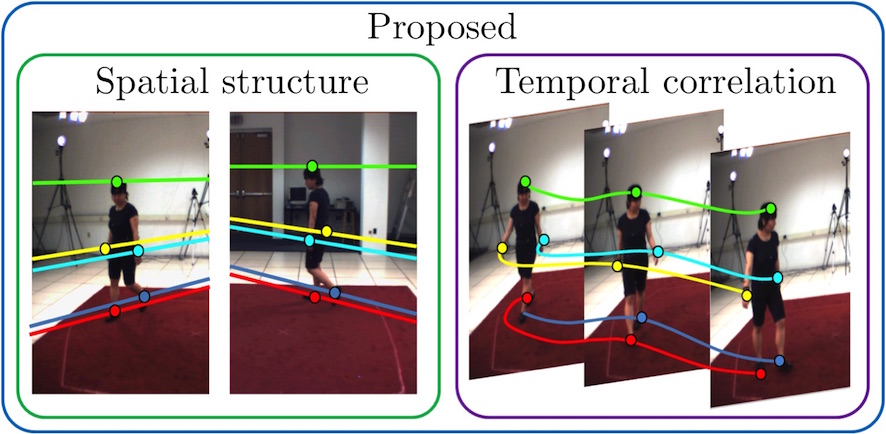}
\caption[Jointly exploiting temporal and spatial information across cameras]{We jointly exploit spatial and temporal structure to solve the multiple assignment problem across multiple cameras and multiple frames. With our proposed method, both tracking and reconstruction are obtained as the solution of one single optimization problem. }
\end{figure}

\section{Related work: reconstruction \vs tracking}

As we argued in previous chapters, we divide the problem of multiple target tracking into two steps: detection and data association. When dealing with multi-view data, data association is commonly split into two optimizations, namely {\it sparse stereo matching} and {\it tracking}. While stereo matching is needed for reconstruction (obtaining 3D positions from 2D calibrated 
cameras), tracking is needed to obtain trajectories across time.

As we have seen in Chapter \ref{LPtracking}, solving the tracking problem as one single optimization problem using Linear Programming is more reliable, as it solves the matching problem jointly for all tracks. 


The sparse stereo matching problem for reconstruction is usually formulated as a linear assignment problem,
and it is well-known that for more than 3 cameras the problem is NP-hard \citep{poore1994}.
In \citep{wuwacv2009}, a comparison of the methods Tracking-Reconstruction \vs Reconstruction-Tracking is presented. In \citep{berclaztpami2011}, first reconstruction is performed using Probabilistic Occupancy Map (POM), and then tracking is done globally using Linear Programming. 
In \citep{yucvpr2007}, the assignments are found using a data-driven MCMC approach, while \citep{wucvpr2011} presented a formulation with two separate optimization problems: linking across-time is solved using network flows and linking across-views is solved using set-cover techniques. In contrast to all previous works, we formulate the problem as a single optimization problem.

In this chapter we present a graph formulation that captures the whole structure of the problem which leads to a problem with a high number of constraints. This rules out standard Linear Programming solvers such as the simplex algorithm \citep{lealiccv2011,zhangcvpr2008} or k-shortest paths \citep{berclaztpami2011,pirsiavashcvpr2011}. In \citep{nilliuscvpr2006}, interactions between objects are modeled in a multiple hypotheses fashion and heuristics are applied to make the problem practical. 
We define our problem as a multi-commodity flow problem, \ie, each object has its own graph with a unique source and sink.  Multi-commodity flows are used in \citep{shitriticcv2011} in order to maintain global appearance constraints during multiple object tracking. 
However, the solution is found by applying several k-shortest paths steps to the whole problem, which would be extremely time consuming for our problem and lead to non-integer solutions.

By contrast, we use decomposition and branching methods, which take advantage of the structure of the problem to reduce computational time and obtain better bounds of the solution.
Decomposition methods are closely related to Lagrangian Relaxation based methods such as Dual Decomposition \citep{DD,komodakisiccv2007} which was used for feature matching in \citep{torresanieccv2008} and for monocular multiple people tracking with groups in \citep{pellegrinieccv2010}.
In our case, we make use of the Dantzig-Wolfe decomposition \citep{dantzigwolfe,LP1} which allows us to take advantage of the special block-angular structure of our problem. It is well-known in the field of traffic flow scheduling \citep{rios2010} as it is able to handle huge linear programs. 
As is usual in multi-commodity flow problems, the solutions found are not integers and therefore branch-and-bound \citep{chardaire1995} is used. The combination of column generation and branch-and-bound methods is known as branch-and-price \citep{branchandprice}. 

Recently, \citep{hofmanncvpr2013} proposed a Linear Programming solution using a simplified graph structure that also includes multi-camera information, but does not constrain the problem as tightly as the formulation presented in this chapter. The advantage is that their problem can be solved in linear time.

In this chapter, we present a global optimization formulation for multi-view multiple object tracking \citep{lealcvpr2012}. We argue that it is not necessary to separate the problem into  two parts, namely, reconstruction (finding the 2D-3D assignments) and tracking (finding the temporal assignments) and propose a new graph structure to solve the problem globally.
To handle this huge integer program, we introduce decomposition and branching methods which can be a powerful tool for a wide range of computer vision problems.

\section{Multi-view Multi-object tracking}

Tracking multiple objects in several calibrated camera views can be expressed as an energy minimization problem. 
We define an energy function that at the 2D level (i) enforces temporal smoothness for each camera view (2D-2D),
and at the 3D level (ii) penalizes inconsistent 2D-3D reconstructions from camera pairs, (iii) enforces coherent reconstructions from different camera pairs and (iv) favors temporal smoothness of the putative 3D trajectories.
In the following section, we detail the proposed graph structure used for multi-view multi-object tracking. 

\subsection{Proposed multi-layer graph}

\begin{figure}[htbp]
\centering
\subfigure[2D layer]{
\includegraphics[width=0.83\linewidth]{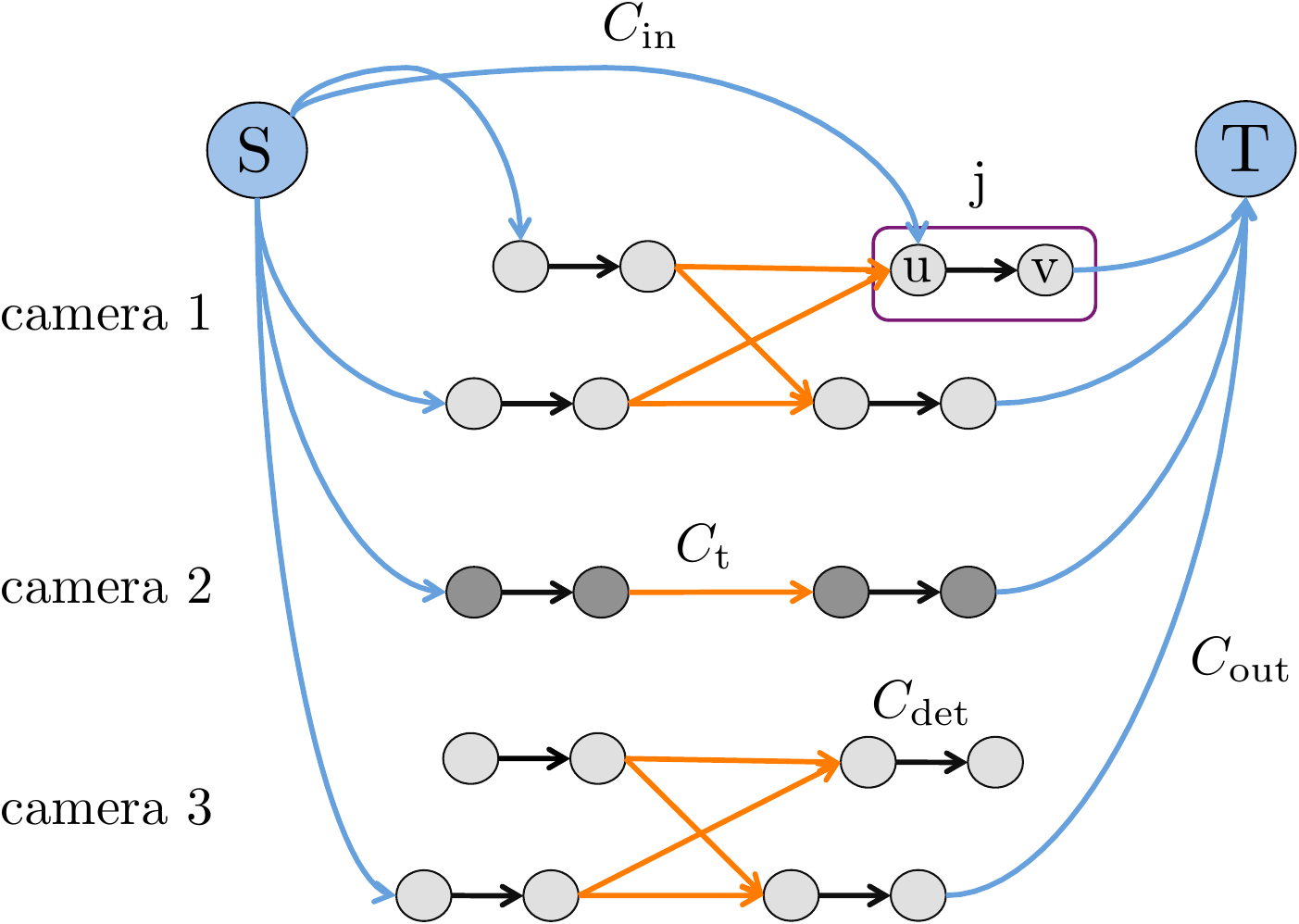} 
\label{fig:graph1}}
\subfigure[3D layer]{
\includegraphics[width=0.8\linewidth]{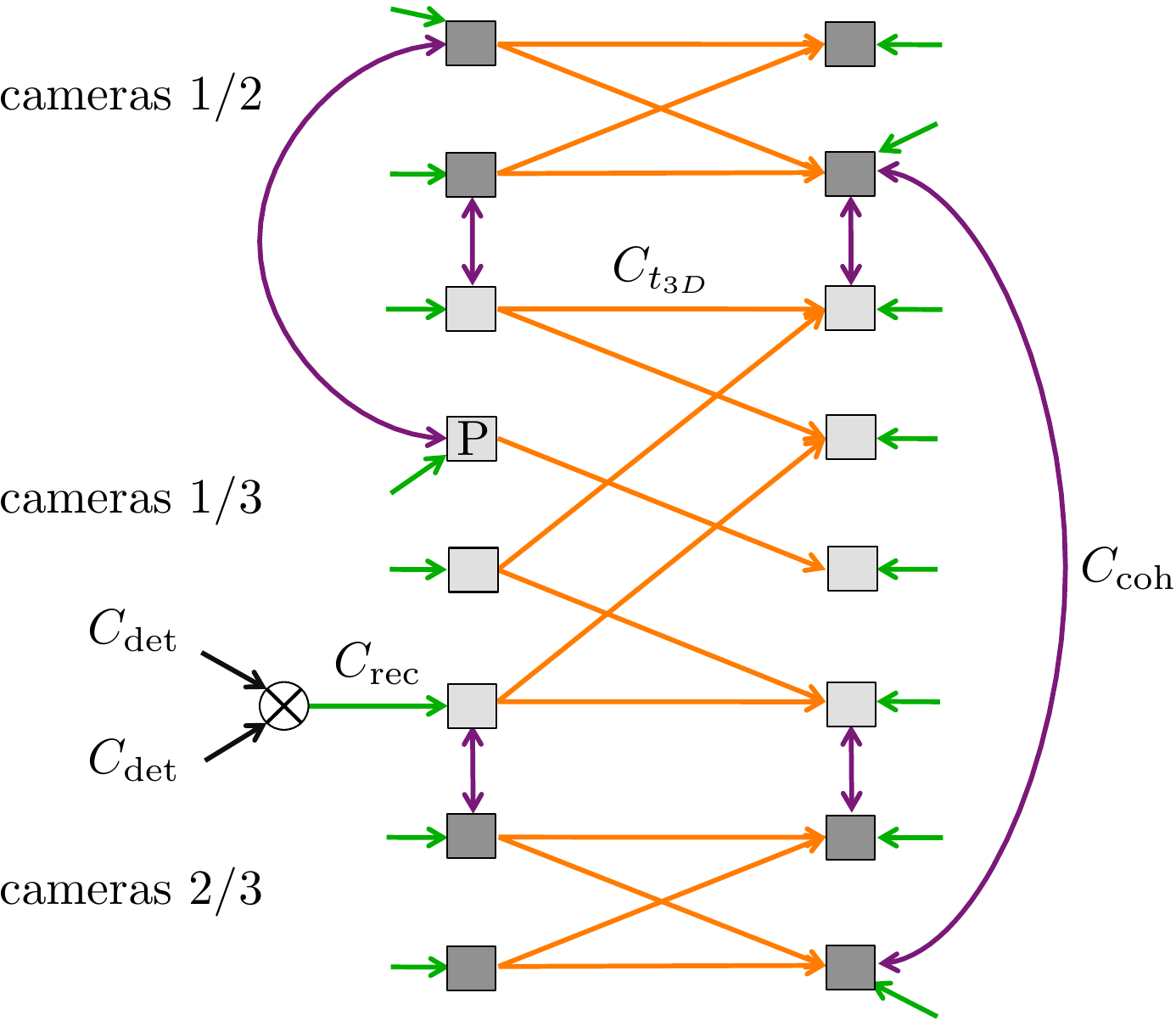} 
\label{fig:graph2}}
\caption{An example of the proposed multi-layer graph structure with three cameras and two frames. Let $u$ and $v$ represent a 2D detection $j$ and $P$ be a 3D reconstructed point.}
\label{fig:graph}
\end{figure}

Matching between more than two cameras (k-partite matching) is an NP-hard problem. In order to be able to handle this problem, we propose to create a \emph{multi-layer graph}. In Figure \ref{fig:graph} an example of the proposed graph with three cameras and two frames is shown. 

The first layer, the 2D layer, depicted in Figure \ref{fig:graph1}, contains 2D detections (circular nodes) and the flow constraints and is where trajectories are matched across time. The second layer, the 3D layer, depicted in Figure \ref{fig:graph2}, contains the putative 3D locations (square nodes) obtained from the 2D detections on each pair of cameras. It is designed as a cascade of prizes and favors consistent matching decisions across camera views. Thereby, the problem is fully defined as a singled global optimization problem.

In the following lines, we define the edges characterizing each of the two layers, namely, the \emph{entrance/exit}, the \emph{detection} and the \emph{temporal 2D} edges that define the 2D layer and the \emph{reconstruction}, the \emph{camera coherence} and the \emph{temporal 3D} edges that form the 3D layer.

\noindent{\bf Entrance/exit edges} ($C_{\textrm{in}},C_{\textrm{out}}$). These edges determine when a trajectory starts and ends; the cost balances the length of the trajectories with the number of identity switches. Shown in blue in Figure \ref{fig:graph1}.

\noindent{\bf Detection edges} ($C_{\textrm{det}}$). 
If all costs of the edges in a graph are positive and we do not know the amount of flow that has to go through that graph (\ie, the number of objects in the scene), then the trivial solution of zero flow is found. 
To avoid the trivial solution, some costs have to be negative so that the solution has a total negative objective cost. Following \citep{lealiccv2011,zhangcvpr2008}, each detection ${\bf p}_{i_v}$ in view $v \in \{1 \ldots V\}$ is divided into two nodes, {\bf b} and {\bf e}, and a new detection edge is created with cost
\begin{align}
\textstyle
C_{\textrm{det}}(i_v)=\log \left(1-P_{\textrm{det}}({\bf p}_{i_v}^t) \right) .
\end{align}
The higher the likelihood of a detection $P_\textrm{det}({\bf p}_{i_v}^t)$ the higher the negative cost of the detection edge (shown in black in Figure \ref{fig:graph1}), hence, flow is likely to be routed through edges of confident detections in order to minimize the total cost.

\noindent{ \bf Temporal 2D edges} ($C_{\textrm{t}}$). The costs of these edges (shown in orange in Figure \ref{fig:graph1}) encode the temporal dynamics of the targets. Assuming temporal smoothness, we define $\mathcal{F}$ to be a decreasing function \citep{lealiccv2011} of the distance between detections in successive frames
\begin{align}
\textstyle
C_\textrm{t}(i_v,j_v) = -\log \left( \mathcal{F}\left(\textstyle \frac{\|{\bf p}_{j_v}^{t+\Delta t} - {\bf p}_{i_v}^t\|}{\Delta t},V^{\textrm{2D}}_{\textrm{max}}\right)  + B_f ^ {\Delta f - 1}\right),
\end{align}
where $V^{\textrm{2D}}_{\textrm{max}}$ is the maximum allowed speed in pixels and $B_f ^ {\Delta f - 1}$ is a bias that depends on the frame difference $\Delta f$ and favors matching detections in consecutive frames.
The function $\mathcal{F}$ maps a distance to a probability which is then converted to a cost by the negative logarithm.

Note, that the 2D layer alone is a special case of our multi-layer graph and would be suited to find the trajectories on each camera independently. 
Finding a global optimum match between trajectories for $k$ cameras means solving a $k$-partite matching problem, which in the case of $k>2$ is well-known to be NP-complete.
We take a slightly different approach and decide to track independently on each camera, but we introduce a series of edges that bind the 2D layers with 3D information. For this, we create the \emph{3D layer} which contains three types of edges.

\noindent{\bf Reconstruction edges} ($C_{\textrm{rec}}$). These edges connect the 2D layer (Figure \ref{fig:graph1}) with the 3D layer (Figure \ref{fig:graph2}). For each camera pair, all plausible 2D-2D matches create new 3D hypothesis nodes (marked by squares in Figure \ref{fig:graph2}). The reconstruction edges, shown in green, connect each newly created 3D detection with the 2D detections that have originated it. 
The costs of these edges encode how well 2D detections match in 3D, which is implemented by computing the minimum distance between pairs of projection rays. Let  $\mathcal{C}_v$ be the set of all possible camera pairs and $m_{k}$ a new 3D hypothesis node generated from the 2D nodes $i_{v_1}$ and $j_{v_2}$, where $k=(v_1,v_2) \in {\mathcal{C}_v}$ and $v_1,v_2$ are two different views. Given the camera calibration, each 2D point defines a line in 3D, ${{\bf L}}(i_{v_1})$ and ${{\mathbf L}}(j_{v_2})$. Now let ${\bf P}_{m_k}$ define the 3D point corresponding to the 3D node, which is the midpoint between the two closest points on the lines.
The reconstruction cost is
\begin{align}
\textstyle
C_\textrm{rec}(m_k) = \log \left( 1- \mathcal{F}\left(\textstyle {\mathrm{dist}\left({\bf L}(i_{v_1}),{\bf L}(j_{v_2})\right)},{\textrm{E}}_{\textrm{3D}}\right) \right),
\end{align}
where $\textrm{E}_\textrm{3D}$ is the maximum allowed 3D error. These edges are active, \ie, have a positive flow, when both originating 2D detections are also active. This constraint can be expressed in linear form as explained in Sect. \ref{sec:LP}. 
Essentially, the 3D layer is a model of possible 3D events in the scene which is supported by 2D evidence (detections). The reconstruction edges are the link to that evidence.

\noindent{\bf Camera coherency edges} ($C_{\textrm{coh}}$). Their purpose is to verify the evidence coming from two different cameras. Their cost is related to the 3D distance between two 3D nodes from different camera pairs. We show a few of these edges in Figure~\ref{fig:graph2} in purple. 
Considering two camera pairs $k,l \in {\mathcal{C}_v}$, two 3D nodes $m_k$ and $n_l$ and their corresponding 3D points ${\bf P}_{m_k}^t$ and ${\bf P}_{n_l}^t$, we define the camera coherency edge cost as
\begin{align}
\textstyle
C_\textrm{coh}(m_k,n_l) = \log \left( 1- \mathcal{F}\left(\textstyle {\|{\bf P}_{m_k}^t - {\bf P}_{n_l}^t\|},{\textrm{E}}_{\textrm{3D}}\right) \right). 
\end{align}
These edges are active when the two 3D nodes they connect are also active. 

\noindent{\bf Temporal 3D edges} ($C_{\textrm{t}_{\textrm{3D}}}$). The last type of edges are the ones that connect 3D nodes in several frames (shown in orange in Figure \ref{fig:graph2}). The connection is exactly the same as for the 2D nodes and their cost is defined as
\begin{align}
\textstyle
C_{\textrm{t}_\textrm{3D}}(m_k,n_k) = \log \left( 1- \mathcal{F}\left(\textstyle \frac{\|{\bf P}_{m_k}^{t+\Delta t} - {\bf P}_{n_k}^t\|}{\Delta t},V^{\textrm{3D}}_{\textrm{max}}\right)  \right),
\end{align}
where $V^{\textrm{3D}}_{\textrm{max}}$ is the maximum allowed speed in world coordinates. These edges are active when the two 3D nodes they connect are also active. \\
It is important to note that the 3D layer costs are always negative. To see this, recall that $\mathcal{F}$ maps a distance to a probability, and the lower the distance it evaluates, the higher the probability will be and hence the higher the negative cost. If the costs were positive, the solution would favor a separate trajectory for each camera and frame, because finding a common trajectory for all cameras and frames activates these edges.
Instead, these edges act as {\it prizes} for the graph, so that having the same identity in 2 cameras is beneficial if the reconstruction, camera coherence and temporal 3D edges are sufficiently negative. 

\begin{figure*}[htbp]
\centering
\subfigure[]{
\includegraphics[width=0.45\linewidth]{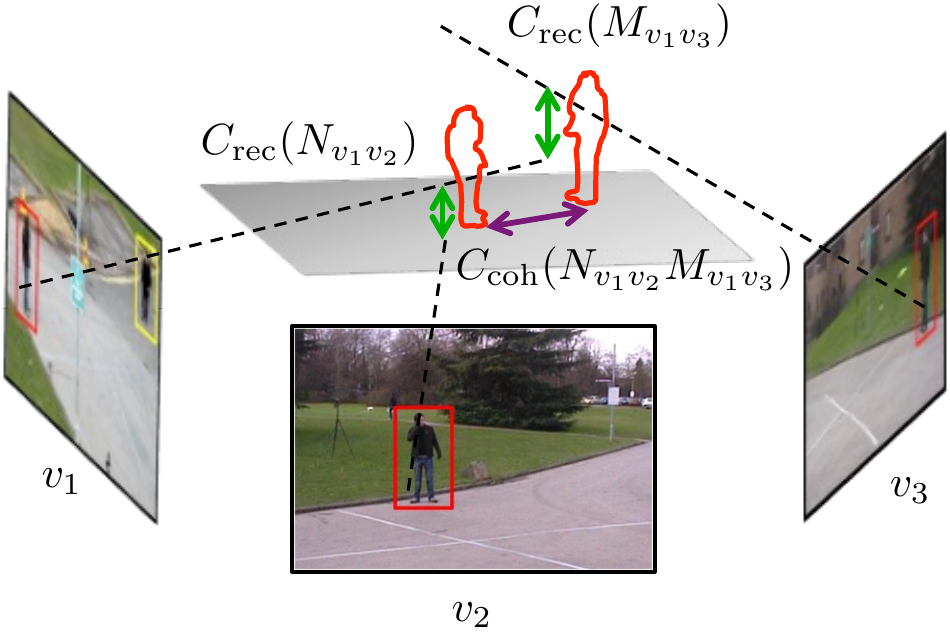} 
\label{fig:rec_edge1}}
\hspace{1 cm}
\subfigure[]{
\includegraphics[width=0.275\linewidth]{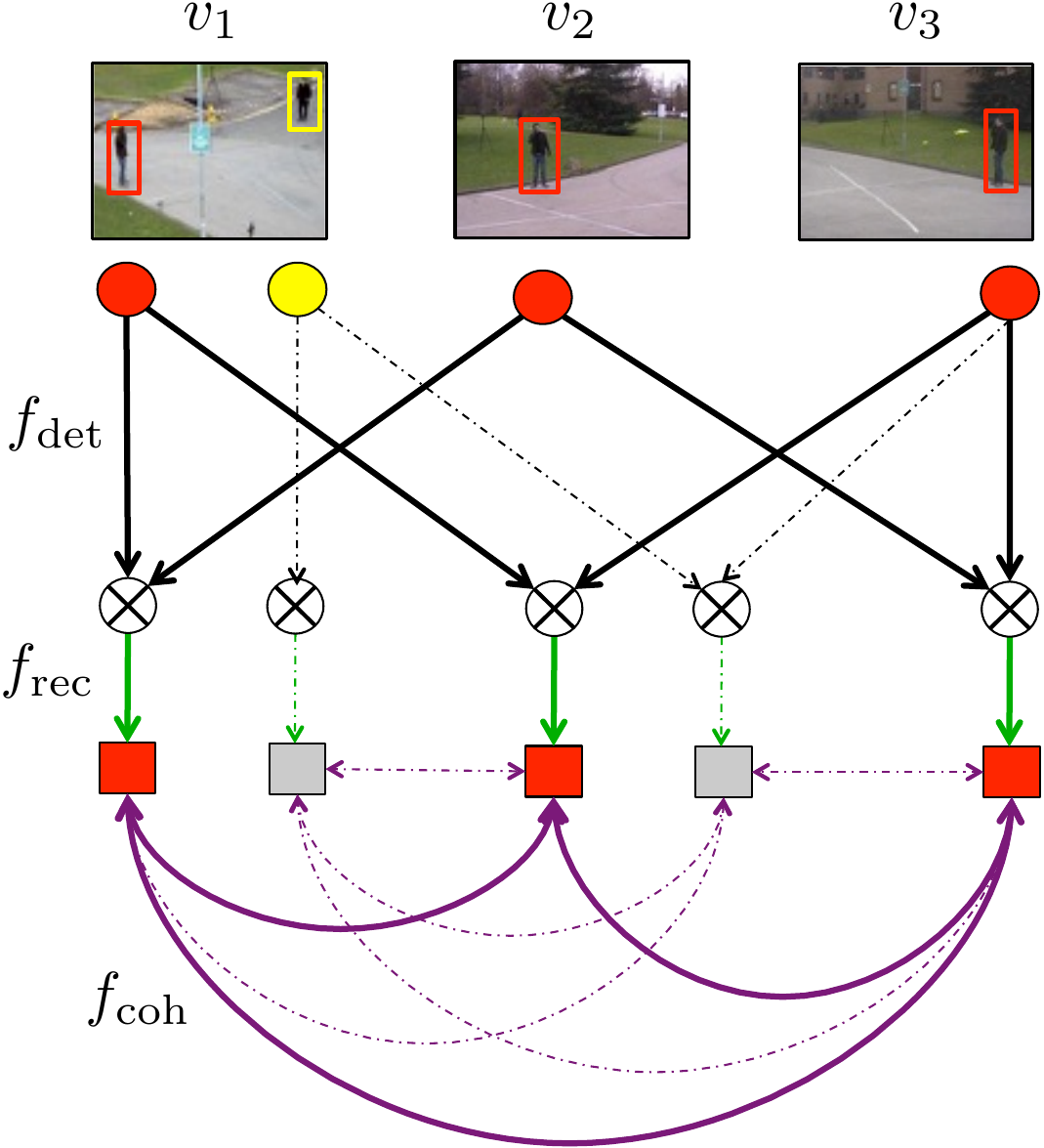} 
\label{fig:rec_edge2}}
\caption[Edges of the 3D layer and depiction of flow activation]{3D layer edges: \textbf{(a)} The 2D nodes in each camera activate the reconstruction and camera coherency edges because they are assigned the same trajectory ID visualized in red. The reconstruction error $C_{rec}$ is defined as the minimum line distance between projection rays. The camera coherency edges $C_\textrm{coh}$ are defined as the 3D distance between putative reconstructions (illustrated as red silhouettes in 3D) from different camera pairs. \textbf{(b)} graph structure of the 3D layer: active edges are shown in continuous lines. The red 2D nodes (circles) activate the 3D nodes (square nodes) since they are assigned the same ID (product of flows equals one).
}
\label{fig:rec_edge}
\end{figure*}

\subsection{Linear programming}
\label{sec:LP}
In the literature, multiple object tracking is commonly formulated as a Maximum A-Posteriori (MAP) problem. To convert it to a Linear Program (LP), its objective function is linearized with a set of flow flags $f(i)\in\{0,1\}$ which indicate whether an edge $i$ is in the path of a trajectory or not \citep{lealiccv2011,zhangcvpr2008}.
The proposed multi-layer graph can be expressed as an LP with the following objective function:
\begin{align}
\textstyle
\mathcal{T}*&=\underset{{f}} \argmin \; \mathbf{C}^\mathrm{T}\mathbf{f}  = \sum\limits_i C(i)f(i) \nonumber\\ 
&= \sum_{v=1}^{V} \sum_{i_v} C_{\textrm{in}}(i_v)f_{\textrm{in}}(i_v) +\sum_{v=1}^{V} \sum_{i_v} C_{\textrm{out}}(i_v)f_{\textrm{out}}(i_v) \nonumber \\
&+ \sum_{v=1}^{V} \sum_{i_v} C_{\textrm{det}}(i_v)f_{\textrm{det}}(i_v) +\sum_{v=1}^{V} \sum_{i_v,j_v}  C_{\textrm{t}}(i_v,j_v) f_{\textrm{t}}(i_v,j_v)  \nonumber \\
&+\sum_{k \in \mathcal{C}_v} \  \sum_{m_{k}}  C_{\textrm{rec}}(m_{k}) f_{\textrm{rec}}(m_{k})\nonumber \\ 
& + \sum_{k \in \mathcal{C}_v} \sum_{l \in \mathcal{C}_v} \sum_{m_k,n_l} C_{\textrm{coh}}(m_{k},n_{l})f_{\textrm{coh}}(m_{k},n_{l}) \nonumber \\
&+ \sum_{k \in \mathcal{C}_v} \sum_{m_k,n_k} C_{\textrm{t}_{\textrm{3D}}}(m_{k},n_{k})f_{\textrm{t}_{\textrm{3D}}}(m_{k},n_{k}) 
\label{eq:mincost}
\end{align}
where $k,l \in {\mathcal{C}_v}$ are the indices of different camera pairs. The problem is subject to the following constraints: 

\begin{itemize}
\item{Edge capacities: we assume that each detection belongs to only one trajectory, thus the flow that goes through detection edges can only assume the values $f(i)=\{0,1\}$. Since integer programming is NP-hard, we relax the conditions to obtain a linear program: $0 \leq f(i)\leq 1$. In the remainder of this chapter, all conditions will be expressed in their relaxed form.}
\item{Flow conservation at the 2D nodes: $f_{\textrm{in}}(i_v),f_{\textrm{out}}(i_v)$ indicate whether a trajectory starts or ends at node $i_v$. }
\begin{align}
  f_{\textrm{det}}(i_v) =  f_{\textrm{in}}(i_v) + \sum\limits_{j_v} f_{\textrm{t}}(j_v,i_v) \nonumber\\ 
  f_{\textrm{det}}(i_v) = \sum\limits_{j_v} f_{\textrm{t}}(i_v,j_v) + f_{\textrm{out}}(i_v) 
 \label{eq:flow1mv}
\end{align}
\item{Activation for reconstruction edges: these 2D-3D connections have to be activated, \ie, have a positive flow, if their 2D originating nodes are also active. More formally, this imposes the following relationship:}
\begin{align}
f_{\textrm{rec}}(m_{k}) = f_{\textrm{det}}(i_{v_1})f_{\textrm{det}}(j_{v_2}) 
 \label{eq:flow2mv}
\end{align}
\item{Activation for the camera coherency edges: for 3D-3D connections we take a similar approach as for the reconstruction edges and define the flow to be dependent on the 3D nodes it connects:}
\begin{align}
\textstyle
 f_{\textrm{coh}}(m_{k},n_{l})=  f_{\textrm{rec}}(m_{k}) f_{\textrm{rec}}(n_{l})
 \label{eq:flow3mv}
 \end{align}
\item{Activation for temporal 3D edges:}
\begin{align}
f_{\textrm{t}_{\textrm{3D}}}(m_{k},n_{k})= f_{\textrm{rec}}(m_{k}) f_{\textrm{rec}}(n_{k})
 \label{eq:flow4mv}
 \end{align}
\end{itemize}
As we can see, the pairwise terms in Eqs. \eqref{eq:flow2mv}, \eqref{eq:flow3mv} and \eqref{eq:flow4mv} are non-linear. Let $f_{ab}=f_a f_b$ be a pairwise term consisting of two flows $f_a$ and $f_b$ . Using the fact that the flows are binary, we can encode the pairwise term with the following linear inequations:
\begin{align} 
\textstyle
f_{ab}-f_a \leq 0 \qquad
f_{ab}-f_b \leq 0 \qquad
f_a + f_b - f_{ab} \leq 1.
 \end{align}
We can now express the constraints in Eqs. \eqref{eq:flow2mv}, \eqref{eq:flow3mv} and \eqref{eq:flow4mv} in linear form.
These constraints define the 3D layer of the graph as a {\it cascade of prizes}. Consider two 2D nodes on different cameras which belong to different trajectories. The question will be whether it is favorable to assign the same trajectory ID to both 2D nodes. The answer depends on the prize costs this assignment activates. When both 2D nodes are assigned the same trajectory ID, the corresponding 3D reconstruction edge is activated. If two 3D nodes from different camera pairs are activated, the camera coherency edge between them is activated, and the same will happen across time. This means that trajectories are assigned the same ID only if the reconstruction, camera coherency and temporal 3D costs are sufficiently negative to be beneficial to minimize the overall solution.


\subsection{Multi-commodity flow formulation}

The goal of the flow constraints defined in the previous section is to activate certain prize edges when two 2D nodes are activated {\it by the same object}. This means that in one graph we can only have a total flow of 1, which corresponds to one object.
To that end, we create one more condition on the number of objects per camera:
\begin{align}
  0 \leq \sum_{i_v}  f_{in}(i_v) \leq 1 \qquad 0 \leq \sum_{i_v}  f_{out}(i_v) \leq 1 \quad \forall v
  \label{eq:flow5mv}
 \end{align}
In order to deal with several objects, we use the multi-commodity flow formulation, well-known in traffic scheduling \citep{dantzigwolfe}. We create one graph for each object $n$ to be tracked on the scene. Each graph has its own source and sink nodes, and each object is a commodity to be sent through the graph. The problem has now a much larger set of variables $\mathbf{f}=\left[ \mathbf{f}^1 \ldots \mathbf{f}^{N_{\textrm{obj}}}\right]$.  
 Obviously, with no further restrictions, computing the global optimum would result in the same solution for all instances of the graph, \ie, we would find the same trajectory for all objects. Therefore, we need to create a set of \emph{binding constraints} which prevent two trajectories from going through the same edges:
\begin{align}
\sum_{n}  f^n(i) \leq 1 \qquad \quad n=1\ldots N_{\textrm{obj}}
\label{eq:binding}
\end{align}
where $f^n(i)$ is the flow of object $n$ going through the edge $i$. 
This set of binding constraints creates a much more complex linear program which cannot be solved with standard techniques. Nonetheless, the problem still has an interesting block-angular structure, which can be exploited. The problem consists of a set of small problems (or subproblems), one for each object, with the goal to minimize Eq.~\eqref{eq:mincost} subject to the constraints in Eq.~\eqref{eq:flow1mv}-\eqref{eq:flow5mv}. On the other hand, the set of complex binding constraints in Eq.~\eqref{eq:binding} defines the master problem.
This structure is fully exploited by the Dantzig-Wolfe decomposition method which is explained in the next section, allowing the algorithm to find a solution with less computation time. 

\section{Branch-and-price for multi-commodity flow}

Branch-and-price is a combinatorial optimization method for solving large scale integer linear problems. It is a hybrid method of column generation and branching. 

\noindent{\bf Column generation: Dantzig-Wolfe decomposition}. The principle of decomposition is to divide the constraints of an integer problem into a set of ``easy constraints" and a set of ``hard constraints". The idea is that removing the hard constraints results in several subproblems which can be easily solved by k-shortest paths, simplex, etc. 
Let us rewrite our original minimum cost flow problem:
\begin{align}
\textstyle
\underset{\mathbf{f}} \min \; \mathbf{C}^\mathrm{T}\mathbf{f}  = \sum\limits_{n=1}^{N_{\textrm{obj}}} ({\mathbf{c}^{n})^\textrm{T} \mathbf{f}^n} 
\end{align}
subject to:
\begin{align}
\textstyle
\mathbf{A}_1 \mathbf{f}\leq \mathbf{b}_1 \qquad   \mathbf{A}^n_2\mathbf{f}^n \leq \mathbf{b}^n_2 \qquad  0 \leq \mathbf{f} \leq 1 
\end{align}
where $(\mathbf{A}_1,\mathbf{b}_1)$ represent the set of hard constraints Eq.~\eqref{eq:binding}, and $(\mathbf{A}_2,\mathbf{b}_2)$ the set of easy constraints, Eqs. \eqref{eq:flow2mv}-\eqref{eq:flow5mv}, which are defined independently for each object $n=1\ldots N_\textrm{obj}$. The idea behind Dantzig-Wolfe decomposition is that the set $\mathcal{T}^*=\{f \in \mathcal{T} : f \mbox{ integer}\}$, with $\mathcal{T}$ bounded, is represented by a finite set of points, \ie, a bounded convex polyhedron is represented as a linear combination of its extreme points. The {\it master problem} is then defined as:
\vspace{-0.1cm}
\begin{align}
\textstyle
\underset{\mathbf{\lambda}} \min \;    \sum\limits_{n=1}^{N_{\textrm{obj}}} (\mathbf{c}^{n})^\textrm{T} \sum\limits_{j=1}^J   \mathbf{x}_j^n\lambda^n_j 
\end{align}
\vspace{-0.2cm}
subject to:
\begin{align}
\textstyle
 \sum\limits_n \mathbf{A}_1^n \sum\limits_{j=1}^J  \mathbf{x}_j^n \lambda_j^n \leq \mathbf{b}_1 \qquad   \sum\limits_{j=1}^J \lambda_j^n=1 \qquad 0 \leq \lambda_j^n \leq 1
\end{align}
where $\mathbf{f}^n =  \sum_{j=1}^J \lambda_j^n \mathbf{x}_j^n$ and $\{\mathbf{x}_j\}_{j=1}^J$ are the extreme points of a polyhedron. This problem is solved using column generation (Algorithm \ref{alg2}).
The advantage of this formulation is that the $N_\textrm{obj}$ column generation subproblems can be solved independently and therefore in parallel. We use the parallel implementation found in \citep{DWcode}, which is based on \citep{dantzigwolfe}.

\begin{algorithm}     
\caption{ Column generation}          
\label{alg2}                           
\begin{algorithmic}                    
\vspace{0.13cm}
\WHILE{Restricted master problem new columns $>0$}
\vspace{0.13cm}
\STATE 1. Select a subset of columns corresponding to $\lambda^n_j$ which form what is called the {\it restricted master problem}
\vspace{0.13cm}
\STATE 2. Solve the restricted problem with the chosen method (\eg, simplex).
\vspace{0.13cm}
\STATE 3. Calculate the optimal dual solution $\mu$
\vspace{0.13cm}
\STATE 4. Price the rest of the columns with $\mu( \mathbf{A}_1^n  \mathbf{f}^n -  \mathbf{b}_1^n)$ \\
\vspace{0.13cm}
\STATE 5. Find the columns with negative cost and add them to the restricted master problem. This is done by solving $N_\textrm{obj}$ column generation {\it subproblems}.
\vspace{0.13cm}
\STATE  $ \quad\underset{\mathbf{f}} \min  \quad   (\mathbf{c}^n)^\textrm{T}  \mathbf{f}^n + \mu( \mathbf{A}_1^n  \mathbf{f}^n -  \mathbf{b}_1^n)\quad    \textrm{s.t.} \quad \mathbf{A}_2^n\mathbf{f}^n \leq \mathbf{b}_2^n$ 
\vspace{0.13cm}
\ENDWHILE
\vspace{0.13cm}
\end{algorithmic}
\end{algorithm}

\noindent{\bf Branching}. Typically in multi-commodity flow problems, the solution is not guaranteed to be composed of all integers. Nonetheless, once we find the fractional solution, we can use branching schemes to find the integer optimal solution.
This mixture of column generation and branching is called {\it branch-and-price}. 
One important thing is that branching must be done on the original variables, not on the $\lambda_j^n$ of the master problem. 
For more details we refer to \citep{networkflows,branchandprice}.



\section{Experimental results}

In this section, we show the tracking results of the proposed method on two key problems in computer vision, namely multi-camera multiple people tracking and 3D human pose tracking. 
We compare our method with the following approaches for multi-view multiple object tracking:

\begin{itemize}
\item{Greedy Tracking-Reconstruction (GTR): first tracking is performed in 2D on a frame-by-frame basis using bipartite graph matching, and then 3D trajectories are reconstructed from the information of all cameras.}
\item{Greedy Reconstruction-Tracking (GRT): first 3D positions are reconstructed from all cameras. In a second step, 3D tracking is performed on a frame-by-frame basis using bipartite graph matching.}
\item{Tracking-Reconstruction (TR): first tracking is performed in 2D using \citep{zhangcvpr2008} and then 3D trajectories are recovered as in GTR.}
\item{Reconstruction-Tracking (RT): first the 3D positions are reconstructed as in GRT and then 3D tracking is performed using \citep{zhangcvpr2008}.}
\end{itemize}

Tests are performed on two publicly available datasets \citep{pets2009,sigal2010humaneva} and a comparison with existing state-of-the-art tracking approaches is made using the CLEAR metrics \citep{clear}, DA (detection accuracy), TA (tracking accuracy), DP (detection precision) and TP (tracking precision).

\begin{figure*}[htbp]
\centering
\subfigure[Tracking-Reconstruction]{
\includegraphics[trim=50mm 20mm 50mm 30mm, clip, width=0.28\linewidth]{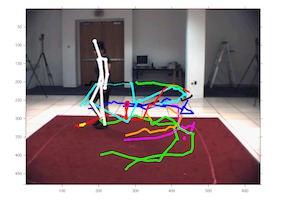} 
\label{fig:TR_outliers}}
\subfigure[Reconstruction-Tracking ]{
\includegraphics[trim=50mm 20mm 50mm 30mm, clip, width=0.28\linewidth]{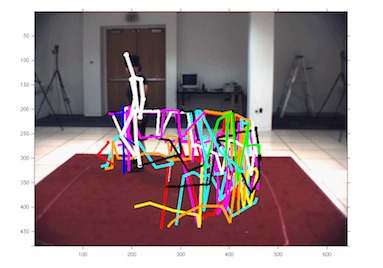} 
\label{fig:RT_outliers}}
\subfigure[Proposed method]{
\includegraphics[trim=50mm 20mm 50mm 30mm, clip, width=0.28\linewidth]{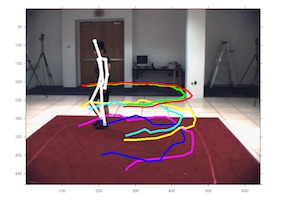} 
\label{fig:RT_DW}}
\caption[Visual results on the HumanEva dataset]{Even with $40\%$ of outliers our method \ref{fig:RT_DW} can recover the trajectories almost error free on the entire the sequence. This is in contrast to \ref{fig:TR_outliers} and \ref{fig:RT_outliers} that struggle with the ambiguities generated by the outliers. }
\label{fig:outliers}
\end{figure*}

\subsection{Multi-camera multiple people tracking}

\begin{figure}[ht]
\centering
\includegraphics[width=0.951\linewidth]{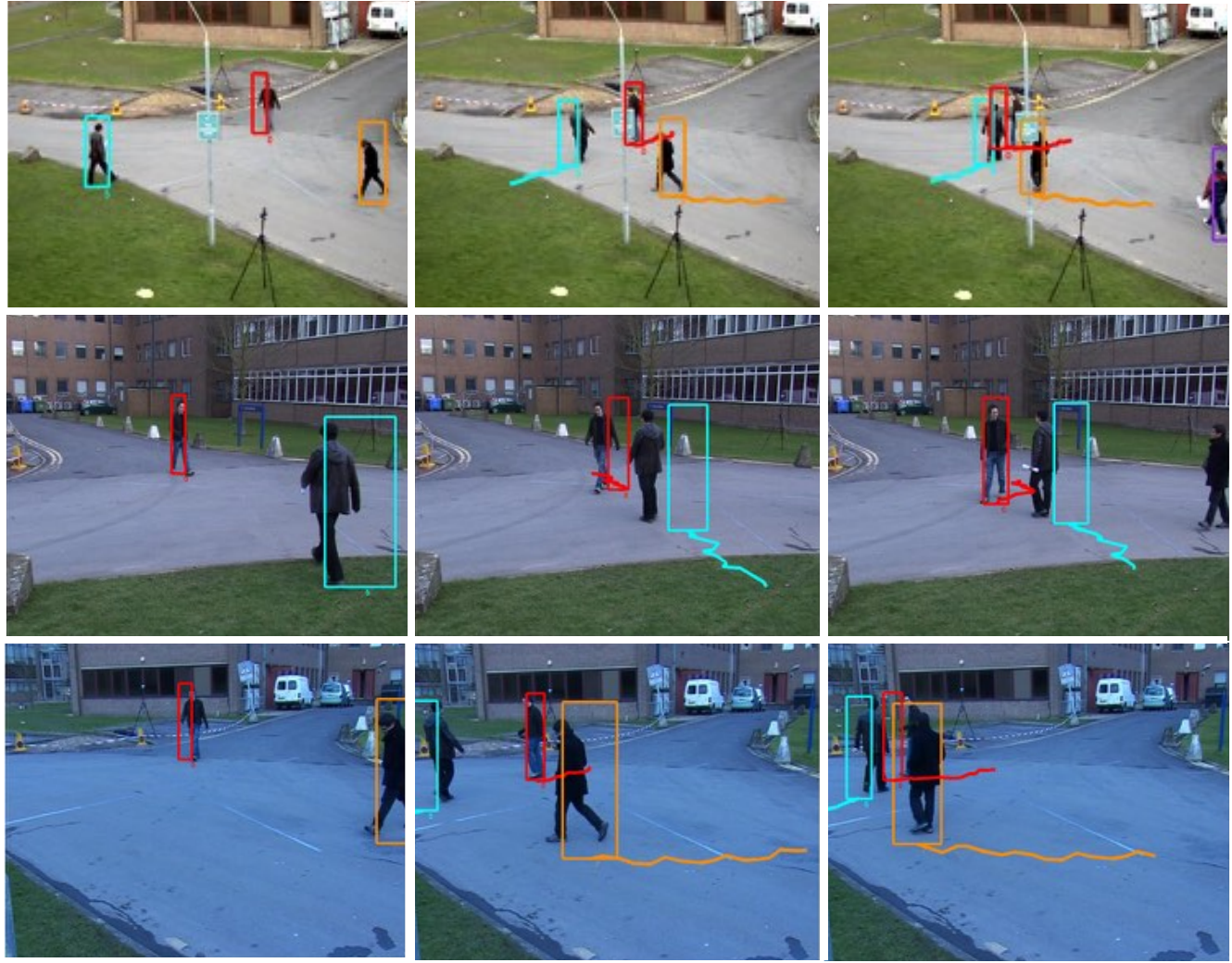} 
\caption[Tracking results on the PETS2009 sequence, 3 camera views]{Results on the PETS sequence, tracking with 3 camera views. Although there are clear 2D-3D inaccuracies, the proposed method is able to track the red pedestrian which is occluded in 2 cameras during 22 frames.}
\label{fig:PETS}
\end{figure}

In this section, we show the tracking results of our method on the publicly available PETS2009 dataset \citep{pets2009}, a scene with several interacting targets. Detections are obtained using the Mixture of Gaussians (MOG) background subtraction. 
For all experiments, we set $B_f=0.3$, $\textrm{E}_\textrm{3D}=0.5\textrm{ m}$ which represents the diameter of a person, $V_\textrm{max}^\textrm{2D}=$\unit[250]{pix/s} and $V_\textrm{max}^\textrm{3D}=$\unit[6]{m/s} which is the maximum allowed speed for pedestrians. 
Note that for this particular dataset, we can infer the 3D position of a pedestrian with only one image since we can assume $z=0$. Since we evaluate on view 1 and the second view we use does not show all the pedestrians, it would be unfair towards the RT and GRT methods to only reconstruct pedestrians visible in both cameras. Therefore, we consider the detections of view 1 as the main detections and only use the other cameras to further improve the 3D position. 
We also compare our results to monocular tracking using \citep{zhangcvpr2008} and multi-camera tracking with Probability Occupancy Maps and Linear Programming \citep{berclaztpami2011}. 
\begin{table}[htbp]
\begin {center}
  \begin{tabular}{ cccccc}
     & {DA} & {TA} & {DP} & {TP} & {miss}   \\ \hline
    Zhang \etal \citep{zhangcvpr2008} (1) & 68.9 & 65.8 & 60.6 & 60.0  & 28.1\\ 
    Greedy tracking-reconstruction (2) & 51.9 & 49.4 & 56.1 & 54.4 & 31.6\\ 
    Greedy reconstruction-tracking (2) & 64.6 & 57.9 & 57.8 & 56.8 & 26.8\\ 
    Tracking-reconstruction  (2) & 66.7 & 62.7 & 59.5 & 57.9 & 24.0\\ 
    Reconstruction-tracking  (2) &  69.7 & 65.7 & 61.2 & 60.2  & 25.1\\ 
     Leal-Taix{\'e} \etal \citep{lealcvpr2012} (2) & {\bf 78.0} & {\bf 76} &  {\bf 62.6} &  60  & {\bf 16.5} \\ \hline
    Tracking-reconstruction (3) & 48.5 & 46.5 & 51.1 & 50.3 & 20\\ 
    Reconstruction-tracking (3) &  56.6 & 51.3 & 54.5 & 52.8 & 23.5\\ 
    Leal-Taix{\'e} \etal \citep{lealcvpr2012} (3) & {73.1} & {71.4} & {55.0} & { 53.4} & {\bf 12.9} \\ \hline
    Berclaz \etal \citep{berclaztpami2011} (5) & 76 & 75 &  62 & {\bf 62} & $-$\\ \hline
    \end{tabular}
  \end{center}
    \caption{PETS2009 L1 sequence. Comparison of several methods tracking on a variable number of cameras (indicated in parenthesis).}
\label{tab:PETS2009mv}
\end{table}
As we can see in the results with 2 camera views, Table \ref{tab:PETS2009mv}, our algorithm \citep{lealcvpr2012} outperforms all other methods. 
In general, TR and RT methods perform better than their counterparts GRT and GTR, since matching across time with Linear Programming is robust to short occlusions and false alarms. Nonetheless, it still suffers from long term occlusions. In contrast, our method is more powerful than existing approaches when dealing with missing and noisy data, with misdetection rates 8.5 to 15 percentage points lower than other methods.
Notably, our method also outperforms \citep{berclaztpami2011} in accuracy, even though our results are computed using only 2 cameras instead of 5. 
When using 3 cameras, the 2D-3D inaccuracies become more apparent since the detections of the third camera project badly on the other two views (see Figure~\ref{fig:PETS}). Interestingly, RT and TR methods are greatly affected by these inaccuracies, while our method is more robust and still able to further reduce the missed detections by 4.6 percentage points.

\begin{figure}[ht]
\centering
\subfigure[Reconstruction-Tracking]{
\includegraphics[width=1\linewidth]{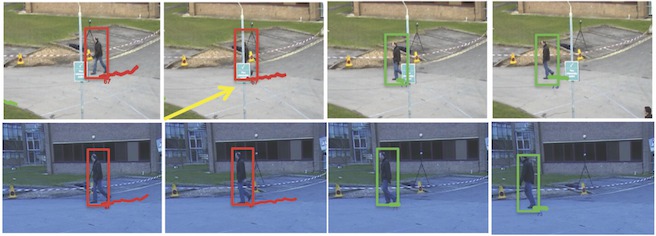} 
\label{fig:PETS1}}
\subfigure[Proposed method ]{
\includegraphics[width=1\linewidth]{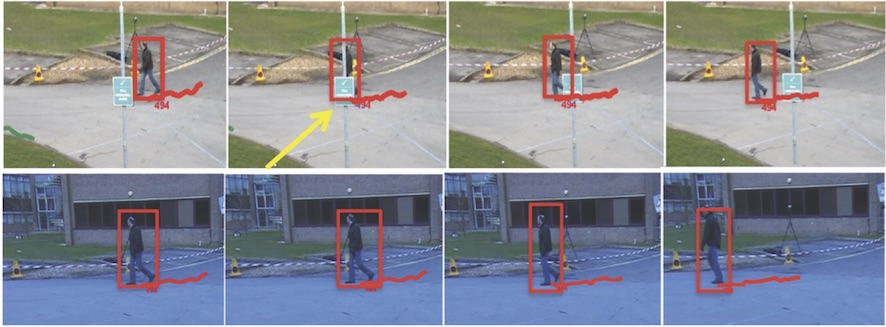} 
\label{fig:PETS2}}
\caption{Results on the PETS sequence, tracking with 2 camera views. Identity switch appears when using Reconstruction-Tracking, while the proposed method is able to correctly track the pedestrian even behind the pole.}
\label{fig:PETS_2cams}
\end{figure}

In Figure \ref{fig:PETS_2cams}, we can see an example with 2 camera views. A pedestrian hides behind a pole and therefore goes undetected for a number of frames in view 1. In this case, the RT method is not able to reconstruct any 3D position, and so a new track is initiated when the pedestrian is visible again in view 1. The advantage of the proposed approach is that, during the occlusion, the pedestrian can be tracked in view 2 using only 2D information. When he reappears in view 1 and therefore 3D information is available again, the method is able to correctly assign the same identity as he had before. It combines the power of RT methods to correctly identify pedestrians with the power of TR methods to track by usings only one view.

In Figure~\ref{fig:PETS}, we show an example with three camera views where a pedestrian (red) is occluded in two of the three views for a length of 22 frames. The RT method is unable to recover any 3D position, and therefore loses track of the pedestrian. The TR method tries to track the pedestrian in one view, but the gap is too large and TR fails to finally recover the whole 3D trajectory. The proposed method overcomes the long occlusion and the noisy 2D-3D correspondences to recover the full trajectory. We obtain a better accuracy than RT(3) by 13.5 percentage points which further proves the advantages of our approach.

\subsection{Human Motion}

\begin{figure*}[htp]
\centering
\subfigure[Tracking accuracy (\%) with missing data]{
\includegraphics[viewport = 58 258 540 532,width=0.5\linewidth]{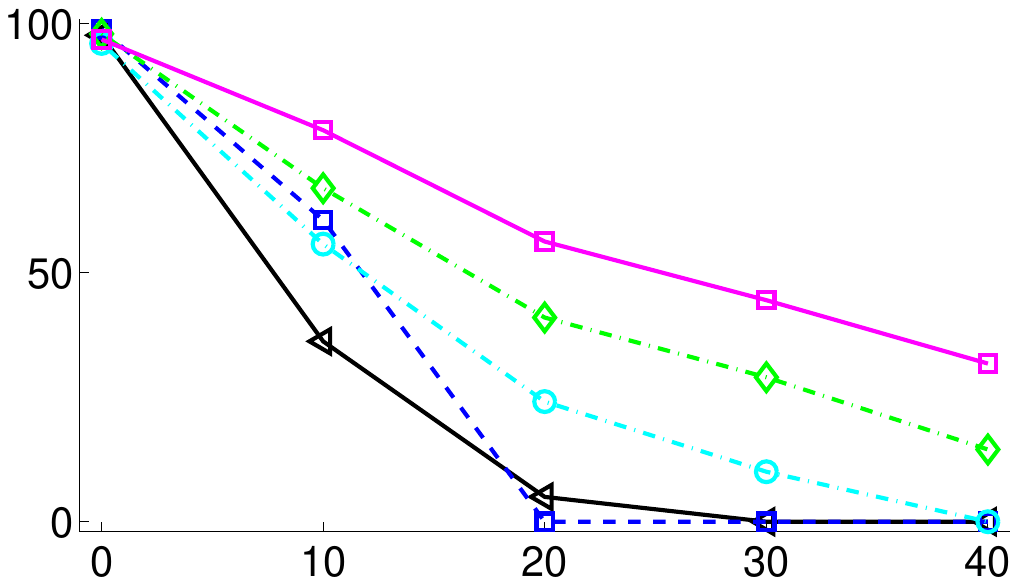} 
\label{fig:missing_graph}}
\hspace{0.1cm}
\subfigure[Tracking accuracy (\%) with outliers]{
\includegraphics[viewport = 58 258 540 532,width=0.5\linewidth]{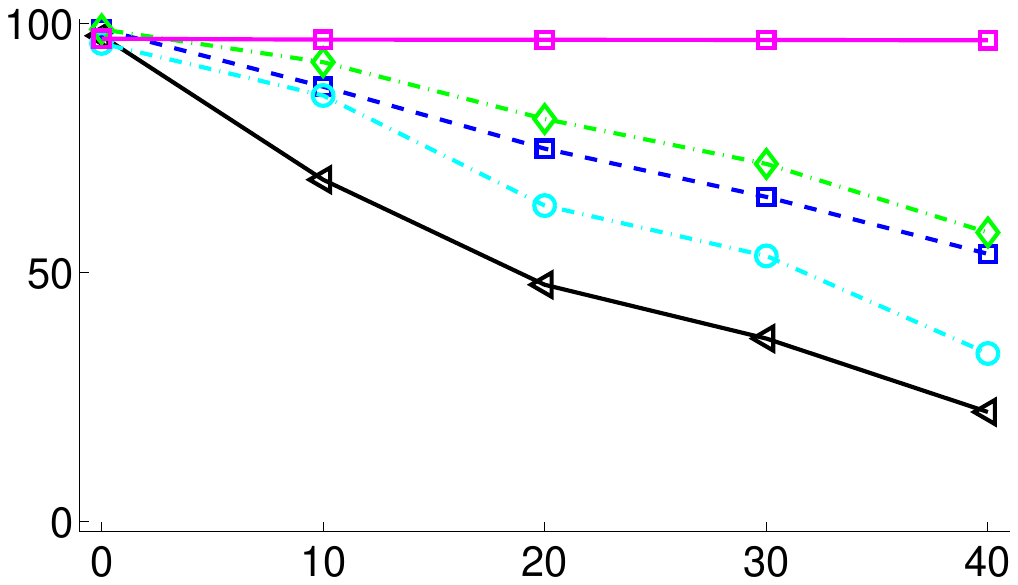} 
\label{fig:outliers_graph}}
\hspace{0.1cm}
\subfigure[ID switches with outliers]{
\includegraphics[viewport=57 256 540 529, width=0.5\linewidth]{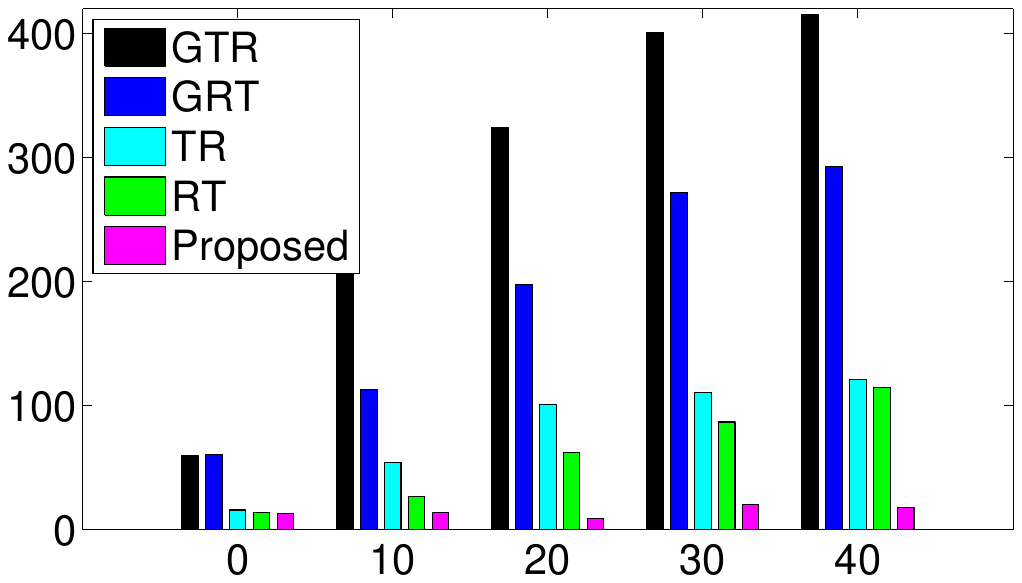} 
\label{fig:idsw}}
\caption[Evaluation of the robustness of the method]{Robustness evaluation: simulation of increasing rate of missing data \ref{fig:missing_graph} and increasing rate of outliers \ref{fig:outliers_graph},\ref{fig:idsw}.}
\label{fig:out_miss_graph}
\end{figure*}

We also tested our algorithm on the problem of human pose tracking using the publicly available human motion database \emph{HumanEva} \citep{sigal2010humaneva}. 
The problem we consider here is the following: given a set of 2D joint locations in two cameras, the goal is to link the locations across time and across cameras at every frame to reconstruct the sequence of poses. 
In these experiments, we use only two cameras at a reduced frame rate of 10 fps to reconstruct the 3D poses.
To obtain joint locations in the image, we project the ground truth 3D data using the known camera parameters. The parameters used are: $B_f=0.3$, $\textrm{E}_\textrm{3D}=0.01\textrm{ mm}$, $V_\textrm{max}^\textrm{2D}=$\unit[400]{pix/s} and $V_\textrm{max}^\textrm{3D}=$\unit[3]{m/s}.
We study the robustness of our algorithm to missing data and outliers. Missing data often occurs due to occlusions, while outliers appear as the result of false detections. 

\begin{figure*}[htbp]
\centering
\subfigure[Camera 1]{
\includegraphics[trim=50mm 20mm 50mm 30mm, clip, height=3.8cm]{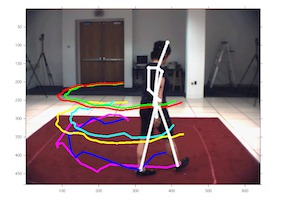} 
\label{fig:missTR_outliers}}
\subfigure[Camera 2 ]{
\includegraphics[trim=50mm 20mm 50mm 30mm, clip, height=3.8cm]{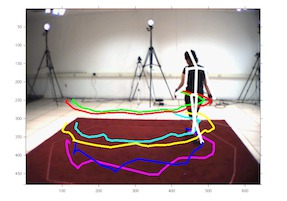} 
\label{fig:missRT_outliers}}
\subfigure[3D trajectories]{
\includegraphics[height=3.9cm]{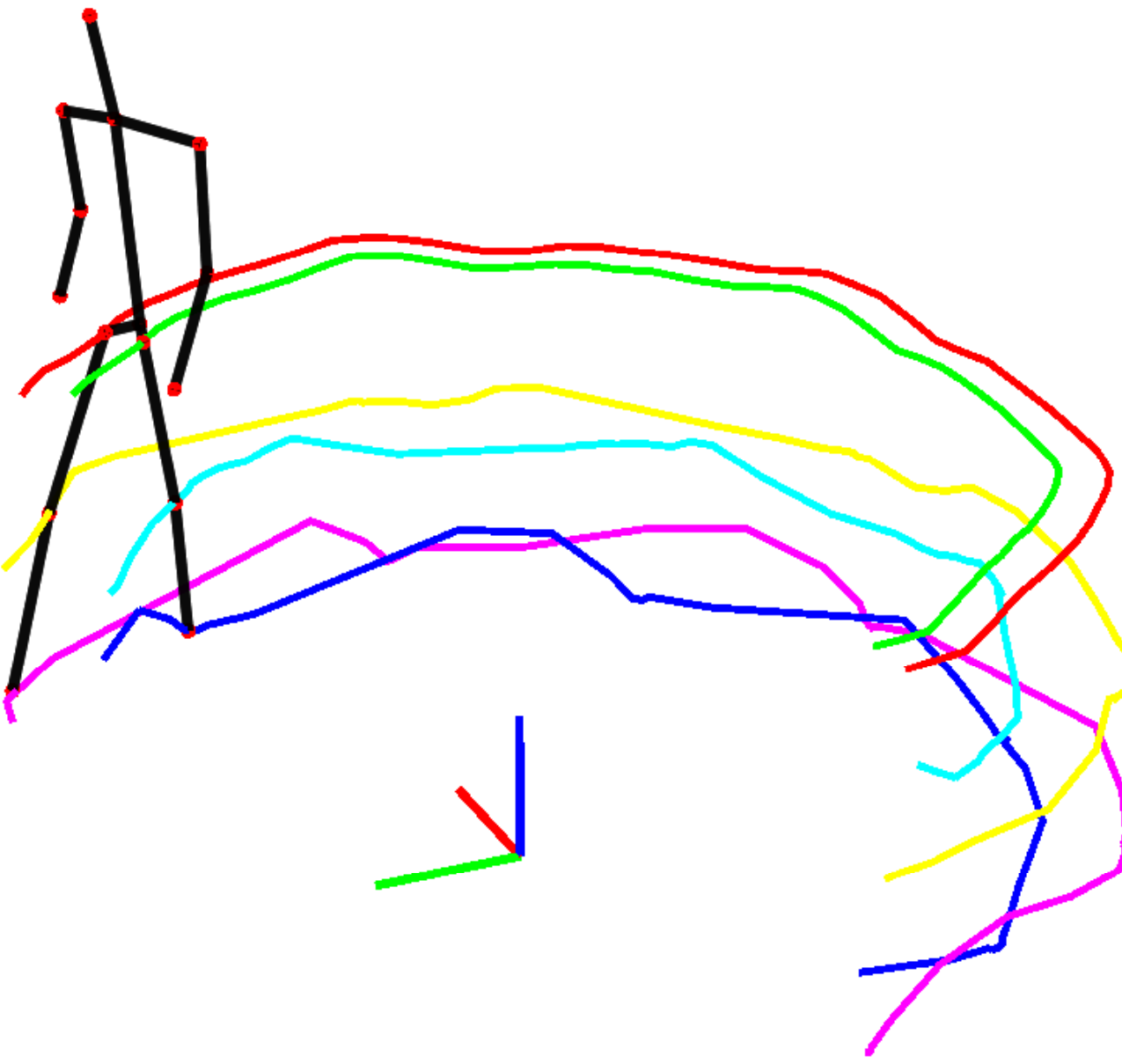} 
\label{fig:missRT_DW}}
\caption{Proposed method with $20 \%$ of missing data. Note that the trajectories are assigned the same ID in both views.}
\label{fig:missing_data}
\end{figure*}

\noindent{\bf Missing data:} To simulate missing data, we increasingly removed percentages of the 2D locations ranging from 0 to 40\%. As can be seen in Figure~\ref{fig:missing_graph}, our proposed method outperforms all other baselines and brings significant improvement. In Figure~\ref{fig:missing_data}, we show the trajectories of the lower body reconstructed with our method with $20\%$ of missing data. The 3D error for our method stays below 5\,mm, whereas it goes up to 10\,mm for the other methods. 

\noindent{\bf Outliers:} We added from 0\% to 40\% of uniformly distributed outliers in windows of $15\times15$ pixels centered at randomly selected 2D joint locations. Again, our method shows a far superior performance as the percentage of outliers increases, see Figure~\ref{fig:outliers_graph}. Notably, our method performs equally well independently from the number of outliers. 
Since outliers are uncorrelated across cameras, they produce lower prizes in the 3D layer of our graph and are therefore correctly disregarded during optimization. This clearly shows the advantage of globally exploiting temporal and 3D coherency information together. Here, the 3D error is only 2\,mm for our method. Furthermore, in Figure~\ref{fig:idsw}, we plot the count of the identity switches for an increasing number of outliers. Our method is the only one that is virtually unaffected by outliers, an effect that is also shown in Figure~\ref{fig:outliers}. This last result is particularly important for pose tracking, as ID switches result in totally erroneous pose reconstructions.

\section{Conclusions}

In this chapter, we presented a formulation to jointly track multiple targets in multiple views. 
The proposed graph structure captures both temporal correlations between objects as well as spatial correlations enforced by the configuration of the cameras and allows us to solve the problem as one global optimization. To find the global optimum, we used the powerful tool of branch-and-price, which allows us to exploit the special block-angular structure of the program to reduce computational time.
We tested the performance of the proposed approach on two key problems in computer vision: multiple people tracking and 3D human pose tracking. We outperform state-of-the-art approaches, which proves the strength of combining 2D and 3D constraints in a single global optimization. 




\chapter{Conclusions} 

\label{conclusions} 

\fancyhead[RE,LO]{Chapter 7. \emph{Conclusions}} 


In a world where video cameras are becoming an inherent part of our lives, it is becoming more important to develop methods to automatically analyze such data streams. 
Many tasks such as surveillance, animation or activity recognition need to have information about where people are located and how they are moving. Hence, multiple people tracking has become a classical problem in computer vision. Though a lot of research has been done in this field, there are still major challenges to overcome, especially in crowded environments.

In this thesis, we approached the problem of multiple people tracking using the paradigm of tracking-by-detection. Recent advances in detectors make it possible to have a reasonably stable detection rate even in moderately crowded scenarios. Nonetheless, occlusions and false alarms are still a big problem that has to be faced during the tracking step. 
We argued that classical tracking methods fail to fully exploit two sources of context, namely {\it social context} and {\it spatial context} coming from different views. Including this context in an efficient way within a global optimization tracker has been the main scope of this thesis. 


We first presented our tracking framework based on Linear Programming. Multiple people tracking is formulated as a unique optimization problem for all pedestrians in all frames, and a globally optimum solution is found for all trajectories. This already provides the perfect setup to introduce any kind of context, since the trajectories are inherently linked to each other. 

The first source of context we explored was that of the {\it social context}. In a scenario where a pedestrian walks alone, it is obvious he or she will follow a straight path towards his or her destination. Nonetheless, this path is affected by all kinds of obstacles, static or moving, in a real life scenario. This effect becomes increasingly apparent in crowded scenarios, therefore, we argued it is far more natural to include the environment and other moving targets during the multiple people tracking task. 
Most pedestrian movements and reactions to the environment are captured by what is called the Social Force Model. We presented a method to efficiently introduce social and grouping behavior into the Linear Programming tracker. The observation that people interaction is persistent rather than transient made it clear that the probabilistic formulation fully exploits the power of behavioral models, as opposed to standard predictive and recursive approaches, such as Kalman filtering.
Experiments were shown on several public datasets revealing the importance of using social interaction models for tracking under difficult conditions, such as crowded scenes with the presence of missed detections, false alarms and noise. Social information was proven to be specially useful in keeping the correct identity of a pedestrian, which is in the end the main goal of tracking. 

Even though the inclusion of social awareness in tracking improved trajectories significantly, there is only so much a tracker can do given a certain set of detections. 
Pedestrian pose, illumination or most commonly occlusions can make it hard to detect pedestrians under certain conditions. Using input from multiple cameras is the most common solution to increase the chances of detecting all pedestrians, specially in surveillance scenarios where the same space is filmed from several angles. 
Nonetheless, information coming from multiple cameras is typically combined in an ad-hoc fashion. Furthermore, object locations in the images are temporally correlated by system dynamics and are geometrically constrained by the spatial configuration of the cameras. 
These two sources of structure have been typically exploited separately, but splitting the problem in two phases has obviously several disadvantages, because the available evidence is not fully exploited. For example, if one object is temporarily occluded in one camera, both data association for reconstruction and tracking become ambiguous and underconstrained when considered separately. 
If, on the other hand, evidence is considered jointly, temporal correlation can potentially resolve reconstruction ambiguities and vice versa. 

The {\it spatial context} is the second source of context that we aimed at fully exploiting in this thesis. We proposed to create a unique graph structure capturing both temporal correlations between objects as well as spatial correlations enforced by the configuration of the cameras, allowing us to solve the problem as one global optimization. Given the large number of constraints and variables, is it intractable to solve this problem using standard Linear Programming solvers. We therefore 
used the powerful tool of branch-and-price to find the global optimum, which allowed us to exploit the special block-angular structure of the program to reduce computational time as well as to find a better lower bound.
Performance was tested for multiple people tracking, outperforming state-of-the-art approaches and proving the strength of combining 2D and 3D constraints in a single global optimization. The main strength was that pedestrians visible in only one view can be tracked in 2D, while pedestrians visible in several views can be tracked using 3D information as well, making the method very flexible and robust at the same time. 
Perhaps the most interesting contribution of our formulation is that it can be of considerable interest to model complex dependencies which arise in a wide range of computer vision problems. We also applied our method to 3D human pose tracking for which we obtained largely better results compared to classical approaches. 

One weakness of the method is that it is very sensitive to noise. For the multiple people tracking sequence, there are large calibration errors which reduced the accuracy of our results significantly. On the 3D human pose tracking dataset though, calibration is extremely accurate and therefore we can see results which are perfect even when we have up to 50\% of outliers present in the data. 
This is because the graph structure contains a high number of constraints that tightly link 2D and 3D information. If calibration is correct, this structure does not allow any tracking error and provides excellent accuracy results. Nonetheless, in practice we know there will be a certain percentage of errors in the 3D position estimation, introduced either by the camera calibration or simply by the detector which can wrongly estimate the 2D bounding box around a pedestrian.  
As future work, we would like to explore ways of relaxing the sensitivity of the method to noise while keeping the tight formulation. 

Another direction for improvement would be to find a solver with better computational complexity. Currently, the methods' complexity increases exponentially with the number of objects and cameras. In practice, it takes about one day to find the solution for one tracking sequence.

The work presented in this thesis has shown that context can be a key source of information that can significantly improve tracking results, especially if introduced in a global optimization framework which guarantees that this information will be fully exploited to improve all trajectories. 
Nonetheless, we believe that the tracking-by-detection framework has reached a saturation point in which results can now only be marginally improved. 
There is only so much that can be done to improve tracking given a certain detection set. 
Long occlusions are still a common unsolved problem, there are just too many assumptions that the tracker needs to make in order to correctly follow a pedestrian occluded for half of the sequence. 
We strongly believe that detection and tracking should not be treated as two separate tasks. Detection can benefit considerably from motion cues, while tracking can benefit from detailed appearance cues used commonly by detectors.


\addtocontents{toc}{\vspace{2em}} 

\appendix 



\chapter{A case study: microorganism tracking and motion analysis} 

\label{microorganism} 

\fancyhead[RE,LO]{Appendix A. \emph{A case study: microorganism tracking and motion analysis}} 

\graphicspath{{./Figures/Holography/}}

Throughout the thesis we have focused on tracking and motion analysis of pedestrians. Humans are usually the center of attention for many computer vision tasks, \eg, detection \citep{felzenszwalbtpami2010,galltpami2011}, tracking \citep{lealiccv2011,lealcvpr2012}, pose estimation \citep{ponsiccv2011,yangcvpr2011}, crowd analysis \citep{rodrigueziccv2011,alieccv2008}. Nonetheless, Computer Vision can be useful in many other fields where huge amounts of data need to be automatically analyzed, for example cell tracking \citep{kanade} for medical purposes.
In this Appendix, we present a case study where Computer Vision is proven to be useful for the field of marine biology and chemical physics. An automatic method is presented for the tracking and motion analysis of swimming microorganisms. This includes early work done by the author at the beginning of the PhD.

Many fields of interest in biology and other scientific research areas deal with intrinsically three-dimensional problems. 
The motility of swimming microorganisms such as bacteria or algae is of fundamental importance for topics like pathogen-host interactions \citep{heid6}, predator-prey interactions \citep{heid6}, biofilm-formation \citep{heid2}, or biofouling by marine microorganisms \citep{heid4,heid5}. 

\begin{figure}[htb]
\centering
\subfigure[]{
\includegraphics[width=0.46\linewidth]{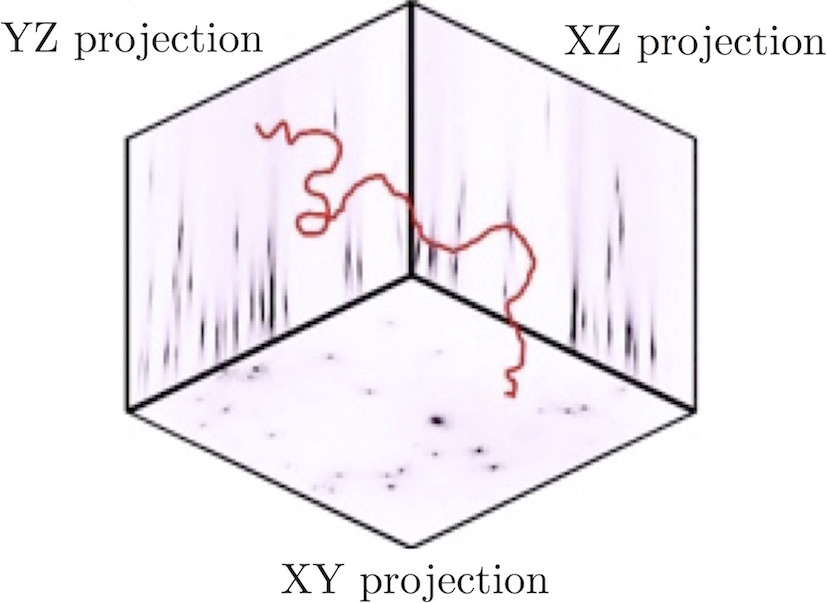} 
\label{fig:speed1}
}
\subfigure[]{
\includegraphics[width=0.49\linewidth]{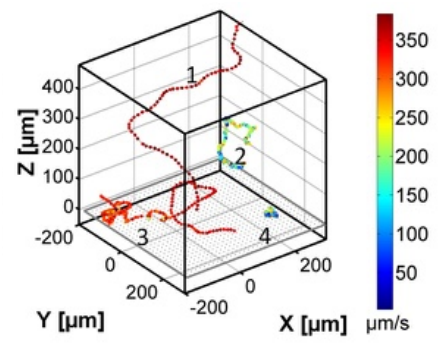} 
\label{fig:speed2}
}
\caption[Digital in-line holography input data and microorganism motion patterns]{\subref{fig:speed1} The input data, the projections obtained with digital in-line holography (inverted colors for better visualization). Sample trajectory in red. \subref{fig:speed2} The output data we want to obtain from each volume, the classification into four motion patterns, colored according to speed: orientation (1), wobbling (2), gyration (3) and intensive surface probing (4).}
\label{fig:speedpattern}

\end{figure}

We present a complete system for the automatic analysis of digital in-line holographic data. This microscopy technique provides videos of a 3D volume, see Figure \ref{fig:speedpattern}, and is used to study complex movements of microorganisms. The huge amount of information that we can extract from holographic images makes it necessary to have an automatic method to analyze this complex 4D data. 
Our system performs the detection of 3D positions, tracking of complete trajectories and classification of motion patterns. 
For multiple microorganism tracking, we propose a geometrically motivated and globally optimal multi-level Hungarian to compensate for leaving and entering particles, recover from missing data and erase outliers to reconstruct the trajectories of the microorganisms \citep{lealwmvc2009}. Afterwards, we focus on the classification of four motion patterns of the green alga {\it Ulva linza} with the use of Hidden Markov Models \citep{lealdagm2010}.
Furthermore, our system is able to find and separate different patterns within a single sequence. Besides classification of motion patterns, a key issue is the choice of features used to classify and distinguish the involved patterns. For this reason, we perform an extensive analysis of the importance of typical motion parameters, such as velocity, curvature, orientation, etc. The system we developed is highly flexible and can easily be extended.
Especially for forthcoming work on cells, microorganisms or human behavior, such automated algorithms are of pivotal importance, as they allow high throughput analysis of individual segments in motion data.

\section{Related work}

Understanding the motility and behavioral patterns of microorganisms allows us to understand their interaction with the environment and thus to control environmental parameters to avoid unwanted consequences such as infections or biofouling. To study these effects in 3D several attempts have been made: tracking light microscopy, capable of tracking one bacterium at a time \citep{heid9}, stereoscopy \citep{heid10} or confocal microscopy \citep{heid11}.

Berg built a pioneering tracking light microscope, capable of tracking one bacterium at a time in 3D. This has been used to investigate bacteria like Escherichia Coli \citep{heid9}.
Another way of measuring 3D trajectories is stereoscopy, which requires two synchronized cameras \citep{heid10}. Confocal microscopy has also been used to study the motion of particles in colloidal systems over time, however the nature of this scanning technique limits the obtainable frame rate \citep{heid11}. 


For any of these techniques, in order to draw statistically relevant conclusions, thousands of images have to be analyzed. 
Nowadays, this analysis is still heavily dependent on manual intervention.
Recent work \citep{kanade} presents a complete vision system for 2D cell tracking, which proves the increasing demand for efficient computer vision approaches in the field of microscopy as an emerging discipline.
Research on the automatic analysis of biological images is extensive \citep{general}, but most of the work focuses on position as well as on the shape of the particle \citep{metaxas}.
Several methods exist for multiple object detection based on methods such as Markov Chain Monte Carlo (MCMC) \citep{khantpami2005}, inference in Bayesian networks \citep{nilliuscvpr2006} or the Nash Equilibrium of game theory \citep{yangiccv2007}. These have been proven useful to track a fairly small number of targets but are less appropriate when the number of targets is very large, as in our case.
Statistical methods like Kalman filters \citep{kanade}, particle filters or recursive Bayesian filters \citep{betkecvpr2007} are widely used for tracking but they need a dynamical model of the target, a requirement that can be challenging to fulfill depending on the microorganism under study and to which we dedicate the second part of this paper. 
In contrast to \citep{betkecvpr2007,kanade}, we do not use the output predictions of the filters to deal with occlusions, but rather use past and future information to complete broken trajectories and detect false alarms. Therefore, we do not need an extra track linking step as in \citep{kanade}. 
Furthermore, we deal with 3D trajectories of random and fast motions which are unsuited for a prediction-based approach. 
In this work we propose a global optimal matching solution and not a local one as suggested in \citep{heid17}.

Besides generating motion trajectories from microscopic data, a subsequent classification allows biologists to get the desired information from large image sets in a compact fashion.  
Indeed, the classification of motion patterns in biology is a well-studied topic \citep{berg}, but identifying these patterns manually is a complicated and time consuming task. Recently, machine learning and pattern recognition techniques have been introduced to analyze such complex movements in detail. These techniques include: Principal Component Analysis (PCA) \citep{hoyle}, a linear transformation used to analyze high dimensional data; Bayesian models \citep{wang}, which use a graph model and the rules of probability theory to select among different hypotheses; Support Vector Machines (SVM) \citep{guyon}, which use training data to find the optimum parameters of the model representing each class. A comparison of machine learning approaches applied to biology can be found in \citep{sbalzariniy}.
In order to classify biological patterns, we need to use an approach able to handle time-varying signals. 
Hidden Markov Models \citep{HMM_rabiner} are statistical models especially known for their application in temporal pattern recognition. 
They were first used in speech recognition, and since then HMMs have been extensively applied to vision. Applications vary from handwritten word recognition \citep{HMM_hand}, face recognition \citep{HMM_Face} or human action recognition \citep{HMM_human1,HMM_human2}.



\section{Detection of 3D positions}

In this section, we present the details of digital in-line holography, how this microscopy technique allows us to obtain 3D positions of microorganisms as well as the image processing methods used to robustly extract these positions from the images.

\subsection{Digital in-line holographic microscopy (DIHM)}
\label{holography}

\begin{figure}[htp] 
   \centering
   \includegraphics[width=0.7\linewidth]{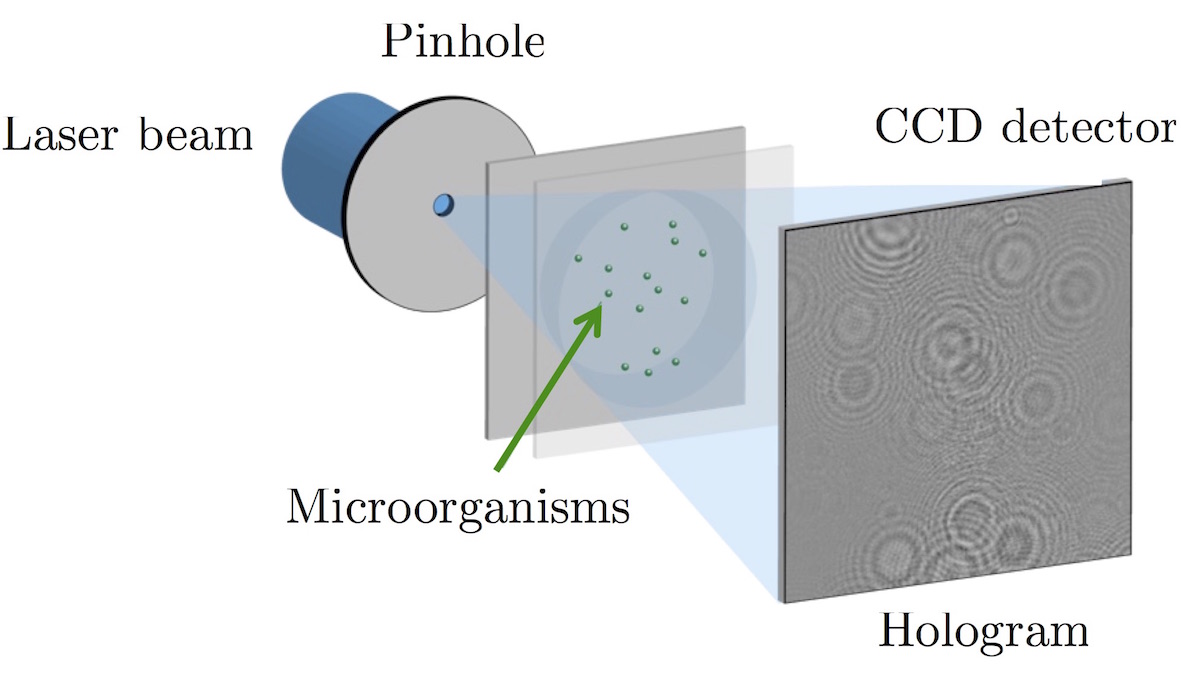} 
   \caption[Digital in-line holography setup]{Schematic setup for a digital in-line holographic experiment consisting of the laser, a spatial filter to create the divergent light cone, the objects of interest (e.g. microorganisms) and a detector which records the hologram.}
   \label{fig:setup}
\end{figure}

Digital in-line holographic microscopy provides an alternative, lensless microscopy technique which intrinsically contains three dimensional information about the investigated volume. It does not require a feedback control which responds to motion and it uses only one CCD chip. This makes the method very straightforward and in practice can be implemented with a very simple setup as shown in Figure \ref{fig:setup}. 

The holographic microscope requires only a divergent wavefront which is produced by diffraction of laser light from a pinhole. A CCD chip finally captures the hologram. The holographic microscope setup follows directly Gabors initial idea \citep{heid12} and has been implemented for laser radiation by Xu et al. \citep{heid13}. 
A hologram recorded without the presence of particles, called the {\it source}, is subtracted from each hologram. This is used to reduce the constant illumination background and other artifacts; there are filtering methods \citep{raupach,fugal} to achieve this in case a {\it source} image is not readily available.  
These resulting holograms can then be reconstructed back into real-world coordinates by a Kirchhoff-Helmholtz transformation \citep{heid13} shown in Equation \eqref{eq:transformation}.

\begin{equation}
\centering
K(\mathbf{r})=\int_{S} d^2 \xi I(\mathbf{\xi}) e^\frac{i k \mathbf{r}\cdot\mathbf{\xi}}{|\mathbf{\xi}|} 
\label{eq:transformation}
\end{equation}

The integral extends over the 2D surface of the screen with coordinates $\mathbf{\xi} = (X,Y,L)$, where $L$ is the distance from the source (pinhole) to the center of the detector (CCD chip), $I(\mathbf{\xi})$ is the contrast image (hologram) on the screen obtained by subtracting the images with and without the object present and $k$ the wave number: $k=2\pi/\lambda$.

\begin{figure}[htp] 
   \centering
   \includegraphics[width=1\linewidth]{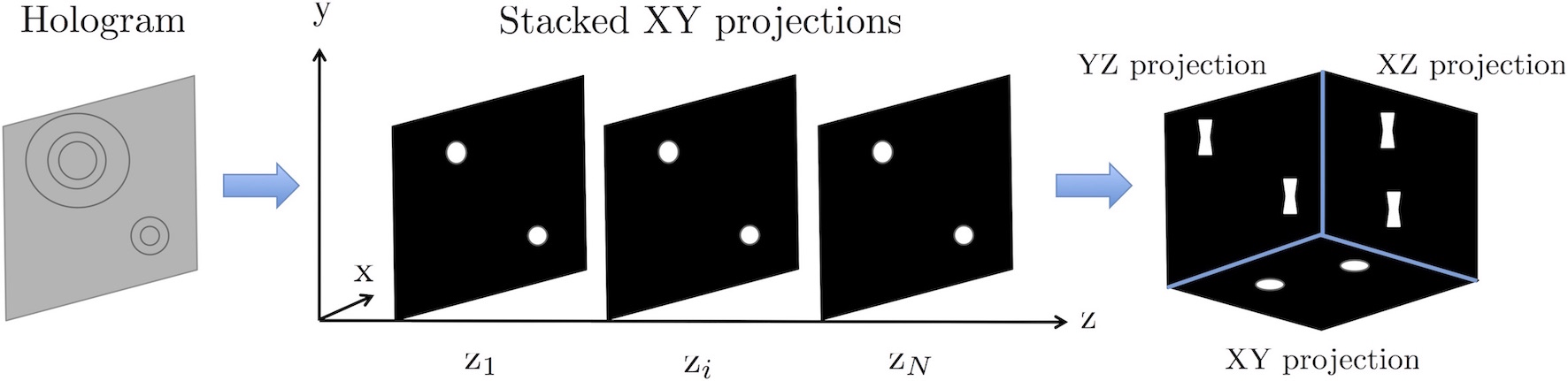} 
   \caption[Reconstruction process for digital in-line holography]{Illustration of the reconstruction process. From the hologram a stack of $XY$ projections is obtained in several depths and from those, the final 3 projections ($XZ$, $XZ$ and $YZ$) are obtained.}
   \label{fig:recon}
\end{figure}

As we can see in Figure \ref{fig:recon}, the idea behind the reconstruction is to obtain a series of stacked $XY$ projections from the hologram image. These projections contain the information at different depth values. 
From these images, we can obtain the 3 final projections $XY$, $XZ$ and $YZ$, as described in \citep{heidelberg1}.
These projections contain the image information of the complete observation volume, i.e. from every object located in the light cone between pinhole and detector. 
The resolution in $X$ and $Y$ is $\delta_{x,y}=\frac{\lambda}{NA}$, where $NA$ stands for the numerical aperture given by $NA=\frac{D}{2L}$, where $D$ is the detector's side length. The resolution in the $Z$ direction, that is the direction of the laser, is worse, $\delta_{z}=\frac{\lambda}{2NA^2}$.
This is because the third dimension, $Z$, is obtained with a mathematical reconstruction, unlike confocal microscopy, where the value of every voxel is returned. 
On the other hand, confocal microscopes take a long time to return the values of all voxels in a volume, and are therefore unsuited for tracking at a high frame rate. 

Using video sequences of holograms, it is possible to track multiple objects in 3D over time at a high frame rate, and multiple spores present in a single frame can be tracked simultaneously \citep{heid4,heid17,heid16}. 
Using this advantage of digital in-line holographic microscopy, a number of 3D phenomena in microbiology have been investigated: Lewis et al. \citep{heid18} examined the swimming speed of Alexandrium (Dinophyceae), Sheng et al. \citep{heid19,sheng2010} studied the swimming behavior of predatory dinoflagellates in the presence of prey, and Sun et al. \citep{heid20} used a submersible device to investigate  in situ  plankton in the ocean.

\subsection{Detection of the microorganisms}

\begin{figure}[htb]
\centering
\subfigure[]{
\includegraphics[width=0.59\linewidth]{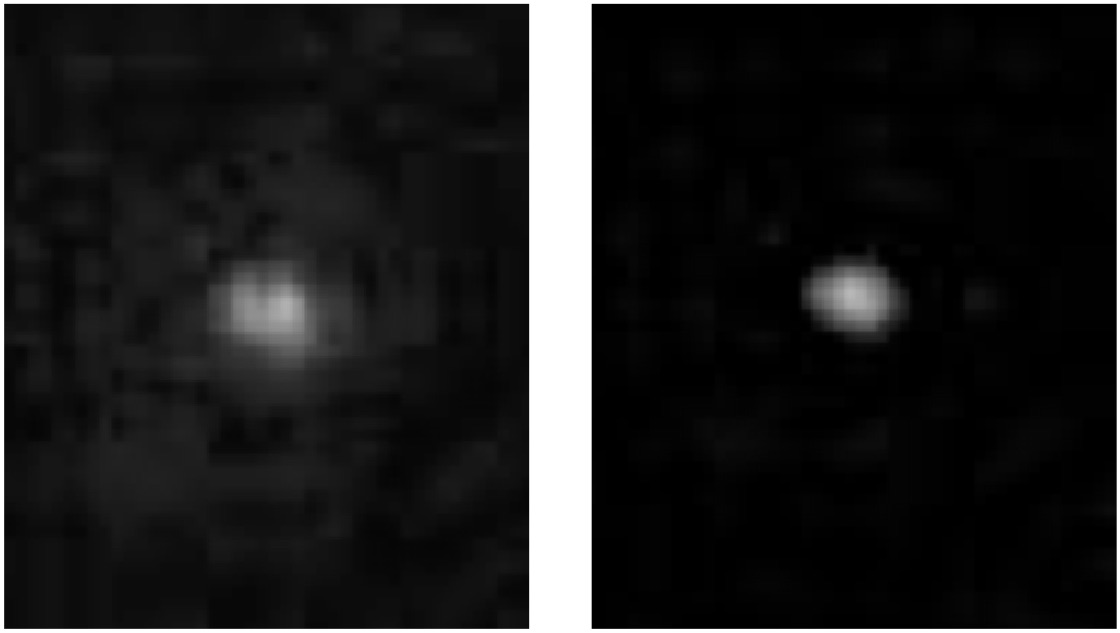} 
\label{fig:LoG1}
}
\subfigure[]{
\includegraphics[width=0.357\linewidth]{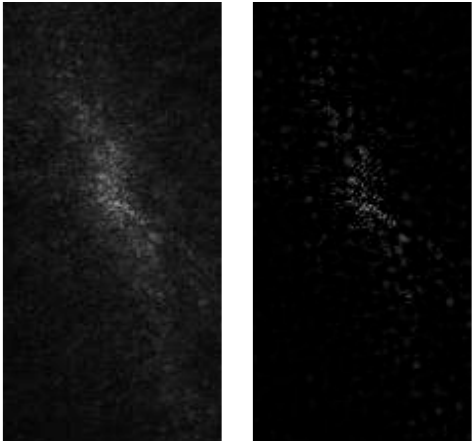} 
\label{fig:LoG2}
}
\caption[Effects of the Laplacian of Gaussian Filter]{\subref{fig:LoG1} Enhancement of the shape of the microorganisms. \subref{fig:LoG2} Reduction of the noise.}
\label{fig:LoG}

\end{figure}

As we saw in Chapter \ref{trackingbydetection}, we can use information such as edges or color histograms in order to detect humans, or we can build more complex models from training data in order to robustly detect humans in single images or in videos with moving cameras. In our case we can use simpler detection methods since the shape of our targets is much more constant than that of a human.
In our sequences, we are observing the green algae {\it Ulva linza} which has a spherical spore body and four flagella. Since the body scatters most of the light, in the projected images the particles have a circular shape. In order to preserve and enhance the particle shape (see Figure \ref{fig:LoG1}) but reduce noise and illumination irregularities of the image (see Figure \ref{fig:LoG2}), we apply the Laplacian of Gaussian filter (LoG), which, for its shape, is a blob detector \citep{blob}:

\begin{equation}
\centering
LoG(x,y)=\frac{-1}{\pi\sigma^4} \left[ 1- \frac{x^2+y^2}{2\sigma^2} \right] e^{-\frac{x^2+y^2}{2\sigma^2}}
\label{eq:LoG}
\end{equation}


Due to the divergent nature of the light cone, the particles can appear smaller or larger in the projections depending on the z-plane. Therefore, the LoG filter is applied in several scales \citep{blob} according to the magnification.
Note that the whole algorithm is extremely adaptable, since we can detect differently shaped microorganisms by just changing the filter. 
After this, we use thresholding on each projection to obtain the positions of candidate particles in the image. 
The final 3D positions (Figure \ref{fig:multidiagram}, green box labeled "Candidate particles") are determined by thresholding each projection $XY$, $XZ$ and $YZ$ to find the particles in each image and crossing the information of the three projections.

\begin{figure}[htp] 
   \centering
   \includegraphics[width=1\linewidth]{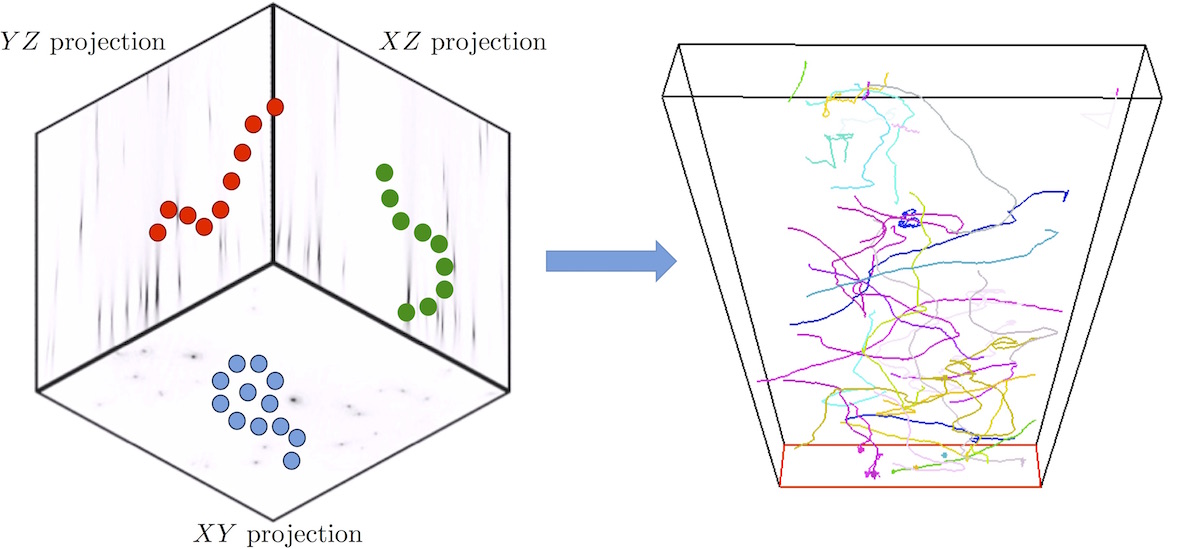} 
   \caption[Overview of the tracking-by-detection process]{From the 3D positions obtained at each time frame, we use the method in Section \ref{tracking} to obtain the full trajectory of each microorganism.}
   \label{fig:link}
\end{figure}

Once we have computed the 3D positions of all microorganisms in all frames, we are interested in linking these 3D positions in order to find their complete 3D trajectories over time (see Figure \ref{fig:link}).


\section{Automatic extraction of 3D trajectories}
\label{tracking}

\begin{figure}[htp] 
   \centering
   \includegraphics[width=0.9\linewidth]{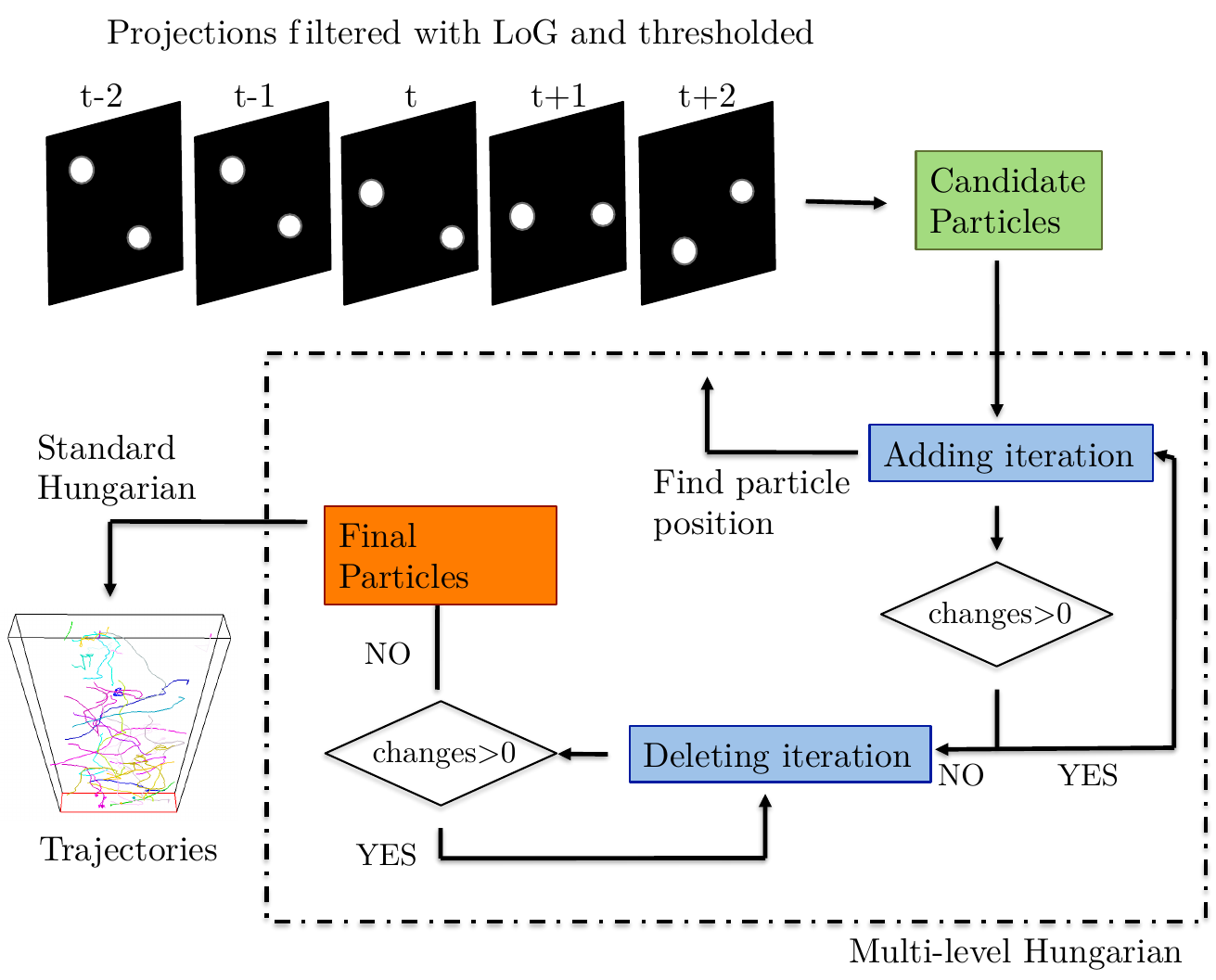} 
   \caption{Diagram of the algorithm described in Section \ref{multihungsection}.}
   \label{fig:multidiagram}
\end{figure}

In this section we present the complete method to estimate the 3D trajectories of microorganisms over time. Our algorithm, the Multi-level Hungarian, is a robust method evolved from the Hungarian-Munkre's assignment method and is capable of dealing with entering and leaving particles, missing data and outliers. The diagram of the method is presented in Figure \ref{fig:multidiagram}.

\subsection{Cost function and bipartite graph matching}

Let us briefly refresh some of the key concepts that we have seen in Chapter \ref{linearprogramming}.
Graph Matching is one of the fundamental problems in Graph Theory and it can be defined as: given a graph $G=(V,E)$, where $E$ represents its set of edges and $V$ its set of nodes or vertices, a matching $M$ in $G$ is a set of pairwise non-adjacent edges, which means that no edges share a common vertex.
For our application, we are specially interested in the {\it Assignment Problem}, which consists in finding a maximum weight matching in a weighted bipartite graph. 
In a general form, the problem can be expressed as: "There are N jobs and N workers. Any worker can be assigned to any job, incurring some cost that varies depending on the job-worker assignment. All jobs must be performed by assigning exactly one worker to each job in such a way that the total cost is minimized (or maximized)". 
For the subsets of vertices $X$ and $Y$, such that $V = X \cup Y$ and $X \cap Y = \emptyset$, we build a cost matrix in which the element $C(i,j)$ will represent the weight or cost related to the edge connecting $i$ in $X$ and $j$ in $Y$.



For numerical optimization, we use the Hungarian or Munkres' assignment algorithm, a combinatorial optimization algorithm \citep{hungarian,munkres} that solves the bipartite graph matching problem in polynomial time. 
For implementation details on the Hungarian, we recommend \citep{implementation1}.
Our initial problem configuration is: there are $M$ particles in frame $t_1$ and $N$ particles in frame $t_2$. The Hungarian will help us to determine which particle in $t_1$ corresponds to which particle in $t_2$, allowing us to reconstruct their full trajectories in 3D space.
Nonetheless, the Hungarian algorithm has some disadvantages which we should be aware of. In the context of our project, we summarize in Table \ref{tab:adv} some of the advantages and disadvantages of the Hungarian algorithm.
%



\begin{table}[htbp]
\begin {center}
  \begin{tabular}{ p{11cm} }
    {\bf ADVANTAGES} \\ \hline
    Finds a global solution for all vertices \\ 
     Cost matrix is versatile  \\ 
    Easy to solve, bipartite matching is the simplest of all graph problems \\ \\
    {\bf DISADVANTAGES} \\ \hline
    Cannot handle missing vertices (a)\\ 
    Cannot handle entering or leaving particles (b) \\ 
    No discrimination of matches even if the cost is very high (c)\\ 
  \end{tabular}
  \end{center}
    \caption{Summary of the advantages and disadvantages of the Hungarian algorithm.}
\label{tab:adv}
\end{table}

In the following sections, we present how to overcome the three disadvantages: (a) is solved with the multi-level Hungarian method explained in Section \ref{multihungsection}, (b) is solved with the IN/OUT states of Section \ref{inoutsection} and finally a solution for (c) is presented in Section \ref{costsection} as a maximum cost restriction.



The cost function $C$, as key input for the Hungarian algorithm, is created using the Euclidean distances between particles, that is, element $C(i,j)$ of the matrix represents the distance between particle $i$ of frame $t_1$ and particle $j$ of frame $t_2$.
With this matrix, we need to solve a {\it minimum assignment problem}, since we are interested in matching those particles which are close to each other. 

Note that it is also possible to include other characteristics of the particle, like speed, size or gray level distribution, in the cost function. Such parameters can act as additional regularizers during trajectory estimation.


\subsubsection{IN and OUT states}
\label{inoutsection}

In order to include more knowledge about the environment in the Hungarian algorithm and avoid matches with very high costs, we have created a variation of the cost matrix.
In our experiments, particles can only enter and leave the scene by crossing the borders of the Field Of View (FOV) of the holographic microscope, therefore, the creation and deletion of particles depends on their distance to the borders of the FOV. 
Nonetheless, the method can be easily extended to situations where trajectories are created (for example by cell division) or terminated (when the predator eats the prey) away from the FOV borders.

\begin{figure}[htp] 
   \centering
   \includegraphics[width=0.7\linewidth]{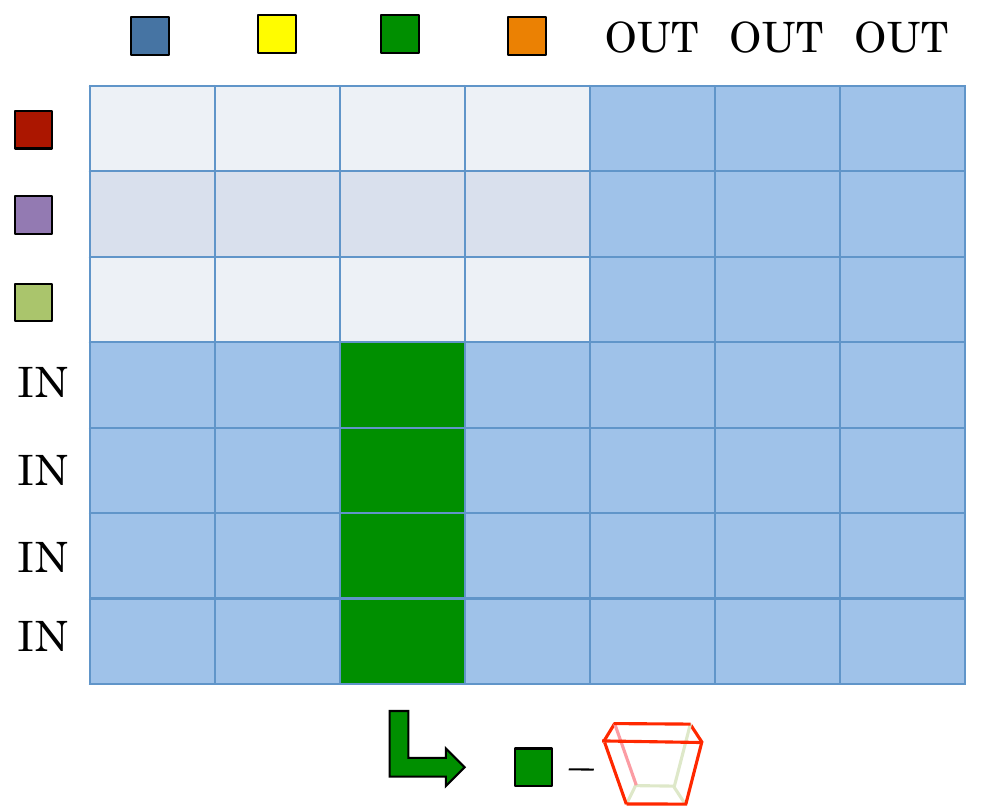} 
   \caption[Change in the cost matrix to include the IN/OUT states]{Change in the cost matrix to include the IN/OUT states. Each particle is represented by a different color. The value of each extra element added is the distance between the particle position and the closest volume boundary.}
   \label{fig:costmatrix}
\end{figure}

As shown in Figure \ref{fig:costmatrix}, we introduce the IN/OUT states in the cost matrix by adding extra rows and columns. If we are matching the particles in frame $f$ to particles in frame $f+1$, we will add as many columns as particles in frame $f$ and as many rows as particles in frame $f+1$. This way, all the particles have the possibility to enter/leave the scene. Additionally, this allows us to obtain a square matrix, needed for the matching algorithm, even if the number of particles is not the same in consecutive frames.
%

The cost of the added elements includes the information of the environment by calculating the distance of each particle to the nearest edge of the FOV. 
Note that the lower border of the $z$ axis is not included, as it represents the surface where the microorganisms might settle and, therefore, no particles can enter or leave from there.

If the distance is small enough, the Hungarian algorithm matches the particle with an IN/OUT state.

\begin{figure}[htb]
\centering
\subfigure[]{
\includegraphics[width=0.48\linewidth]{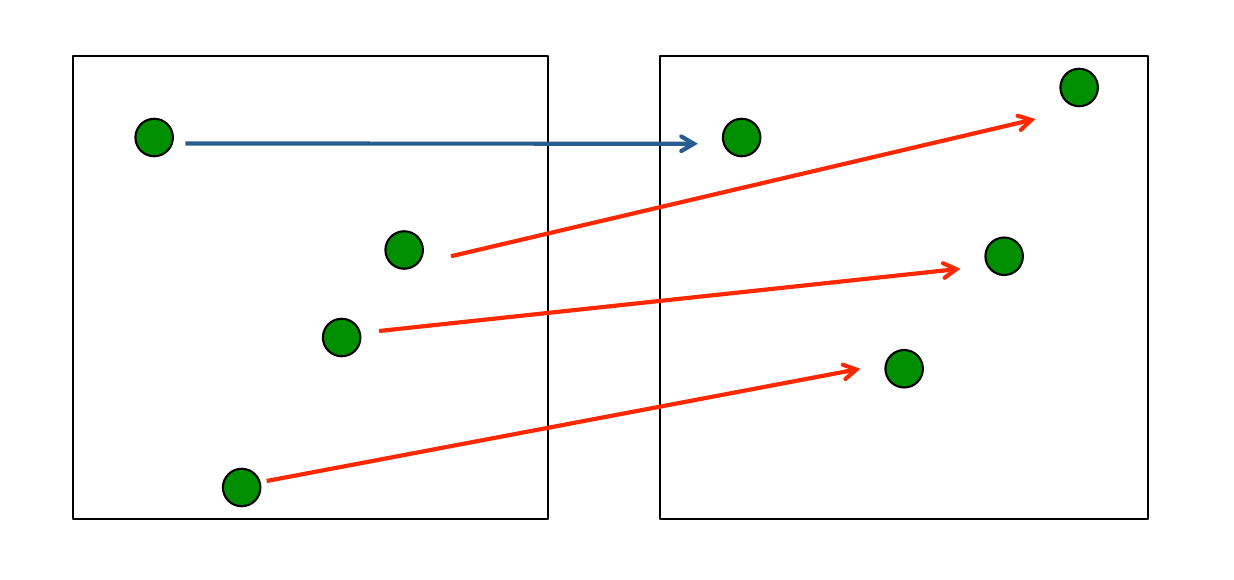} 
\label{fig:wrong}
}
\subfigure[]{
\includegraphics[width=0.48\linewidth]{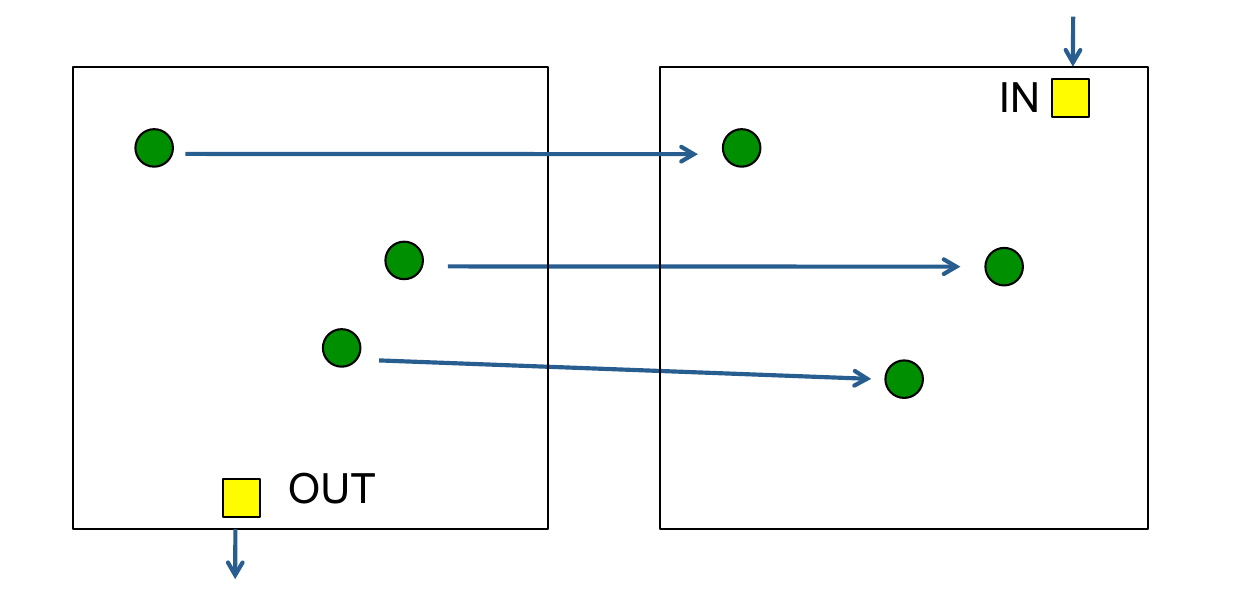} 
\label{fig:right}
}
\caption[Effect of the IN/OUT states on the matchings]{Representation of the particles in frame $t_1$ (left) and $t_2$ (right). The lines represent the matchings. \subref{fig:wrong} Wrongly matched. \subref{fig:right} Correctly matched as a result of the IN/OUT state feature.}
\label{fig:inoutstate}
\end{figure}

In Figure \ref{fig:inoutstate}, we consider the simple scenario in which we have 4 particles in one frame and 4 in the next frame. As we can see, there is a particle which leaves the scene from the lower edge and a particle which enters the scene in the next frame from the right upper corner.
As shown in Figure \ref{fig:wrong}, the Hungarian algorithm finds a wrong matching since the result is completely altered by the entering/leaving particles. 
With the introduction of the IN/OUT state feature, the particles are now correctly matched (see Figure \ref{fig:right}) and the ones which enter/leave the scene are identified as independent particles.

\subsubsection{Maximum cost restriction}
\label{costsection}

Due to noise and illumination irregularities of the holograms, it is common that a particle is not detected in several frames, which means a particle can virtually disappear in the middle of the scene. If a particle is no longer detected, all matches can be greatly affected.
That is why we introduce a maximum cost restriction for the cost matrix which does not allow matches with costs higher than a given threshold $V$.
This threshold is the observed maximum speed of the algae spores under study \citep{heidelberg1}. 
The restriction is guaranteed by using the same added elements as the ones used for the IN/OUT states, therefore
%
if a particle is near a volume border or cannot be matched to another particle which is within a reachable distance, it will be matched to an IN/OUT state. This ensures that the resulting matches are all physically possible.
Still, if we have missing data and a certain particle is matched to an IN/OUT state, we will recover two trajectories instead of the complete one. In the next section, we present a hierarchical solution to recover missing data by extending the matching to the temporal dimension.

\subsection{Multi-level Hungarian for missing data}
\label{multihungsection}


If we consider just the particles detected using thresholding, we see that there are many gaps within a trajectory (see Figure \ref{fig:merging}). 
These gaps can be a result of morphing (different object orientations yield different contrast), changes in illumination, etc.
The standard Hungarian is not capable of filling in the missing data and creating full trajectories, therefore, we now introduce a method based on the standard Hungarian that allows us to deal with missing data, outliers and create full trajectories. 
The general routine of the algorithm, the multi-level Hungarian, is:

\begin{itemize}
\item Find the matchings between particles in frames $[i-2 \ldots i+2]$, so we know the position of each particle in each of these frames (if present). (Section \ref{levels}).
\item Build a table with all these positions and fill the gaps given some strict conditions. Let the algorithm converge until no particles are added. (Section \ref{conditions}).
\item On the same table and given some conditions, erase the outliers. Let the algorithm converge until no particles are deleted. (Section \ref{conditions}).
 \end{itemize}


\subsubsection{The levels of the multi-level Hungarian}
\label{levels}

\begin{figure}[htp] 
   \centering
   \includegraphics[width=0.7\linewidth]{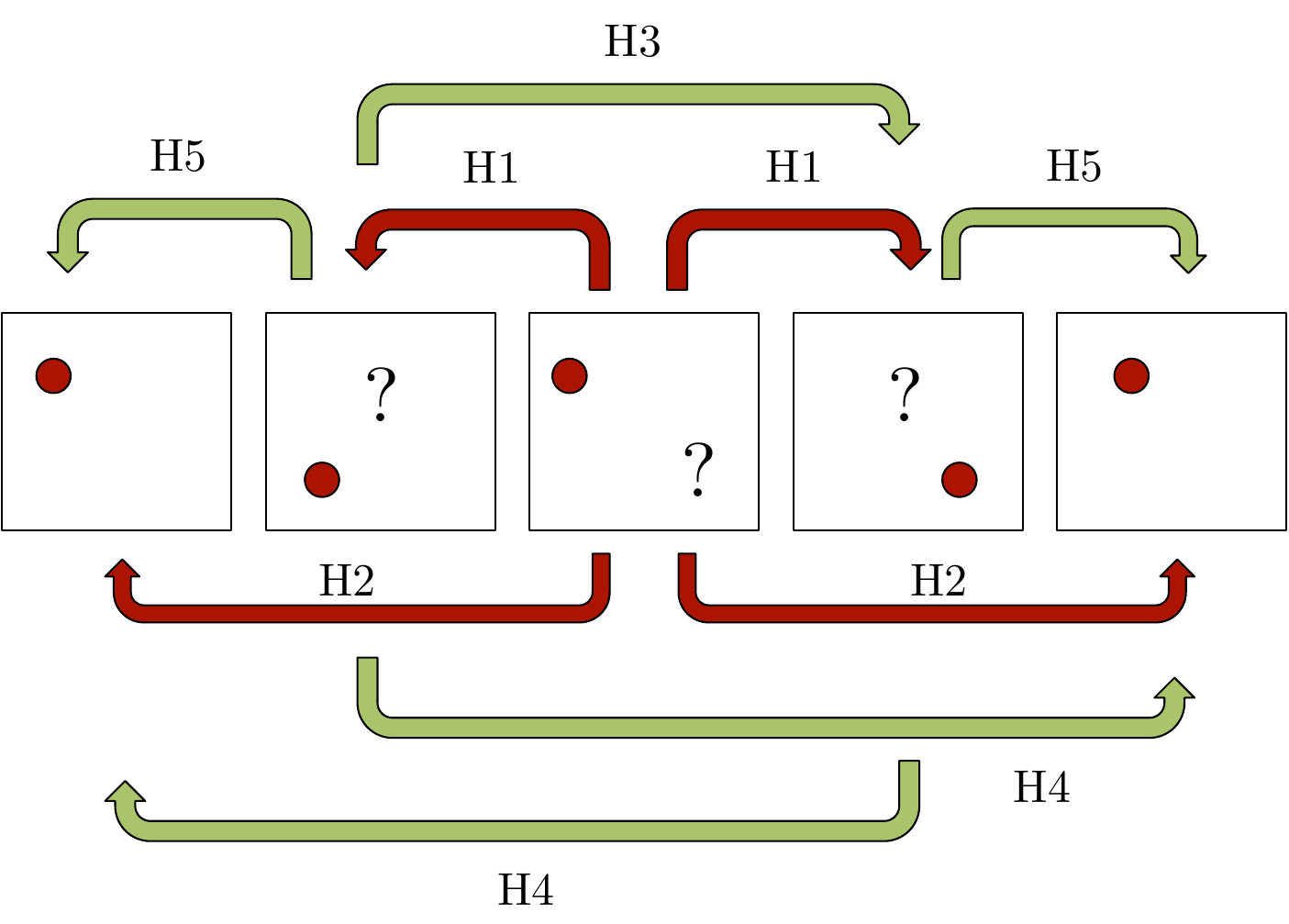} 
   \caption[Levels of the multi-level Hungarian]{Represented frames: [i-2,i-1,i,i+1,i+2]. Levels of the multi-level Hungarian.}
   \label{fig:multi1}
\end{figure}

The multi-level Hungarian takes advantage of the temporal information in 5 consecutive frames and is able to recover from occlusions and gaps in up to two consecutive frames.
%
%
%
The standard Hungarian gives us the matching between the particles in frame $t_1$ and frame $t_2$ and we use this to find matchings of the same particle in 5 consecutive frames, $[i-2,\ldots,i+2]$.
In order to find these matchings, the Hungarian is applied on different levels. 
The first two levels, represented in Figure \ref{fig:multi1} by red arrows, are created to find the matching of the particles in the frame of study, frame $i$. But it can also be the case that a particle is not present in frame $i$ but is present in the other frames. To solve all possible combinations given this fact, we use levels 3, 4 and 5, represented in Figure \ref{fig:multi1} by green arrows.

Below we show a detailed description and purpose of each level of the multi-level Hungarian:

\begin{itemize}
\item Level 1: Matches particles in frame $i$ with frames $i\pm1$. 
\item Level 2: Matches particles in frame $i$ with frames $i\pm2$. With the first two levels, we know, for all the particles in frame $i$, their position in the neighboring frames (if they appear). 
\item Level 3: Matches particles in frame $i-1$ with frame $i+1$.  
\item Level 4: Matches particles in frame $i\pm1$ with frame $i\mp2$. Level 3 and 4 solve the detection of matchings when a particle appears in frames $i\pm1$ and might appear in $i\pm2$, but is not present in frame $i$.
\item Level 5: Matches particles in frame $i\pm1$ with frame $i\pm2$.
\end{itemize}


%

\subsubsection{Conditions to add/delete particles}
\label{conditions}

\begin{figure}[htp] 
   \centering
   \includegraphics[width=0.7\linewidth]{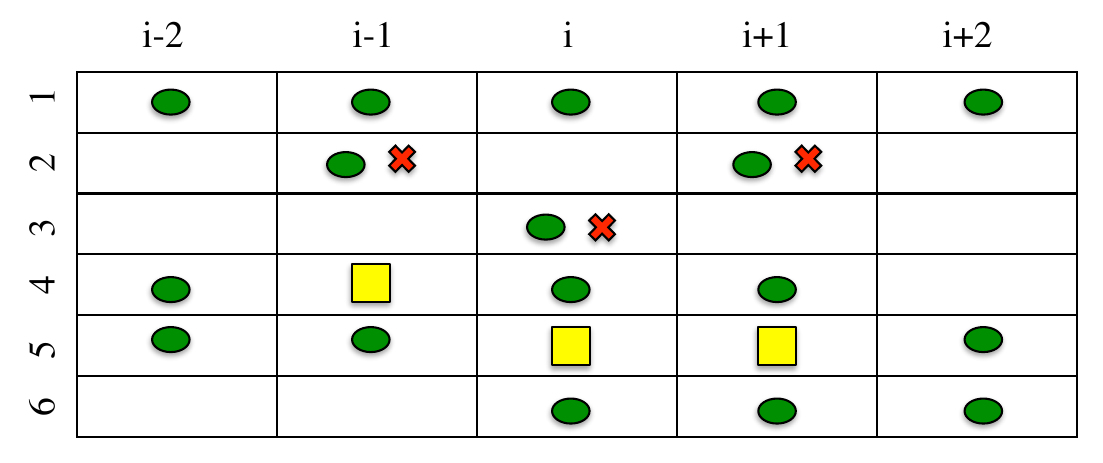} 
   \caption[Illustration of the process of adding and deleting particles]{Table with: the initial particles detected by the multi-level Hungarian (green ellipses), the ones added in the adding iteration (yellow squares) and the ones deleted in the deleting iteration (red crosses). In the blank spaces no position has been added or deleted.}
   \label{fig:table2}
\end{figure}

Once all the levels are applied hierarchically, a table with the matching information is created. The table has a column for each of the 5 frames from $[i-2 \ldots i+2]$ and a row for each detected trajectory, as shown in Figure \ref{fig:table2}. This table will be used to interpolate missing detections and delete false alarms.
 

To change the table information, we use two iterations: the adding iteration and the deleting iteration, which appear in Figure \ref{fig:multidiagram} as blue boxes.
During the {\it adding iteration}, we look for empty cells in the table where there is likely to be a particle. A new particle position is added if, and only if, two conditions are met:

\begin{enumerate}
\item The trajectory (row) consists of at least 3 particles. Trajectories have continuity while noise points do not.
\item It is not the first or last particle of the row. We use this strict condition to avoid the creation of false particle positions or the incorrect elongation of trajectories. 
 \end{enumerate}

Let us look at particle 6 of the table in Figure \ref{fig:table2}. In this case, we do not want to add any particle in frames $i-2$ and $i-1$, since the trajectory could be starting at frame $i$.
In the case of particle 4, we do not want to add a particle in frame $i+2$ because the trajectory could be ending at $i+1$.
This process is repeated iteratively until no particles are added to the table.

After convergence, the {\it deleting iteration} starts and we erase the outliers considered as noise. 
A new particle position is deleted if, and only if, two conditions are met:

\begin{enumerate}
\item The particle is present in the frame of study $i$. 
\item There are less than 3 particles in the same row.
 \end{enumerate}

We only erase particles from the frames [$i-1$,$i$,$i+1$] because it can be the case that a particle appears blurry in the first frames but is later correctly detected and has more continuity. 
Therefore, only particles whose complete neighborhood is known are removed. This process is repeated iteratively until no particles are deleted from the table.

The resulting particles are shown in Figure \ref{fig:table2}. 

\subsubsection{Missing data interpolation}

During the adding iteration, we use the information of the filtered projection in order to find the correct position of the new particle (Figure \ref{fig:multidiagram}). 
For example, if we want to add a particle in frame $i-1$, we go to the filtered projections XY, XZ, YZ in $t=i-1$, take the positions of the corresponding particle in $t=i$ or $t=i-2$ and search for the maximum likelihood within a window $w$.
If the position found on that frame is already present in the particles' table, we go back to the projection and determine the position of the second maximum value. This allows us to distinguish two particles which are close to each other.


There are many studies on how to improve the particle depth-position resolution ($z$-position). As in \citep{masuda}, we use the traditional method of considering the maximum value of the particle as its center. Other more complex methods \citep{fugal} have been developed which also deal with different particle sizes, but the flexibility of using morphological filtering already allows us to easily adjust our algorithm.

\subsection{The final Hungarian}
\label{finalhungsection}

Once the final particle positions are obtained (in Figure \ref{fig:multidiagram}, orange box labeled "Final particles"), we perform one last step to determine the trajectories. We use the standard Hungarian to match particles in all pairs of consecutive frames.




\section{Motion pattern classification}
\label{class}

In this section we describe the different types of motion patterns as well as the design of the combined HMM and the features used for their classification.


\subsection{Hidden Markov Models}
\label{HMM}

Hidden Markov Models \citep{HMM_rabiner} are statistical models of sequencial data widely used in many applications in artificial intelligence, speech and pattern recognition and modeling of biological processes.

In an HMM, the system is modeled as a Markov process $N$ unobserved states $S_1, S_2, ..., S_N$, with the condition that the system can only be in one of the states at any given time. 
The only observable variables are the sequence of symbols $O={o_1,o_2,...,o_M}$ produced by a set of stochastic processes.
Every HMM can be defined by the triple $\lambda=(\Pi,A,B)$. $\Pi=\{\pi_i\}$ is the vector of initial state probabilities.
Each transition from $S_i$ to $S_j$ can occur with a probability of $a_{ij}$, where $\sum_{j} a_{ij}=1$. $A=\{a_{ij}\}$ is the state transition matrix. 
In addition, each state $S_i$ generates an output $o_k$ with a probability distribution $b_{ik} = P(o_k|S_i)$. $B=\{b_{ik}\}$ is the emission matrix.

There are three main problems related to HMMs:
\begin{enumerate}
\item{{\it The evaluation problem}: for a sequence of observations $O$, compute the probability $P(O|\lambda)$ that an HMM $\lambda$ generated $O$}. This is solved using the Forward-Backward algorithm.
\item{{\it The estimation problem}: given $O$ and an HMM $\lambda$, recover the most likely state sequence $S_1, S_2, ..., S_N$ that generated $O$}. Problem 2 is solved by the Viterbi algorithm, a dynamic programming algorithm that computes the most likely sequence of hidden states in $O(N^2T)$ time. 
\item{{\it The optimization problem}: find the parameters of the HMM $\lambda$ which maximize $P(O|\lambda)$ for some output sequence $O$}. A local maximum likelihood can be derived efficiently using the Baum-Welch algorithm. 
\end{enumerate}

For a more detailed introduction to HMM theory, we refer to \citep{HMM_rabiner}.

\subsection{Types of patterns}
\label{patterns}

In our experimental setup, we are interested in four patterns shown by the green algae {\it Ulva linza} as depicted in Figure \ref{fig:speed2}: Orientation(1), Wobbling(2), Gyration(3) and intensive surface probing or Spinning(4). These characteristic swimming patterns are highly similar to the patterns observed before in \citep{Iken} for the brown algae {\it Hincksia irregularis}.


\noindent{\bf Orientation. }
Trajectory 1 in Figure \ref{fig:speed2} is an example of the Orientation pattern. 
This pattern typically occurs in solution and far away from surfaces. The most important characteristics of the pattern are the high swimming speed (a mean of approximately \unit[150]{$\mu$ m/s}) and a straight swimming motion with moderate turning angles.

\noindent{\bf Wobbling. }
Trajectory 2 shows the Wobbling pattern and its main characteristic is a much slower mean velocity of around \unit[50]{$\mu$ m/s}. 
The spores assigned to the pattern often change their direction of movement and only swim in straight lines for very short distances, which leads to zig-zag trajectories.

\noindent{\bf Gyration. }
Trajectory 3 is an example of the Gyration pattern. This pattern is extremely important for the exploration of surfaces, as occasional surface contacts are observable. The behavior in solution is similar to the Orientation pattern. Since in this pattern spores often switch between swimming towards and away from the surfaces, it can be interpreted as a pre-stage to surface probing. 

\noindent{\bf Intensive surface probing and Spinning.}
Trajectory 4 is an example of the Spinning pattern, which involves swimming in circles close to the surface within a very limited region. After a certain exploration time, the spores can either permanently attach to the surface or start swimming in circular patterns again looking for a better position. This motion is characterized by decreased mean velocities of about \unit[30]{$\mu$ m/s} in combination with a higher tendency to change direction (see Figure \ref{fig:speed2}, case 4). 


\subsection{Features used for classification}
\label{features}

An analysis of the features used for classification is presented in this section. Most of the features are generally used in motion analysis problems. An intrinsic characteristic of digital in-line holographic microscopy is the lower resolution of the $Z$ position compared to the $X$,$Y$ resolutions \citep{fugal}. 
Since many of the following features depend on the depth value, we compute the average measurements within 5 frames in order to reduce the noise of such features.
The four characteristic features used are:

\begin{itemize}
\item  {\it $v$, velocity}: the speed of the particles is an important descriptive feature, as we can see in Figure \ref{fig:speed2}. We use only the magnitude of the speed vector, since the direction is described by the next two parameters. Range is $[0,maxSpeed]$. $maxSpeed$ is the maximum speed of the particles as found experimentally in \citep{heidelberg1}.
\item {\it $\alpha$, angle between velocities}: it measures the change in direction, distinguishing stable patterns from random ones. Range is $[0,180]$.
\item {\it $\beta$, angle to normal of the surface}: it measures how the particles approach the surface or how they swim above it. Range is $[0,180]$.
\item {\it $D$, distance to surface}: this can be a key feature to differentiate surface-induced movements from general movements. Range is $(m_z,M_z]$, where $m_z$ and $M_z$ are the $z$ limits of the volume under study.
\end{itemize}




In order to work with Hidden Markov Models, we need to represent the features for each pattern with a fixed set of symbols.
The total number of symbols will depend on the number of symbols used to represent each feature $N_{symbols}=N_{v}N_{\alpha}N_{\beta}N_D$.


In order to convert every symbol for each feature into a unique symbol for the HMM, we use Equation \eqref{eq:symbol}, where $J$ is the final symbol we are looking for, $J_{1..4}$ are the symbols for each of the features, ranged $[1..N_{J_{1..4}}]$, where $N_{J_{1..4}}$ are the number of symbols per feature.

\begin{eqnarray}
\label{eq:symbol}
J &=&J_1 + (J_2-1)N_{J_1} + (J_3-1)N_{J_1}N_{J_2}+(J_4-1)N_{J_1}N_{J_2}N_{J_3}
\end{eqnarray}

In the next sections, we present how to use the resulting symbols to train the HMMs. The symbols are the observations of the HMM, and the training process gives us the probability of emitting each symbol in each of the states and the probability of going from one state to the others.

\subsection{Building and training the HMMs}

In speech recognition, an HMM is trained for each of the phonemes of a language. Later, words are constructed by concatenating several HMMs of the phonemes that form the word. HMMs for sentences can be created by concatenating HMMs of words, etc.
We take a similar hierarchical approach in this paper. We train one HMM for each of the patterns and then we combine them into a unique Markov chain with a simple yet effective design that will be able to describe any pattern or combination of patterns.
This approach can be used in any problem where multiple motion patterns are present.

\noindent{\bf Individual HMM per pattern. }
In order to represent each pattern, we build a Markov chain with $N$ states and we only allow the model to stay in the same state or move one state forward. 
Finally, from state $N$ we can also go back to state 1. The number of states N is found empirically using training data (we use $N=4$ for all experiments, see Section \ref{experiments_param}).
The HMM is trained using the Baum-Welch algorithm to obtain the transition and emission matrices.

\noindent{\bf Combined HMM. }
The idea behind a combined HMM that represents all patterns is that we can not only classify sequences where there is one pattern present, but sequences where the particle makes transitions between different patterns.
\begin{figure}[htbp]
\centering
\subfigure[]{
\includegraphics[width=0.6\linewidth]{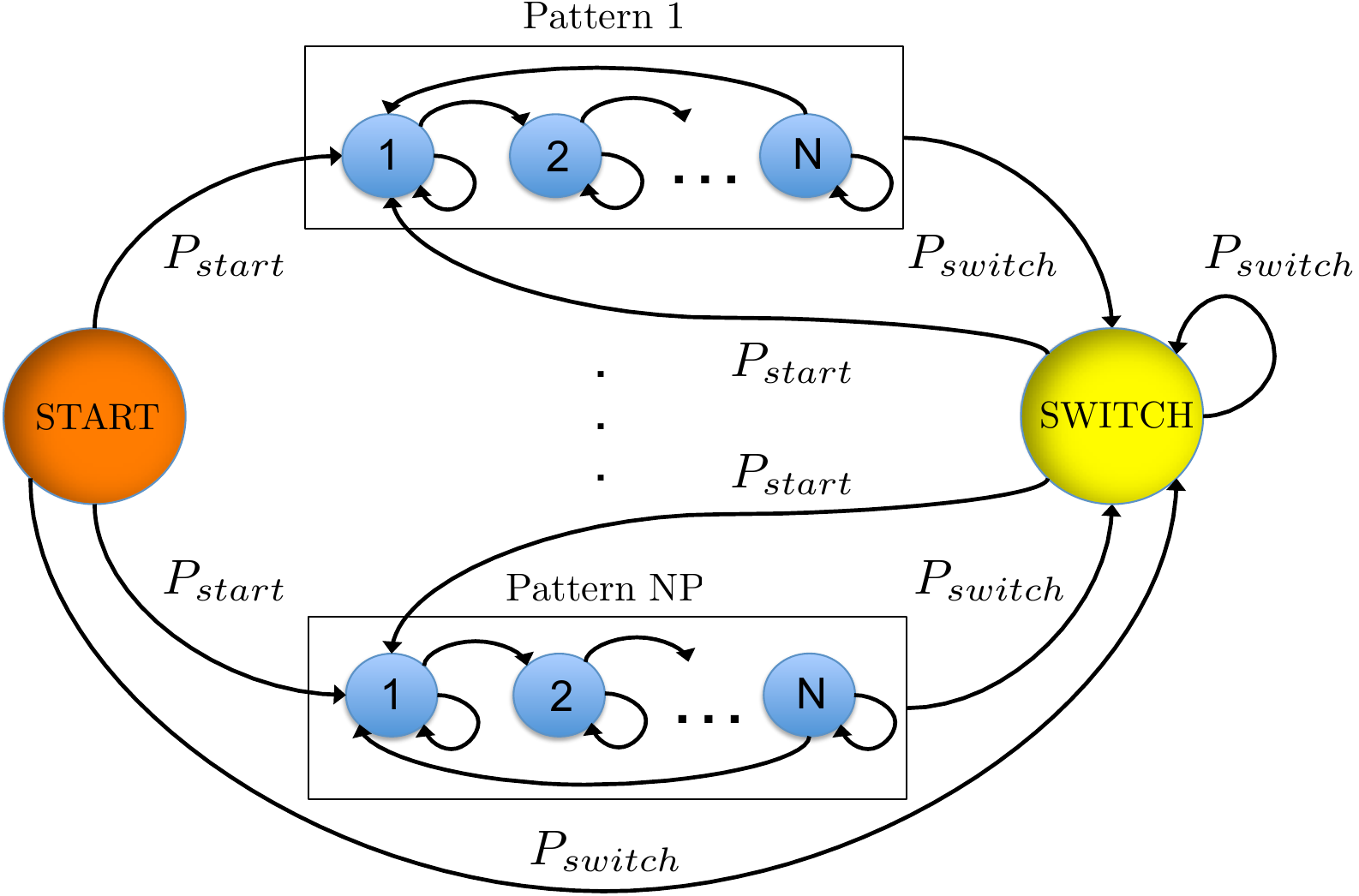}
\label{fig:completeHMM}
}
\subfigure[]{
\includegraphics[width=0.365\linewidth]{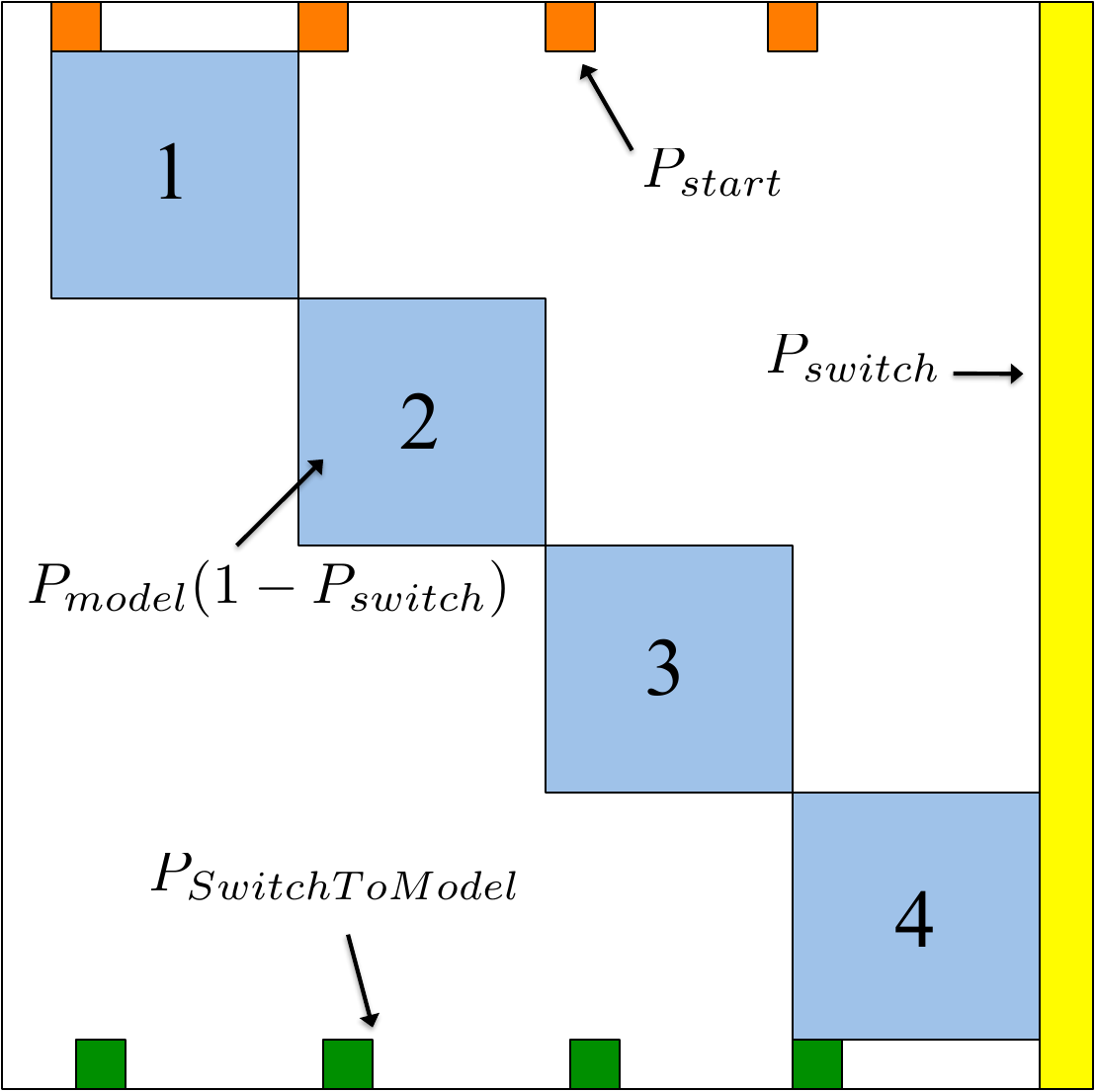}
\label{fig:HMM_matrix}
}
\caption[Graphical depiction of the complete Hidden Markov Model]{\subref{fig:completeHMM} Combined HMM created to include changes between patterns within one trajectory. \subref{fig:HMM_matrix} Transition matrix of the combined HMM}
\label{fig:complete}

\end{figure}
In Figure \ref{fig:completeHMM}, we can see a representation of the combined model, while the design of the transition matrix is depicted in Figure \ref{fig:HMM_matrix}. The four individual HMMs for each of the patterns are placed in parallel (blue). In order to deal with transitions, we create two special states: the START and the SWITCH state.

The START state is just created to allow the system to begin at any pattern (orange). 
We define $P_{start}=P_{SwitchToModel}=\frac{1-P_{switch}}{N_P}$, where $N_P$ is the number of patterns.
As START does not contain any information of a pattern, it does not emit any symbol.

The purpose of the new state SWITCH is to make transitions easier. Imagine a given trajectory which makes a transition from Pattern 1 to Pattern 2. While transitioning, the features create a symbol that neither belongs to Pattern 1 nor 2. The system can then go to state SWITCH to emit that symbol and continue to Pattern 2. Therefore, all SWITCH emission probabilities are $\frac{1}{N_{symbols}}$.
Since SWITCH is such a convenient state, we need to impose restrictive conditions so that the system does not go to or stay in SWITCH too often. This is controlled by the parameter $P_{switch}$, set to the minimum value of all the probabilities in the model minus a small $\epsilon$. This way, we ensure that $P_{switch}$ is the lowest transition probability in the system.

Finally, the sequence of states given by the Viterbi algorithm determines the motion pattern observed. Our implementation uses the standard Matlab HMM functions.

\section{Experimental results}
\label{experiments}

In order to test our algorithm, we use 6 sequences (labeled S1 to S6) in which the swimming motion of {\it Ulva linza} spores is observed \citep{heid4}.
All sequences have some particle positions which have been semi-automatically reconstructed, manually labeled and inspected (our ground truth) for later comparison with our fully-automatic results.

\subsection{Performance of the standard Hungarian}

First of all, we want to show the performance of the final standard Hungarian described in Section \ref{finalhungsection}. For this, we use the ground truth particle positions and apply the Hungarian algorithm to determine the full trajectories of the microorganisms. 
Comparing the automatic matches to the ground truth, we can see that in 67\% of all sequences the total number of particles is correctly detected, while in the remaining 33\%, there is just a 5\% deviation in the number of particles. 
The average accuracy of the matchings reaches 96.61\%.

To further test the robustness of the Hungarian algorithm, we add random noise to each position of our particles. The added noise is in the same order as the noise intrinsically present in the reconstructed images, determined experimentally in \citep{heidelberg1}. 
$N=100$ experiments are performed on each of the sequences and the accuracy is recorded.
Results show that the average accuracy of the matching is just reduced from 96.61\% to 93.51\%, making the Hungarian algorithm very robust to the noise present in holographic images and therefore well suited to find the trajectories of the particles.



\begin{figure*}[htb]
\centering
\subfigure[]{
\includegraphics[width=0.48\linewidth]{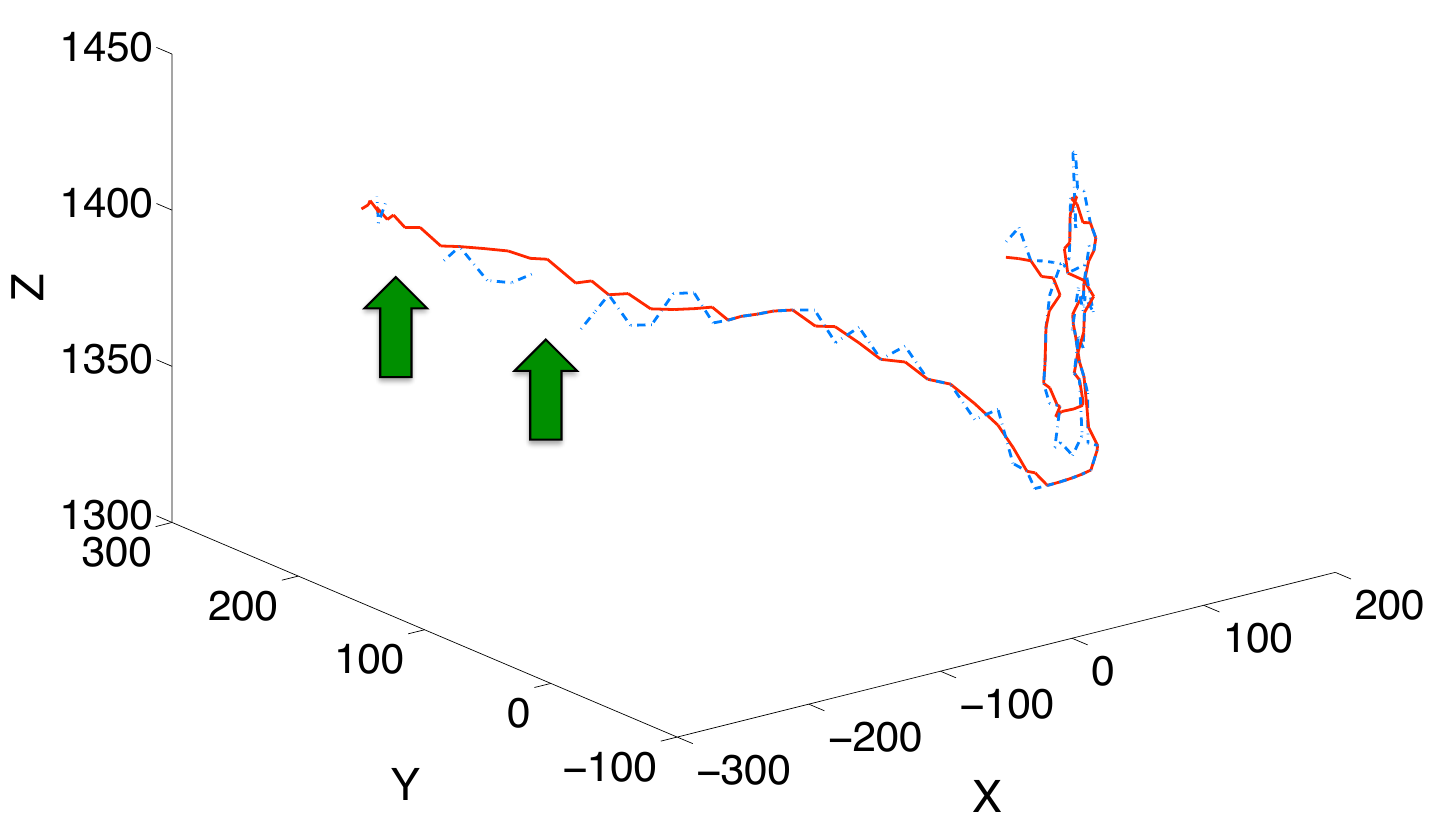} 
\label{fig:merging}
}
\subfigure[]{
\includegraphics[width=0.48\linewidth]{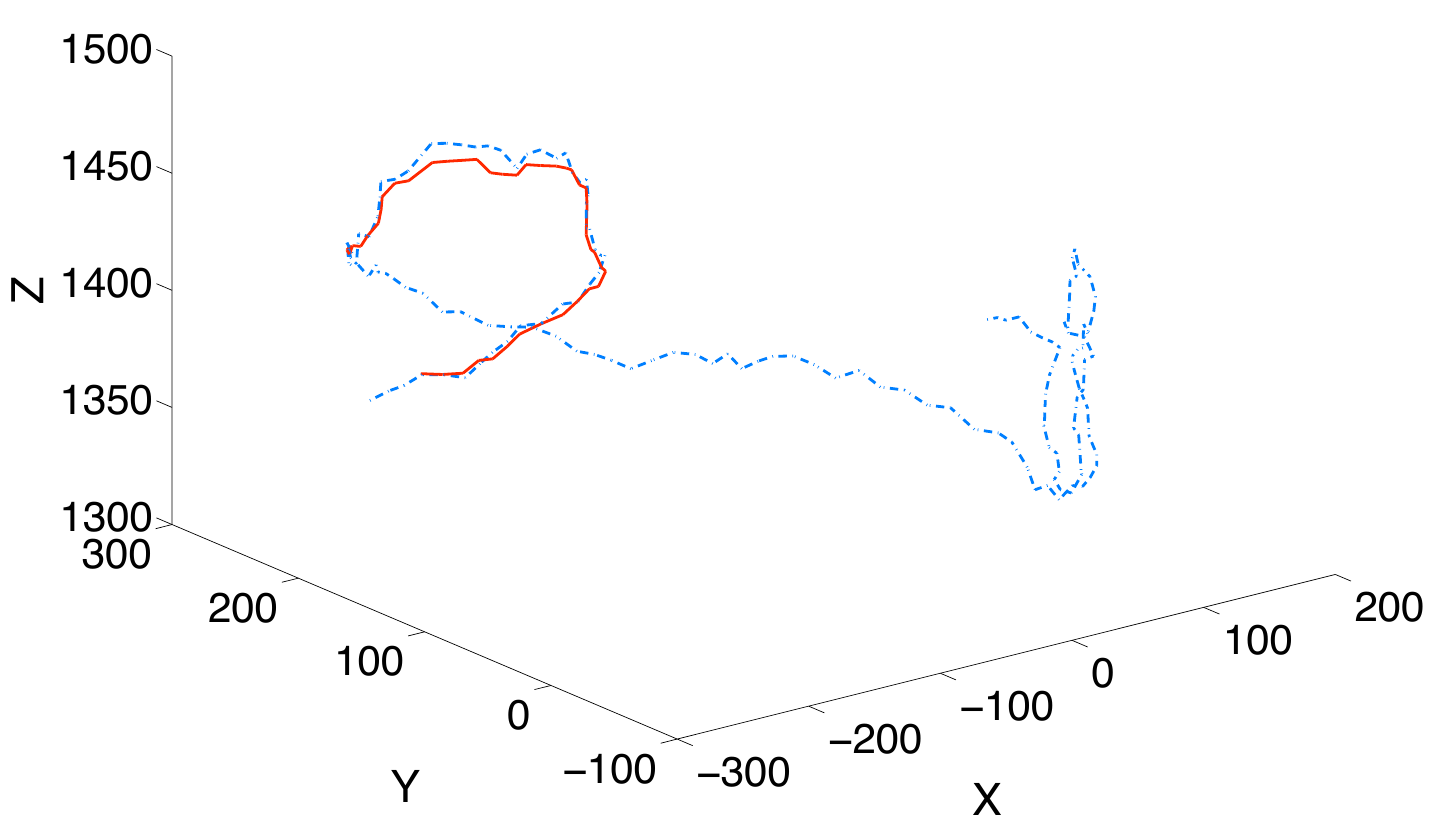} 
\label{fig:incomplete}
}
\subfigure[][]{
\includegraphics[width=0.48\linewidth]{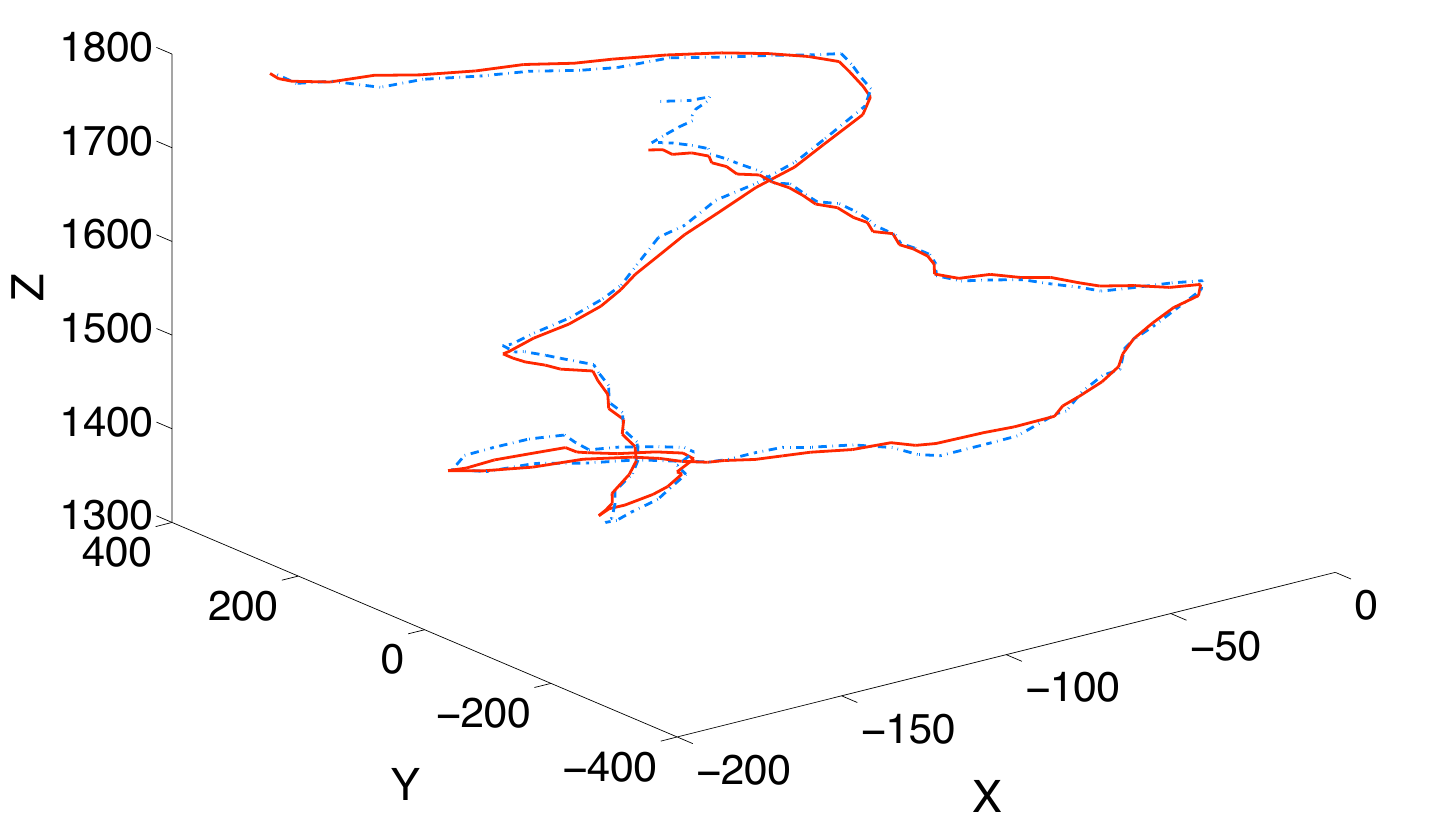}
\label{fig:completed}
}
\caption[Results obtained with multi-level Hungarian compared to the standard Hungarian]{\subref{fig:merging} 3 separate trajectories are detected with the standard Hungarian (blue dashed line). Merged trajectory detected with our method (with a smoothing term, red line). Missing data spots marked by arrows. \subref{fig:incomplete}, \subref{fig:completed} Ground truth trajectories (blue dashed line). Trajectories automatically detected with our method (red line).}
\label{fig:tracks}
\end{figure*}

\subsection{Performance of the multi-level Hungarian}

To test the performance of the multi-level Hungarian, we apply the method to three sets of particles:

\begin{itemize}
\item Set A: particles determined by the threshold (pre multi-level Hungarian)
\item Set B: particles corrected after multi-level Hungarian
\item Set C: ground truth particles, containing all the manually labeled particles 
\end{itemize}

We then start by comparing the number of particles detected, as shown in Table \ref{tab:numpart}. 

\begin{table}[htbp]
\begin {center}
   \begin{tabular}{ ccccccc }
     \hline
    & {\bf S1} & {\bf S2} & {\bf S3} & {\bf S4} & {\bf S5} & {\bf S6}\\ \hline
   {\bf Set A} & 1599 & 1110 & 579 & 668 & 1148 & 2336\\ \hline
    {\bf Set B} & 236 & 163 & 130 & 142 & 189 & 830\\ \hline
    {\bf Set C} & 40 & 143  & 44 & 54 & 49 & 48\\ \hline    
  \end{tabular}
  \end{center}
    \caption{Comparison of the number of particles detected by thresholding, by the multi-level Hungarian and the ground truth for the 6 examined sequences (S1-S6).}
\label{tab:numpart}
\end{table}

As shown in Table \ref{tab:numpart}, the number of particles detected in Set A is drastically reduced in Set B, after applying the multi-level Hungarian, demonstrating its abilities to compensate for missing data and merging trajectories. 
If we compare it to Set C, we see that the number is still too high, indicating possible tracks which were not merged and so detected as independent.

Nonetheless, as we do not know the exact amount of particles present in a volume (not all particle positions have been labeled), it is of great value for us to compare the average length of the trajectories, defined as the number of frames in which the same particle is present. The results are shown in Table \ref{tab:length}, where we can clearly see that the average length of a trajectory is greatly improved with the multi-level Hungarian, which is crucial since long trajectories give us more information on the behavior of the particles.
%
%
\begin{table}[htbp]
\begin {center}
   \begin{tabular}{ ccccccc }
     \hline
    & {\bf S1} & {\bf S2} & {\bf S3} & {\bf S4} & {\bf S5} & {\bf S6}\\ \hline
   {\bf Set A} & 3 & 5 & 5 & 4 & 6 & 7\\ \hline
    {\bf Set B} & 19 & 31 & 27 & 23 & 38 & 23\\ \hline
    {\bf Set C} & 58 & 54  & 54 & 70 & 126 & 105\\ \hline    
    \end{tabular}
  \end{center}
    \caption{Comparison of the trajectories' average length.}
\label{tab:length}
\end{table}

Now let us consider just useful trajectories for particle analysis, that is, trajectories with a length of more than 25 frames which are the trajectories that will be useful later for motion pattern classification. Tracking with the standard Hungarian returns 20.7\% of useful trajectories from a volume, while the multi-level Hungarian allows us to extract 30.1\%. In the end, this means that we can obtain more useful information from each analyzed volume.


Ultimately, this means that fewer volumes have to be analyzed in order to have enough information to draw conclusions about the behavior of a microorganism.

\subsection{Performance of the complete algorithm}

Finally, we are interested in determining the performance of the complete algorithm, including detection and tracking.
For this comparison, we are going to present two values:

\begin{itemize}
\item Missing: percentage of ground truth particles which are not present in the automatic determination.
\item Extra: percentage of automatic particles that do not appear in the ground truth data. 
\end{itemize}

In Table \ref{tab:finalresult} we show the detailed results for each sequence.
%

\begin{table}[htbp]
\begin {center}
  \begin{tabular}{ ccccccc }
     \hline
    & {\bf S1} & {\bf S2} & {\bf S3} & {\bf S4} & {\bf S5} & {\bf S6}\\ \hline
     {\bf Missing (\%)}  & 8.9 & 20.7 & 19.1 & 23.6 & 11.5 & 12.9  \\ \hline
    {\bf Extra (\%)}   & 54.9 &  34.1 & 46.5 & 13.3 & 25.8 & 74.6 \\ \hline
  \end{tabular}
  \end{center}
    \caption{Missing labeled and extra automatic particles.}
\label{tab:finalresult}
\end{table}

Our automatic algorithm detects between 76\% and 91\% of the particles present in the volume.
This gives us a measure of how reliable our method is, since it is able to detect most of the verified particle positions.
Combining this information with the percentage of particles detected by our algorithm but not labeled, we can see that our method extracts much more information from the volume of study. This is clear in the case of S6, where we have a volume with many crossing particles which are difficult to label manually, and where our algorithm gives us almost 75\% more information.

We now consider the actual trajectories and particle position and measure the position error of our method.
The error is measured as the Euclidean distance between each point of the ground truth and the automatic trajectories, both at time $t$.
In Figure \ref{fig:merging}, we can see the 3 independent trajectories found with the standard Hungarian and the final merged trajectory, which proves the power of our algorithm to fill in the gaps (pointed by arrows).
In Figure \ref{fig:incomplete}, we can see that the automatic trajectory is much shorter (there is a length difference of 105 frames), although the common part is very similar with an error of just \unit[4,2]{$\mu$m}. 
Figure \ref{fig:completed}, on the other hand, shows a perfectly matched trajectory with a length difference of 8 frames and error of \unit[6,4]{$\mu$m} for the whole trajectory which is around twice the diameter of the spore body. 
This proves that the determination of the particle position is accurate but the merging of trajectories can be improved.



\subsection{Comparison with a Linear Programming tracker}

In order to compare the Multi-level Hungarian with the Linear Programming formulation introduced in Chapter \ref{LPtracking}, we perform several experiments with simulated and real data. 
For the first experiment, we simulate 15 randomly moving particles and an increasing number of missing data, from 2\% to 10\%. Four methods are compared:

\begin{itemize}
\item{Standard Hungarian (SH): matching frame by frame, shown in black.}
\item{Multi-level Hungarian (MLH): matching taking into account several frames, as presented in Section \ref{multihungsection}, shown in blue.}
\item{Linear programming, 1 level (LP1Lev): matching using Linear Programming but only allowing matching of particles which are at a maximum distance of one frame, shown in cyan.}
\item{Linear Programming, 2 levels (LP): first, matching using Linear Programming, 1 level and second, creating another graph with the found trajectories, allowing particles to be matched when they are up to 5 frames apart, shown in pink.}
\end{itemize}

\begin{figure}[htbp]
\centering
\subfigure[]{
\includegraphics[width=0.475\linewidth]{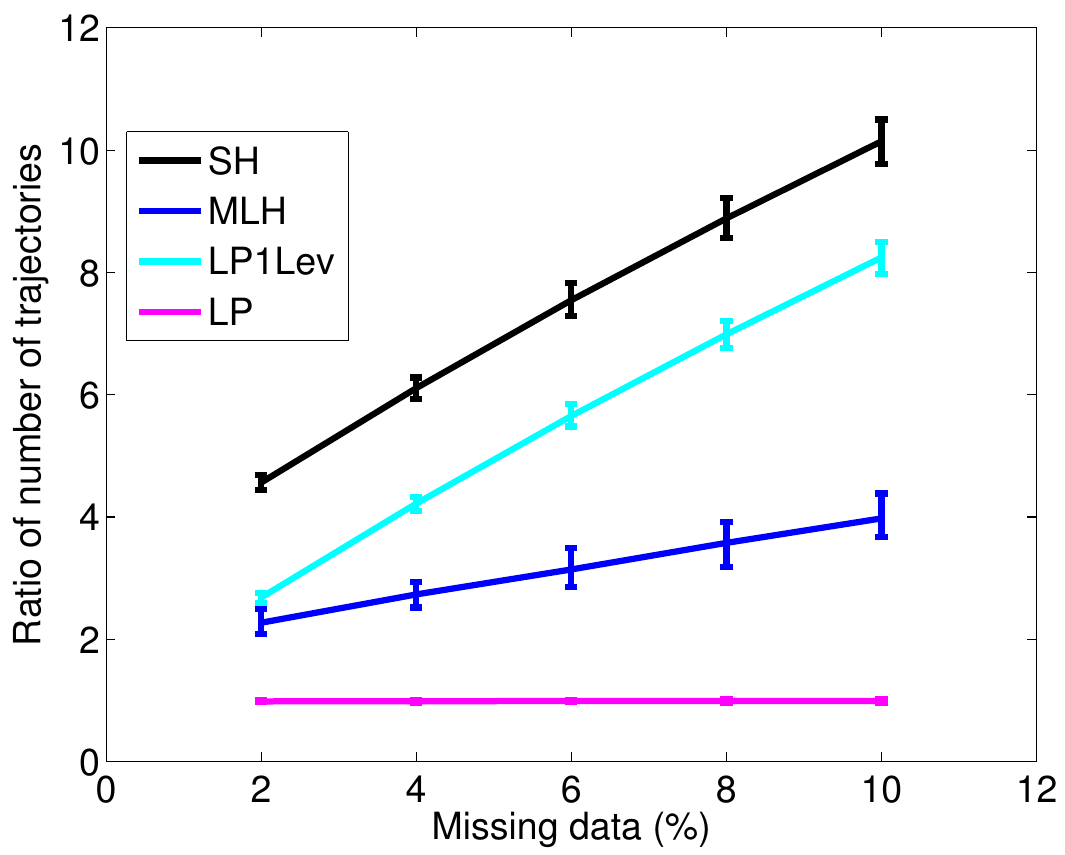}
\label{fig:missexps1}
}
\subfigure[]{
\includegraphics[width=0.475\linewidth]{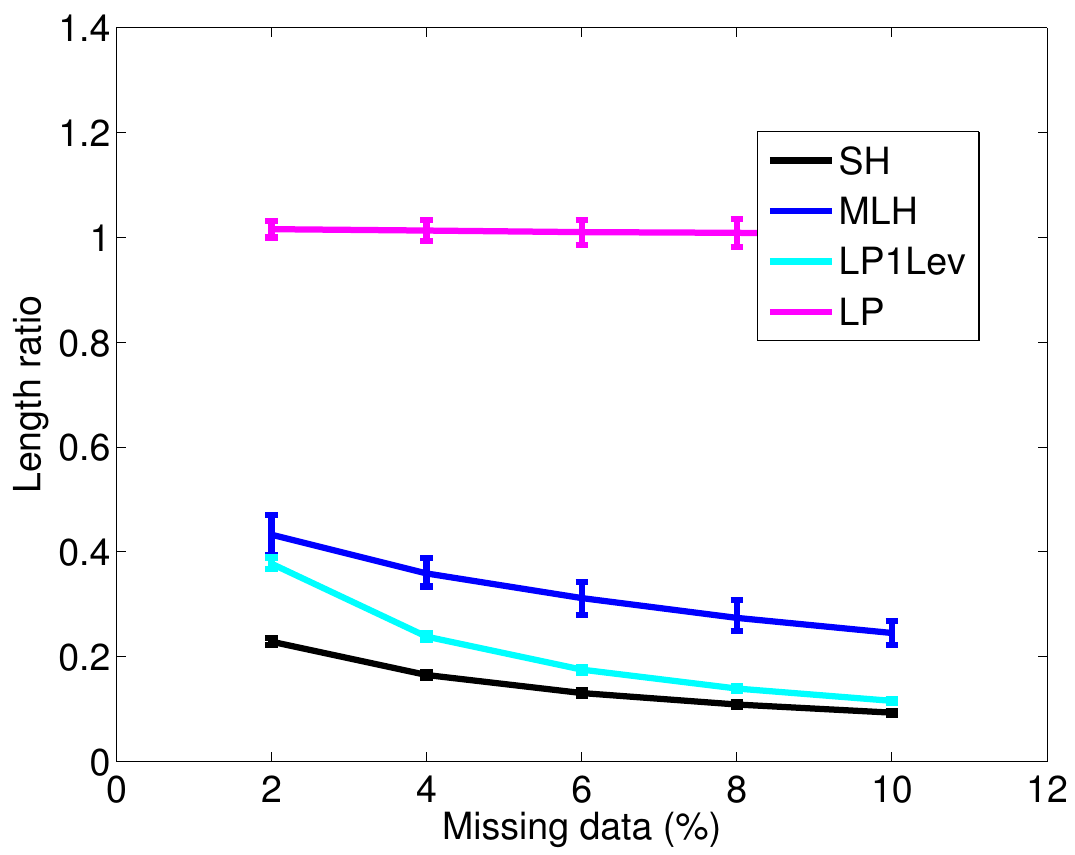}
\label{fig:missexps2}
}
\subfigure[]{
\includegraphics[width=0.475\linewidth]{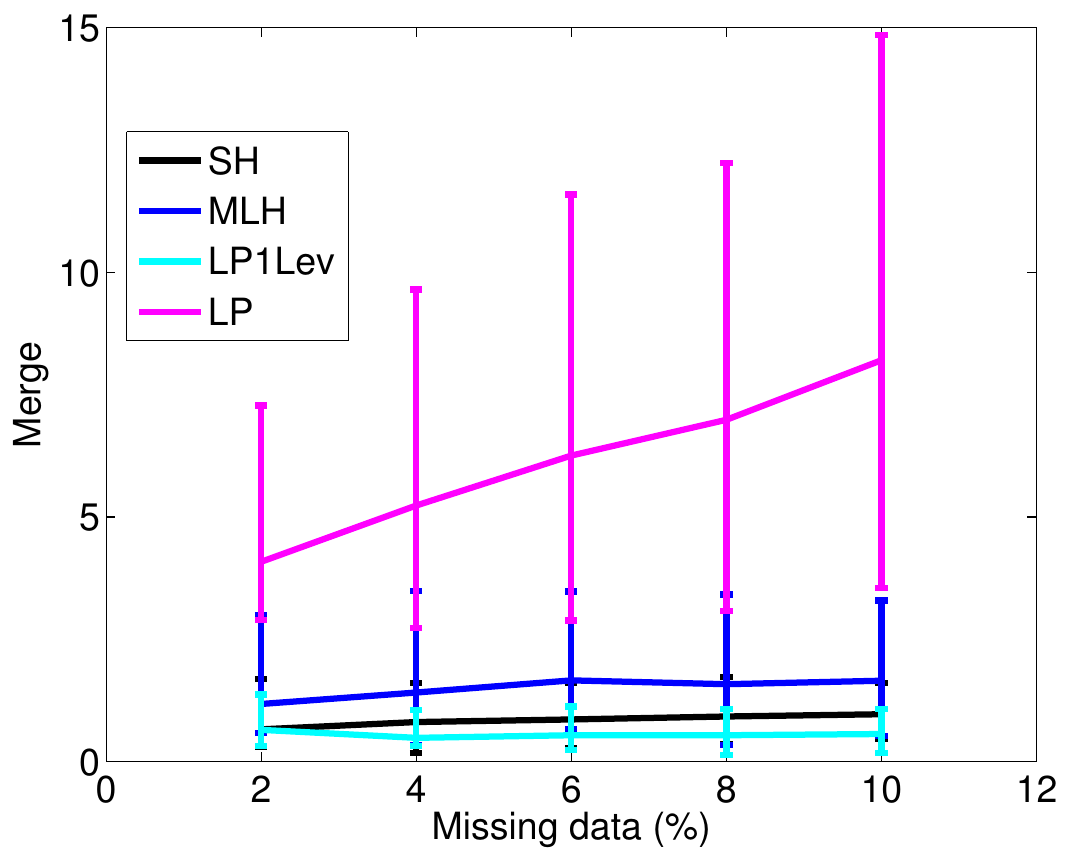}
\label{fig:missexps3}
}
\subfigure[]{
\includegraphics[width=0.475\linewidth]{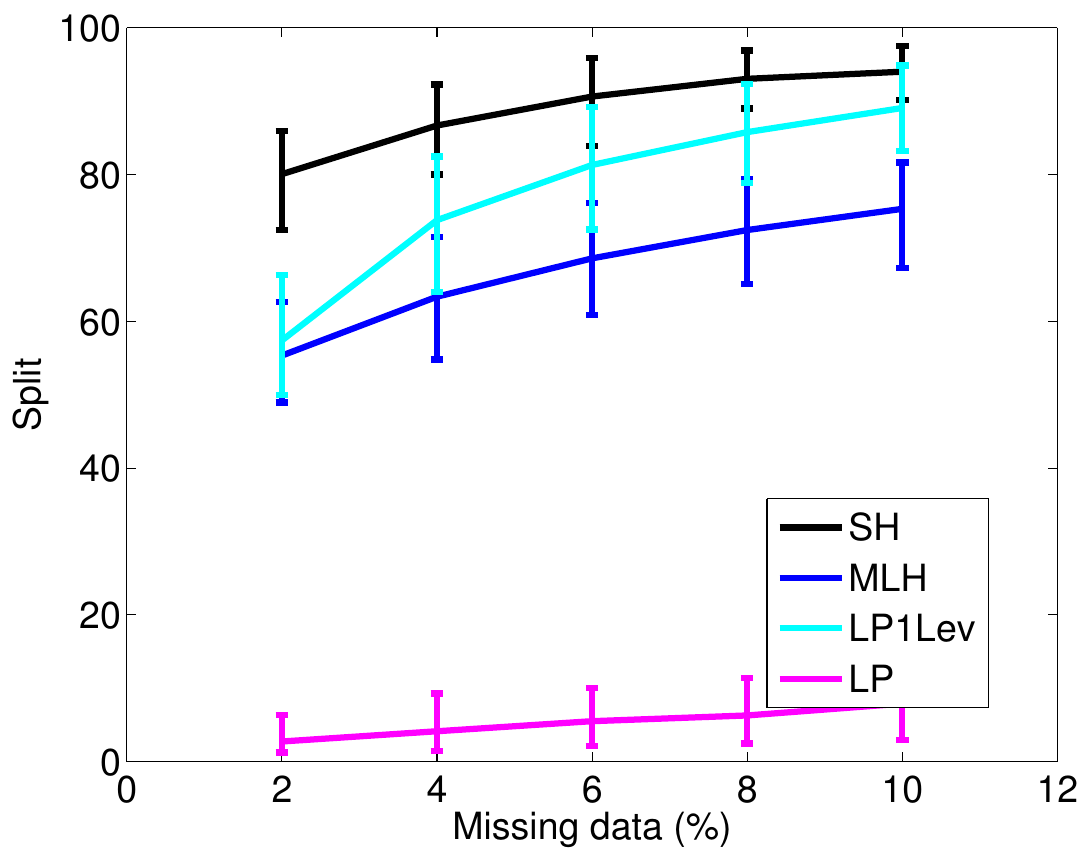}
\label{fig:missexps4}
}
\caption[Comparison of Multi-level Hungarian and Linear Programming methods]{Comparative experiments with a simulation of 15 randomly moving particles and an increasing number of missing data from 2\% to 10\% (N=100 repetitions of the experiment are performed and average results shown). Compared methods: standard Hungarian (black), Multi-level Hungarian (blue), Linear Programming with maximum matching distance of 1 frame (cyan) and Linear Programming 2 levels (pink).}
\label{fig:simexps}
\end{figure}

In Figure \ref{fig:missexps1}, we can see the ratio between the number of trajectories found by each algorithm and the ground truth number of trajectories. The closer this ratio is to 1, the better the algorithm performs. A similar measure is the one plotted in Figure \ref{fig:missexps2}, where we plot the length ratio. Again, if this ratio is 1, it means that the average length of the trajectories found automatically is the same as the length of ground truth trajectories.

Note that, as the percentage of missing data increases, the SH and LP1Lev have an increasing ratio of the number of trajectories and a decreasing ratio of length. This means that these algorithms are not capable of filling the gaps found in the data, and therefore the trajectories found are shorter as the missing data percentage increases.
The MLH performs better on both ratios, but cannot achieve the superior performance of LP 2 levels, which scores an almost perfect 1 on both ratios. This means that the LP is virtually unaffected by up to 10\% of missing data.

In Figure \ref{fig:missexps3}, we show the percentage of automatically found trajectories which contain two or more ground truth trajectories, \ie two or more trajectories have been merged. In Figure \ref{fig:missexps4}, on the other hand, we show the number of ground truth trajectories that are split into two or more. As can be seen, the MLH is superior to SH and LP1Lev in terms of splitting fewer trajectories, but the LP 2 levels improves the number of split trajectories by 60\% to 70\%. Though it also merges more trajectories than the other methods, the percentage of merged trajectories ranges from 4\% to 7\% on average, which means overall the LP 2 levels is far superior than the other methods.

\begin{figure}[htbp]
\centering
\includegraphics[width=0.475\linewidth]{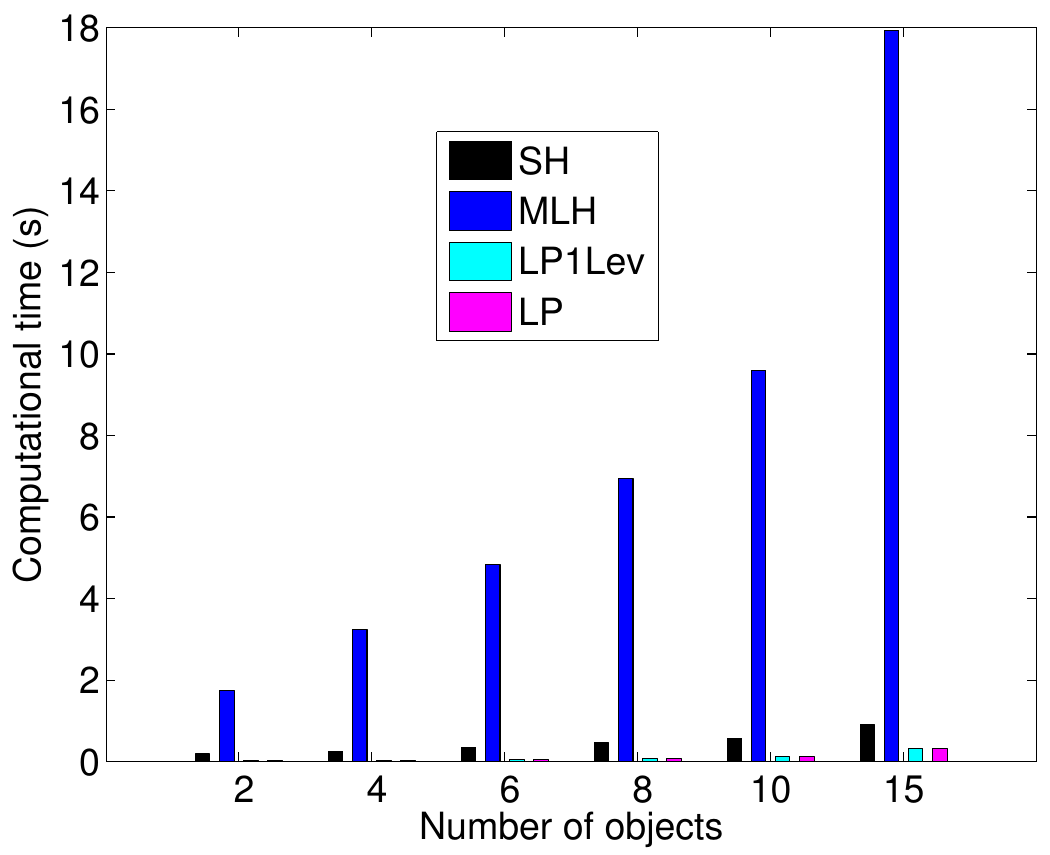}
\caption[Computational time \vs number of objects to be tracked]{Computational time \vs number of objects to be tracked. Compared methods: standard Hungarian (black), Multi-level Hungarian (blue), Linear Programming with maximum matching distance of 1 frame (cyan) and Linear Programming 2 levels (pink).}
\label{fig:runtimexobj}
\end{figure}

While the MLH performs much better than SH and LP 1 level, it is also computationally more expensive than the other methods, as can be seen in Figure \ref{fig:runtimexobj}. Using the same experimental setup as before but with increasing number of objects, we observe that the computational cost increases exponentially with the number of objects to be tracked. Since our datasets contain up to 25 objects per sequence, the algorithm takes only a few minutes to track each sequence.

\begin{figure}[htbp]
\centering
\subfigure[]{
\includegraphics[width=0.475\linewidth]{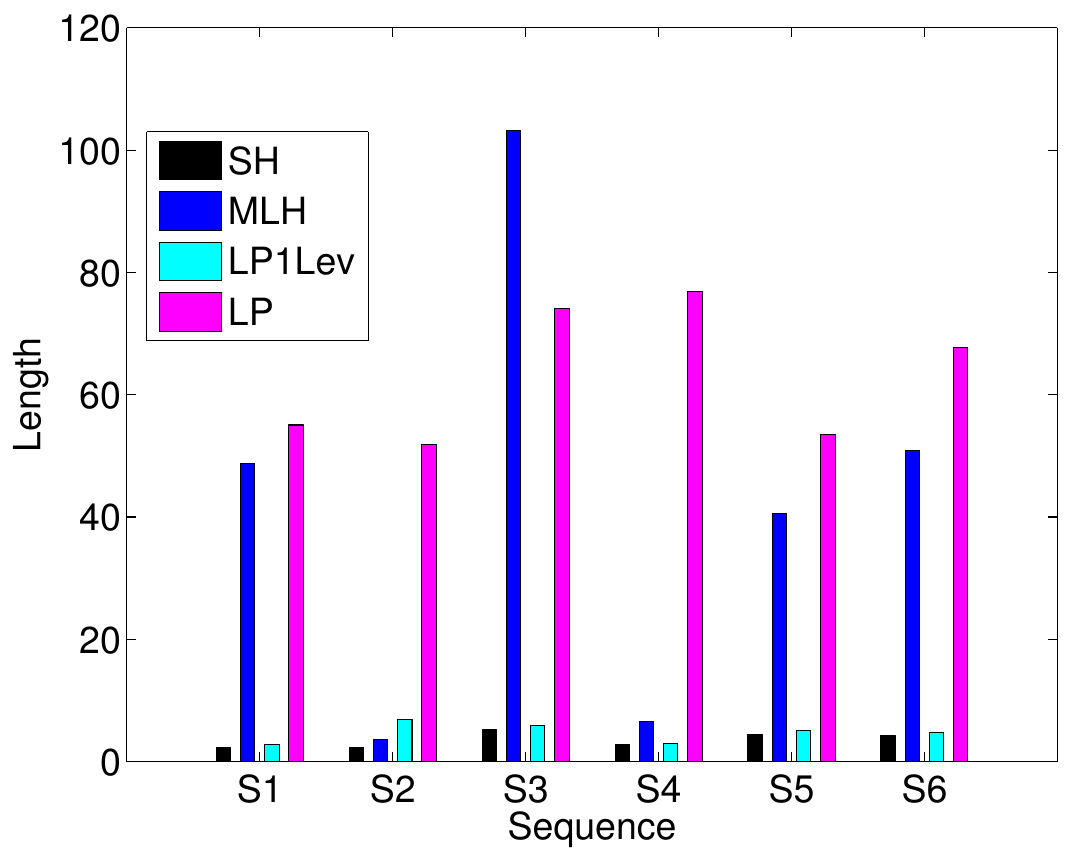}
\label{fig:length}
}
\subfigure[]{
\includegraphics[width=0.475\linewidth]{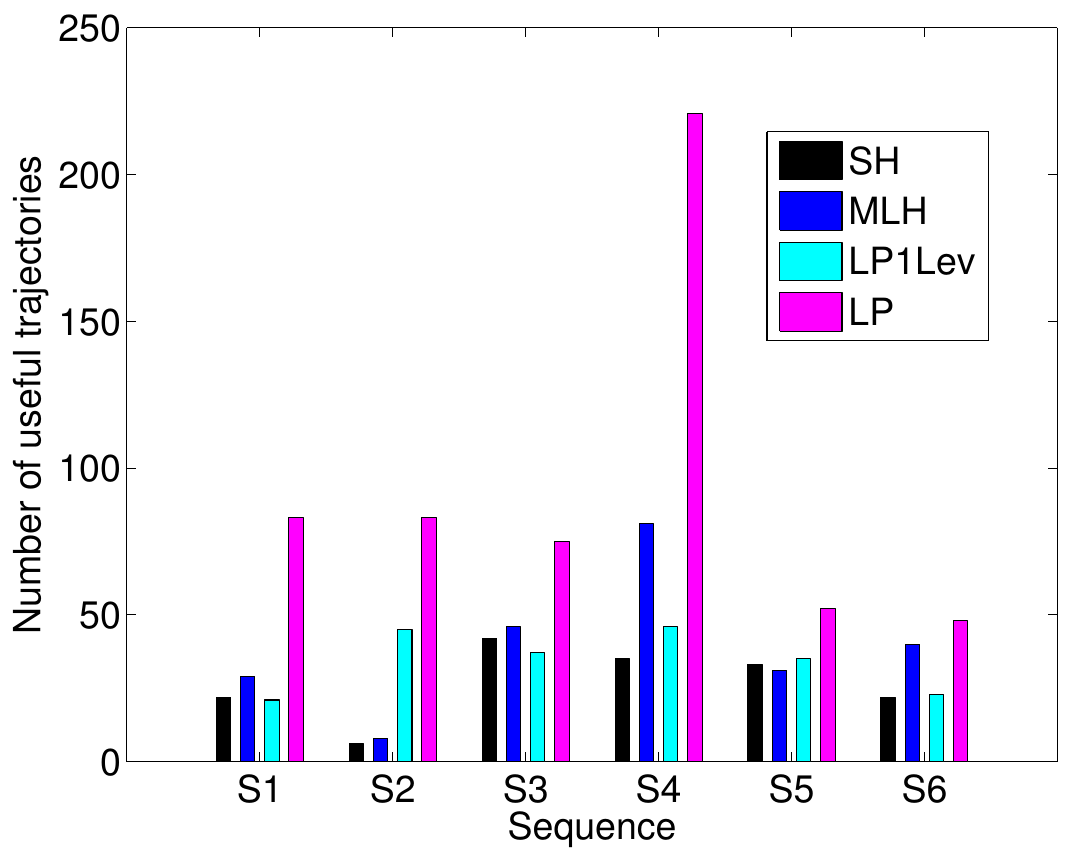}
\label{fig:usefultraj}
}
\caption[Experiments on real data to compare the Multi-level Hungarian \vs Linear Programming]{Experiments on real data comparing 4 methods: standard Hungarian (black), Multi-level Hungarian (blue), Linear Programming with maximum matching distance of 1 frame (cyan) and Linear Programming 2 levels (pink).}
\label{fig:realdata}
\end{figure}

Finally, we apply the four algorithms to the 6 sequences of real data and plot the average length of the trajectories in Figure \ref{fig:length}. As we can see, the MLH and LP 2 levels obtain much longer trajectories. Obtaining long trajectories is specially important for conducting accurate motion analysis, as will be presented in the experiments in the next sections. To this end, we plot in Figure \ref{fig:usefultraj} the number of useful trajectories obtained by each method. A useful trajectory for motion analysis is defined as having a length of 25 frames or more. The LP 2 levels (pink) obtains a much larger number of these trajectories for each sequence, which means this method is able to extract much more useful information from each sequence than SH, MLH or LP1Lev.

The next sections are dedicated to several experimental results on the automatic classification of biological motion patterns. All trajectories used from now on are obtained automatically with the method described in Section \ref{tracking} and are classified manually by experts, which we refer to as our ground truth classification data.

\subsection{Evaluation of the features used for classification}
\label{experiments_param}

The experiments in this section have the purpose of determining the impact of each feature on the correct classification of each pattern. We perform leave-one-out tests on our training data which consists of 525 trajectories: 78 for wobbling, 181 for gyration, 202 for orientation and 64 for intensive surface probing. To perform these tests, all training sequences except one are used to train the HMM. The remaining sequence is then tested with the combined HMM and, using the Viterbi algorithm, the sequence of likely states is obtained. With this information, we can classify the sequence into one of the four patterns.

\begin{figure}[htp] 
   \centering
   \includegraphics[width=0.95\linewidth]{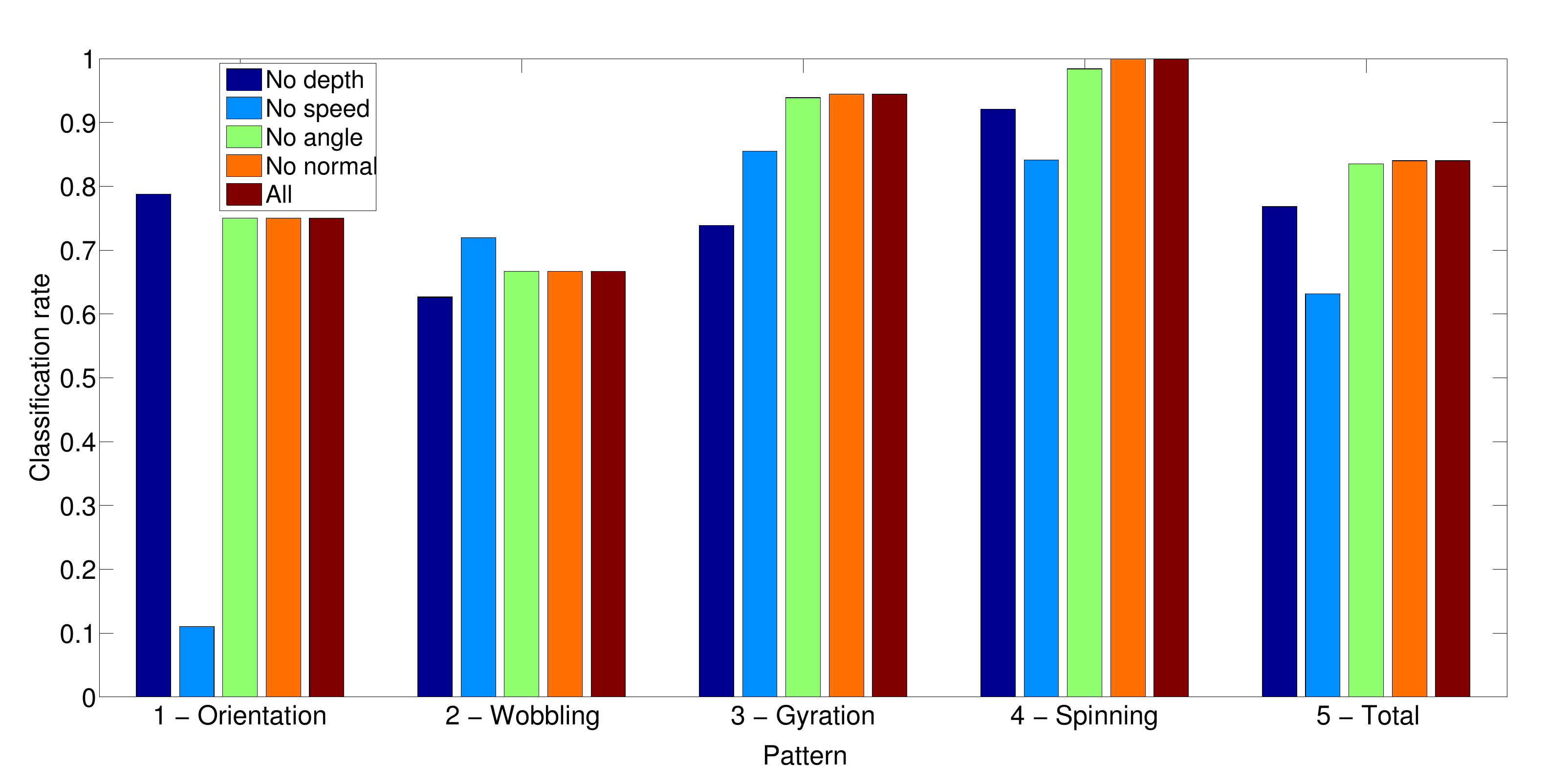} 
   \caption[Classifications rates]{Classification rate for parameters $N=4$, ${N_{v}=4}$, ${N_{\alpha}=3}$, ${N_{\beta}=3}$ and ${N_D=3}$. On each experiment, one of the features is not used. In the last experiment all features are used.}
   \label{fig:classrate}
\end{figure}

The first experiment that we conduct (see Figure \ref{fig:classrate}) is to determine the effect of each parameter on the recognition of the patterns. 
The number of symbols and states can only be determined empirically, since they depend heavily on the amount of training data. In our experiments, we found the best set of parameters to be $N=4$, ${N_{v}=4}$, ${N_{\alpha}=3}$, ${N_{\beta}=3}$ and ${N_D=3}$ for which we obtain a classification rate of 83.86\%.

For each test, we set one parameter to 1, which means that the corresponding feature has no effect in the classification process. For example, the first bar in blue labeled "No Depth" is done with ${N_D=1}$. The classification rate for each pattern (labeled from 1 to 4), as well as the mean for all the patterns (labeled Total), is recorded.

As we can see, the angles $\alpha$ and $\beta$ (see section \ref{features}) are the less relevant features, since the classification rate with and without these features is almost the same. 
The angle $\alpha$ depends on the $z$ component, hence the lower resolution in $z$ can result in noisy measurements. 
In this case, the trade-off is between having noisy angle data which can be unreliable, or an average measure which is less discriminative for classification.
The most distinguishing feature, according to Figure \ref{fig:classrate}, is the speed. Without it, the total classification rate decreases to 55.51\% and down to just 11.05\% for the Orientation pattern.





Based on the previous results, we could think of just using the depth and speed information for classification. But if ${N_\alpha=N_\beta=1}$, the rate goes down to 79.69\%. That means that we need one of the two measures for correct classification. 
The final set of parameters used for all experiments is: $N=4$, ${N_{v}=4}$, ${N_{\alpha}=1}$, ${N_{\beta}=3}$ and ${N_D=3}$, for which we obtain a classification rate of 83.5\%. This rate is very close to the result with ${N_{\alpha}=3}$, with the advantage that we now use fewer symbols to represent the same information.
Several tests lead us to choose $N=4$ number of states of the HMM.

\begin{figure}[htp] 
   \centering
   \includegraphics[width=0.5\linewidth]{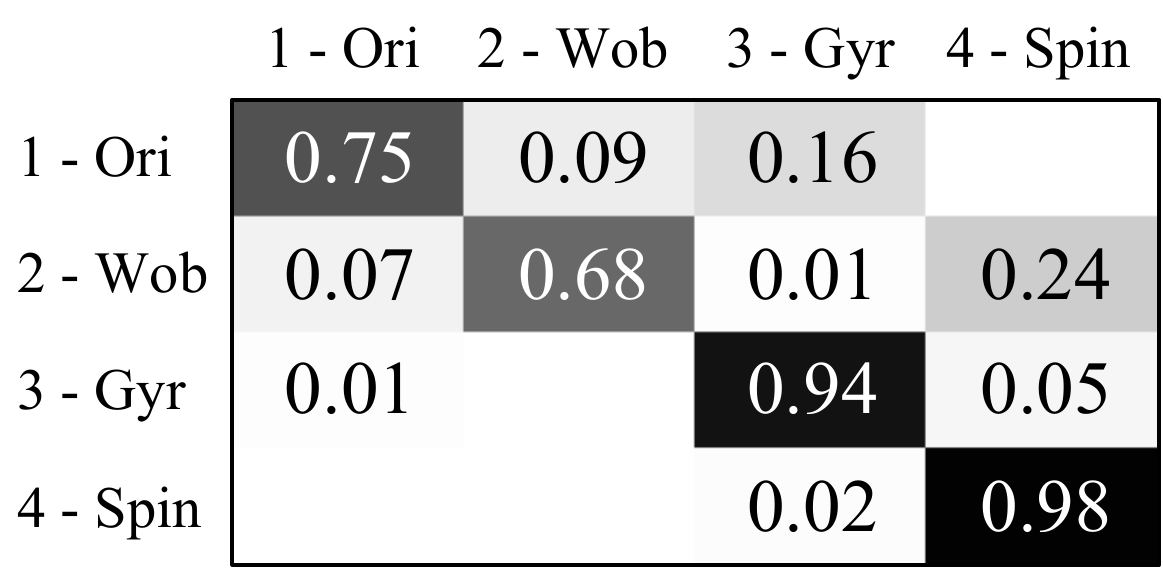} 
   \caption[Confusion matrix]{Confusion matrix; parameters $N=4$, ${N_{v}=4}$, ${N_{\alpha}=1}$, ${N_{\beta}=3}$ and ${N_D=3}$.}
   \label{fig:confusion}
\end{figure}

The confusion matrix for these parameters is shown in Figure \ref{fig:confusion}. As we can see, patterns 3 and 4 are correctly classified. The common misclassifications occur when Orientation (1) is classified as Gyration (3), or when Wobbling (2) is classified as Spinning (4). In the next section we discuss these misclassifications in detail.



\subsection{Classification on other sequences}

In this section, we present the performance of the algorithm when several patterns appear within one trajectory and also analyze the typical misclassifications.
As test data we use four sequences which contain 27, 40, 49 and 11 trajectories, respectively. We obtain classification rates of 100\%, 85\%, 89.8\% and 100\%, respectively. Note that for the third sequence, 60\% of the misclassifications are only partial, which means that the model detects that there are several patterns but only one of them is misclassified.

\begin{figure}[htp] 
   \centering
   \subfigure[]{
   \includegraphics[width=0.4\linewidth]{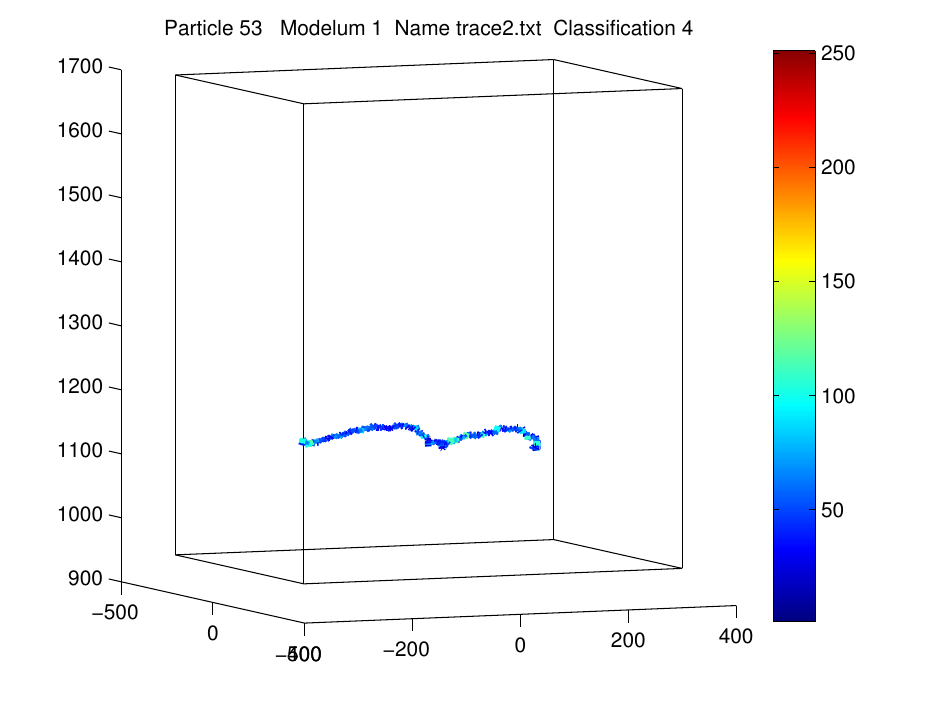}  
   \label{fig:class1miss4}
   }
   \qquad
    \subfigure[]{
    \includegraphics[width=0.3\linewidth]{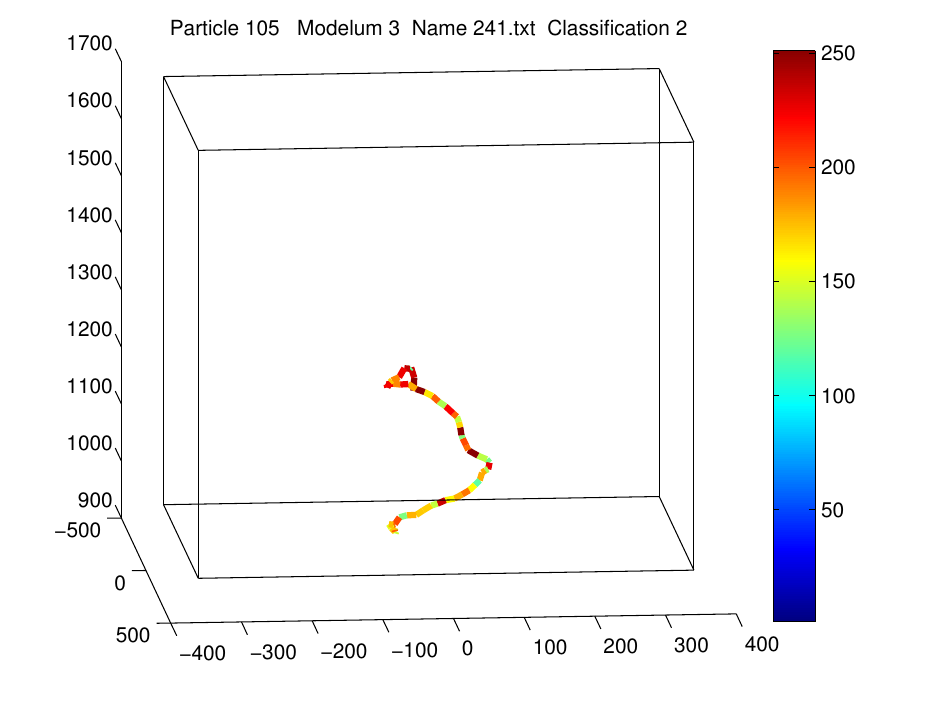}  
   \label{fig:class3miss2}
   }
   \caption[Missclassifications]{\subref{fig:class1miss4} Wobbling (pattern 2) misclassified as Spinning (4). \subref{fig:class3miss2} Gyration (3) misclassified as Orientation (1). Color coded according to speed as in Figure \ref{fig:speed2}}
   \label{fig:miss}
\end{figure}

One of the misclassifications that can occur is that Wobbling (2) is classified as Spinning (4). Both motion patterns have similar speed values and the only truly differentiating characteristics are the depth and the angle $\alpha$. Since we use 3 symbols for depth, the fact that the microorganism touches the surface or swims near the surface leads to the same classification. 
That is the case of Figure \ref{fig:class1miss4}, in which the model chooses pattern Spinning (4), because the speed is very low (dark blue), and sometimes the speed in the Wobbling pattern can be a little higher (light blue).

As commented in section \ref{patterns}, Gyration (3) and Orientation (1) are two linked patterns. The behavior of Gyration in solution is similar to the Orientation pattern, that is why the misclassification shown in Figure \ref{fig:class3miss2} can happen. In this case, since the microorganism does not interact with the surface and the speed of the pattern is high (red color), the model detects it as an Orientation pattern. We note that this pattern is difficult to classify, even for a trained expert, since the transition from Orientation into Gyration usually occurs gradually as spores swim towards the surface and interrupt the swimming pattern (which is very similar to the Orientation pattern) by short surface contacts.

\begin{figure}[htb] 
   \centering
   \subfigure[Gyration (3) + Spinning (4). Zoom on the spinning part. Color coded according to speed as in Figure \ref{fig:speed2}.]{
   \includegraphics[width=0.7\linewidth]{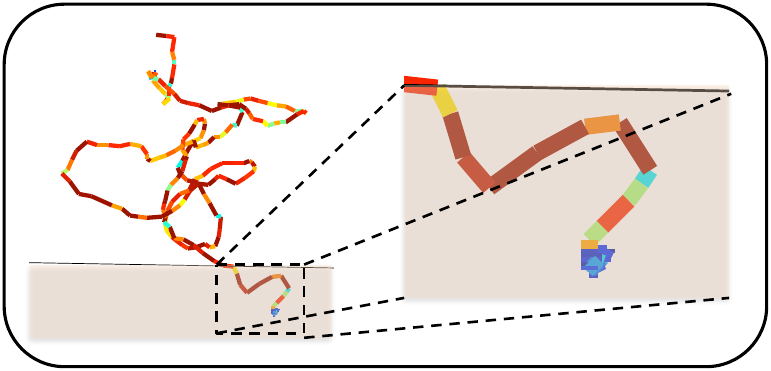}  
   \label{fig:miss1}
   }
    \subfigure[Orientation (1, red) + Gyration (3, yellow). Transition marked in blue and pointed to by an arrow.]{
    \includegraphics[width=0.51\linewidth]{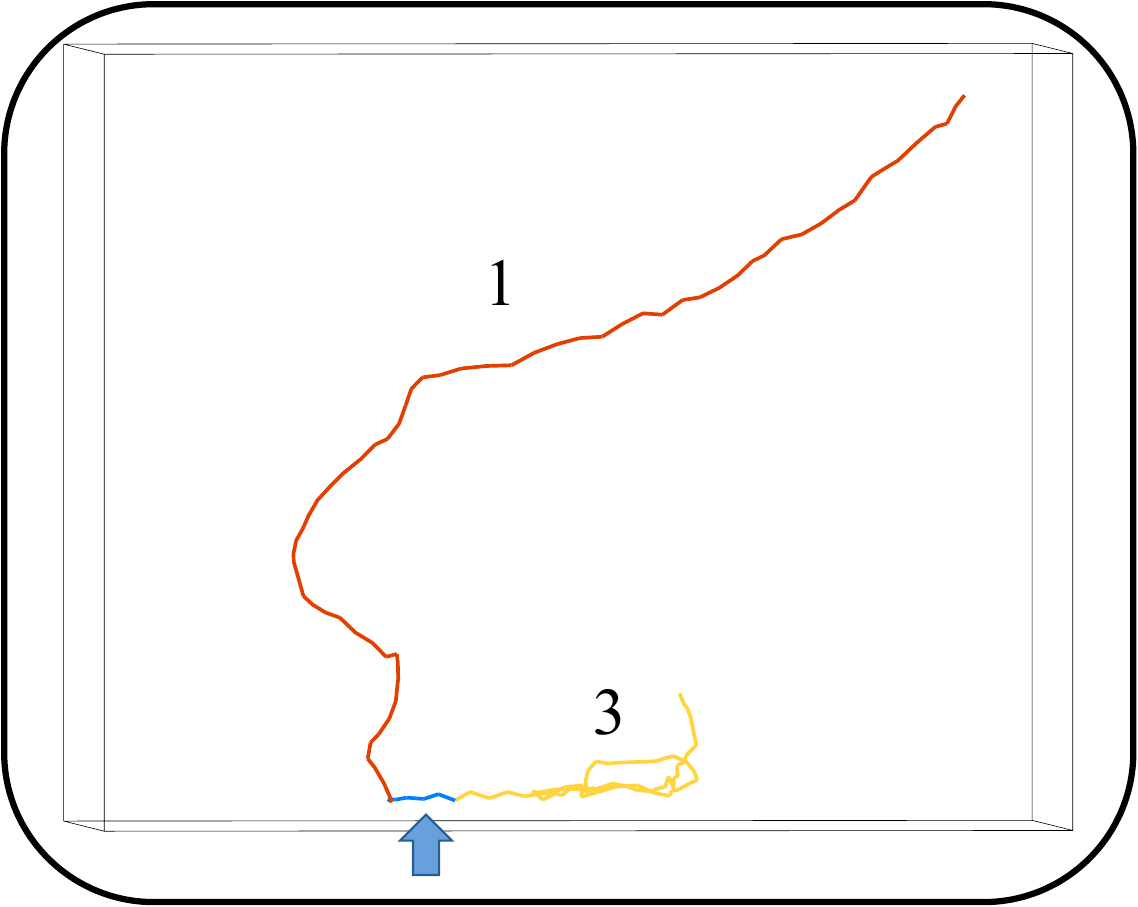}  
   \label{fig:miss2}
   }
   \caption{Sequences containing two patterns within one trajectory.}
   \label{fig:misses}
\end{figure}

In general, the model has been proven to handle changes between patterns extremely well. In Figure \ref{fig:miss1}, we see the transition between Gyration (3) and Spinning (4). 
In Figure \ref{fig:miss2}, color coded according to classification, we can see how the HMM detects the Orientation part (red) and the Gyration part (yellow) perfectly well. 
The model performs a quick transition (marked in blue) and during this period the model stays in the SWITCH state. 
We have verified that all transition periods detected by the model lie within the manually annotated transition boundaries marked by experts, even when there is more than one transition present in a trajectory.

\begin{figure}[htpb] 
   \center
   \includegraphics[width=0.7\linewidth]{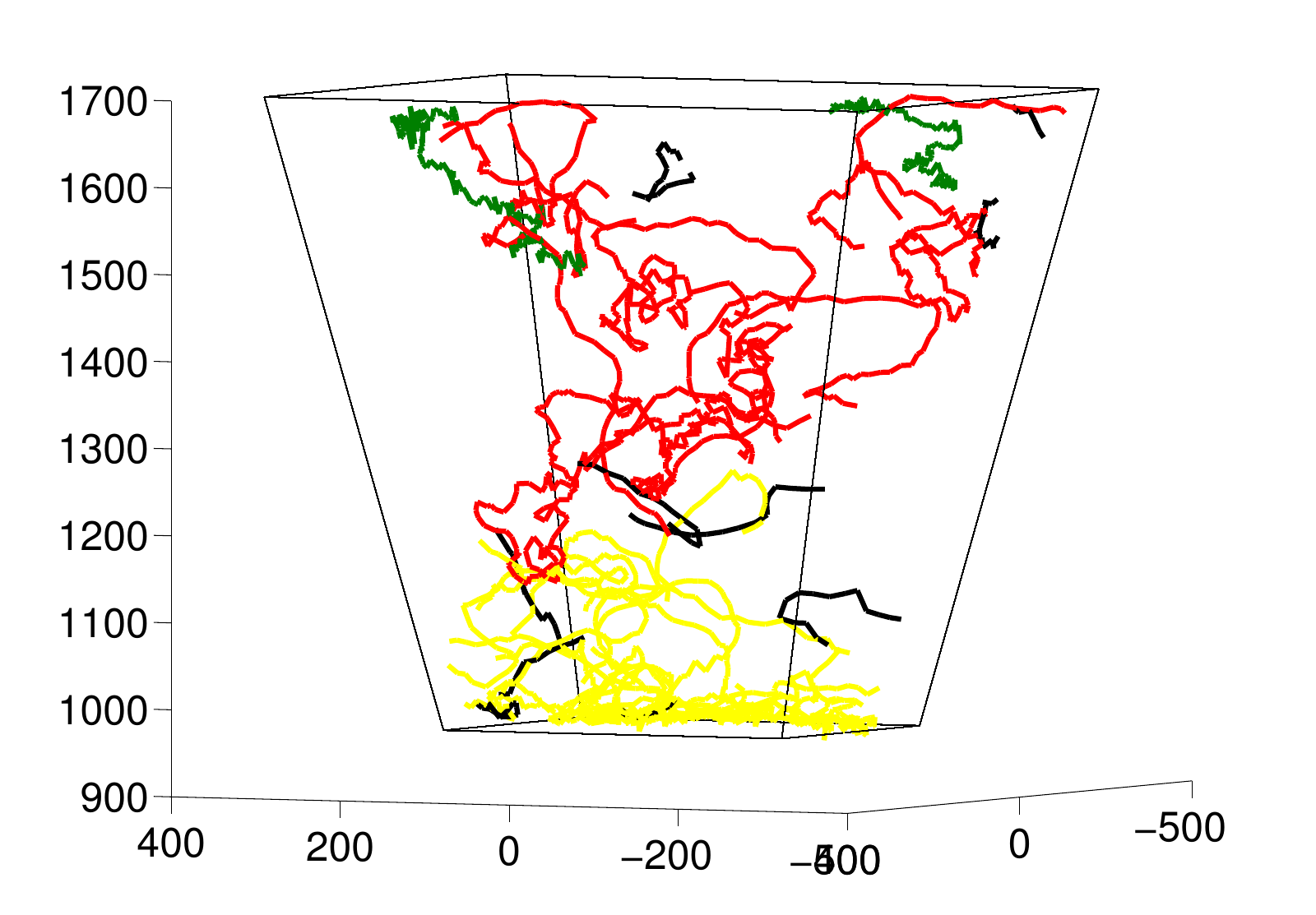}   
   \caption[Complete volume with automatically classified color-coded patterns]{Complete volume with patterns: Orientation (1, red), Wobbling (2, green), Gyration (3, yellow). The Spinning (4) pattern is not present in this sequence. Trajectories which are too short to be classified are plotted in black.}
   \label{fig:volume}
\end{figure}

The classification results on a full sequence are shown in Figure \ref{fig:volume}.

Finally, we can obtain the probability of each transition (e.g. from Orientation to Spinning) for a given dataset under study. This is extremely useful for experts to understand the behavior of a certain microorganism under varying conditions.


%
%

\section{Conclusions}

In this chapter, we presented a fully-automatic method to analyze 4D digital in-line holographic microscopy videos of moving microorganisms by detecting the microorganisms, tracking their full trajectories and classifying the obtained trajectories into meaningful motion patterns.

The detection of the microorganisms is based on a simple blob detector and can be easily adapted for any microorganism shape. To perform multiple object tracking, we modified the standard Hungarian graph matching algorithm, so that it is able to overcome the disadvantages of the classical approach. The new multi-level Hungarian recovers from missing data, discards outliers and is able to incorporate geometrical information in order to account for entering and leaving particles.
The automatically determined trajectories are compared with ground truth data, proving the method detects between 75\% and 90\% of the labeled particles.
Nonetheless, we have seen that the proposed tracking approach does not outperform the Linear Programming formulation presented in Chapter \ref{LPtracking}.

For motion pattern classification, we presented a simple yet effective hierarchical design which combines multiple trained Hidden Markov Models (one for each of the patterns) and has proven successful to identify different patterns within one single trajectory. 
The experiments performed on four full sequences result in a total classification rate between 83.5\% and 100\%.

As future work, we plan on including the motion pattern information into the tracking framework, in a similar fashion as we included social behaviors to improve pedestrian tracking in Chapter \ref{SFM}.

\addtocontents{toc}{\vspace{2em}} 

\backmatter


\label{Bibliography}

\fancyhead[LO,RE]{Bibliography}

\bibliographystyle{unsrtnat} 

\bibliography{thesis} 

\end{document}